%% file: main.tex
\documentclass[lettersize,journal]{IEEEtran}
\usepackage{amsmath,amsfonts,amssymb}
\usepackage{algorithmic}
\usepackage{algorithm}
\usepackage{array}
\usepackage{textcomp}
\usepackage{stfloats}
\usepackage{url}
\usepackage{verbatim}
\usepackage{graphicx}
\usepackage{cite}

\usepackage{times}
\usepackage{soul}
\usepackage[hidelinks]{hyperref}
\usepackage[utf8]{inputenc}
\usepackage[small]{caption}
\usepackage{amsthm}
\usepackage{booktabs}
\usepackage{balance}
\usepackage{wrapfig}
\usepackage[inline]{enumitem}
\usepackage{bm}
\usepackage{makecell}
\usepackage[normalem]{ulem}

\usepackage{outlines}
\usepackage{multirow}
\usepackage{subcaption}

\usepackage{relsize}

\usepackage{tikz}
\usepackage{pgfplots}
\usepackage{pgfplotstable}
\usepackage{xifthen}
\usepackage{upgreek}
\usepackage{mathtools}
\usepackage{xcolor}
\usepackage{varwidth}
\usepackage{needspace}

\usepgfplotslibrary{groupplots,fillbetween}
\pgfplotsset{compat=1.18}
\usetikzlibrary{shapes,fit,backgrounds,arrows,decorations.markings,arrows.meta,positioning,decorations.pathreplacing,calligraphy,shapes.misc,matrix,shadows,shapes.geometric,calc}

\pgfdeclarelayer{background}
\pgfdeclarelayer{foreground}
\pgfsetlayers{background,main,foreground}
\pgfplotsset{
    line and fill/.style={
        legend image code/.code={%
          \draw [##1,fill=none, thick] (0mm,0mm) -- (4mm,0mm);
        },
    },
}

\UseRawInputEncoding

\newtheorem{theorem}{Theorem}
\newtheorem{lemma}[theorem]{Lemma}

\newtheorem{definition}{Definition}
\newtheorem{remark}{Remark}

\newcommand{\ppw}[1]{$\text{PP}_{#1}^{\text{Obs}}$}

\newcommand{\lj}[3]{$\text{LJ}_{#1}^{#2,#3}$}
\newcommand*{\eg}{\emph{e.g.}}

\newcommand{\std}[1]{\footnotesize{\color{black}$\pm$#1}}
\newcommand{\errorband}[6][]{
    \addplot [draw=none, stack plots=y, forget plot]
      table [x={#3},y expr=\thisrow{#4}-\thisrow{#5}] {#2};
    \addplot [fill=gray!40, stack plots=y, opacity=0.16, #1, draw=none, forget plot]
      table [x={#3},y expr=2*\thisrow{#5}] {#2} \closedcycle;
    \addplot [stack plots=y, draw=none, forget plot]
      table [x={#3},y expr=-(\thisrow{#4}+\thisrow{#5})] {#2};
    \addplot [forget plot, thick, #1, fill=none]
      table [x={#3},y expr=\thisrow{#4}] {#2};
    \addlegendimage{line and fill,#1}
    \addlegendentry{#6}
}
\newcommand{\parag}[1]{\textbf{#1.}\quad}
\newcommand{\subparag}[2]{
    \ifthenelse{\isempty{#2}}
        {\textit{#1}\hspace{.2cm}}
        {(#2) \textit{#1}\hspace{.2cm}}
}

\newcommand{\coolname}{{\fontfamily{lmtt}\selectfont CLOVER}}

\input{math_commands}

\definecolor{color_env_inner}{HTML}{AAF0D1}
\definecolor{color_env_outer}{HTML}{64AA8B}
\definecolor{color_nn_inner}{HTML}{AED6F1}
\definecolor{color_nn_outer}{HTML}{21618C}
\definecolor{color_light_purple}{HTML}{C241C1}
\definecolor{color_nn_background}{HTML}{E0E0E0}

\definecolor{color_commnet_}{HTML}{2ca02c}
\definecolor{color_ic3net_}{HTML}{1f77b4}
\definecolor{color_laurel_on}{HTML}{ff7f0e}
\definecolor{color_qmix_}{HTML}{9467bd}
\definecolor{color_tarmac_}{HTML}{8c564b}
\definecolor{color_laurel_off}{HTML}{e377c2}

\definecolor{color_vdn_tarmac}{HTML}{d62728}
\definecolor{color_vdn}{HTML}{3d3d3d}

\begin{document}

\input{data/ablation_arch_a.tex}
\input{data/ablation_arch_b.tex}
\input{data/ablation_noise.tex}
\input{data/ablation_rss.tex}
\input{data/ablation_when.tex}
\input{data/sota_steps_a.tex}
\input{data/sota_steps_b.tex}
\input{data/sota_pp_tarmac.tex}
\input{data/sota_lj.tex}
\input{data/traj_comm.tex}
\input{data/dummy.tex}
\input{data/bw_limit_steps.tex}

\input{data_new/sota_steps_pp.tex}
\input{data_new/sota_steps_lj.tex}
\input{data_new/sota_steps_pp_grid7.tex}
\input{data_new/sota_steps_lj_T2.tex}
\input{data_new/sota_steps_lj_Gseven.tex}
\input{data_new/ablation_gnnmixer.tex}

\title{Wireless Communication Enhanced Value Decomposition for Multi-Agent Reinforcement Learning}

\author{%
\begin{tabular}[t]{c}
Diyi Hu\\
\textit{IEEE Member}\\
diyihu.hu@gmail.com
\end{tabular}
\hspace{1.5cm}
\begin{tabular}[t]{c}
Bhaskar Krishnamachari\\
\textit{University of Southern California}\\
bkrishna@usc.edu
\end{tabular}%
}

\maketitle

\begin{abstract}
Cooperation in multi-agent reinforcement learning (MARL) benefits from inter-agent communication, yet most approaches assume idealized channels that deliver messages reliably, and existing value decomposition methods treat agent utilities as an unstructured collection regardless of who successfully shared information with whom.
We propose \coolname{} (Communication-enhanced vaLue decOmposition oVer rEalistic wiReless channels), a cooperative MARL framework in which the centralized value mixer is conditioned on the inter-agent communication graph realized under a realistic wireless channel.
Since value decomposition serves as an implicit credit-assignment mechanism, the communication graph introduces a graph-structured relational inductive bias, architecturally constraining how utilities are mixed based on the realized communication structure.
We realize this through a GNN-based mixer whose node-specific weights are generated by a Permutation-Equivariant Hypernetwork (PEHypernet): multi-hop propagation along communication edges reshapes the mixing of agents' utility, so that structurally different topologies induce different mixing and correspondingly different credit assignment.
We prove that this mixer is permutation invariant, monotonic (preserving the IGM condition for decentralized execution), and strictly more expressive than QMIX-style monotonic mixers that do not condition on the realized communication structure.
To operate under realistic channels, we formulate an augmented MDP that isolates stochastic channel effects (e.g., medium contention, signal fading, interference) from the agent computation graph, and employ a stochastic receptive field encoder for variable-size received message sets, enabling end-to-end differentiable training.
On Predator-Prey and Lumberjacks benchmarks under $p$-CSMA wireless channels, \coolname{} achieves consistent improvements in convergence speed and final task performance over VDN, QMIX, TarMAC+VDN, and TarMAC+QMIX, with larger gains in more challenging settings.
Behavioral analysis confirms that agents learn genuine positive signaling and listening strategies, adapting communication frequency to channel conditions. A mixer ablation further indicates the communication-graph inductive bias as the important source of improvement.
\end{abstract}


\input{1_intro.tex}
\input{2_related.tex}
\input{4_method.tex}
 \input{6_exp.tex}

\needspace{5\baselineskip}
 \section{Conclusion}
 \label{sec: conclusion}

The realized communication graph encodes which agents successfully shared information and which acted in isolation, providing structural information that standard value-decomposition mixers discard. This paper showed how to exploit that structure through a communication-graph-conditioned GNN mixer that propagates individual utilities along realized communication edges. The resulting mixer provably extends the monotone function class of QMIX-style graph-agnostic mixers (Theorem~\ref{thm: exp power}) while preserving the IGM condition (Theorem~\ref{theorem: monotonicity}) and permutation invariance (Theorem~\ref{theorem: PI}). Combined with causal state alignment and a stochastic receptive-field encoder, the complete \coolname{} framework remains end-to-end trainable under realistic channel stochasticity without relying on idealized communication assumptions.

Experiments on Predator-Prey and Lumberjacks showed that communication-graph conditioning improved both convergence speed and terminal performance, with advantages that became more pronounced as coordination difficulty increased. In PP~$10{\times}10$ with 4 agents, \coolname{} converged to $19.7 \pm 0.1$ steps versus $25.0 \pm 0.9$ for the strongest graph-agnostic baseline. Behavioral analysis further indicated that agents learned semantically meaningful communication strategies, while the mixer ablation isolated graph conditioning as the main differentiating factor over graph-agnostic alternatives.

Promising directions for future work include:
(i)~scaling to larger agent populations;
(ii)~extending the framework to richer factorization classes such as QPLEX~\cite{qplex}; and
(iii)~validating the approach in more realistic large-scale or physical multi-agent systems.
\input{7_appendix}

\balance
\bibliographystyle{IEEEtran}
\bibliography{citation_l}

\end{document}

%% file: math_commands.tex
\newcommand{\paren}[1]{\left( #1 \right)}
\newcommand{\func}[2][]{
    {\text{\fontfamily{lmtt}\selectfont #1}\left(#2\right)}}

\newcommand{\att}{\delta}

\DeclarePairedDelimiterX\set[1]\lbrace\rbrace{\def\given{\;\delimsize\vert\;}#1}

%% file: data/ablation_arch_a.tex
\pgfplotstableread[col sep=comma]{
0.0,40.0,0.0,40.0,0.0
40000.0,39.95156,0.09687999999999876,39.91406,0.10528421724076313
80000.0,39.79376,0.16316122823759305,39.91406,0.16800716175211108
120000.0,39.77656,0.14171182872293994,39.5875,0.25182901341982084
160000.0,38.62814,0.8646382147464923,39.31562,0.5541002992238858
200000.0,38.829699999999995,0.4296204278197213,38.843740000000004,0.4544916043228951
240000.0,38.80466,0.45782357125862355,38.742180000000005,0.25670242227139156
280000.0,38.959379999999996,0.5333649234811009,38.149980000000006,0.3577284355485322
320000.0,37.310959999999994,0.9092813483185492,37.818760000000005,0.815869236091175
360000.0,37.05624,0.7574045381432571,37.61248,0.883235976169448
400000.0,36.19534,1.123759055313904,37.5953,1.0637240563228791
440000.0,36.74218,0.9769602497543084,36.781220000000005,1.3599940402810602
480000.0,35.44376,0.6590425649379551,36.45624,0.26218043100124844
520000.0,35.28752,1.9444043976498302,35.10938,1.0274376504683882
560000.0,33.445319999999995,1.3497894078707249,35.23906,0.7658466676822465
600000.0,33.781240000000004,1.6427902381010189,35.92342,0.7218701140787014
640000.0,33.89532,1.615513997958545,35.05782,1.5587492709220445
680000.0,32.415620000000004,0.4217670892803284,34.5047,0.7412388548909198
720000.0,31.34686,1.8061872655956814,34.1625,1.2451807884801314
760000.0,31.545299999999997,1.5516874259979028,34.09846,1.434144788506376
800000.0,30.735940000000006,1.4260263414116865,34.21718,1.5387858244733088
840000.0,30.507800000000003,2.73113493039066,32.07968,1.3645340412023446
880000.0,30.3375,1.3278418083491723,32.54218,0.4606297576145061
920000.0,29.565620000000003,2.28296337894413,31.93282,0.894980246485921
960000.0,28.448439999999998,1.9652020075300143,30.931220000000003,0.853665886398187
1000000.0,28.078120000000002,1.2332626636690174,31.190639999999995,1.1731644839492885
1040000.0,27.296879999999998,1.1616103622127352,30.81096,1.0861661504576545
1080000.0,25.868759999999998,0.9810898401267856,31.11404,1.4915461569793946
1120000.0,25.13124,1.689044978205139,30.837500000000006,1.1889867669574807
1160000.0,25.0203,1.7342750312450443,29.66406,1.3780526486313942
1200000.0,23.87656,1.2033925387835838,28.9375,0.66942874452775
1240000.0,23.94842,1.1665470353140504,29.27812,0.6372065266457974
1280000.0,23.3594,0.8601975610288604,29.879700000000003,0.5755541156138143
1320000.0,22.87342,0.7914075824756792,29.020339999999997,0.6055687990641526
1360000.0,21.776559999999996,0.8260173281475404,29.54842,0.773312866568248
1400000.0,21.829659999999997,0.4265210925616693,27.60626,0.9756211285124983
1440000.0,21.8672,0.8807864031648083,28.082819999999998,1.7231819699613853
1480000.0,21.69844,1.0958030308408542,27.67502,1.3895923667032717
1520000.0,21.14532,0.41633896478710647,27.56094,1.2406928541746345
1560000.0,20.97656,0.7244183283158979,27.518759999999997,2.074775716649874
1600000.0,21.087500000000002,0.868156969677719,27.107840000000003,1.1133705198180883
1640000.0,20.63282,0.5788149942771011,27.696859999999997,1.4742552704331777
1680000.0,20.73594,0.3434005509605362,27.66096,1.848329805635347
1720000.0,20.30312,0.4122994125632482,27.6031,2.800051610952913
1760000.0,20.25626,0.5159866843243146,26.506259999999997,2.241899670904119
1800000.0,20.017180000000003,0.7856509693241656,26.41562,1.6351976827283
1840000.0,20.2594,0.6882086689369726,26.514059999999994,1.8828705729284743
1880000.0,19.803140000000003,0.3587140231437853,26.531240000000004,2.099029081837601
1920000.0,19.91562,0.5100360512748092,26.665640000000003,1.9404588576932007
1960000.0,19.96252,0.3686757024811914,26.070299999999996,2.9178365608786248
2000000.0,19.53282,0.3181152646447506,26.31248,2.5677284921891568
2040000.0,20.32814,0.9394887452226351,25.51408,1.9806442430684013
2080000.0,19.671879999999998,0.642977540509776,25.3875,2.550890469620364
2120000.0,19.29218,0.29580057065529713,25.81094,2.5404790104230344
2160000.0,19.51248,0.6593391521819404,25.35314,2.1278274371762382
2200000.0,19.203120000000002,0.3074192602944715,24.3453,2.065951884241257
2240000.0,19.7,0.5155495786052009,24.55312,2.3529930874526594
2280000.0,19.29844,0.5407850131059472,24.31716,2.9223971164781832
2320000.0,19.94062,0.5596436094515859,25.31566,2.626033207406182
2360000.0,19.1125,0.8148507347974837,24.44688,2.248256760603646
2400000.0,19.595299999999998,0.3587086227009326,23.882799999999996,2.2946914537688947
2440000.0,19.445320000000002,0.412946681304015,23.64064,2.1559713046327866
2480000.0,19.05936,1.193079599356221,24.390639999999998,2.7355368603621484
2520000.0,19.09062,0.7209798621320852,23.64846,2.0461531649414715
2560000.0,18.90312,0.4594472827213148,23.484379999999998,2.4377561645086656
2600000.0,18.781240000000004,0.4828187096623331,23.6125,2.459696087731165
2640000.0,18.41252,0.2524219118856371,23.64844,2.56867578226603
2680000.0,18.86562,0.5918415916442513,23.510939999999998,2.3802386683692034
2720000.0,18.91718,0.6558594739728933,23.740640000000003,2.8135679445145807
2760000.0,19.085919999999998,0.8748736077857194,24.00156,3.3167075177651704
2800000.0,19.10782,0.4373684003217424,23.2297,2.578981001868762
2840000.0,19.30468,0.4282154569839811,23.040660000000003,3.0533363297219647
2880000.0,19.010920000000002,0.9427185760342268,23.28282,2.3564796756178477
2920000.0,18.782819999999997,0.6782473513402024,22.7328,2.2301167816955236
2960000.0,18.70626,0.9991030629519659,23.69218,3.364004849223616
3000000.0,18.639059999999997,0.5343507971361136,23.240640000000003,3.823204604307753
3040000.0,19.02812,0.5653059256721097,23.393760000000004,3.1647022928547317
3080000.0,18.81094,0.3799433147194465,22.36408,2.613155747673682
3120000.0,18.59374,0.25454325054889965,23.571859999999997,2.7904768593199267
3160000.0,18.63124,0.6609563150466146,22.38438,3.016018449810942
3200000.0,18.79376,0.20577640875474487,22.787520000000004,3.2954159484957284
3240000.0,18.4375,0.4236990205322634,22.72344,3.4447381036009106
3280000.0,18.6422,0.4839833096295781,22.959380000000003,2.8039052526075117
3320000.0,18.49844,0.6387068610873072,22.59842,2.9004524311906925
3360000.0,18.20626,0.7658374046754315,22.690640000000002,4.033215667231298
3400000.0,18.529680000000003,0.5998981493553712,22.21874,3.4020553990786224
3440000.0,18.545319999999997,0.6873030289472035,22.45468,3.104293420667576
3480000.0,18.53124,0.316271184903083,22.114060000000002,3.570112894349421
3520000.0,18.54218,0.2974981102460983,22.751560000000005,3.715978485190677
3560000.0,18.33594,0.4484561632980424,22.450000000000003,3.2513062039740275
3600000.0,18.91094,0.8151657809304805,22.489060000000002,3.163892422064947
3640000.0,18.910919999999997,0.49060044598430524,22.10312,3.259150012748722
3680000.0,18.1578,0.5815671517546362,22.598419999999997,4.436992049305475
3720000.0,18.05626,0.5969382232693765,22.12344,3.6965445500358842
3760000.0,18.935899999999997,0.6149039242027976,22.517200000000003,3.330350159968167
3800000.0,18.40624,0.6677361772436775,22.132820000000002,2.982218357129472
3840000.0,18.828120000000002,0.47808696865737815,22.4875,4.256578268515687
3880000.0,18.71562,0.5457531068166266,22.298440000000003,3.7711884877847197
3920000.0,19.10938,0.4692482771412167,22.457820000000005,2.8593988811636613
3960000.0,18.29842,0.5875508093773676,22.20938,4.3278801348466205
4000000.0,18.732799999999997,0.6035989396942308,22.17344,2.680349898502059
}\dataAblationArchA

%% file: data/ablation_arch_b.tex
\pgfplotstableread[col sep=comma]{
0,0,0,0,0
}\dataAblationArchB

%% file: data/ablation_noise.tex
\pgfplotstableread[col sep=comma]{
8000.0,40.670925,0.06113053144706004,40.76526,0.09613296208897339
1608000.0,39.511250000000004,0.053749999999998875,40.13974,0.5992720971311779
3208000.0,39.436575000000005,0.2489985680179698,39.97098,0.9751651991329475
4808000.0,39.465325,0.12289046698177863,40.05748000000001,1.1719897190675346
6408000.0,39.586575,0.13416980612268772,39.46674,1.9716082040811238
8008000.0,39.4156,0.20196283073872817,38.93052,1.5756114386485018
9608000.0,39.816275,0.3955150526528689,38.75726,1.8439991611711757
11208000.0,39.41437500000001,0.19367954686801625,37.27274,2.4041828770707085
12808000.0,39.7506,0.5766335101257997,36.43876,2.8797273875143117
14408000.0,39.868425,0.4386707955574428,35.30924,3.4613916869374957
16008000.0,39.286225,0.9506295292462799,35.6665,4.462902267807352
17608000.0,39.41035,0.6042125722789932,34.522980000000004,4.116422927931482
19208000.0,38.949400000000004,0.9437037167458864,33.34778,4.93210798600355
20808000.0,38.973150000000004,0.6718050665929821,32.4865,3.8632661896379847
22408000.0,38.888725,0.6031570084770619,30.942500000000003,4.36004634746008
24008000.0,38.8447,1.060909524888905,29.66452,2.8513294347724902
25608000.0,38.935649999999995,2.1956204743762067,28.86676,1.9579188099612301
27208000.0,38.68905,1.042302764315629,27.427000000000003,2.022960330802362
28808000.0,38.106575,1.3049614714140036,28.338480000000004,2.1929876747487653
30408000.0,37.275025,2.2055741104925497,26.104000000000003,1.0438759868873313
32008000.0,36.5569,3.4169747562719857,25.844279999999998,1.0892504971768426
33608000.0,37.368424999999995,2.591981609092743,25.53824,1.4627601534086165
35208000.0,36.308425,3.841166481926421,25.4105,1.4068615610642008
36808000.0,36.436875,3.858643594823835,24.90128,0.8427855751019954
38408000.0,36.145325,3.4030641989058927,25.077499999999997,1.6311123125033415
40008000.0,35.452825,4.181292896566205,24.754260000000002,0.9477385010645082
41608000.0,34.551525,4.959640001237487,24.156259999999996,1.1770100366606906
43208000.0,36.345,4.529298717792853,24.252000000000002,1.2746371452299665
44808000.0,34.393125,4.993965785513053,23.8195,0.8550015649108473
46408000.0,34.032174999999995,5.561226745231217,23.954,1.1787187433819835
48008000.0,33.34155,6.151527926661797,23.471,0.6295879414347136
49608000.0,33.005,6.436668543586815,23.27524,0.503658148350644
51208000.0,33.2128,6.400097137153466,23.33474,0.9523827331488117
52808000.0,33.19685,6.459780350174454,23.078480000000003,1.1646241735426923
54408000.0,32.5428,7.250214970261227,23.27202,0.6682592472985318
56008000.0,32.3947,7.199642770304651,22.92226,1.0201139183444168
57608000.0,33.114025,6.496995378009362,22.735000000000003,0.7673428177809444
59208000.0,31.869675,7.469286932959197,22.95176,0.6917685786446216
60808000.0,31.9212,7.541159766574369,22.62378,0.9667098456103562
62408000.0,31.745599999999996,7.694442490200312,22.436,0.8169009585010902
64008000.0,31.632524999999998,7.685700889110568,22.675500000000003,0.7172350047229984
65608000.0,31.819999999999997,7.473169681935503,22.567500000000003,0.8155965767461266
67208000.0,31.5744,8.072906949172648,22.50726,0.8376147720760427
68808000.0,31.53595,8.09875912887277,22.63824,0.9694818546006938
70408000.0,31.93655,7.730257213217423,22.72674,1.4895200859337212
72008000.0,31.359075,8.005210789659133,22.43324,0.9879958837970929
73608000.0,31.461550000000003,8.475084945444499,22.08,0.8058916651759097
75208000.0,31.2428,8.231580960848285,22.14976,0.7418645633806747
76808000.0,31.257175000000004,8.361562729949169,22.022,0.9481776078351571
78408000.0,30.98155,8.551098384564405,21.775,0.774366112378376
80008000.0,31.788724999999996,7.957415873063,22.03648,1.1904049149764127
81608000.0,31.328749999999996,8.32294401233722,22.06148,0.6659908870247391
83208000.0,31.12595,8.39834623914137,21.88148,0.7560997034783173
84808000.0,30.873425,8.519225423527365,22.248,0.883614990818966
86408000.0,31.2072,8.375482144032068,21.79948,0.6263712011259774
88008000.0,30.9553,8.450324534596291,22.15774,0.7880546341466432
89608000.0,31.1625,8.392418600737216,21.797739999999997,0.6903648357209401
91208000.0,30.981875,8.554777598913661,21.65326,0.7895317513564599
92808000.0,30.839350000000003,8.504042637034459,22.003500000000003,0.806147242133843
94408000.0,30.865000000000002,8.242603356646493,21.90826,0.783616798186461
96008000.0,31.010650000000002,8.304187914691,21.81326,0.5924175979830443
97608000.0,30.6644,8.644698576873575,21.788220000000003,0.6402562155887281
99208000.0,30.927199999999996,8.823334383610314,21.771759999999997,0.7597580314810762
100808000.0,31.018725000000003,8.538226332610012,21.775760000000002,0.8786665957005529
102408000.0,30.932499999999997,8.443373798133067,21.834780000000002,0.7831094212177507
104008000.0,30.84815,8.719886878423367,21.864259999999998,0.5468218781285185
105608000.0,30.7641,9.02451739125146,21.73752,0.739654598850031
107208000.0,30.814725,8.555774399017016,21.40102,0.7441399637165046
108808000.0,30.805,8.816025101767803,21.51376,0.4304331845943112
110408000.0,30.575650000000003,9.16595718119499,21.58102,0.7077918914483262
112008000.0,30.512175,9.167229306713944,21.55248,0.8264774053777879
113608000.0,30.62,8.85631563320775,21.28174,0.39871168831625653
115208000.0,30.48875,8.938812928040278,21.63074,1.084729397776238
116808000.0,30.437825,9.078494157175793,21.44198,0.855997419154987
118408000.0,30.251575,9.203544049271185,21.19726,0.749373825537028
120008000.0,30.366549999999997,9.082757804351054,21.282,0.7939666718446061
121608000.0,30.31155,9.337536175699672,21.387539999999998,0.822920075341464
123208000.0,30.424725,9.127621243614076,21.12038,0.7190771012902579
124808000.0,30.083099999999998,9.187074329458753,21.25976,0.7946416213614793
126408000.0,30.375024999999997,9.11181050378436,21.271,0.8747642402384767
128008000.0,30.415000000000003,8.968480492536068,21.28588,0.8206520490439291
129608000.0,30.33185,9.077572606291838,21.27478,0.39309002225953227
131208000.0,30.584675000000004,9.114346125031405,21.33348,0.7970972598121254
132808000.0,30.248425,9.28142477138478,21.228019999999997,0.5672750704904986
134408000.0,30.281899999999997,9.306851935805147,21.28676,0.738524468382734
136008000.0,30.626575,9.168704349136524,21.326019999999996,0.7208566165334129
137608000.0,30.356275,9.296519209191954,21.3115,0.7021476169581433
139208000.0,30.529075000000006,9.11692394351708,21.308999999999997,0.7252892360982619
140808000.0,30.2528,9.014045582866773,21.279239999999998,0.6300566723716214
142408000.0,30.284099999999995,9.165808455613721,21.355240000000002,0.5985290355529962
144008000.0,30.328125,9.207756880308851,21.35874,0.5598139426630961
145608000.0,30.212175000000002,8.976264502111944,21.23726,0.5955500670808468
147208000.0,30.383775,9.155894250802321,21.260119999999997,0.6944238141077831
148808000.0,30.410925,9.110133418994202,20.87886,0.7157674275908348
150408000.0,30.244699999999998,9.202409311968252,21.178620000000002,0.7140018862720184
152008000.0,30.259050000000002,9.315644164924937,21.136740000000003,0.7544094951682403
153608000.0,30.302825,9.215442247492792,21.196,0.7948035455381409
155208000.0,30.34,9.12376842209402,21.137999999999998,0.7114808866020222
156808000.0,30.118425000000002,9.294556616206876,21.24874,0.829276568100173
158408000.0,30.305325000000003,9.269296133842905,21.121019999999998,0.8480616496458264
}\dataAblationNoise

%% file: data/ablation_rss.tex
\pgfplotstableread[col sep=comma]{
8000.0,40.7665,0.09527066704920384,40.76974,0.08713680278733911
1592000.0,39.40652,0.25284514153924353,39.974000000000004,0.5895398001831605
3176000.0,39.37274,0.28121612044831473,40.15926,0.6813516466553844
4760000.0,39.490759999999995,0.25577643832065433,39.50224000000001,1.4548415537095436
6344000.0,39.756780000000006,0.4703005906864231,39.36326,1.3973704542461176
7928000.0,39.66302,0.6653329855042522,38.54174,2.0242786148156577
9512000.0,39.88374,0.31073513866313773,37.794,2.4991335050372943
11096000.0,39.7085,0.6029438215953463,37.30574,3.400582121696225
12680000.0,39.40574,0.19907006404781205,36.114,3.3669647684524393
14264000.0,39.407740000000004,0.44286951396545876,35.55574,2.018757970238135
15848000.0,39.34674,0.8856318142433663,35.6295,4.004442917060998
17432000.0,39.61376,0.43509220448084324,35.46476,3.883346933561306
19016000.0,39.393980000000006,0.49980252260267793,32.8805,3.4538768038249423
20600000.0,38.768739999999994,1.3296234994914922,31.52326,3.460967723975478
22184000.0,38.56924,1.7829382452569704,29.831259999999997,2.901694406101373
23768000.0,38.03654,2.0462646041995654,29.21148,2.4634512095026353
25352000.0,38.084720000000004,1.717225985594209,28.21524,2.0275188794188823
26936000.0,38.220240000000004,1.53017695264306,28.879,3.387671066086551
28520000.0,38.09776,1.8445040922697886,27.761020000000002,1.9507520758927837
30104000.0,38.141479999999994,1.7393558536423772,26.708,1.987002837441356
31688000.0,38.10752,1.726494606304924,26.509000000000004,1.4680557128392633
33272000.0,37.69802,2.33846280825674,26.05402,1.482748213824586
34856000.0,38.61376,2.1495646420612706,25.823,1.7227868922185352
36440000.0,37.80048,2.4608531336916477,25.38598,1.1292553039946283
38024000.0,37.55998,2.79475464497333,25.316,1.4878067750887547
39608000.0,37.593239999999994,3.0754472628221072,24.322499999999998,0.9509369463849844
41192000.0,38.08374,3.871324696069808,24.649520000000003,1.1700002758974035
42776000.0,37.2105,3.779412054275108,24.50926,1.1448950565008131
44360000.0,37.23026,3.639673923636567,24.544239999999995,1.213989780187626
45944000.0,36.662279999999996,3.8723608485780363,24.191,1.0956193116224264
47528000.0,36.63376,3.723867579600542,24.855240000000002,0.8837131923876655
49112000.0,36.186,4.141262066085653,23.496000000000002,1.2422782602943674
50696000.0,35.985240000000005,4.011681618274312,23.532980000000002,1.1643137230145497
52280000.0,35.963,4.068561152545208,23.21926,1.0069772045086223
53864000.0,36.132259999999995,4.717816457896598,24.30274,2.838141769961465
55448000.0,35.00926,4.579752991854474,23.61022,1.0638539211752713
57032000.0,35.9275,4.64464800345516,23.23702,1.1304576708572511
58616000.0,35.155240000000006,4.2758050856417675,23.090980000000002,1.0413147293685996
60200000.0,34.53148,4.806063241947613,22.999519999999997,1.1928352868690633
61784000.0,36.370259999999995,4.753249051585661,22.83396,1.0534324232716594
63368000.0,34.7205,4.681004654558676,22.188499999999998,0.9404739634886228
64952000.0,34.35224,5.2244274467543335,22.446,0.6242693136779989
66536000.0,34.4615,5.130386140633081,22.067479999999996,0.7950104411892968
68120000.0,33.416740000000004,5.693094816213762,21.82,0.5248951095218927
69704000.0,32.210480000000004,6.3318452085944115,21.711,0.6815645794787161
71288000.0,31.924,6.520144285213327,21.92652,0.4906469867430145
72872000.0,31.800739999999998,6.720880445774943,21.9495,0.5793931963701328
74456000.0,31.23448,7.207418088719428,21.753,0.6472369550636002
76040000.0,30.780520000000003,6.5184622424004255,21.66624,0.4819875790930721
77624000.0,30.704040000000003,6.6162896573835095,21.64748,0.621668662230935
79208000.0,30.69198,6.175118612755547,21.54222,0.7530748207183667
80792000.0,30.596719999999998,5.902830347011508,21.64902,0.45943752741803845
82376000.0,30.9005,6.899230245179529,21.34698,0.6362484432986844
83960000.0,30.61074,7.274318441641115,21.3865,0.5527279403106014
85544000.0,30.84074,6.848504865326445,21.33452,0.6200279200165095
87128000.0,31.28752,6.904673730568302,21.155,0.5952415476090366
88712000.0,29.78426,6.3560382603002,21.37972,0.4998762462850184
90296000.0,30.05378,6.496726177514333,21.08026,0.34319272486461566
91880000.0,30.64626,7.82500232756515,21.42048,0.6194594592061692
93464000.0,29.57448,6.5351519802985445,21.416019999999996,0.6704537162250646
95048000.0,29.28548,6.511280184848445,21.29878,0.6275610931216177
96632000.0,29.15874,6.72979699919693,21.370759999999997,0.5143154541718536
98216000.0,28.745240000000003,6.646952813613167,21.316,0.6363848489711241
99800000.0,29.275239999999997,7.30196821455695,21.38352,0.6211523046725331
101384000.0,28.746979999999997,6.7717337574361265,21.369500000000002,0.5964560972946787
102968000.0,28.806239999999995,6.848448170965448,21.2595,0.7287034046853352
104552000.0,28.710240000000006,6.719778858444673,21.301840000000002,0.6307738504408696
106136000.0,28.39624,6.504642497970199,21.40002,0.41154878398556827
107720000.0,28.351260000000003,6.390377314869601,21.32226,0.5390544558762131
109304000.0,27.86974,6.378187536157901,21.348779999999998,0.719370854010642
110888000.0,27.67826,6.639360736275746,21.337020000000003,0.5461895692889057
112472000.0,27.674500000000002,6.53645673710153,21.1875,0.6686583402605548
114056000.0,27.633259999999996,6.217316495916868,21.08498,0.6814229417916595
115640000.0,27.42526,6.2900172294199646,21.3695,0.5327673038015758
117224000.0,27.15724,6.506886787888658,21.393,0.7464507190699191
118808000.0,26.8705,6.4588656338400465,20.997120000000002,0.5040018388855333
120392000.0,26.862479999999998,6.672075794053901,20.987740000000002,0.634620442784504
121976000.0,26.726980000000005,6.540182014409081,20.92752,0.43130580983798517
123560000.0,26.87038,6.519604287500892,21.059019999999997,0.5767173845134204
125144000.0,26.743239999999997,6.3828440716031905,21.30574,0.39433560123326505
126728000.0,26.58876,6.534262791348386,21.036,0.652710560662228
128312000.0,27.06176,6.581377241155532,21.06322,0.36402904499503835
129896000.0,26.96126,6.4776789613564505,20.988500000000002,0.4309853547395784
131480000.0,26.55898,6.646135844353468,21.22826,0.43244694749761037
133064000.0,26.05036,6.920988694283499,21.075480000000002,0.6116793454090139
134648000.0,26.180380000000003,6.94135094758938,20.947499999999998,0.5330657107711956
136232000.0,25.910379999999996,6.829906562581951,21.194499999999998,0.4419372127350219
137816000.0,26.16126,7.16461518508845,20.87774,0.5406385173107814
139400000.0,25.83088,7.001157031919797,21.03076,0.5205810084895537
140984000.0,26.019479999999998,7.0155147410293415,21.06636,0.4484844594854989
142568000.0,25.93398,6.921849172121564,21.058,0.5951126582421188
144152000.0,25.602359999999997,7.005471239709716,20.75514,0.4519351617212368
145736000.0,25.58648,7.135614401409314,20.79352,0.558420796174354
147320000.0,25.730740000000004,7.244877175660053,20.686020000000003,0.6341446440678971
148904000.0,25.66476,7.089041644002381,20.94236,0.5743952388382061
150488000.0,25.64024,7.1865064187266965,20.86286,0.508150991733756
152072000.0,25.759500000000003,7.055856932506496,20.81622,0.5482073746311699
153656000.0,25.616859999999996,6.99327066017039,20.52888,0.5085606095638943
155240000.0,25.36988,7.1227367867695355,20.675900000000002,0.2959898444203778
156824000.0,25.299,6.9829714789622335,20.77178,0.3406834976925056
158408000.0,25.871019999999998,6.751558367191978,20.532519999999998,0.15822042093231806
}\dataAblationRss

%% file: data/ablation_when.tex
\pgfplotstableread[col sep=comma]{
0.0,40.0,0.0,1.0,0.0,40.0,0.0,0.0,0.0,40.0,0.0,0.83026,0.3393800088396487
40000.0,39.923825,0.13193897026655788,1.0,0.0,39.72136666666666,0.22966947748642808,0.0,0.0,40.0,0.0,0.40408,0.11682471313895874
80000.0,39.6289,0.3644928051416103,1.0,0.0,39.95053333333333,0.06995643088539019,0.0,0.0,39.676579999999994,0.175523416101669,0.48768,0.16649699576869248
120000.0,39.552749999999996,0.5422518257230686,1.0,0.0,39.57293333333333,0.46291218989734606,0.0,0.0,39.40468,0.48468282164731236,0.60634,0.016512492240724888
160000.0,39.005874999999996,0.6447425081961006,1.0,0.0,38.98696666666667,0.4200473015691875,0.0,0.0,39.39378,0.7518042415416412,0.52684,0.13187482853069424
200000.0,39.08985,0.7152378013639925,1.0,0.0,38.375033333333334,1.1324150309061696,0.0,0.0,37.86406,1.2450487662738359,0.60476,0.13920061206761988
240000.0,38.209,0.8356188245845112,1.0,0.0,38.080733333333335,0.21206634706043426,0.0,0.0,38.51404,0.7030079959715952,0.5876,0.15300709787457575
280000.0,37.794925,0.5731010442103563,1.0,0.0,39.0026,0.4272813826976308,0.0,0.0,38.38438,1.01504800162357,0.51512,0.15857225986912085
320000.0,37.480475,1.5833983601339885,1.0,0.0,38.27863333333334,0.447281017805236,0.0,0.0,36.493739999999995,1.728187180371386,0.5253599999999999,0.16815592288111653
360000.0,37.322275,1.5442587985422005,1.0,0.0,38.08076666666667,0.31180478223116176,0.0,0.0,36.26094,1.493811539117301,0.5720000000000001,0.14895591294070876
400000.0,36.12695000000001,2.001828176317836,1.0,0.0,38.140633333333334,0.21001302076035383,0.0,0.0,35.62656,1.4824707816344995,0.55018,0.07913258241710555
440000.0,36.25585,1.7603226472723692,1.0,0.0,37.2578,0.47664251453963435,0.0,0.0,34.981260000000006,0.6348252912416156,0.5608199999999999,0.10885203535074571
480000.0,35.513675,1.4098407044325967,1.0,0.0,37.5807,0.33903121783497486,0.0,0.0,35.04376,1.2182393420013993,0.55696,0.12484980736869401
520000.0,35.695325,3.0543664395214605,1.0,0.0,37.916666666666664,0.7608654078671786,0.0,0.0,35.4625,1.2633156248539008,0.58714,0.09041905993760385
560000.0,34.537125,2.3560783532970637,1.0,0.0,36.91143333333333,0.3963811998008427,0.0,0.0,33.818740000000005,1.3857626399928675,0.62532,0.04051317810293336
600000.0,33.982425,3.1200279793416916,1.0,0.0,36.802099999999996,0.22100381595498977,0.0,0.0,33.6172,0.5336223870865976,0.55454,0.1149687018279323
640000.0,34.541,3.567308786045862,1.0,0.0,36.830733333333335,0.9653237464993579,0.0,0.0,34.28438,1.6130342406781069,0.5748599999999999,0.166910737821148
680000.0,34.46095,2.958975454866091,1.0,0.0,37.393233333333335,0.666623713616276,0.0,0.0,32.63438,1.2771804060507665,0.6257999999999999,0.10338458298992166
720000.0,34.072275,3.499279331787476,1.0,0.0,37.080733333333335,0.4316997207421959,0.0,0.0,31.584339999999997,2.1728534774346855,0.53026,0.14774887613785764
760000.0,33.04295,3.0217285587060916,1.0,0.0,36.60413333333333,1.1274108429888765,0.0,0.0,31.246879999999997,0.9983059839548197,0.60828,0.12339905023945685
800000.0,31.703125000000004,3.5451892815299733,1.0,0.0,36.60156666666666,1.1883622969822354,0.0,0.0,31.99842,0.5914790322572734,0.59202,0.10749994232556594
840000.0,31.52345,3.08958234920191,1.0,0.0,36.0599,1.775417455135551,0.0,0.0,29.66406,1.296725021891688,0.58002,0.13804583876379614
880000.0,31.248075,2.4573858461533873,1.0,0.0,36.2474,0.9066790538369507,0.0,0.0,31.07654,0.91202837148852,0.6104,0.13645954711928368
920000.0,29.773425000000003,3.0877125986521166,1.0,0.0,35.1901,0.7672080856369173,0.0,0.0,30.35,1.441572993642708,0.59512,0.12054077152565434
960000.0,29.425775,3.4427807273881106,1.0,0.0,35.21353333333334,0.40150717165312305,0.0,0.0,28.13906,1.248387456841825,0.6060000000000001,0.14068914670293514
1000000.0,28.24415,4.533859602204284,1.0,0.0,35.51303333333333,1.9077924316398311,0.0,0.0,29.34686,0.8813294380650176,0.6051200000000001,0.13175708557796809
1040000.0,28.341799999999996,3.5158832588696676,1.0,0.0,35.1328,1.1842630225869053,0.0,0.0,27.89064,1.7342202185420392,0.6025,0.1414705340344766
1080000.0,28.334000000000003,3.6829558244703393,1.0,0.0,34.55206666666667,0.9444554139232254,0.0,0.0,27.095319999999997,0.9668569085443821,0.6044999999999999,0.1158973856478221
1120000.0,27.9043,3.656403553356767,1.0,0.0,34.645833333333336,0.8034853009787359,0.0,0.0,26.021859999999997,1.0134242657446089,0.6062399999999999,0.15076268238526402
1160000.0,26.2539,3.227680220374999,1.0,0.0,33.393233333333335,0.44436525766785834,0.0,0.0,26.48594,1.1315795413491703,0.59628,0.13911433283454294
1200000.0,26.740225,3.6129679941669837,1.0,0.0,33.890633333333334,0.6308626598203075,0.0,0.0,25.70624,1.160170223027638,0.63832,0.11792813743971368
1240000.0,27.164099999999998,4.845275470497007,1.0,0.0,33.60156666666666,1.1638656575777493,0.0,0.0,24.80624,1.1980525507672866,0.5997199999999999,0.11871246606822722
1280000.0,25.951175,4.167155158123465,1.0,0.0,33.1745,0.305431268864209,0.0,0.0,25.0625,1.8954536480747826,0.62296,0.11250442835728731
1320000.0,25.68945,4.7001136467643,1.0,0.0,32.95053333333333,1.4769764821719036,0.0,0.0,24.234379999999998,1.0161886132013087,0.63964,0.1050384805678376
1360000.0,25.755875000000003,4.215259705744712,1.0,0.0,32.9427,0.22212217358921874,0.0,0.0,24.3625,0.6291870691614699,0.6359,0.12044253401518917
1400000.0,24.650375,4.4780468450960855,1.0,0.0,32.015633333333334,0.780979412162845,0.0,0.0,23.118740000000003,0.7444272337844716,0.63294,0.11174597263436384
1440000.0,24.2637,3.594743277342626,1.0,0.0,31.807333333333332,1.8031426017434724,0.0,0.0,22.91874,1.2033304277711918,0.6281399999999999,0.10298669040220682
1480000.0,24.265625,4.359374350394217,1.0,0.0,31.666666666666668,1.2804392848636836,0.0,0.0,22.96876,1.0235134422175407,0.6377599999999999,0.11726942653564908
1520000.0,23.1543,3.746503628051092,1.0,0.0,31.877633333333335,0.11417417493558639,0.0,0.0,22.85624,0.2658325232171563,0.64828,0.12389116836966226
1560000.0,23.466825,3.479229318667426,1.0,0.0,31.86196666666667,1.8263516388934777,0.0,0.0,21.895319999999998,0.9060112877884021,0.64324,0.10221214409256858
1600000.0,23.9824,3.8791811971084824,1.0,0.0,29.763033333333336,0.38083501531357244,0.0,0.0,21.775,0.6363373287808913,0.64796,0.12567546459034876
1640000.0,23.1504,2.308075017195065,1.0,0.0,30.362,1.3835580893719877,0.0,0.0,21.83592,1.0956938977652477,0.6042799999999999,0.10163879967807572
1680000.0,22.24415,2.2857120919967158,1.0,0.0,29.4375,1.4270884508910666,0.0,0.0,22.13126,1.0530865502891957,0.6311,0.1151733476113289
1720000.0,22.955075,3.7306216274067516,1.0,0.0,30.3203,1.0405229870919086,0.0,0.0,21.473419999999997,0.9637699256565335,0.64118,0.08477717617377922
1760000.0,22.289025,2.7858625113014814,1.0,0.0,29.223966666666666,0.7590399213626536,0.0,0.0,21.1547,0.8189381710483402,0.6457200000000001,0.12431584613395028
1800000.0,22.125,3.1247512949033234,1.0,0.0,29.3776,0.9589163084788305,0.0,0.0,21.0125,0.763851236825601,0.65196,0.12015324548259194
1840000.0,22.214824999999998,3.306399944331447,1.0,0.0,28.843733333333333,1.3346747077679846,0.0,0.0,20.72498,0.7102854676818323,0.67884,0.11064152204303772
1880000.0,21.720699999999997,2.824769803895532,1.0,0.0,28.97136666666667,0.7855000374849692,0.0,0.0,21.329700000000003,1.161752918653532,0.6549600000000001,0.10817098686801374
1920000.0,21.625025,2.7002973469370004,1.0,0.0,28.42186666666667,1.8884391461969032,0.0,0.0,21.318760000000005,0.44221856406080523,0.63558,0.0968481987442203
1960000.0,21.742199999999997,2.0110233203521037,1.0,0.0,28.17446666666667,1.9058636280932826,0.0,0.0,20.635939999999998,0.42074230402943824,0.6481,0.0982220341878542
2000000.0,21.2969,2.8461090061696517,1.0,0.0,27.65886666666667,2.4486036270132048,0.0,0.0,20.9625,0.9410568569432988,0.6540000000000001,0.0970197917952827
2040000.0,20.578125,2.4540059029829164,1.0,0.0,27.992199999999997,1.184184929251622,0.0,0.0,21.13906,0.5512060580218627,0.6461,0.09159181186110471
2080000.0,20.302725,1.8355772870884526,1.0,0.0,27.880200000000002,2.3022206033885344,0.0,0.0,20.72656,0.7885135346967731,0.66686,0.09033410430175308
2120000.0,20.392575,1.736777905166633,1.0,0.0,26.718733333333333,2.612932193702869,0.0,0.0,20.7172,0.3743053993732928,0.6382199999999999,0.12368298832094898
2160000.0,20.623025,1.861096781974275,1.0,0.0,26.562466666666666,1.9749742383692561,0.0,0.0,20.00626,0.7571136548762014,0.664,0.11331739495770277
2200000.0,20.527350000000002,0.9359868388497782,1.0,0.0,26.406266666666667,2.192964424902714,0.0,0.0,19.73594,0.27197353253579715,0.6607800000000001,0.085031697619182
2240000.0,20.61915,1.4549222358944132,1.0,0.0,25.2474,0.9441391881850194,0.0,0.0,20.84686,1.8627629785885262,0.65176,0.09869114651274448
2280000.0,20.304675,2.1405652937658792,1.0,0.0,26.7422,2.1240175579939704,0.0,0.0,20.521859999999997,1.134369986556414,0.6300399999999999,0.113639334739341
2320000.0,20.41015,1.3111909862792674,1.0,0.0,25.059933333333333,1.068735755720541,0.0,0.0,20.16718,0.41053800506165067,0.6548,0.10213569405452726
2360000.0,20.4707,1.3268707981563241,1.0,0.0,26.309866666666665,1.9325469452397666,0.0,0.0,20.0172,0.4727464648202033,0.63714,0.09751571360555178
2400000.0,20.351575,1.7652797092458177,1.0,0.0,25.2396,1.8386919970457267,0.0,0.0,20.071859999999997,0.7283598660003168,0.62586,0.08554085807378838
2440000.0,20.011725,1.612465229663884,1.0,0.0,25.6354,1.504782624390203,0.0,0.0,20.15626,0.5493564074442022,0.63462,0.08798105250563894
2480000.0,20.00195,1.4107071143579033,1.0,0.0,25.773433333333333,2.75068768654111,0.0,0.0,19.851580000000002,0.4104066685618059,0.63706,0.0889039616665084
2520000.0,20.527325,1.04634481738813,1.0,0.0,24.927066666666665,1.6208466354209938,0.0,0.0,19.75624,0.7341053292273529,0.6364,0.10936260786941757
2560000.0,19.52735,0.7955166890141272,1.0,0.0,25.4245,0.917242054567204,0.0,0.0,20.06406,1.1362443656185934,0.6346,0.059411413044969705
2600000.0,19.64065,1.01673184886675,1.0,0.0,23.916666666666668,1.279853745116561,0.0,0.0,20.0344,0.803107830867064,0.61816,0.06323421858456069
2640000.0,19.556625,0.9802247736488812,1.0,0.0,23.60676666666667,1.0590335919548952,0.0,0.0,19.665640000000003,0.6832510127325097,0.6279399999999999,0.07150277197423888
2680000.0,19.3125,1.0727959102271045,1.0,0.0,24.552066666666665,1.985416114795306,0.0,0.0,19.80312,0.7093772745161775,0.62428,0.055857439969980716
2720000.0,19.882825,1.1632867173981658,1.0,0.0,24.632800000000003,0.7631479061536279,0.0,0.0,20.13124,0.6540832748205692,0.6615400000000001,0.08586208942251519
2760000.0,20.015625,0.9905697688073263,1.0,0.0,23.34113333333333,0.47324032184739045,0.0,0.0,19.784380000000002,0.6505676671953498,0.6213200000000001,0.08871395380660246
2800000.0,19.75,1.536078754491448,1.0,0.0,24.450533333333336,1.8549857741293387,0.0,0.0,20.1578,0.6119636688562491,0.6618999999999999,0.06948677572027646
2840000.0,20.082025,1.3594888513242769,1.0,0.0,23.846333333333334,1.0918472736096791,0.0,0.0,19.81094,0.8068658069344612,0.62866,0.09257398338626245
2880000.0,19.009775,0.845909783531909,1.0,0.0,24.286466666666666,1.707356116599255,0.0,0.0,19.39844,0.9357448575332922,0.6514800000000001,0.06450427582726588
2920000.0,19.730475000000002,0.8169975133836086,1.0,0.0,23.726566666666667,1.5916394322277343,0.0,0.0,19.6172,0.8196571551569594,0.64144,0.058428369821517355
2960000.0,19.40625,0.7616349798295765,1.0,0.0,23.1016,1.9182511566528508,0.0,0.0,19.86564,0.9729456564474706,0.6301599999999999,0.07711216765206387
3000000.0,20.228525,0.9934323061361542,1.0,0.0,23.82813333333333,1.5186423790858592,0.0,0.0,19.3953,1.0685760094630614,0.63498,0.06691610867347263
3040000.0,19.1836,0.9276884363836815,1.0,0.0,23.763033333333336,1.7321259506424156,0.0,0.0,19.56252,0.39518997659353655,0.6444799999999999,0.06875089526689818
3080000.0,18.99805,0.8589027578835692,1.0,0.0,22.895833333333332,1.4761081720373879,0.0,0.0,19.140620000000002,0.404559174410864,0.6386000000000001,0.06718258702967608
3120000.0,19.36135,0.7498258147730039,1.0,0.0,23.458333333333332,1.008557310661566,0.0,0.0,19.16252,0.4961859062891649,0.63464,0.0670067638376903
3160000.0,19.4375,0.7982543830133337,1.0,0.0,22.822933333333335,1.2186390177388695,0.0,0.0,19.19842,0.5446334708774327,0.6365999999999999,0.035215223980545676
3200000.0,18.814475,1.303094548325255,1.0,0.0,23.843766666666667,0.891066740236418,0.0,0.0,19.7531,0.6695935393953559,0.64258,0.08982479390457848
3240000.0,19.148425,0.7139194680599488,1.0,0.0,23.078100000000003,1.60111185742908,0.0,0.0,19.53594,0.5598677793908132,0.6147599999999999,0.05875173529352133
3280000.0,18.6973,0.6307464585711124,1.0,0.0,22.533866666666665,1.3705162295848805,0.0,0.0,19.62186,0.874277572856585,0.62958,0.06617060979014776
3320000.0,19.1289,0.5416591225115658,1.0,0.0,22.166666666666668,1.0073758693865074,0.0,0.0,19.6125,1.0664698082927624,0.64206,0.08439781039813771
3360000.0,19.066375,0.4624179352869007,1.0,0.0,22.244833333333332,0.10980243875049184,0.0,0.0,19.445320000000002,0.44431236489658854,0.62032,0.06623749391394576
3400000.0,18.912125,1.1607237276264328,1.0,0.0,22.51823333333333,1.0998921290542798,0.0,0.0,18.996879999999997,0.8250147208383611,0.64578,0.062497276740670855
3440000.0,18.847649999999998,0.6908853975153914,1.0,0.0,22.343733333333333,1.246824300194519,0.0,0.0,19.06408,0.6686016374493852,0.67256,0.08635229238416316
3480000.0,18.269550000000002,0.6090331374399918,1.0,0.0,22.3828,1.1588802296469938,0.0,0.0,19.92344,0.4245860035375642,0.65838,0.06769831312521753
3520000.0,19.001949999999997,1.0187827503938212,1.0,0.0,22.208333333333332,0.5049103506784369,0.0,0.0,19.559379999999997,0.6058005361503072,0.62936,0.08219015999497747
3560000.0,18.3965,0.9119337832320943,1.0,0.0,22.541666666666668,1.0017346066809423,0.0,0.0,19.198439999999998,0.6943498991142711,0.64086,0.06941321488016526
3600000.0,19.427725,0.9582413562746074,1.0,0.0,22.8047,0.9407265312866795,0.0,0.0,19.08906,0.9294583317179961,0.6373599999999999,0.051682863697748
3640000.0,19.0332,0.80317399111276,1.0,0.0,22.26823333333333,1.1133143381613104,0.0,0.0,19.14686,0.756566638439735,0.63184,0.04038804773692335
3680000.0,19.06445,0.662677329097654,1.0,0.0,22.539033333333332,0.5889443682461777,0.0,0.0,19.604680000000002,0.7969448799007365,0.64028,0.05881728997497248
3720000.0,18.652324999999998,0.6000383086728709,1.0,0.0,22.611966666666664,1.502924283595891,0.0,0.0,18.93438,0.48079827953103244,0.65746,0.06915873336029225
3760000.0,18.490225000000002,0.5177302018184761,1.0,0.0,22.580766666666666,1.0913167378090662,0.0,0.0,19.065620000000003,0.5375108944012201,0.6215200000000001,0.04939159442658234
3800000.0,19.083975,1.238190555558795,1.0,0.0,22.742199999999997,1.066993836283353,0.0,0.0,19.4875,1.0339974584108036,0.63748,0.038263998745557165
3840000.0,18.554675,0.8856092320402944,1.0,0.0,21.76046666666667,0.5621140651346683,0.0,0.0,18.88438,0.5038152335926332,0.64302,0.05724269036304987
3880000.0,19.290975,0.8745753608895006,1.0,0.0,21.60936666666667,0.6590755158216363,0.0,0.0,19.004700000000003,0.9543455307172563,0.65826,0.05182407162699587
3920000.0,18.615225,0.5775553626060449,1.0,0.0,22.16926666666667,0.33356648845803205,0.0,0.0,19.76248,0.11998450566635702,0.6498999999999999,0.08104435822436994
3960000.0,18.195325,0.5059765378700875,1.0,0.0,22.341166666666666,0.683229396973585,0.0,0.0,18.97344,0.5880135529050335,0.6619999999999999,0.08458186566871177
4000000.0,18.949225,0.6228959318176677,1.0,0.0,21.385433333333335,0.8007652187473904,0.0,0.0,19.1922,0.21409000910831832,0.64378,0.08471395162545543
}\dataAblationWhen

%% file: data/sota_steps_a.tex
\pgfplotstableread[col sep=comma]{
8000.0,40.73623333333334,0.06309550080807992,40.784533333333336,0.10410998457827715,40.7404,0.06744988262900739
1608000.0,39.57916666666667,0.2133502962006127,39.42166666666667,0.060770515511681854,39.85536666666667,0.7288242052938565
3208000.0,39.818333333333335,0.03723370635444326,39.7029,0.22661160605758848,40.5054,0.489363348307437
4808000.0,39.5833,0.0651753020706462,39.73333333333334,0.7346692377450489,40.861666666666665,0.04237155754617485
6408000.0,39.577933333333334,0.20181520149769747,40.085433333333334,0.5464778149405738,40.43833333333333,0.33277453161095455
8008000.0,40.3029,0.7112323717792014,40.0,0.5572684511675368,39.12253333333334,0.9370822850149739
9608000.0,40.3054,0.6567438516397893,39.82083333333333,0.5203415694415452,38.050000000000004,1.5384461858208336
11208000.0,40.086666666666666,0.3957710477311629,38.45706666666666,1.2782436526560785,37.28496666666666,0.8248207131788658
12808000.0,39.4904,0.20730127833662873,39.6592,0.8164182873992969,38.52833333333333,1.7745097131946688
14408000.0,39.861266666666666,0.6753725210742757,39.43873333333333,0.5364187191198899,37.74626666666666,2.1252765409601557
16008000.0,38.9546,1.3237679126896336,39.785000000000004,0.37437564913688737,36.581266666666664,2.7941643358176993
17608000.0,39.50083333333333,0.4782990858819965,39.1096,1.5260850194752165,35.46586666666667,0.871429309939838
19208000.0,38.369566666666664,1.6004180710746245,38.9754,0.5015291483719247,32.99543333333333,1.456289134142743
20808000.0,37.961666666666666,2.760204048415423,39.77956666666666,0.9525632309137744,31.260433333333335,1.5948976066053764
22408000.0,37.46746666666666,3.153136434656064,39.29333333333333,1.2904930151776184,30.41333333333333,0.9145472553248536
24008000.0,35.52583333333333,3.4336917288668882,38.35833333333333,1.3164092634469293,29.65206666666667,1.4895072033692516
25608000.0,36.4996,3.9112351561452634,39.319966666666666,1.319493605727424,28.523733333333336,0.9649977766238051
27208000.0,34.83083333333334,3.71199987278496,39.15876666666667,1.4309084674507366,29.870833333333337,1.8462281410727355
28808000.0,35.793366666666664,3.317084042080065,38.21876666666667,1.3654830801669515,27.337100000000003,1.1586679794775836
30408000.0,33.861266666666666,4.297263048603018,37.9879,2.507275510722081,26.329166666666666,1.183554556785994
32008000.0,33.067499999999995,4.928155042880313,36.0425,2.4955234921755394,26.638766666666665,1.2511511534938091
33608000.0,34.56126666666666,5.656723083395741,37.329166666666666,2.821693738795115,27.734166666666667,1.5646809223892544
35208000.0,35.24373333333333,6.370995045952833,36.8183,2.7210259682700566,26.0625,1.3084775351529727
36808000.0,32.95246666666667,4.865532503459638,36.0713,3.750000555555515,25.887533333333334,0.9258310008971512
38408000.0,31.830433333333335,5.744929217047752,35.21876666666667,3.8910141405836876,25.436233333333334,0.6031332430639921
40008000.0,30.941266666666667,6.330759094207337,37.54913333333334,4.331396430970297,25.455833333333334,0.5354543387525115
41608000.0,31.977500000000003,5.618643471515165,35.98413333333333,3.79267061258364,25.03,0.4682057952083328
43208000.0,30.4104,6.323608301173206,35.8987,3.6541010066316817,24.7546,0.6548243937626835
44808000.0,29.932100000000002,6.587666012076407,36.21413333333333,4.4273740571835845,24.475866666666665,0.2032700393291871
46408000.0,30.459999999999997,6.71683429600582,36.444199999999995,4.954873542281376,28.932533333333335,6.845141576979172
48008000.0,31.833766666666666,6.159676455969277,33.655433333333335,4.1871013890545115,24.138333333333332,0.774742788858914
49608000.0,31.819166666666664,5.444908923225643,34.0829,4.134490112053318,23.642033333333334,0.6644368960923899
51208000.0,32.92293333333333,6.308032488467031,31.8429,4.703950231454411,23.657933333333336,0.8108883249594584
52808000.0,31.148333333333337,6.128822050135102,30.6046,4.7014157782523345,23.9808,0.4625024180116973
54408000.0,29.89623333333334,6.735036624663267,30.94083333333333,3.9308657514033145,23.963333333333335,0.40417268036763226
56008000.0,29.560433333333332,7.293423021843417,30.63416666666667,3.916663297870892,23.848366666666667,0.8447855993616888
57608000.0,29.854133333333333,7.072687021838935,30.37086666666667,3.374693424429676,23.5904,0.7546287078203863
59208000.0,29.586233333333336,6.970988595760447,30.205,4.143781988312932,23.178766666666665,0.9479928140843441
60808000.0,29.1825,7.216630524466849,29.733733333333333,4.013079184643909,23.444999999999997,0.6130175038283981
62408000.0,30.065033333333332,6.430810773836288,29.222466666666666,3.473047539105806,24.013333333333332,0.7812850625020872
64008000.0,29.02376666666667,7.325026214439251,28.491233333333337,2.779452171841703,23.450833333333335,0.6960373952278388
65608000.0,28.746266666666667,7.496198483824242,28.95873333333334,3.1514454422622578,23.605433333333334,1.0009867009883575
67208000.0,28.788299999999996,7.462417100823743,28.3346,3.8344928591231824,24.16586666666667,1.6372764098411194
68808000.0,28.87,7.444408416164892,28.4525,3.237206376903806,24.067966666666667,2.2031202620122414
70408000.0,29.286666666666665,7.518107630839618,27.734166666666667,3.041409516801197,23.446233333333335,0.9947811328238106
72008000.0,29.819600000000005,6.9916070270003035,27.574166666666667,2.8512601335004297,23.6867,0.7584401668336567
73608000.0,29.240400000000005,7.6401391204610904,27.046266666666668,2.7697088671715826,23.5017,0.581125620383981
75208000.0,28.60043333333333,7.659145734059089,28.0204,3.9413552855162215,23.162900000000004,0.7908827262411708
76808000.0,28.19496666666667,7.874513426802249,27.9983,3.1758549221692522,22.90706666666667,0.28074016852282135
78408000.0,27.857066666666665,8.220905944129406,27.416666666666668,3.1385309284582323,23.485833333333332,0.6287710358751875
80008000.0,27.816666666666666,8.278321169711218,27.75916666666667,2.962698491053196,23.218766666666667,0.7689514780241706
81608000.0,28.106666666666666,8.147836431566082,27.179199999999998,2.812824467802189,22.949166666666667,0.7371520753699481
83208000.0,28.065866666666665,8.316949042500836,27.41333333333333,2.977776248963123,23.11293333333333,0.8253026973312423
84808000.0,27.858766666666668,8.143932365605425,27.0296,2.255167285738835,23.0712,0.8230431337420894
86408000.0,28.0275,8.358004396983766,26.7971,2.378481612289655,22.98913333333333,0.46859226294177075
88008000.0,28.015866666666668,8.208115316089453,26.632900000000003,2.4804412550995845,23.101266666666664,1.005396854757144
89608000.0,27.673333333333336,8.436045796593463,26.3446,2.262355250618257,22.605866666666667,0.7407955468428667
91208000.0,28.160433333333334,8.281622166513568,26.272066666666664,2.371732138238962,22.7142,0.6162122091184713
92808000.0,27.37793333333333,8.563654284760032,26.1771,2.419619870971472,22.711233333333336,0.6465243296960209
94408000.0,27.692899999999998,8.263784412725201,26.4646,2.6281758857935418,22.391266666666667,0.6382079093489488
96008000.0,27.557466666666667,8.48117354549999,25.9954,2.136685865228359,22.830000000000002,0.8690604850450094
97608000.0,27.796666666666667,8.47796188256483,25.749133333333333,2.1641178053783383,22.53916666666667,0.41049443628656235
99208000.0,27.642933333333332,8.283942036789542,26.156666666666666,2.460660762658861,22.541633333333333,0.32935397708578223
100808000.0,27.481266666666667,8.587461256713508,26.2821,2.175982188958969,22.5925,1.050272405934131
102408000.0,27.675833333333333,8.76495325461326,25.492499999999996,2.297921697244418,23.167066666666667,0.310363296083083
104008000.0,27.305866666666663,8.629411164281272,25.084999999999997,2.274833327227881,22.649600000000003,0.21065252589671676
105608000.0,27.481233333333336,8.434134842544447,25.1829,2.256001068262158,22.789599999999997,0.6985082819838292
107208000.0,27.445800000000002,8.421284135253163,25.3175,2.281077961549467,22.168366666666667,0.21162004210901686
108808000.0,27.547066666666666,8.320686710575968,25.549566666666667,2.4926543834947426,22.442133333333334,0.654294573482685
110408000.0,27.6425,8.439796391303918,25.11876666666667,2.2566259198684704,22.406666666666666,0.7218538741016463
112008000.0,27.8275,8.478234879187216,24.8733,2.8716102602314724,22.355433333333334,0.8610369730092252
113608000.0,27.532500000000002,8.721581708612263,24.71666666666667,2.4389225795192617,22.617500000000003,0.7745934159286401
115208000.0,27.474966666666663,8.15275182840466,25.4029,2.7750889751501653,22.3471,0.4979984805867053
116808000.0,27.746700000000004,8.271265611742862,24.753766666666667,2.9491335239731375,22.344166666666666,0.3737227344198147
118408000.0,27.365800000000004,8.551387011473636,25.059166666666666,2.947305502017431,22.159599999999998,0.5501970737835661
120008000.0,27.497933333333332,8.351500745907222,25.23,2.9732345394648347,22.445466666666665,0.22684735445277324
121608000.0,27.4592,8.679577235480233,25.09293333333333,2.9634189998865983,22.388733333333334,0.4767529082822219
123208000.0,27.374166666666667,8.349522524605158,25.101233333333337,2.2649618191533003,22.4729,0.585743533183821
124808000.0,27.462933333333336,8.65315043065562,24.550833333333333,2.74817194068267,22.665833333333335,0.2735288934565329
126408000.0,27.3079,8.600388275343542,24.85996666666667,3.1153917424441007,22.3404,0.3483895520821473
128008000.0,27.514166666666668,8.115568804190842,24.5596,2.746230110533347,22.38123333333333,0.46695186285335893
129608000.0,27.648333333333337,8.837172767098966,24.499166666666667,2.6830380272287524,22.172933333333333,0.2545937984755768
131208000.0,27.659566666666663,8.584635463560593,24.51546666666667,2.7806326718132968,22.209166666666665,0.5892096872553568
132808000.0,27.585800000000003,8.85885051723228,24.606233333333336,2.6587315275438312,22.662033333333337,0.5696502338179886
134408000.0,27.757499999999997,8.389270677478466,24.61836666666667,2.696202831555684,22.592933333333335,0.39679003963753423
136008000.0,27.87166666666667,9.140773201552602,24.47666666666667,2.7869117344393155,22.46623333333333,0.04343626237245712
137608000.0,28.093766666666667,8.960237462007107,24.58543333333333,3.0417371750731887,22.638733333333334,0.5685294265813231
139208000.0,27.6217,9.387611614249922,24.857933333333335,2.7483739511856005,22.3467,0.08664767740684151
140808000.0,27.784966666666666,8.864529647734026,24.777500000000003,3.0756788074613155,22.100833333333338,0.21006282129136689
142408000.0,27.708366666666667,8.91155888208617,24.66163333333333,2.77883261620575,22.465033333333334,0.19454744636948756
144008000.0,27.786699999999996,8.919253546121446,24.58083333333333,3.003640571410339,22.451266666666665,0.24899339125544956
145608000.0,27.74,9.007081628363318,24.296633333333332,2.350673883993458,22.47583333333333,0.4817826434768656
147208000.0,27.701233333333334,9.244703711254836,24.775833333333335,2.815371026269106,22.1775,0.22151552240569217
148808000.0,27.438366666666667,8.626047165932311,25.39,2.932014687321102,22.390833333333333,0.26148082572575476
150408000.0,27.739166666666666,9.049396005013568,24.941666666666663,2.9196053983753063,22.3625,0.18981164347847537
152008000.0,27.674599999999998,9.26952924838509,24.87,2.889120396706698,22.432500000000005,0.035484174876509435
153608000.0,27.7825,9.0919568392435,24.50333333333333,2.455707920109574,22.270033333333334,0.2301116588866269
155208000.0,27.762533333333334,9.13807553822041,24.6837,3.2894648754268023,22.233766666666668,0.38068321330062527
156808000.0,27.242900000000002,8.504866223913618,24.230433333333334,2.875665031876202,22.2871,0.22856469543654342
158408000.0,27.690033333333332,8.6488604476094,24.284166666666668,2.7279642621477938,22.2129,0.25371996373955213
}\dataSotaStepsA

%% file: data/sota_steps_b.tex
\pgfplotstableread[col sep=comma]{
0.0,45.0,0.0,45.0,0.0
40000.0,45.0,0.0,45.0,0.0
80000.0,45.0,0.0,45.0,0.0
120000.0,44.7969,0.14488411461118336,45.0,0.0
160000.0,44.78123333333334,0.15794311915651302,44.86976666666666,0.09405829161866774
200000.0,44.72393333333333,0.02946278254943948,44.091133333333325,0.42121988979101443
240000.0,43.770833333333336,0.2115958621733612,44.28906666666666,0.5390464069900546
280000.0,43.32293333333333,0.6185165281183328,43.359399999999994,1.0006783832314303
320000.0,43.1875,0.18501812523822084,43.33596666666667,0.6222617098581237
360000.0,43.9974,0.10170732520325194,42.872366666666665,1.1131963239049782
400000.0,43.96876666666666,0.3382510934531062,42.625,1.8416830400478816
440000.0,43.34116666666666,0.6324431797051435,41.934866666666665,0.6417085856437412
480000.0,43.22656666666666,1.0717516638143791,40.47656666666666,1.0164623794754486
520000.0,43.46873333333334,0.23237465916537106,40.427099999999996,0.7080753820509983
560000.0,42.52603333333334,0.2475887629832082,39.638,0.8169291809371632
600000.0,43.7578,0.21312958499466736,41.38803333333333,1.2770944896739458
640000.0,43.020833333333336,0.3305202599270148,40.7526,1.0559491275624957
680000.0,42.97656666666666,1.1153531707739743,39.22656666666666,0.6041848962767016
720000.0,43.562466666666666,0.733664638797743,39.31513333333333,0.8140088956652867
760000.0,43.0703,0.7213101598249311,37.59113333333334,0.4280378825394904
800000.0,43.15626666666666,0.6225386431557671,38.21353333333334,1.0343103896907473
840000.0,43.4297,0.6473916022521965,35.674499999999995,0.8398964340917293
880000.0,43.1927,1.036343157453168,37.177099999999996,0.9923480572191723
920000.0,42.796899999999994,0.6608019219100394,34.205733333333335,0.6118570711821135
960000.0,42.333333333333336,0.38386783605240443,34.49216666666667,1.1344456188915453
1000000.0,42.51303333333333,1.2509467596806623,33.8099,0.8193187698732803
1040000.0,41.705733333333335,0.8089172612545467,32.61976666666667,1.5082406093046163
1080000.0,41.356766666666665,1.2963538362996772,31.895866666666667,1.2874883697425088
1120000.0,41.4323,0.8625282178958958,30.885433333333335,0.7174406100081652
1160000.0,40.458333333333336,2.027559360961405,30.4896,1.0915951294626898
1200000.0,39.018233333333335,1.9166778138806264,29.6771,1.3748071282910928
1240000.0,40.0104,0.5456940046094221,29.265633333333337,0.16278120967182455
1280000.0,40.63543333333333,0.6827832517636095,27.567700000000002,1.3082649604214982
1320000.0,40.34116666666666,1.8929067670883544,26.666666666666668,1.1214133027370223
1360000.0,40.0703,1.903442724819074,25.968766666666667,1.275437144232866
1400000.0,38.919266666666665,1.5090106965234615,24.51823333333333,1.4627783526183618
1440000.0,38.640633333333334,2.653861168603629,25.09113333333333,0.6346169150667901
1480000.0,36.7474,2.062273940742758,24.60936666666667,0.9221963071325369
1520000.0,37.390633333333334,2.9199650732309976,24.15363333333333,1.1491664873096308
1560000.0,38.08593333333334,3.2673102213016487,23.067733333333337,1.3864725537212124
1600000.0,38.86456666666667,1.782004317864827,23.0625,0.8683283979386303
1640000.0,38.161500000000004,3.0338808886748776,23.317700000000002,0.7718692505858745
1680000.0,36.88803333333333,2.8811141321062275,22.7526,0.8307970911520259
1720000.0,36.34376666666667,1.5117059730281173,23.374966666666666,0.7188686497236861
1760000.0,36.770833333333336,1.241772830360778,22.119799999999998,1.4883521648678
1800000.0,35.94266666666667,2.547338953147425,22.447933333333335,0.8035054628867744
1840000.0,35.59636666666666,1.1332723160045093,21.739566666666665,1.1159860821513667
1880000.0,35.01303333333333,1.911977092494107,23.1901,1.9312508709814655
1920000.0,33.50000000000001,1.7791139423881772,21.48696666666667,1.1247164096883353
1960000.0,34.138,0.9010820199441695,21.6901,0.7055320687254417
2000000.0,34.828133333333334,1.4261743379482827,21.52603333333333,0.5811357978610119
2040000.0,33.421866666666666,1.1060086839120602,22.02606666666667,0.3067405019809995
2080000.0,30.625,2.847555340989882,21.351566666666667,0.36483167199256294
2120000.0,31.526033333333334,2.027837775124583,21.46356666666667,0.6496809285248336
2160000.0,30.127599999999997,2.1855337395397636,21.062466666666666,0.574345925572927
2200000.0,30.088533333333334,2.3924002805179216,20.90103333333333,0.3352675979307013
2240000.0,29.651033333333334,1.3595515690435909,20.838533333333334,0.3147017247419452
2280000.0,28.609366666666663,2.779131344303267,20.648433333333333,0.4783500972672172
2320000.0,28.1198,1.5200816710514824,20.367166666666666,0.5513487845476962
2360000.0,28.528666666666666,0.4762307937217921,20.5599,0.3904462660426745
2400000.0,28.630200000000002,1.4336602177643063,20.4844,0.9863108941910765
2440000.0,27.109366666666663,1.2763146564316432,20.929666666666666,0.634727785922613
2480000.0,26.609333333333336,2.048058341182909,21.3698,0.6611437413049198
2520000.0,26.320333333333334,2.2926136065392457,20.606766666666665,0.6166422968157659
2560000.0,26.835966666666668,3.5961376580375157,20.40886666666667,0.4807769152343125
2600000.0,26.885433333333335,2.2962760132198587,20.309866666666668,0.5584523932758779
2640000.0,25.513033333333336,2.034969772639278,20.4974,0.6655414988313406
2680000.0,25.473966666666666,0.925830813677939,20.51823333333333,1.0093922538945015
2720000.0,25.947933333333335,2.4634201432606297,20.3021,0.5981837566054984
2760000.0,25.255200000000002,0.49388643094001594,19.854166666666668,0.5364448423546347
2800000.0,25.994766666666663,0.8510661718626161,21.3073,0.5007077457626022
2840000.0,25.700499999999995,1.6307433294870988,20.523433333333333,0.22378855099301947
2880000.0,26.419300000000003,2.2431369240418646,20.588533333333334,0.7849951903603533
2920000.0,24.763,0.7963562310088791,19.908833333333334,0.9645571568802374
2960000.0,25.023433333333333,1.4198810427481425,19.778666666666666,0.5195587957317457
3000000.0,25.583333333333332,1.4546126868994678,20.0625,0.6622172654549767
3040000.0,25.317700000000002,0.8346437483541507,20.182299999999998,0.8245646608968873
3080000.0,25.273433333333333,1.3421907622324867,19.820333333333334,0.2180237958470493
3120000.0,24.656266666666667,1.2504554379194095,20.085933333333333,0.33973143641542736
3160000.0,24.8125,1.003384994240331,19.888,1.1048957597891305
3200000.0,24.479166666666668,1.590815401679975,20.218766666666667,0.28406656888052795
3240000.0,24.536466666666666,1.200620174002679,20.3177,0.5688410908739511
3280000.0,24.60676666666667,1.714488410252133,19.58073333333333,0.649260647882565
3320000.0,25.1224,0.9325663122087715,20.3073,0.24358189587898307
3360000.0,24.3646,1.2212802708633261,20.364566666666665,0.45668123334431837
3400000.0,23.8802,1.3279066407947016,19.721366666666665,0.5943654898760148
3440000.0,24.51823333333333,1.340435228656059,19.6198,1.0585406211698571
3480000.0,24.51303333333333,0.7700256330509753,19.76563333333333,0.9549889225651901
3520000.0,24.65363333333333,1.4714325091178628,19.95573333333333,0.45953894527256656
3560000.0,24.854200000000002,1.4380309407890592,19.9349,0.5888061537269007
3600000.0,25.40366666666667,1.0238632862296069,19.08073333333333,0.8733724075228286
3640000.0,24.97133333333333,1.2095523395960273,19.984333333333332,0.871150049583244
3680000.0,23.5625,0.9030402575005542,20.0,0.4783136209643203
3720000.0,24.04946666666667,1.1059412923940495,19.997400000000003,0.9443189397655851
3760000.0,24.403666666666666,1.678563277197365,19.729166666666668,0.15286251629777914
3800000.0,23.807299999999998,0.8830197091043137,19.411466666666666,0.4411789533612043
3840000.0,24.52863333333333,1.1741778580020237,19.28643333333333,0.5478501031811124
3880000.0,23.66926666666667,0.7773092277565958,19.453133333333334,0.8138757986053864
3920000.0,23.585933333333333,1.2445885995871175,19.5677,0.4663733268530696
3960000.0,24.10676666666667,1.7305350701123894,19.627566666666663,0.19993819600622176
4000000.0,25.0651,1.9826009734689425,19.2005,0.29650741418498644
}\dataSotaStepsB

%% file: data/sota_pp_tarmac.tex
\pgfplotstableread[col sep=comma]{
0.0,45.0,0.0,44.893233333333335,0.15099086800936892,45.0,0.0
40000.0,45.0,0.0,44.90103333333334,0.1399600022228577,44.83593333333334,0.23202530513334432
80000.0,45.0,0.0,44.78906666666666,0.22068120798009871,44.643233333333335,0.5045442586026427
120000.0,44.7969,0.14488411461118336,44.7682,0.19142211645122653,44.05466666666666,1.0186511648035146
160000.0,44.78123333333334,0.15794311915651302,44.625,0.3724341910548332,43.770833333333336,0.9325861616434648
200000.0,44.72393333333333,0.02946278254943948,44.604166666666664,0.2157789815734819,41.89066666666667,1.328697009186905
240000.0,43.770833333333336,0.2115958621733612,44.44793333333333,0.7642535763010143,42.5651,1.2031834855914534
280000.0,43.32293333333333,0.6185165281183328,43.890633333333334,0.6086405306546352,42.8099,0.3161503123515764
320000.0,43.1875,0.18501812523822084,43.34633333333334,0.9331523145886846,41.911433333333335,0.46020137862558486
360000.0,43.9974,0.10170732520325194,44.0052,0.39226328743162664,40.8802,0.7064869472726748
400000.0,43.96876666666666,0.3382510934531062,42.395833333333336,1.8835453420492843,41.4141,1.7860862707794007
440000.0,43.34116666666666,0.6324431797051435,41.96616666666667,1.199372964881603,40.312533333333334,0.4106359404089657
480000.0,43.22656666666666,1.0717516638143791,42.57816666666667,0.27739546179096597,39.46093333333334,1.6211574966328492
520000.0,43.46873333333334,0.23237465916537106,41.5,0.816496580927726,38.4427,1.3196324968212414
560000.0,42.52603333333334,0.2475887629832082,41.958333333333336,0.575734605139864,38.18226666666667,0.9881516291653917
600000.0,43.7578,0.21312958499466736,41.7396,1.3129920131770296,36.4271,1.347765093280972
640000.0,43.020833333333336,0.3305202599270148,41.182300000000005,0.9100506396166455,35.54683333333333,0.988267798839069
680000.0,42.97656666666666,1.1153531707739743,41.0469,1.9284686688320007,35.90106666666667,1.101202507362848
720000.0,43.562466666666666,0.733664638797743,39.79946666666667,0.7538166060492145,35.2448,2.5187605933606867
760000.0,43.0703,0.7213101598249311,39.96353333333334,0.89654672431998,33.958333333333336,0.9761323999449161
800000.0,43.15626666666666,0.6225386431557671,39.9948,1.2940906021862093,34.14843333333334,2.956610894626182
840000.0,43.4297,0.6473916022521965,40.169266666666665,1.879405917363842,32.38023333333333,0.8838965261210648
880000.0,43.1927,1.036343157453168,39.624966666666666,0.5113852841927395,32.52343333333334,0.7655422580686774
920000.0,42.796899999999994,0.6608019219100394,38.7552,1.228035213935931,31.367166666666666,1.519023230749142
960000.0,42.333333333333336,0.38386783605240443,37.89843333333334,2.3122359226995464,32.28386666666666,2.3574296940146953
1000000.0,42.51303333333333,1.2509467596806623,38.90103333333334,1.7348823770567936,30.625033333333334,2.5494141867408575
1040000.0,41.705733333333335,0.8089172612545467,37.77863333333334,2.8153203492478243,30.526033333333334,3.393128827825762
1080000.0,41.356766666666665,1.2963538362996772,36.080733333333335,2.022071812660359,29.54686666666667,3.5099172092167024
1120000.0,41.4323,0.8625282178958958,35.627633333333335,0.9716085848849949,26.637999999999995,3.415630562575526
1160000.0,40.458333333333336,2.027559360961405,35.46353333333334,3.4946637800445925,27.091133333333335,2.4116716834778495
1200000.0,39.018233333333335,1.9166778138806264,34.69533333333334,3.1696556377976175,24.54946666666667,1.3348235572122955
1240000.0,40.0104,0.5456940046094221,34.916666666666664,3.7432859542860935,24.854166666666668,2.0452908682032374
1280000.0,40.63543333333333,0.6827832517636095,32.27603333333334,2.8954633496020783,25.148433333333333,2.1437469631983666
1320000.0,40.34116666666666,1.8929067670883544,30.924466666666664,2.7711831664872353,23.458333333333332,0.4235577043200706
1360000.0,40.0703,1.903442724819074,31.067733333333337,4.03364472103773,22.721333333333334,0.8463998516593029
1400000.0,38.919266666666665,1.5090106965234615,29.8229,4.1833253885714745,22.096333333333334,1.1397167903572463
1440000.0,38.640633333333334,2.653861168603629,30.01563333333333,4.673076064958593,23.28643333333333,2.1024383373174635
1480000.0,36.7474,2.062273940742758,28.1771,2.26336694476761,22.960966666666668,0.7957944472175097
1520000.0,37.390633333333334,2.9199650732309976,27.645833333333332,4.100070902096967,22.20053333333333,1.0042180683939572
1560000.0,38.08593333333334,3.2673102213016487,28.718733333333333,3.992514350075095,21.76563333333333,0.5399124640984767
1600000.0,38.86456666666667,1.782004317864827,27.710966666666668,3.3011743590560148,21.804666666666666,0.8573171577012138
1640000.0,38.161500000000004,3.0338808886748776,27.625,3.9923810497830328,20.8854,0.013257450735340535
1680000.0,36.88803333333333,2.8811141321062275,26.718766666666667,3.4348841846883604,20.984366666666663,0.40339969701966305
1720000.0,36.34376666666667,1.5117059730281173,25.440133333333335,3.3718290756732543,21.039033333333332,0.8456716160675037
1760000.0,36.770833333333336,1.241772830360778,24.960966666666668,1.7905183762127546,21.8021,0.13279023558480005
1800000.0,35.94266666666667,2.547338953147425,25.6797,2.598117056639289,21.786466666666666,0.5436818084954554
1840000.0,35.59636666666666,1.1332723160045093,23.9974,1.29690312925317,21.088533333333334,0.9130904202517713
1880000.0,35.01303333333333,1.911977092494107,25.645833333333332,3.498737433544976,20.984400000000004,0.5516797863495342
1920000.0,33.50000000000001,1.7791139423881772,23.76823333333333,2.3119654788849147,20.9297,0.23385358667337133
1960000.0,34.138,0.9010820199441695,22.84116666666667,1.274396788898793,20.6224,0.6899033120662633
2000000.0,34.828133333333334,1.4261743379482827,22.960933333333333,1.6620325354483558,21.04426666666667,0.2606102624396987
2040000.0,33.421866666666666,1.1060086839120602,22.098966666666666,1.087643818331882,20.361966666666664,0.2862841517715492
2080000.0,30.625,2.847555340989882,22.893199999999997,1.751001047020436,20.1328,0.4651681918044996
2120000.0,31.526033333333334,2.027837775124583,22.445300000000003,1.5587558265060844,20.549466666666667,0.803428944517739
2160000.0,30.127599999999997,2.1855337395397636,20.9375,1.185137370378078,20.156233333333333,0.7707841908659574
2200000.0,30.088533333333334,2.3924002805179216,21.73696666666667,0.6118467146452795,19.7604,0.43195222729680055
2240000.0,29.651033333333334,1.3595515690435909,21.619799999999998,0.6646790854740848,20.0495,0.8473271269114429
2280000.0,28.609366666666663,2.779131344303267,20.78386666666667,0.7168974465632371,20.487000000000002,0.11455691453014487
2320000.0,28.1198,1.5200816710514824,21.351566666666667,0.45506402724110073,20.554666666666666,1.419842297189688
2360000.0,28.528666666666666,0.4762307937217921,21.023433333333333,0.6324067379639653,19.6927,0.7934933017990758
2400000.0,28.630200000000002,1.4336602177643063,21.2604,0.4543801345422863,20.117166666666666,0.4425261147347372
2440000.0,27.109366666666663,1.2763146564316432,21.182266666666667,0.7259549358526934,19.557299999999998,0.6740630880464132
2480000.0,26.609333333333336,2.048058341182909,20.958333333333332,0.4671049156476755,19.2266,0.1823199385695374
2520000.0,26.320333333333334,2.2926136065392457,21.041666666666668,0.254840425015777,19.161466666666666,0.3274199783492489
2560000.0,26.835966666666668,3.5961376580375157,20.726566666666667,0.4530415163708021,19.450533333333336,0.20503323849778268
2600000.0,26.885433333333335,2.2962760132198587,20.765633333333334,0.8686537719687607,20.3021,1.3780672625093457
2640000.0,25.513033333333336,2.034969772639278,20.489566666666665,1.2954479851473082,19.90103333333333,0.36097118382989446
2680000.0,25.473966666666666,0.925830813677939,21.3203,0.2917248475304529,20.1198,0.3026116433098154
2720000.0,25.947933333333335,2.4634201432606297,20.5365,0.40188088616736495,19.73436666666667,0.7863798672114856
2760000.0,25.255200000000002,0.49388643094001594,21.145866666666667,0.9273811202640596,20.067733333333333,0.32022476828350105
2800000.0,25.994766666666663,0.8510661718626161,20.6354,1.0336152701400405,19.999966666666666,0.0710431949982235
2840000.0,25.700499999999995,1.6307433294870988,20.679733333333335,0.8579738198544029,18.981800000000003,0.6539150760356174
2880000.0,26.419300000000003,2.2431369240418646,20.541700000000002,0.7106033633469514,19.450533333333336,0.34271020151466514
2920000.0,24.763,0.7963562310088791,20.835933333333333,0.39516305776504856,19.51823333333333,0.5585250178421337
2960000.0,25.023433333333333,1.4198810427481425,20.7526,0.4259051850666619,20.112,0.4580889506053016
3000000.0,25.583333333333332,1.4546126868994678,20.184900000000003,0.7509153125796982,19.8203,1.2651459230723803
3040000.0,25.317700000000002,0.8346437483541507,20.651033333333334,0.3912642272883575,18.7474,0.4338764878011561
3080000.0,25.273433333333333,1.3421907622324867,20.320333333333334,0.7305457792332769,19.007833333333334,0.49460810300232094
3120000.0,24.656266666666667,1.2504554379194095,19.971366666666665,0.66673481151721,19.648433333333333,1.0164153656628554
3160000.0,24.8125,1.003384994240331,20.09633333333333,0.3431713889913057,19.3802,0.3218798222939734
3200000.0,24.479166666666668,1.590815401679975,20.700533333333333,0.681708864219585,18.875,0.36925680314202286
3240000.0,24.536466666666666,1.200620174002679,19.9375,0.8739659070390938,19.132833333333334,0.315326078140639
3280000.0,24.60676666666667,1.714488410252133,19.3151,0.5965171637653573,19.1432,0.3572361777125417
3320000.0,25.1224,0.9325663122087715,20.3854,0.7916013685351138,18.682266666666667,0.48586379664355434
3360000.0,24.3646,1.2212802708633261,20.612000000000002,0.14690861104782146,19.40363333333333,0.4953621054900706
3400000.0,23.8802,1.3279066407947016,20.28386666666667,0.31719169107794937,18.937466666666666,0.2906963402284623
3440000.0,24.51823333333333,1.340435228656059,20.10676666666667,0.3941456358025837,19.093733333333333,0.8037944278372556
3480000.0,24.51303333333333,0.7700256330509753,20.614566666666665,0.33810247296082035,19.174466666666664,0.4748681524615244
3520000.0,24.65363333333333,1.4714325091178628,19.822933333333335,0.5618268554951377,19.257833333333334,0.1053957726329139
3560000.0,24.854200000000002,1.4380309407890592,20.1875,0.6411368392680834,19.1172,0.29732006323152943
3600000.0,25.40366666666667,1.0238632862296069,19.2422,0.35223532853288136,18.890599999999996,0.2801691036975112
3640000.0,24.97133333333333,1.2095523395960273,20.04426666666667,0.7199779826880502,19.049500000000002,0.6174939729800343
3680000.0,23.5625,0.9030402575005542,19.4844,0.412352009816855,19.0755,0.18336653638727723
3720000.0,24.04946666666667,1.1059412923940495,19.734366666666666,0.7765600484758974,18.7474,0.27252659809029023
3760000.0,24.403666666666666,1.678563277197365,19.640600000000003,0.6804004164215856,18.518233333333335,0.6835657799769946
3800000.0,23.807299999999998,0.8830197091043137,20.135433333333335,0.9716085848849948,19.09636666666667,0.22612023841802079
3840000.0,24.52863333333333,1.1741778580020237,20.080699999999997,0.4668296548706669,19.015633333333334,0.2008265807993449
3880000.0,23.66926666666667,0.7773092277565958,20.23436666666667,0.3156671805698032,18.73176666666667,0.20829713285486073
3920000.0,23.585933333333333,1.2445885995871175,19.710933333333333,0.6744514083477198,18.75,0.3181400425389222
3960000.0,24.10676666666667,1.7305350701123894,19.364566666666665,0.4564786109142717,19.499966666666666,0.21545539883903742
4000000.0,25.0651,1.9826009734689425,19.747400000000003,0.405152177171327,18.791666666666668,0.4774405745453793
}\dataSotaPpTarmac

%% file: data/sota_lj.tex
\pgfplotstableread[col sep=comma]{
0.0,-19.50235,0.1910500000000006,-19.64395,0.14475000000000016,-19.6125,0.29235557973125814,0.0,-19.703425000000003,0.08598384659341494,-19.201833333333337,0.22234133418887397,-19.560533333333332,0.18738945778482033
40000.0,-19.366600000000002,0.2350000000000012,-18.825,0.16639999999999944,-19.443849999999998,0.352255720322609,40000.0,-19.18545,0.3354428006382015,-19.181766666666665,0.12165024546716834,-19.335800000000003,0.18530689859437618
80000.0,-19.32795,0.4482499999999998,-18.94455,0.10545000000000115,-18.4703,0.5554729471360421,80000.0,-19.4123,0.24275695664594205,-19.2159,0.13135496437769942,-19.279833333333332,0.23208396947848162
120000.0,-19.0258,0.4297000000000004,-19.05565,0.19784999999999897,-17.454875,0.5954310345245701,120000.0,-18.9924,0.4388073609227627,-19.191533333333332,0.07468718468087761,-18.738,0.10011566643970773
160000.0,-18.879900000000003,0.22049999999999947,-18.9348,0.29959999999999987,-17.824025,0.4072436210365975,160000.0,-19.067775,0.35846385713904216,-18.764466666666667,0.3422178286153746,-19.124066666666668,0.21558250289751266
200000.0,-18.65625,0.6925500000000007,-18.506999999999998,0.09999999999999964,-18.099825000000003,0.19546230295123404,200000.0,-18.896774999999998,0.2394627776398653,-18.792199999999998,0.05337995878604689,-19.0862,0.13290066465848283
240000.0,-18.95235,0.5394500000000004,-17.3822,0.6661999999999999,-17.45495,0.5032090296288417,240000.0,-18.809874999999998,0.38104490795049284,-18.916433333333334,0.09204438542838471,-18.0754,0.2311770028931662
280000.0,-19.10605,0.2021499999999996,-17.999450000000003,0.7224500000000003,-17.120524999999997,0.5575669124643254,280000.0,-18.541499999999996,0.6219555570939139,-18.819100000000002,0.2009815414409985,-18.2332,0.6434472990592679
320000.0,-19.068550000000002,0.07364999999999888,-18.3723,0.3995999999999995,-16.128325,0.38063817711180753,320000.0,-19.0122,0.3265325634603693,-19.07683333333333,0.0958845254575639,-17.843233333333334,0.07551235366187116
360000.0,-18.694100000000002,0.4961000000000002,-17.969549999999998,0.08904999999999852,-16.31445,0.14921113396794483,360000.0,-18.093149999999998,0.6081755441482337,-18.869666666666667,0.11745593026984782,-17.780333333333335,1.0070402386311206
400000.0,-16.74145,0.3562499999999993,-18.27445,0.6357499999999998,-16.66425,0.566164099444675,400000.0,-18.695300000000003,0.33060202661205745,-18.9276,0.13342446052604634,-17.7565,1.0134223831486395
440000.0,-17.2576,0.7735999999999983,-18.438650000000003,0.08985000000000021,-15.62715,0.9250505297009459,440000.0,-18.1749,0.5455366486314186,-18.816433333333332,0.18812216479960228,-17.498966666666664,0.8883177522085715
480000.0,-16.2281,0.6340000000000012,-18.37305,0.11524999999999963,-15.556550000000001,0.31116700098178846,480000.0,-17.445425,1.0553663448656105,-18.86236666666667,0.36704092717600556,-17.159733333333335,0.17258416948131608
520000.0,-16.5705,1.1123000000000003,-18.8541,0.15140000000000065,-15.058200000000003,0.24509929620461982,520000.0,-18.080274999999997,0.4772072263440689,-18.67343333333333,0.23796427369577097,-16.635299999999997,1.8717497286407354
560000.0,-16.13165,0.6812499999999995,-18.37675,0.695549999999999,-15.199875,0.5435218228139511,560000.0,-17.1862,0.8162674745449562,-18.616799999999998,0.17156946115203586,-17.046466666666664,0.8974116570572399
600000.0,-16.0041,1.056400000000001,-18.34845,0.5906500000000001,-15.558975,0.7435941210600043,600000.0,-16.802925000000002,0.9019463076453053,-18.630233333333333,0.2896086593241908,-15.743733333333333,0.6219961914853029
640000.0,-15.907,0.6961000000000004,-18.76265,0.035349999999999326,-14.845324999999999,1.4580232240520041,640000.0,-16.69905,1.1192660597462956,-18.665133333333333,0.19751938864042837,-15.929433333333336,1.0727467092572336
680000.0,-17.111150000000002,0.7267500000000009,-18.39845,0.13945000000000007,-14.555575,1.2415449173811637,680000.0,-16.335924999999996,1.3221612427669323,-18.5389,0.3127587035825984,-15.739333333333333,0.31493258890682535
720000.0,-15.371899999999998,0.6714999999999991,-18.4096,0.07949999999999946,-14.325875000000003,0.7325468120707375,720000.0,-16.646950000000004,1.447757221532671,-18.7383,0.27217308218607245,-15.732166666666666,0.7167815861350111
760000.0,-16.9369,0.6799,-18.647299999999998,0.5327999999999999,-14.477999999999998,0.5735232296951884,760000.0,-16.98575,0.7093356698348103,-18.644133333333333,0.5220783168154836,-15.733966666666667,1.0207432237780898
800000.0,-17.4992,1.798799999999999,-18.4203,0.12069999999999936,-15.389150000000003,0.7446921159109979,800000.0,-16.153225,1.0007103461416795,-19.04296666666666,0.14127331272709892,-15.133200000000002,0.40127526711722494
840000.0,-15.816999999999998,0.7794999999999997,-17.8547,0.39259999999999984,-14.737474999999998,0.7854189212611319,840000.0,-16.6236,0.8414774269105505,-18.906666666666666,0.2170610717952172,-14.9285,0.800180733751236
880000.0,-15.3793,1.3840000000000003,-18.4217,0.13339999999999996,-14.211025,1.2534995958814665,880000.0,-16.682425,0.7307550765304335,-18.811433333333337,0.39281768742707507,-15.546500000000002,0.796156630318432
920000.0,-15.693950000000001,0.97675,-18.879649999999998,0.07654999999999923,-13.990249999999998,1.5094719614818948,920000.0,-16.418175,1.3494024017597566,-18.521600000000003,0.4804410959385835,-14.5608,1.137577622259979
960000.0,-16.149,1.0334000000000003,-18.6553,0.583400000000001,-13.522450000000001,1.0106569558955203,960000.0,-16.058,1.3583533284826887,-18.747266666666665,0.4424135797594313,-14.9595,0.8848995460879542
1000000.0,-15.50195,1.1507499999999995,-18.053700000000003,0.0923999999999996,-13.72035,1.1242132326654053,1000000.0,-15.737200000000001,1.613806458346229,-18.76353333333333,0.27908565391689716,-13.967966666666667,1.4027763360168604
1040000.0,-14.8168,1.0554999999999994,-18.47715,0.4048500000000015,-13.6952,1.869493788435789,1040000.0,-15.841425,1.586104699215976,-18.8069,0.3471475766874953,-13.377999999999998,0.7718399315920371
1080000.0,-14.941999999999998,0.5834000000000001,-18.121850000000002,0.1941499999999987,-13.556724999999998,1.1049188191333335,1080000.0,-15.088275,1.1731052433072664,-18.249333333333336,0.1663673512308093,-13.895466666666666,1.533910247990048
1120000.0,-14.559149999999999,0.2693500000000002,-18.218700000000002,0.025000000000000355,-13.115425,0.6532697236785122,1120000.0,-15.510925,1.3795585315147016,-18.9836,0.2719207727752079,-13.808966666666663,1.4528517298364922
1160000.0,-14.400400000000001,1.1078000000000001,-18.366799999999998,0.10740000000000016,-13.179975,0.7468862308779032,1160000.0,-15.39415,1.3444539811760008,-19.05886666666667,0.1258788394537462,-13.627733333333333,1.4308030526790034
1200000.0,-14.79315,0.6052499999999998,-18.564449999999997,0.14605000000000068,-12.917775,0.6915335254888227,1200000.0,-14.798624999999998,2.219986554435635,-19.078,0.19360022382907083,-13.250533333333331,1.4205897398224756
1240000.0,-14.89495,0.3816500000000005,-18.821450000000002,0.185550000000001,-13.163574999999998,1.0613758697440787,1240000.0,-15.486775,1.5529942117970041,-18.994133333333334,0.22950390459036238,-12.935533333333332,1.5651151210764727
1280000.0,-15.4922,0.0007999999999999119,-18.51285,0.1816500000000012,-13.19755,0.8177358512991838,1280000.0,-14.9621,1.6583012738944636,-18.896766666666668,0.14250511881644337,-12.783166666666668,0.8746533992896208
1320000.0,-14.0789,0.6223000000000001,-18.208399999999997,0.5264000000000006,-13.780175,1.413960587100998,1320000.0,-15.526174999999997,2.162256409835568,-19.033333333333335,0.06780488346883408,-12.214566666666665,1.9781332956996491
1360000.0,-14.556649999999998,0.6265499999999999,-18.7816,0.31680000000000064,-13.2218,0.7650246172509746,1360000.0,-14.519625000000001,1.2554868585035046,-18.752333333333333,0.19863551769230336,-12.2931,1.165013736685824
1400000.0,-15.969149999999999,1.80785,-18.5324,0.19650000000000034,-12.668849999999999,1.1622237510479638,1400000.0,-14.852250000000002,1.5385361053611968,-18.9629,0.11832945533551636,-12.921100000000001,1.4474547477094641
1440000.0,-16.5117,0.8906000000000001,-18.8006,0.5014000000000003,-12.556650000000001,0.8669569928779624,1440000.0,-14.500875,1.092119662342456,-19.0754,0.33374973657917206,-11.4902,1.3058601941504564
1480000.0,-14.707,0.6034999999999995,-18.525550000000003,0.64785,-12.858200000000002,0.8692698344012637,1480000.0,-13.715625000000003,1.3164152087677354,-19.1056,0.14016506935277076,-11.896866666666668,1.5920335870270526
1520000.0,-14.68285,1.40585,-18.7498,0.12209999999999965,-12.99005,0.31442447026273207,1520000.0,-14.037599999999998,1.4334939065095462,-18.483966666666664,0.5502979213319107,-11.683333333333335,1.119607348830632
1560000.0,-14.059,1.1969000000000003,-18.71895,0.04005000000000081,-12.7758,1.693432381289551,1560000.0,-13.721,1.6654778098191523,-18.687766666666665,0.2613916899129653,-12.408333333333333,1.0559593878975124
1600000.0,-13.024600000000001,1.3516000000000004,-18.6186,0.3291000000000004,-12.60995,1.024501372619871,1600000.0,-13.937699999999998,1.3385295289981465,-18.400133333333333,0.40171560697698655,-11.942333333333332,1.596085146712278
1640000.0,-15.04705,1.3650499999999992,-17.9369,0.03220000000000134,-12.367475,1.0198357842687225,1640000.0,-13.326274999999999,1.7292740809007112,-18.791933333333336,0.27415701502768214,-11.372133333333332,1.8427803456251162
1680000.0,-14.53105,1.555250000000001,-18.199,0.3994,-12.833850000000002,0.823866665486594,1680000.0,-13.769349999999998,2.003798957605278,-18.6457,0.11026250495975629,-11.552600000000004,0.8286647130574987
1720000.0,-13.468150000000001,1.1923500000000002,-18.20645,0.3126500000000014,-12.215225,1.5298758403462025,1720000.0,-13.370999999999999,1.9899858730654345,-18.803,0.3060680425439197,-10.864333333333335,0.5039171514798402
1760000.0,-13.5242,1.2949000000000002,-18.257849999999998,0.3351500000000005,-12.598925,0.8522157484316991,1760000.0,-13.065700000000001,2.016115190161515,-18.688833333333335,0.36817535978872246,-11.2875,1.1780657904661638
1800000.0,-14.4047,0.6254,-18.33225,0.4681499999999996,-11.80675,1.3890521165528666,1800000.0,-12.521775,2.372370190732256,-18.938266666666667,0.2250234111277207,-11.385033333333332,1.253438950337121
1840000.0,-13.716999999999999,1.2045000000000003,-18.7652,0.0995999999999988,-11.380575,1.475628319352472,1840000.0,-12.8888,2.0962078057768982,-18.755733333333332,0.3375789224594583,-11.217733333333333,1.391573201651841
1880000.0,-12.78595,1.2257499999999997,-18.79065,0.42735000000000056,-12.1718,1.546342022322358,1880000.0,-12.841775,2.204584367148375,-18.502333333333336,0.6157462915483574,-11.775,0.93700168623114
1920000.0,-13.66095,0.4054500000000001,-18.8701,0.21229999999999905,-11.7925,0.9194272347499828,1920000.0,-12.907600000000002,2.5537584782042333,-18.73266666666667,0.5131325386508082,-10.316133333333333,0.9498130950642637
1960000.0,-12.551,0.6290999999999993,-18.688850000000002,0.5513500000000011,-11.572824999999998,0.9415190289500259,1960000.0,-12.70625,2.232735591712552,-19.0918,0.11404104524249042,-10.786233333333335,0.5874845605536276
2000000.0,-12.119150000000001,1.0441500000000001,-18.2457,0.13480000000000025,-11.788675000000001,1.6049282326244367,2000000.0,-12.97325,2.4684818061918143,-18.695966666666667,0.08942976138971925,-10.951166666666667,0.816070186251703
2040000.0,-12.38965,0.6955499999999999,-18.62775,0.5890500000000003,-12.319825,1.7790475603184415,2040000.0,-12.61545,2.5956386155433884,-18.938933333333335,0.11797065548497836,-10.924100000000001,1.0129695388641593
2080000.0,-12.878699999999998,2.2221,-18.8332,0.15080000000000027,-12.073725000000001,1.5803565030951081,2080000.0,-12.604575,2.097954356480379,-18.753666666666664,0.5002136165901748,-10.658733333333334,0.5369345666735278
2120000.0,-11.4016,0.35389999999999944,-18.8361,0.42440000000000033,-11.559549999999998,1.8278412643060669,2120000.0,-12.477250000000002,2.130717420846791,-18.9275,0.15838777309712626,-10.621633333333335,0.9474539930196549
2160000.0,-13.35275,0.90625,-19.09235,0.11425000000000018,-11.893675,0.7590821871675029,2160000.0,-11.841425,1.7971336850315283,-18.950233333333333,0.2701594631982301,-10.508366666666667,0.6579846215697012
2200000.0,-11.2947,0.6989999999999998,-18.0092,0.02990000000000137,-12.253799999999998,1.1647347874087046,2200000.0,-11.917275,1.8612953223158866,-18.770566666666667,0.05021675240616688,-10.246200000000002,0.6536254024031397
2240000.0,-10.6299,0.7825999999999995,-18.51935,0.2947500000000005,-11.0046,1.4392489412884768,2240000.0,-11.96945,1.841372131726773,-18.7156,0.3732460939737571,-10.328366666666666,0.9582772087913229
2280000.0,-11.67365,1.4490499999999997,-18.24555,0.07285000000000075,-11.708499999999999,0.7316265884998988,2280000.0,-12.138649999999998,2.4043918456649283,-18.65273333333333,0.2981974439118411,-10.102466666666666,0.680553002262784
2320000.0,-10.8469,0.8672000000000004,-17.96075,0.2259499999999992,-11.08155,0.9631234253718477,2320000.0,-11.892675,2.9602032957003135,-18.766966666666665,0.08701978063763642,-10.081766666666667,0.6321378242827181
2360000.0,-11.28105,0.49904999999999955,-17.582050000000002,0.13085000000000058,-11.921575,1.1007707172136263,2360000.0,-11.790625,2.1736368928307686,-18.522633333333335,0.24505455628401512,-9.761999999999999,1.0366350659706622
2400000.0,-11.581050000000001,1.0912499999999996,-18.4551,0.12929999999999886,-11.629574999999999,0.7691130000689105,2400000.0,-11.9168,2.5841645535840017,-18.587233333333334,0.2439007630619933,-9.681233333333333,0.5491136211103205
2440000.0,-11.260900000000001,0.3379000000000003,-18.60195,0.42965000000000053,-10.75655,1.1555476158514628,2440000.0,-12.09835,2.6914637936446413,-18.434366666666666,0.17992769535442776,-9.707166666666668,0.34952970625621493
2480000.0,-11.320699999999999,0.1003999999999996,-18.5238,0.024199999999998667,-11.149225000000001,0.9173939104196187,2480000.0,-12.402825,2.569275009000594,-18.8767,0.40462838094561115,-9.810033333333333,0.45312315716100404
2520000.0,-11.3711,0.8179999999999996,-18.474600000000002,0.40310000000000024,-11.11765,0.5632726537832274,2520000.0,-12.813175,3.0745774737148848,-19.007033333333332,0.18170579395152875,-10.0682,0.45051109494291763
2560000.0,-11.15215,0.9337499999999999,-18.8928,0.21189999999999998,-10.7042,0.7823586741386584,2560000.0,-12.756374999999998,3.5690492265973295,-18.719633333333334,0.23310593204711827,-9.986833333333335,0.14391410246702344
2600000.0,-11.246299999999998,0.18689999999999962,-18.8471,0.42050000000000054,-10.941225000000001,1.0405640426590765,2600000.0,-12.784749999999999,3.8338730948611213,-18.883300000000002,0.232870793932315,-9.775933333333333,0.5314662757148592
2640000.0,-10.7039,0.4358999999999993,-18.301949999999998,1.0542500000000015,-10.973149999999999,1.6068621417221831,2640000.0,-12.325899999999999,3.6190816715017635,-18.946866666666665,0.17171995289488612,-10.0578,0.1897898311290672
2680000.0,-10.2287,0.5197000000000003,-18.9791,0.254900000000001,-11.057125000000001,0.765511167047875,2680000.0,-12.297450000000001,3.6236791590178075,-18.377333333333336,0.16631985917368872,-9.738433333333335,0.3857728721526295
2720000.0,-11.257200000000001,0.4950999999999999,-18.3293,0.2922000000000011,-10.7914,1.285683961943992,2720000.0,-12.638975,3.741337890631505,-18.779400000000006,0.2819465197515301,-9.159,0.6167343890741513
2760000.0,-11.55745,0.2937500000000002,-18.31565,0.21834999999999916,-10.947350000000002,1.5099931332625327,2760000.0,-12.061449999999999,3.310794968357297,-18.672533333333334,0.27556335428032136,-9.345966666666667,0.7223399307497518
2800000.0,-10.91425,0.023249999999999993,-18.4348,0.6164000000000005,-10.565225,0.9485797208853877,2800000.0,-12.112125,3.810491837266549,-18.371499999999997,0.5066428525105238,-9.616033333333334,0.5508893072922079
2840000.0,-11.2469,1.3956999999999997,-18.728150000000003,0.3808500000000006,-10.875599999999999,1.2647717817851571,2840000.0,-11.9872,4.105759756731998,-18.437766666666665,0.12174619318711957,-8.975800000000001,0.8399068559469354
2880000.0,-10.37285,0.01034999999999986,-18.951349999999998,0.023249999999999105,-10.490025,1.5204496166841575,2880000.0,-12.492775,3.7787446476144697,-18.3087,0.17667203136508888,-9.562199999999999,1.0752477326954317
2920000.0,-10.22695,0.1609499999999997,-18.92675,0.016250000000001208,-10.337875,0.988860008734806,2920000.0,-12.166699999999999,3.975156366106873,-18.367866666666668,0.3068363009090605,-9.260133333333334,0.5300232720257562
2960000.0,-10.054499999999999,0.22750000000000004,-18.541,0.7555000000000014,-10.074025000000002,1.1488940973279476,2960000.0,-12.250175,3.6700604241449484,-18.817333333333334,0.23053447078955935,-9.276533333333333,0.5979539354238659
3000000.0,-11.0908,0.28690000000000015,-18.9033,0.22089999999999854,-10.066425000000002,0.8739333680979345,3000000.0,-12.0447,3.4931176010835934,-18.7978,0.42409674210805587,-9.381,0.4063871635111849
3040000.0,-10.8951,0.42989999999999995,-18.9,0.01720000000000077,-9.780874999999998,0.869758783729719,3040000.0,-12.263275,3.4879662027999925,-18.3703,0.5302880789407456,-9.2746,0.5240445019270781
3080000.0,-10.500399999999999,0.41330000000000044,-18.609750000000002,0.22855000000000025,-10.625574999999998,1.1229146369493095,3080000.0,-12.140025000000001,3.920189531371538,-18.807833333333335,0.5251298400290058,-9.082300000000002,0.48822077656186147
3120000.0,-10.5424,0.9783,-19.234,0.0,-10.244625000000001,1.314144184203164,3120000.0,-12.34795,3.4955638175407415,-18.254299999999997,1.518348978331397,-9.530866666666668,0.5606046279588577
3160000.0,-10.4756,0.1369000000000007,-18.6133,0.03790000000000049,-9.62815,1.0941016371891594,3160000.0,-11.991599999999998,3.1242481135466824,-18.452866666666665,0.6096867902638391,-9.594133333333334,0.9878575887017088
3200000.0,-10.21735,0.006050000000000111,-18.9508,0.02030000000000065,-9.654574999999998,0.6976387187326973,3200000.0,-12.018825000000001,3.5364813666800234,-18.386466666666667,0.25309339953639404,-9.3298,0.38322455558066765
3240000.0,-10.920100000000001,0.2153999999999998,-19.19555,0.07714999999999961,-9.875300000000001,1.1295658834260178,3240000.0,-11.352350000000001,3.3338340131896187,-18.44113333333333,0.2284364584639574,-9.353766666666667,1.4044619570813903
3280000.0,-10.74065,0.17304999999999993,-19.1965,0.2944999999999993,-10.108275,1.1175910396361448,3280000.0,-11.7384,2.881198801714315,-18.200633333333332,0.5801523095034802,-9.436833333333334,0.49883433678482453
3320000.0,-10.83945,0.59415,-18.60665,0.5523499999999988,-9.28965,0.7043931732349479,3320000.0,-10.962299999999999,3.385088819366487,-18.18386666666667,0.406237839476115,-9.106233333333334,0.3000697067164372
3360000.0,-11.44065,0.25234999999999985,-18.58245,0.7695500000000006,-9.595699999999999,0.5937637619457765,3360000.0,-10.994725,2.409153125451971,-18.4147,0.589623088647881,-8.878233333333332,0.7984275476766126
3400000.0,-11.21115,0.048650000000000304,-17.976200000000002,0.8785000000000007,-9.610349999999999,0.44252462360867534,3400000.0,-11.315800000000001,2.358702371008263,-18.634500000000003,0.4438694477734039,-8.941,1.1231616298051972
3440000.0,-10.68185,0.06585000000000019,-18.4508,0.013700000000000045,-9.434075,0.6743585410410401,3440000.0,-11.174525000000001,2.7082011736344476,-18.767733333333332,0.21860537250691844,-8.734766666666665,0.6578116717994255
3480000.0,-11.0234,0.1288999999999998,-18.2586,1.1805000000000003,-9.471,0.7973270219928585,3480000.0,-11.161024999999999,2.538073498123134,-18.391499999999997,0.5478069428792119,-9.245066666666666,0.2569151524444509
3520000.0,-10.9027,0.4375,-18.03885,0.5892499999999998,-9.62415,1.2337821211624038,3520000.0,-11.15,2.352059394445642,-18.16,0.7921480459274435,-8.669166666666667,0.8984693069635468
3560000.0,-10.67775,0.6925499999999998,-18.8582,0.23399999999999999,-9.024424999999999,0.9326819993304253,3560000.0,-11.405575,3.454486540989124,-18.153666666666663,0.9981445531028493,-9.1697,0.4375423179533612
3600000.0,-10.4881,0.013499999999999623,-18.090600000000002,0.7172000000000001,-9.7823,1.1862039011063825,3600000.0,-11.349525,3.046719108627345,-18.442566666666668,0.5853671573370768,-8.999233333333335,0.6567128359404049
3640000.0,-9.8869,0.10920000000000041,-17.21755,1.673449999999999,-9.223825000000001,0.6770720654959854,3640000.0,-11.371699999999997,2.8625412983920424,-17.807566666666663,0.8087369013193073,-8.5233,0.7645340149398197
3680000.0,-10.54555,0.018950000000000244,-18.142000000000003,0.35880000000000045,-9.631250000000001,0.7995100671661368,3680000.0,-11.028875,2.396246120888879,-18.287133333333333,0.9524188936013859,-8.761566666666665,0.5928079246733765
3720000.0,-10.6115,0.3783000000000003,-18.1791,0.20610000000000106,-9.588075,0.42315547600734166,3720000.0,-11.543624999999999,2.573975329500071,-17.963000000000005,0.9351557766846471,-8.217333333333332,0.12335207965638624
3760000.0,-10.07675,0.07554999999999978,-18.46095,0.04684999999999917,-8.921,0.5536205379138317,3760000.0,-10.710175,2.174425272773245,-17.92486666666667,0.7941360812578379,-8.3987,0.5051465199985713
3800000.0,-10.3379,0.37929999999999975,-18.2879,0.18440000000000012,-9.474525,1.3725220897584853,3800000.0,-11.10105,2.3281377337906792,-18.364733333333334,0.5847350758154413,-8.687733333333332,0.5995328255307535
3840000.0,-10.5125,0.05430000000000046,-18.659550000000003,0.3888499999999997,-9.31475,0.7605170395855707,3840000.0,-10.93155,2.167538117888587,-18.08566666666667,0.7657750946299804,-8.925133333333333,0.9745654153974933
3880000.0,-10.0549,0.28030000000000044,-18.677149999999997,0.2361500000000003,-9.341525,0.6014811816466081,3880000.0,-10.283100000000001,1.724285836223217,-18.464066666666668,0.6602089887974028,-8.611333333333333,0.7775792964207721
3920000.0,-10.17035,0.28125,-18.241,0.06170000000000009,-9.677725,0.3574253864445001,3920000.0,-10.536925000000002,1.6833292790999035,-18.008200000000002,0.6819025492448799,-8.734,0.4700184889980394
3960000.0,-9.424999999999999,0.6391,-17.893,0.4089999999999989,-9.849625,0.5743370324774467,3960000.0,-10.534875,1.5333104991732762,-17.907566666666664,1.0806452712255867,-8.3005,0.35419422167317527
4000000.0,-10.0293,0.32730000000000015,-18.05505,0.2472499999999993,-8.936425,0.9561370492115658,4000000.0,-10.902650000000001,1.7724762530144096,-18.287233333333333,1.0413973347809615,-8.247933333333334,0.2711672587578051
}\dataSotaLj

%% file: data/traj_comm.tex
\pgfplotstableread[col sep=comma]{
0.0,0.6995,0.45847546281125834,0.7335,0.44212865774568194,0.6845,0.4647146974219786,0.0
1.0,0.8245,0.3803942034258698,0.8055,0.3958152978347431,0.81,0.3923009049186594,0.0
2.0,0.769,0.42147241902643934,0.771,0.42018924307982314,0.7535,0.4309730269982055,0.0
3.0,0.6405,0.47985388401054463,0.6305,0.4826694003145448,0.6175,0.48599768517967956,0.0
4.0,0.5725,0.4947158275212163,0.579,0.49371955602345224,0.577,0.4940354238311182,0.0
5.0,0.5675,0.4954227992331458,0.5775,0.493957234991051,0.567,0.4954906659060347,0.0
6.0,0.6135,0.48694737908730434,0.598,0.4903019477832021,0.587,0.492372826220129,0.0
7.0,0.6075,0.48830702431974404,0.6195,0.4855097836295304,0.608,0.4881966816765523,0.0
8.0,0.6175,0.4859976851796796,0.6285,0.48320570153920034,0.624,0.48438001610305875,0.0
9.0,0.604,0.4890644129355555,0.6025,0.48938098655343276,0.6205,0.48526255779732347,0.0
10.0,0.6035,0.4891704713083068,0.603,0.4892759957324638,0.613,0.4870636508712128,0.0
11.0,0.569,0.49521611443894126,0.5775,0.49395723499105104,0.5915,0.491556456574413,0.0
12.0,0.5505,0.49744321283942483,0.5515,0.4973406780065405,0.5735,0.4945682460490181,0.0
13.0,0.52,0.49959983987186446,0.529,0.49915829152683877,0.5395,0.4984373079936843,0.0
14.0,0.4895,0.4998897378422605,0.495,0.49997499937496426,0.52,0.4995998398718645,0.0
15.0,0.454,0.497879503494571,0.473,0.49927046778274503,0.4795,0.4995795732413483,0.0
16.0,0.419,0.49339537898119873,0.446,0.4970754469896888,0.464,0.49870231601628906,0.0
17.0,0.396,0.48906441293555547,0.426,0.4944936804449518,0.437,0.4960151207372534,0.0
18.0,0.3785,0.48501314415179175,0.39,0.48774993593028887,0.421,0.4937195560234521,0.0
19.0,0.3485,0.4764952780458615,0.3595,0.4798538840105446,0.383,0.4861182983595639,0.0
20.0,0.3045,0.46019533895945114,0.3085,0.4618741711765268,0.33,0.47021271782035096,0.0
21.0,0.271,0.444476096095166,0.274,0.44600896851969957,0.286,0.45188936699152016,0.0
22.0,0.2235,0.41659062639477806,0.228,0.4195426080864742,0.245,0.4300872004605528,0.0
23.0,0.191,0.39308904843559206,0.1875,0.3903123748998999,0.2015,0.40112061776977026,0.0
24.0,0.1575,0.36427153333743306,0.1585,0.3652091866314462,0.177,0.38166870450693346,0.0
25.0,0.124,0.3295815528818369,0.124,0.3295815528818369,0.1395,0.3464675309462678,0.0
26.0,0.0935,0.29113184298527195,0.0975,0.2966374049239196,0.1015,0.3019896521405944,0.0
27.0,0.081,0.2728351150420381,0.082,0.27436472076416973,0.086,0.28036404905051604,0.0
28.0,0.067,0.2500219990320915,0.065,0.2465258607124251,0.075,0.2633913438213165,0.0
29.0,0.0475,0.21270578271406063,0.0495,0.21690954335851453,0.0535,0.22502833154961246,0.0
30.0,0.0455,0.20839805661282315,0.047,0.21163884331567895,0.049,0.21586801523153235,0.0
31.0,0.0355,0.1850398605706378,0.0375,0.18998355191963628,0.038,0.19119623427254026,0.0
32.0,0.032,0.17600000000000207,0.035,0.18377975949488906,0.0315,0.17466467874187178,0.0
33.0,0.026,0.1591351626762578,0.0265,0.1606167799453133,0.0265,0.1606167799453127,0.0
34.0,0.0205,0.14170303454760644,0.024,0.15304901175767266,0.023,0.14990330216509642,0.0
35.0,0.0215,0.14504395885385973,0.0205,0.1417030345476061,0.023,0.1499033021650964,0.0
36.0,0.0175,0.1311249404194359,0.018,0.13295111883696015,0.0215,0.14504395885386004,0.0
37.0,0.0155,0.12353036064061435,0.0155,0.12353036064061432,0.0175,0.1311249404194366,0.0
38.0,0.0135,0.11540255629751213,0.0115,0.10661965109678447,0.0135,0.11540255629751374,0.0
39.0,0.0085,0.09180277773575347,0.0085,0.09180277773575342,0.0125,0.11110243021644435,0.0
}\dataTrajComm

%% file: data/dummy.tex
\pgfplotstableread[col sep=comma]{
0,0,0,0,0
0,0,0,0,0
0,0,0,0,0
0,0,0,0,0
0,0,0,0,0
0,0,0,0,0
0,0,0,0,0
0,0,0,0,0
0,0,0,0,0
0,0,0,0,0
0,0,0,0,0
0,0,0,0,0
0,0,0,0,0
0,0,0,0,0
0,0,0,0,0
0,0,0,0,0
0,0,0,0,0
0,0,0,0,0
0,0,0,0,0
0,0,0,0,0
0,0,0,0,0
0,0,0,0,0
0,0,0,0,0
0,0,0,0,0
0,0,0,0,0
0,0,0,0,0
0,0,0,0,0
0,0,0,0,0
0,0,0,0,0
0,0,0,0,0
0,0,0,0,0
0,0,0,0,0
0,0,0,0,0
0,0,0,0,0
0,0,0,0,0
0,0,0,0,0
0,0,0,0,0
0,0,0,0,0
0,0,0,0,0
0,0,0,0,0
0,0,0,0,0
0,0,0,0,0
0,0,0,0,0
0,0,0,0,0
0,0,0,0,0
0,0,0,0,0
0,0,0,0,0
0,0,0,0,0
0,0,0,0,0
0,0,0,0,0
0,0,0,0,0
0,0,0,0,0
0,0,0,0,0
0,0,0,0,0
0,0,0,0,0
0,0,0,0,0
0,0,0,0,0
0,0,0,0,0
0,0,0,0,0
0,0,0,0,0
0,0,0,0,0
0,0,0,0,0
0,0,0,0,0
0,0,0,0,0
0,0,0,0,0
0,0,0,0,0
0,0,0,0,0
0,0,0,0,0
0,0,0,0,0
0,0,0,0,0
0,0,0,0,0
0,0,0,0,0
0,0,0,0,0
0,0,0,0,0
0,0,0,0,0
0,0,0,0,0
0,0,0,0,0
0,0,0,0,0
0,0,0,0,0
0,0,0,0,0
0,0,0,0,0
0,0,0,0,0
0,0,0,0,0
0,0,0,0,0
0,0,0,0,0
0,0,0,0,0
0,0,0,0,0
0,0,0,0,0
0,0,0,0,0
0,0,0,0,0
0,0,0,0,0
0,0,0,0,0
0,0,0,0,0
0,0,0,0,0
0,0,0,0,0
0,0,0,0,0
0,0,0,0,0
0,0,0,0,0
0,0,0,0,0
0,0,0,0,0
0,0,0,0,0
}\dataDummy

%% file: data/bw_limit_steps.tex
\pgfplotstableread[col sep=comma]{
0.0,44.893233333333335,0.15099086800936892,1.0,0.0,45.0,0.0,1.0,0.0,45.0,0.0,0.107,0.13961002829309938,45.0,0.0,0.3333333333333333,0.4714045207910317
40000.0,44.90103333333334,0.1399600022228577,1.0,0.0,44.8164,0.18359999999999843,1.0,0.0,44.83593333333334,0.23202530513334432,0.4907333333333333,0.171655397688379,45.0,0.0,0.44693333333333335,0.08371978393559207
80000.0,44.78906666666666,0.22068120798009871,1.0,0.0,44.886700000000005,0.011700000000001154,1.0,0.0,44.643233333333335,0.5045442586026427,0.48269999999999996,0.20376535524961056,45.0,0.0,0.4733,0.08402289370562446
120000.0,44.7682,0.19142211645122653,1.0,0.0,44.41015,0.5898499999999984,1.0,0.0,44.05466666666666,1.0186511648035146,0.44980000000000003,0.08351099727979941,45.0,0.0,0.6003333333333333,0.15140476287826027
160000.0,44.625,0.3724341910548332,1.0,0.0,44.51175,0.2929499999999976,1.0,0.0,43.770833333333336,0.9325861616434648,0.5124333333333334,0.04524395601133434,44.86976666666666,0.09405829161866774,0.5144333333333333,0.05812000994111714
200000.0,44.604166666666664,0.2157789815734819,1.0,0.0,44.84765,0.15234999999999843,1.0,0.0,41.89066666666667,1.328697009186905,0.4764,0.11343229992672575,44.091133333333325,0.42121988979101443,0.5453,0.06409560567360811
240000.0,44.44793333333333,0.7642535763010143,1.0,0.0,44.328149999999994,0.10154999999999959,1.0,0.0,42.5651,1.2031834855914534,0.49763333333333337,0.029680109313964628,44.28906666666666,0.5390464069900546,0.49720000000000003,0.1608838918806562
280000.0,43.890633333333334,0.6086405306546352,1.0,0.0,44.90625,0.09375,1.0,0.0,42.8099,0.3161503123515764,0.6234666666666667,0.10675911618634208,43.359399999999994,1.0006783832314303,0.5030333333333333,0.10277160221686835
320000.0,43.34633333333334,0.9331523145886846,1.0,0.0,44.8125,0.1875,1.0,0.0,41.911433333333335,0.46020137862558486,0.48216666666666663,0.08616667311411966,43.33596666666667,0.6222617098581237,0.6182,0.06588914933431757
360000.0,44.0052,0.39226328743162664,1.0,0.0,44.457049999999995,0.5429500000000012,1.0,0.0,40.8802,0.7064869472726748,0.5236333333333333,0.18437144271521252,42.872366666666665,1.1131963239049782,0.49720000000000003,0.06984110537498675
400000.0,42.395833333333336,1.8835453420492843,1.0,0.0,44.480450000000005,0.347649999999998,1.0,0.0,41.4141,1.7860862707794007,0.6278,0.12507920157510868,42.625,1.8416830400478816,0.4968333333333333,0.03150347423521556
440000.0,41.96616666666667,1.199372964881603,1.0,0.0,44.66015,0.15234999999999843,1.0,0.0,40.312533333333334,0.4106359404089657,0.5582333333333334,0.046692635631566365,41.934866666666665,0.6417085856437412,0.6177,0.10014392975446221
480000.0,42.57816666666667,0.27739546179096597,1.0,0.0,44.33985,0.15234999999999843,1.0,0.0,39.46093333333334,1.6211574966328492,0.5072,0.09722976224730094,40.47656666666666,1.0164623794754486,0.5461666666666667,0.07603605869731953
520000.0,41.5,0.816496580927726,1.0,0.0,44.0625,0.04690000000000083,1.0,0.0,38.4427,1.3196324968212414,0.5936,0.11851838113417963,40.427099999999996,0.7080753820509983,0.6328333333333332,0.022808818957197753
560000.0,41.958333333333336,0.575734605139864,1.0,0.0,44.4453,0.46089999999999876,1.0,0.0,38.18226666666667,0.9881516291653917,0.5623333333333334,0.16449478884011964,39.638,0.8169291809371632,0.5657666666666666,0.04850136309653805
600000.0,41.7396,1.3129920131770296,1.0,0.0,44.96875,0.01565000000000083,1.0,0.0,36.4271,1.347765093280972,0.5912333333333333,0.11307520604545553,41.38803333333333,1.2770944896739458,0.5990333333333334,0.03392052803572231
640000.0,41.182300000000005,0.9100506396166455,1.0,0.0,44.68355,0.15235000000000198,1.0,0.0,35.54683333333333,0.988267798839069,0.6156333333333334,0.07370266088970076,40.7526,1.0559491275624957,0.5725333333333333,0.06202329311548113
680000.0,41.0469,1.9284686688320007,1.0,0.0,44.074200000000005,0.8632999999999988,1.0,0.0,35.90106666666667,1.101202507362848,0.6491666666666668,0.14348594201368842,39.22656666666666,0.6041848962767016,0.6246666666666667,0.04975046621780429
720000.0,39.79946666666667,0.7538166060492145,1.0,0.0,43.6797,0.28910000000000124,1.0,0.0,35.2448,2.5187605933606867,0.6144333333333333,0.12953921757093056,39.31513333333333,0.8140088956652867,0.6212333333333334,0.03939630552334682
760000.0,39.96353333333334,0.89654672431998,1.0,0.0,43.1758,0.04299999999999926,1.0,0.0,33.958333333333336,0.9761323999449161,0.6470333333333333,0.15489085045784842,37.59113333333334,0.4280378825394904,0.5986666666666667,0.08973829852533546
800000.0,39.9948,1.2940906021862093,1.0,0.0,42.82425,0.7304499999999976,1.0,0.0,34.14843333333334,2.956610894626182,0.6533666666666667,0.07383451466323566,38.21353333333334,1.0343103896907473,0.5662999999999999,0.09686334015852784
840000.0,40.169266666666665,1.879405917363842,1.0,0.0,43.699200000000005,0.6992000000000012,1.0,0.0,32.38023333333333,0.8838965261210648,0.6361,0.11264066761165792,35.674499999999995,0.8398964340917293,0.5825,0.10388147091757993
880000.0,39.624966666666666,0.5113852841927395,1.0,0.0,43.417950000000005,0.29295000000000115,1.0,0.0,32.52343333333334,0.7655422580686774,0.6797,0.07431769641209289,37.177099999999996,0.9923480572191723,0.5827333333333333,0.10833839988152347
920000.0,38.7552,1.228035213935931,1.0,0.0,43.875,0.28119999999999834,1.0,0.0,31.367166666666666,1.519023230749142,0.6805666666666667,0.08999412326492336,34.205733333333335,0.6118570711821135,0.5881666666666666,0.09261174631522481
960000.0,37.89843333333334,2.3122359226995464,1.0,0.0,44.4297,0.0,1.0,0.0,32.28386666666666,2.3574296940146953,0.6343,0.1220705806763721,34.49216666666667,1.1344456188915453,0.5583666666666666,0.13729846644769522
1000000.0,38.90103333333334,1.7348823770567936,1.0,0.0,44.015649999999994,0.5234500000000004,1.0,0.0,30.625033333333334,2.5494141867408575,0.6718999999999999,0.09684671737682528,33.8099,0.8193187698732803,0.5808666666666666,0.07199635793257568
1040000.0,37.77863333333334,2.8153203492478243,1.0,0.0,44.015649999999994,0.3359500000000004,1.0,0.0,30.526033333333334,3.393128827825762,0.6735333333333333,0.08481919332058961,32.61976666666667,1.5082406093046163,0.5454,0.08960773776112568
1080000.0,36.080733333333335,2.022071812660359,1.0,0.0,43.28905,0.7578500000000012,1.0,0.0,29.54686666666667,3.5099172092167024,0.6832666666666666,0.07035340945698525,31.895866666666667,1.2874883697425088,0.5507,0.09641441800892646
1120000.0,35.627633333333335,0.9716085848849949,1.0,0.0,44.13675,0.2070500000000024,1.0,0.0,26.637999999999995,3.415630562575526,0.6851666666666666,0.093667722413979,30.885433333333335,0.7174406100081652,0.5365666666666666,0.07317423651046104
1160000.0,35.46353333333334,3.4946637800445925,1.0,0.0,42.35155,1.53125,1.0,0.0,27.091133333333335,2.4116716834778495,0.7206,0.07367554999229166,30.4896,1.0915951294626898,0.5617333333333333,0.10635814757485933
1200000.0,34.69533333333334,3.1696556377976175,1.0,0.0,44.53905,0.20314999999999728,1.0,0.0,24.54946666666667,1.3348235572122955,0.6866,0.06085195148883886,29.6771,1.3748071282910928,0.5563666666666667,0.10319858956831189
1240000.0,34.916666666666664,3.7432859542860935,1.0,0.0,42.9922,0.6327999999999996,1.0,0.0,24.854166666666668,2.0452908682032374,0.6729333333333333,0.07385446650150702,29.265633333333337,0.16278120967182455,0.5239999999999999,0.08445795798305014
1280000.0,32.27603333333334,2.8954633496020783,1.0,0.0,43.0078,0.7968999999999973,1.0,0.0,25.148433333333333,2.1437469631983666,0.6846000000000001,0.08586116700814171,27.567700000000002,1.3082649604214982,0.5377666666666667,0.13522382268750657
1320000.0,30.924466666666664,2.7711831664872353,1.0,0.0,43.15235,0.29295000000000115,1.0,0.0,23.458333333333332,0.4235577043200706,0.6844,0.07136306234086837,26.666666666666668,1.1214133027370223,0.5067,0.13145845985202576
1360000.0,31.067733333333337,4.03364472103773,1.0,0.0,43.886700000000005,0.6992000000000012,1.0,0.0,22.721333333333334,0.8463998516593029,0.6758333333333334,0.07149826726727176,25.968766666666667,1.275437144232866,0.4808666666666667,0.10382884420472419
1400000.0,29.8229,4.1833253885714745,1.0,0.0,43.7461,0.6445000000000007,1.0,0.0,22.096333333333334,1.1397167903572463,0.6798000000000001,0.0627094889151554,24.51823333333333,1.4627783526183618,0.5081333333333333,0.10946528622759313
1440000.0,30.01563333333333,4.673076064958593,1.0,0.0,43.980450000000005,1.0195499999999988,1.0,0.0,23.28643333333333,2.1024383373174635,0.6809666666666666,0.0806106830526961,25.09113333333333,0.6346169150667901,0.5455333333333333,0.11639600031310735
1480000.0,28.1771,2.26336694476761,1.0,0.0,43.1797,0.07029999999999959,1.0,0.0,22.960966666666668,0.7957944472175097,0.6753,0.07389591779433198,24.60936666666667,0.9221963071325369,0.4839,0.11309335376876337
1520000.0,27.645833333333332,4.100070902096967,1.0,0.0,43.7461,0.13670000000000115,1.0,0.0,22.20053333333333,1.0042180683939572,0.7080333333333333,0.07060284854184158,24.15363333333333,1.1491664873096308,0.5415666666666666,0.11807278357954566
1560000.0,28.718733333333333,3.992514350075095,1.0,0.0,43.6172,0.8281000000000027,1.0,0.0,21.76563333333333,0.5399124640984767,0.6941666666666667,0.0796617990144726,23.067733333333337,1.3864725537212124,0.5165,0.09563601134858493
1600000.0,27.710966666666668,3.3011743590560148,1.0,0.0,43.078149999999994,0.023450000000000415,1.0,0.0,21.804666666666666,0.8573171577012138,0.6943333333333334,0.07378822549853209,23.0625,0.8683283979386303,0.5367666666666667,0.09410410311045009
1640000.0,27.625,3.9923810497830328,1.0,0.0,44.171899999999994,0.5547000000000004,1.0,0.0,20.8854,0.013257450735340535,0.6802,0.05025740409797015,23.317700000000002,0.7718692505858745,0.5651,0.10500688866291898
1680000.0,26.718766666666667,3.4348841846883604,1.0,0.0,43.28125,0.32815000000000083,1.0,0.0,20.984366666666663,0.40339969701966305,0.6542333333333333,0.059020919078652825,22.7526,0.8307970911520259,0.5321333333333333,0.13337866729312034
1720000.0,25.440133333333335,3.3718290756732543,1.0,0.0,43.6172,0.8047000000000004,1.0,0.0,21.039033333333332,0.8456716160675037,0.6855000000000001,0.05706353184536224,23.374966666666666,0.7188686497236861,0.48163333333333336,0.10146350192173649
1760000.0,24.960966666666668,1.7905183762127546,1.0,0.0,42.863299999999995,0.441399999999998,1.0,0.0,21.8021,0.13279023558480005,0.6688999999999999,0.054533903827496746,22.119799999999998,1.4883521648678,0.5052666666666666,0.08746356701824796
1800000.0,25.6797,2.598117056639289,1.0,0.0,43.8203,0.5702999999999996,1.0,0.0,21.786466666666666,0.5436818084954554,0.6677,0.02894627206855256,22.447933333333335,0.8035054628867744,0.5433666666666667,0.11437366635530904
1840000.0,23.9974,1.29690312925317,1.0,0.0,43.265649999999994,0.4765499999999996,1.0,0.0,21.088533333333334,0.9130904202517713,0.6744666666666667,0.04246483512533898,21.739566666666665,1.1159860821513667,0.5507333333333334,0.09020888106069282
1880000.0,25.645833333333332,3.498737433544976,1.0,0.0,43.8047,1.0781000000000027,1.0,0.0,20.984400000000004,0.5516797863495342,0.6686,0.02126703238974981,23.1901,1.9312508709814655,0.5218666666666666,0.09518124231637709
1920000.0,23.76823333333333,2.3119654788849147,1.0,0.0,43.480450000000005,0.6679500000000012,1.0,0.0,20.9297,0.23385358667337133,0.6439666666666666,0.0039718453589679984,21.48696666666667,1.1247164096883353,0.5739,0.10213729322175452
1960000.0,22.84116666666667,1.274396788898793,1.0,0.0,43.0078,0.24220000000000041,1.0,0.0,20.6224,0.6899033120662633,0.6398333333333334,0.050935863811485765,21.6901,0.7055320687254417,0.5598,0.11305193496796066
2000000.0,22.960933333333333,1.6620325354483558,1.0,0.0,42.9844,0.25,1.0,0.0,21.04426666666667,0.2606102624396987,0.6305666666666666,0.03628492187611204,21.52603333333333,0.5811357978610119,0.5821333333333333,0.07712899728521189
2040000.0,22.098966666666666,1.087643818331882,1.0,0.0,43.324200000000005,1.6757999999999988,1.0,0.0,20.361966666666664,0.2862841517715492,0.6770666666666667,0.03828945314603256,22.02606666666667,0.3067405019809995,0.5772,0.07398436771823265
2080000.0,22.893199999999997,1.751001047020436,1.0,0.0,43.292950000000005,1.3554500000000012,1.0,0.0,20.1328,0.4651681918044996,0.6503,0.019617509186098732,21.351566666666667,0.36483167199256294,0.5762333333333334,0.11627671974886271
2120000.0,22.445300000000003,1.5587558265060844,1.0,0.0,43.54685,0.65625,1.0,0.0,20.549466666666667,0.803428944517739,0.6802666666666667,0.032954750121273184,21.46356666666667,0.6496809285248336,0.5834,0.0917751963586386
2160000.0,20.9375,1.185137370378078,1.0,0.0,44.2578,0.22660000000000124,1.0,0.0,20.156233333333333,0.7707841908659574,0.6603333333333333,0.02772920642371305,21.062466666666666,0.574345925572927,0.6012666666666667,0.09505652119777068
2200000.0,21.73696666666667,0.6118467146452795,1.0,0.0,43.88285,0.59375,1.0,0.0,19.7604,0.43195222729680055,0.6638666666666667,0.05915754859318932,20.90103333333333,0.3352675979307013,0.6406666666666667,0.12821443843117758
2240000.0,21.619799999999998,0.6646790854740848,1.0,0.0,43.0195,0.8789000000000016,1.0,0.0,20.0495,0.8473271269114429,0.6687,0.019337700656144916,20.838533333333334,0.3147017247419452,0.5729666666666666,0.08333747989683606
2280000.0,20.78386666666667,0.7168974465632371,1.0,0.0,42.99215,0.40625,1.0,0.0,20.487000000000002,0.11455691453014487,0.6648000000000001,0.032254405383864475,20.648433333333333,0.4783500972672172,0.5898,0.05284814723967782
2320000.0,21.351566666666667,0.45506402724110073,1.0,0.0,43.47655,0.44535000000000124,1.0,0.0,20.554666666666666,1.419842297189688,0.614,0.047305602205235664,20.367166666666666,0.5513487845476962,0.586,0.0721295131459146
2360000.0,21.023433333333333,0.6324067379639653,1.0,0.0,43.453149999999994,0.10154999999999959,1.0,0.0,19.6927,0.7934933017990758,0.6482333333333333,0.04027077793581289,20.5599,0.3904462660426745,0.5934333333333334,0.07485596539725847
2400000.0,21.2604,0.4543801345422863,1.0,0.0,43.828100000000006,0.4922000000000004,1.0,0.0,20.117166666666666,0.4425261147347372,0.6742333333333334,0.03187102480658919,20.4844,0.9863108941910765,0.5673,0.106696422932855
2440000.0,21.182266666666667,0.7259549358526934,1.0,0.0,43.1719,1.2030999999999992,1.0,0.0,19.557299999999998,0.6740630880464132,0.6756666666666667,0.017721988852521332,20.929666666666666,0.634727785922613,0.6066666666666666,0.09692589379979373
2480000.0,20.958333333333332,0.4671049156476755,1.0,0.0,43.207049999999995,0.8945499999999988,1.0,0.0,19.2266,0.1823199385695374,0.6764000000000001,0.01774711244118322,21.3698,0.6611437413049198,0.5946666666666667,0.09252136089694217
2520000.0,21.041666666666668,0.254840425015777,1.0,0.0,43.46095,0.039049999999999585,1.0,0.0,19.161466666666666,0.3274199783492489,0.6861,0.06349887138104633,20.606766666666665,0.6166422968157659,0.6118333333333333,0.03514506445513444
2560000.0,20.726566666666667,0.4530415163708021,1.0,0.0,42.86325,0.3320500000000024,1.0,0.0,19.450533333333336,0.20503323849778268,0.6723666666666666,0.03634742845863455,20.40886666666667,0.4807769152343125,0.5481,0.082256468843895
2600000.0,20.765633333333334,0.8686537719687607,1.0,0.0,43.7344,0.8125,1.0,0.0,20.3021,1.3780672625093457,0.6957999999999999,0.023076105968439866,20.309866666666668,0.5584523932758779,0.6051333333333333,0.08028965617617807
2640000.0,20.489566666666665,1.2954479851473082,1.0,0.0,42.476600000000005,0.3672000000000004,1.0,0.0,19.90103333333333,0.36097118382989446,0.6830999999999999,0.055278265771156986,20.4974,0.6655414988313406,0.6269666666666667,0.0872877361883609
2680000.0,21.3203,0.2917248475304529,1.0,0.0,42.7461,0.5585999999999984,1.0,0.0,20.1198,0.3026116433098154,0.6914333333333333,0.04583858151770794,20.51823333333333,1.0093922538945015,0.5885000000000001,0.053434321055541294
2720000.0,20.5365,0.40188088616736495,1.0,0.0,44.14845,0.46875,1.0,0.0,19.73436666666667,0.7863798672114856,0.6953999999999999,0.04729545432702811,20.3021,0.5981837566054984,0.6223,0.07900649762308581
2760000.0,21.145866666666667,0.9273811202640596,1.0,0.0,43.16015,0.8242499999999993,1.0,0.0,20.067733333333333,0.32022476828350105,0.6571666666666667,0.007286669716376293,19.854166666666668,0.5364448423546347,0.6257,0.07649710234163559
2800000.0,20.6354,1.0336152701400405,1.0,0.0,42.52345,0.44535000000000124,1.0,0.0,19.999966666666666,0.0710431949982235,0.7264333333333334,0.014693384754900951,21.3073,0.5007077457626022,0.6104333333333333,0.062239876463744875
2840000.0,20.679733333333335,0.8579738198544029,1.0,0.0,42.6836,0.8242000000000012,1.0,0.0,18.981800000000003,0.6539150760356174,0.6818333333333334,0.02899498040849293,20.523433333333333,0.22378855099301947,0.5855,0.08984835372262905
2880000.0,20.541700000000002,0.7106033633469514,1.0,0.0,41.7422,1.2031000000000027,1.0,0.0,19.450533333333336,0.34271020151466514,0.6998666666666667,0.033490330279383966,20.588533333333334,0.7849951903603533,0.5972666666666667,0.06350434805761115
2920000.0,20.835933333333333,0.39516305776504856,1.0,0.0,43.03125,0.04684999999999917,1.0,0.0,19.51823333333333,0.5585250178421337,0.6797333333333334,0.01306403034629395,19.908833333333334,0.9645571568802374,0.6035666666666667,0.05758850193880333
2960000.0,20.7526,0.4259051850666619,1.0,0.0,43.542950000000005,0.285149999999998,1.0,0.0,20.112,0.4580889506053016,0.6814333333333332,0.014121457274500955,19.778666666666666,0.5195587957317457,0.5754,0.039430276015603394
3000000.0,20.184900000000003,0.7509153125796982,1.0,0.0,43.2148,0.19139999999999802,1.0,0.0,19.8203,1.2651459230723803,0.6813333333333333,0.03593784758286014,20.0625,0.6622172654549767,0.5987333333333335,0.059576524086440444
3040000.0,20.651033333333334,0.3912642272883575,1.0,0.0,44.2539,0.26950000000000074,1.0,0.0,18.7474,0.4338764878011561,0.6679666666666667,0.036618786920862845,20.182299999999998,0.8245646608968873,0.6152000000000001,0.07983612382040271
3080000.0,20.320333333333334,0.7305457792332769,1.0,0.0,44.0703,0.13279999999999959,1.0,0.0,19.007833333333334,0.49460810300232094,0.6917333333333334,0.039492812285556775,19.820333333333334,0.2180237958470493,0.5827000000000001,0.011203868379567244
3120000.0,19.971366666666665,0.66673481151721,1.0,0.0,43.7617,0.23049999999999926,1.0,0.0,19.648433333333333,1.0164153656628554,0.6674000000000001,0.05357854048030796,20.085933333333333,0.33973143641542736,0.5983666666666666,0.074851868528596
3160000.0,20.09633333333333,0.3431713889913057,1.0,0.0,44.375,0.07030000000000314,1.0,0.0,19.3802,0.3218798222939734,0.6822333333333334,0.03405342208289141,19.888,1.1048957597891305,0.5755666666666667,0.07781560826009706
3200000.0,20.700533333333333,0.681708864219585,1.0,0.0,43.6875,0.6171999999999969,1.0,0.0,18.875,0.36925680314202286,0.6763666666666667,0.044581112094199196,20.218766666666667,0.28406656888052795,0.5655,0.01925062769539387
3240000.0,19.9375,0.8739659070390938,1.0,0.0,43.7578,0.14060000000000272,1.0,0.0,19.132833333333334,0.315326078140639,0.6570333333333334,0.026486516485856634,20.3177,0.5688410908739511,0.5708333333333333,0.02029931580675127
3280000.0,19.3151,0.5965171637653573,1.0,0.0,42.921850000000006,0.3515499999999996,1.0,0.0,19.1432,0.3572361777125417,0.6632000000000001,0.036187659038222804,19.58073333333333,0.649260647882565,0.5694333333333333,0.04296799842777048
3320000.0,20.3854,0.7916013685351138,1.0,0.0,43.6875,0.7344000000000008,1.0,0.0,18.682266666666667,0.48586379664355434,0.6660333333333334,0.027682645987854686,20.3073,0.24358189587898307,0.5708666666666667,0.02584625483293251
3360000.0,20.612000000000002,0.14690861104782146,1.0,0.0,43.703100000000006,0.4452999999999996,1.0,0.0,19.40363333333333,0.4953621054900706,0.6280666666666667,0.041325563785896774,20.364566666666665,0.45668123334431837,0.5714333333333333,0.051121902242472254
3400000.0,20.28386666666667,0.31719169107794937,1.0,0.0,43.488299999999995,0.5117000000000012,1.0,0.0,18.937466666666666,0.2906963402284623,0.6886,0.0062631195634976185,19.721366666666665,0.5943654898760148,0.6240666666666667,0.05771656800453592
3440000.0,20.10676666666667,0.3941456358025837,1.0,0.0,42.457049999999995,0.23045000000000115,1.0,0.0,19.093733333333333,0.8037944278372556,0.6865666666666667,0.04531595255046014,19.6198,1.0585406211698571,0.5644999999999999,0.024978791003569423
3480000.0,20.614566666666665,0.33810247296082035,1.0,0.0,43.58985,0.5195499999999988,1.0,0.0,19.174466666666664,0.4748681524615244,0.6617333333333333,0.03238850962232681,19.76563333333333,0.9549889225651901,0.5641999999999999,0.04432245480566255
3520000.0,19.822933333333335,0.5618268554951377,1.0,0.0,42.257850000000005,0.16404999999999959,1.0,0.0,19.257833333333334,0.1053957726329139,0.6598,0.03269342441531632,19.95573333333333,0.45953894527256656,0.5734333333333334,0.03353011913025197
3560000.0,20.1875,0.6411368392680834,1.0,0.0,43.73825,0.0429499999999976,1.0,0.0,19.1172,0.29732006323152943,0.6424333333333333,0.04482486908947853,19.9349,0.5888061537269007,0.5957,0.025515616133393027
3600000.0,19.2422,0.35223532853288136,1.0,0.0,42.855450000000005,0.285149999999998,1.0,0.0,18.890599999999996,0.2801691036975112,0.6603,0.013305888420795785,19.08073333333333,0.8733724075228286,0.5974666666666666,0.023374820260746876
3640000.0,20.04426666666667,0.7199779826880502,1.0,0.0,43.66405,0.53125,1.0,0.0,19.049500000000002,0.6174939729800343,0.6419666666666667,0.027399553929864547,19.984333333333332,0.871150049583244,0.5621666666666667,0.030605700267905794
3680000.0,19.4844,0.412352009816855,1.0,0.0,43.320350000000005,0.14845000000000041,1.0,0.0,19.0755,0.18336653638727723,0.6823,0.03882275964774614,20.0,0.4783136209643203,0.5881333333333334,0.015390545438316623
3720000.0,19.734366666666666,0.7765600484758974,1.0,0.0,42.480450000000005,0.7929500000000012,1.0,0.0,18.7474,0.27252659809029023,0.6697666666666667,0.04269335103060219,19.997400000000003,0.9443189397655851,0.5618,0.024476927911811153
3760000.0,19.640600000000003,0.6804004164215856,1.0,0.0,42.93355,0.652350000000002,1.0,0.0,18.518233333333335,0.6835657799769946,0.6846333333333333,0.04718822828724225,19.729166666666668,0.15286251629777914,0.5688000000000001,0.006011655346075648
3800000.0,20.135433333333335,0.9716085848849948,1.0,0.0,42.82035,0.34375,1.0,0.0,19.09636666666667,0.22612023841802079,0.6712333333333333,0.02341457286012757,19.411466666666666,0.4411789533612043,0.5779,0.04677456573822999
3840000.0,20.080699999999997,0.4668296548706669,1.0,0.0,42.55465,0.28125,1.0,0.0,19.015633333333334,0.2008265807993449,0.6231,0.06627805066535979,19.28643333333333,0.5478501031811124,0.5848666666666666,0.03467874789486431
3880000.0,20.23436666666667,0.3156671805698032,1.0,0.0,43.58985,0.11325000000000074,1.0,0.0,18.73176666666667,0.20829713285486073,0.6659666666666667,0.0041354833118055255,19.453133333333334,0.8138757986053864,0.5964333333333333,0.04391843449952293
3920000.0,19.710933333333333,0.6744514083477198,1.0,0.0,42.88675,0.0429499999999976,1.0,0.0,18.75,0.3181400425389222,0.6562666666666667,0.06073825446582705,19.5677,0.4663733268530696,0.5508000000000001,0.026501320721805565
3960000.0,19.364566666666665,0.4564786109142717,1.0,0.0,43.21875,1.0546500000000023,1.0,0.0,19.499966666666666,0.21545539883903742,0.6289,0.024544245761481467,19.627566666666663,0.19993819600622176,0.5631,0.003053959178945702
4000000.0,19.747400000000003,0.405152177171327,1.0,0.0,42.578149999999994,1.2265499999999996,1.0,0.0,18.791666666666668,0.4774405745453793,0.6467666666666667,0.043652440430697076,19.2005,0.29650741418498644,0.5720999999999999,0.023394016328967517
}\dataBwLimitSteps

%% file: data_new/sota_steps_pp.tex
\pgfplotstableread[col sep=comma]{
0.0,45.0,0.0,-17.321633333333335,0.3313669298862256,45.0,0.0,-17.268733333333333,0.6240553732553612,45.0,0.0,-17.275533333333332,0.07913256107458043,44.97396666666666,0.03681669307377918,-16.293733333333336,0.4686900492033318,45.0,0.0,-17.460433333333334,0.31304479267705865
40000.0,45.0,0.0,-16.585933333333333,0.10756568019380253,44.90363333333334,0.13628304696068616,-15.5961,0.959356419689784,45.0,0.0,-16.750766666666664,0.20717724027722872,44.82033333333334,0.15715591691763525,-15.4474,0.5036420620506847,45.0,0.0,-16.567466666666665,0.04789609123462562
80000.0,44.9297,0.09941921343482966,-16.325266666666664,0.23911158250676356,44.515633333333334,0.5159599295380317,-15.142966666666666,0.9551642383497307,45.0,0.0,-17.26406666666667,0.1547226622774513,45.0,0.0,-15.8234,0.543234909285721,45.0,0.0,-16.903366666666667,0.15636730974073673
120000.0,44.955733333333335,0.06260252036105049,-16.1724,0.6734264028087995,45.0,0.0,-15.973700000000001,0.7445293412619812,44.7552,0.27252659809029023,-15.842166666666666,0.6211464150173361,44.861999999999995,0.17887045591712367,-14.975766666666665,0.40405763065296657,44.91146666666666,0.12520504072209762,-16.006766666666667,0.22261470950700696
160000.0,44.9401,0.04788200775517404,-15.880733333333332,0.37942398741015537,44.90103333333334,0.1399600022228577,-16.0776,0.774314617366007,44.8099,0.26884199820712684,-15.440900000000001,0.6754534230179509,45.0,0.0,-16.462500000000002,0.6754314818445075,44.78386666666666,0.16831934595352482,-15.054966666666667,0.47950310623486936
200000.0,44.71876666666666,0.23976591825269072,-15.6104,0.4804168398380726,44.8203,0.2541341771584441,-16.1909,1.2386726955360992,44.5026,0.703429825924376,-14.334133333333334,0.2043081713708218,44.97656666666666,0.03313973781161101,-15.121333333333334,0.5300994896138735,44.5,0.38785056400629464,-14.9172,0.4226091653841245
240000.0,44.8698,0.09228531844231758,-15.576066666666668,0.7406269071236586,45.0,0.0,-16.240866666666665,0.6159876099042536,44.143233333333335,0.6125671736407525,-14.347666666666667,0.9103915872975881,44.72656666666666,0.13180759546483875,-15.721366666666668,0.5833664390605812,44.53906666666666,0.16587128209012397,-14.294266666666667,0.6015651382481834
280000.0,45.0,0.0,-15.484633333333333,0.5140862270951138,45.0,0.0,-16.028666666666666,0.49789006372446926,44.24743333333333,0.4310552349241967,-14.0849,0.8442401830442959,44.8802,0.09576401551035091,-16.156499999999998,0.5133155429817678,43.955733333333335,0.5986597995151803,-13.614333333333335,0.7257163648576634
320000.0,44.11196666666666,0.6229417326060447,-15.0401,0.9641397858540358,44.90626666666666,0.13255895124643732,-15.847133333333334,1.1244267349286128,43.3724,0.9511565626471119,-13.569033333333332,0.2580365004327014,44.8099,0.15019034589479985,-14.853633333333333,0.2935172491618771,44.1302,0.4082648486787304,-13.496099999999998,0.6478661795978139
360000.0,44.6276,0.1786435557192038,-14.382299999999999,0.6266419073123022,45.0,0.0,-15.944266666666666,0.4622845756553946,43.49216666666667,0.3090346402733677,-12.973966666666668,0.1729404971530834,44.875,0.1767766952966369,-15.570599999999999,0.3383991430249193,43.59373333333334,0.6963156388369348,-13.107266666666666,0.62271125643342
400000.0,44.10156666666666,0.3278146359691037,-14.029933333333332,0.7651432690824783,44.854166666666664,0.179303547712316,-15.341366666666667,0.18464129789645892,42.46093333333334,0.9465946028909205,-12.809899999999999,0.6853226733931012,44.02863333333334,0.48753267468846506,-14.225766666666667,0.641994352692365,43.8568,0.5676972667422897,-13.247166666666667,1.056062136219056
440000.0,44.20053333333334,0.553703225764691,-13.6984,0.6181478841399256,44.856766666666665,0.15552393885044039,-15.643233333333335,0.9992546233179125,42.36456666666667,0.5646202519294606,-12.751566666666667,0.29117257578434247,44.8672,0.13181747481524062,-14.741666666666667,0.1855177320066441,44.09373333333334,0.8176802811748743,-13.2169,1.1923634596883623
480000.0,43.6875,0.9549349960424882,-13.371366666666667,1.6093556854288678,44.71353333333334,0.2501192426734811,-15.309133333333333,1.6275443841020816,42.05466666666667,0.7020453752350241,-12.368733333333333,0.0064095414985954075,43.65363333333334,0.4515569387008558,-13.498433333333333,0.4682579227543535,44.0703,0.22106380677683857,-13.313833333333335,0.3100683401373895
520000.0,43.21353333333334,0.7032574414031361,-12.730733333333333,0.722020379829329,44.40103333333334,0.45965375616386983,-14.1802,0.9161962780976575,40.3958,1.021421078693798,-11.5362,0.2905575445013713,44.49216666666667,0.20250771727407385,-14.179699999999999,0.4840645480374148,43.484366666666666,0.8211756382704551,-12.495833333333332,0.922100111460549
560000.0,43.875,0.8771264713065415,-13.927333333333332,1.104083820288216,44.362,0.5230606338338473,-14.4794,0.7846466253458731,41.71093333333334,0.17361262114898734,-11.624733333333332,0.18816188656461622,44.53126666666666,0.5252659918767086,-14.108033333333333,0.8016846276574241,43.34113333333334,0.5252359554418262,-13.0672,0.44687154753911124
600000.0,43.1823,0.4761873650850728,-12.890900000000002,0.7092153739638377,44.499966666666666,0.4369737164737575,-14.236233333333333,0.8255152747762392,41.7448,0.3467262897445193,-12.220033333333333,0.5238247692586605,44.6172,0.3125548911791332,-14.085666666666667,0.5657777910884175,43.6849,0.5432087628159179,-12.922633333333332,0.4306078365391058
640000.0,42.86196666666667,0.3457375626428528,-12.380233333333331,0.8219251804283781,44.64846666666667,0.3968601544904972,-14.542466666666664,0.7649576081210132,42.08593333333334,0.6519826037215698,-12.380166666666668,0.18179443213573812,44.609366666666666,0.1486661659184407,-13.548700000000002,0.03244729059053604,43.61976666666667,0.5978765777501448,-12.8414,0.6372399391124195
680000.0,42.96873333333334,0.6021645529993346,-12.7555,0.6109870211387474,44.02606666666666,0.7948481713859301,-13.5995,1.1446160433379684,38.200500000000005,0.8245400657336212,-10.136700000000001,0.4127995881780893,44.044266666666665,0.379744724197132,-13.825266666666666,0.6838282061713712,42.703100000000006,0.5852997921293551,-11.760166666666665,0.36645645792584347
720000.0,41.643233333333335,0.9149507977056598,-11.986200000000002,0.551143592904789,43.6198,0.4484540630506829,-13.305733333333334,0.5089374050138407,39.171866666666666,0.7095448133995631,-10.814566666666666,0.06174713668575187,44.294266666666665,0.4433207291441375,-13.835133333333332,0.8122615438118167,42.794266666666665,0.5935111362804341,-12.031766666666668,0.27218162236924726
760000.0,41.4844,1.5745875163567986,-11.817933333333334,0.6767976424965507,44.309866666666665,0.6281630379306175,-14.1685,0.8942508074733098,38.97396666666666,1.3651839249305915,-11.009633333333333,0.6285705069618064,43.705733333333335,0.2287033206774409,-13.256499999999997,0.20083108989065093,43.57293333333333,0.39291669459172884,-12.9596,0.28653203427656476
800000.0,41.78383333333334,1.4010581009445051,-12.310433333333334,0.4149089244105938,43.66406666666666,0.6630082570291957,-13.678133333333333,0.5060928263558864,38.25783333333333,1.4299367383054244,-9.934366666666667,0.7250918301996119,44.356766666666665,0.5546562499021851,-13.702066666666667,0.5268463048070929,42.867200000000004,0.4995812713329695,-12.268733333333335,0.4678602806631717
840000.0,42.08853333333334,1.0525231346098236,-11.9086,0.7163932765364749,43.544266666666665,0.8325467647859508,-12.9,0.5244458853558359,36.8776,0.5946076409420498,-9.2315,0.24105287110231005,43.825500000000005,0.5407391299570136,-13.543233333333333,0.16793868590120106,42.677099999999996,0.803264120116582,-12.294566666666668,0.6616765691954206
880000.0,42.72393333333334,0.4330200944785619,-12.3667,0.3464086315321828,43.84896666666666,0.9279859960629178,-13.7617,0.8633648745847063,37.52863333333334,0.16649913580022851,-9.655466666666667,0.47810510931755923,43.96876666666666,0.5358787943389998,-13.598966666666668,1.0201560022314664,42.984333333333325,0.1628689725583793,-12.118266666666665,0.25538588232102605
920000.0,42.234366666666666,0.5326703441675322,-12.160966666666667,0.14904859013832442,43.739599999999996,0.7474907535660015,-13.169533333333334,0.7797142824280078,35.802099999999996,0.3242691783071591,-8.750533333333333,0.8562684794437371,44.109399999999994,0.1823199385695382,-13.571866666666665,0.17541315673448102,41.895833333333336,0.8892190781179227,-11.445566666666666,0.2918966750219825
960000.0,41.29423333333333,0.8249900699739106,-11.5456,0.6226257516893003,43.856766666666665,0.6823044693455321,-13.845833333333331,0.8029811136564105,34.763000000000005,1.964042083731066,-8.663033333333333,0.8671135886119853,44.0625,0.3916031920196795,-13.6573,0.49672242416330176,41.71876666666666,0.34838027434910396,-11.422666666666666,0.4005902506163734
1000000.0,41.97656666666666,0.3600894179073987,-11.8302,0.28437546776518297,43.0573,0.8165296973574608,-12.797366666666667,0.7035798809580108,37.3099,0.8718320136356565,-9.240900000000002,0.35234461350596313,44.109366666666666,0.6425651008955334,-13.136433333333335,0.32666221425537134,42.171866666666666,0.42319215756228545,-12.016399999999999,0.4205999603740663
1040000.0,41.65103333333334,0.554674561241895,-11.4052,0.38068202829483155,44.08853333333334,0.406900121514959,-13.581766666666667,0.38596031171899275,34.82293333333333,0.8596802868250227,-8.1224,0.4221536292236117,44.291666666666664,0.14183423031443526,-13.475,0.6560900903585325,41.796866666666666,0.5973377901619454,-11.646066666666664,0.6111788081695536
1080000.0,41.65103333333334,1.4883396618006544,-11.627333333333333,1.2215592503936192,43.510400000000004,0.3994064346001449,-13.014833333333334,0.34529146464335825,33.427099999999996,0.6611055336832894,-7.665366666666666,0.4262922341409575,43.51303333333333,0.08299776436078753,-12.984900000000001,0.5297188373719272,41.79946666666667,0.9029038203977698,-11.423966666666667,0.1851256210132877
1120000.0,40.46356666666666,1.1847530272405118,-11.297633333333332,0.9039547714841094,43.3047,1.219960269844884,-13.333866666666665,1.888611457011621,30.921900000000004,0.9576158206713169,-6.065366666666667,0.40283621430496463,43.82293333333334,0.4394465635571896,-13.143766666666666,0.4222483971418824,41.0026,0.8890789091338679,-10.953133333333334,0.5402019396073618
1160000.0,40.70053333333333,0.44487609011449075,-10.6427,0.12783648409850246,43.4193,0.601772321286603,-12.610166666666666,0.6611932966656235,32.32033333333333,0.39544056724395715,-6.848466666666667,0.14144118526401317,44.375,0.563638258696717,-13.442966666666669,0.2997138338852956,41.31773333333334,1.0182858843282792,-11.171333333333331,1.0274967131604635
1200000.0,42.28386666666666,0.8994101560961449,-11.690633333333333,0.21498791180487867,43.5625,0.5814347828146053,-13.001033333333332,0.5694460602687107,30.0651,2.2246779916802937,-5.694533333333333,1.3054754800029333,43.69793333333333,0.6599395797259694,-13.142966666666666,1.0927024734828577,39.80466666666667,2.188210182368737,-10.2518,0.9899256773448538
1240000.0,40.979166666666664,0.7252699007189574,-10.938,0.6710162789878253,43.71353333333334,0.1033397417367686,-13.128633333333333,0.1211159316064109,30.2266,0.6654000500951791,-5.466133333333334,0.40642327962633035,44.25523333333333,0.07610485456847775,-13.272666666666666,0.43032363273342106,40.822900000000004,0.8857734699120303,-10.910433333333335,0.7034238377788711
1280000.0,43.33853333333334,1.3248531021295225,-12.610433333333333,1.2834586743985525,44.138000000000005,0.5968239327194133,-13.536433333333335,0.23985963024698903,29.388,0.6762177805017163,-4.9911666666666665,0.34333744657730286,43.888000000000005,0.3404897159484651,-12.9643,0.6787128749822465,40.59373333333334,0.4946254767217547,-10.591933333333332,0.5218507470745083
1320000.0,43.27346666666667,0.5800720031934746,-11.924466666666666,0.888828301129576,42.9323,0.47264642175732424,-12.745033333333334,0.4273424102000126,28.143199999999997,1.0630179396416606,-4.792433333333333,0.6300294825552917,42.96876666666666,0.49978934451315493,-12.915366666666666,0.9201380959158008,40.013000000000005,1.1985585870814444,-10.601033333333334,0.556080595677361
1360000.0,42.250033333333334,1.9690348504336372,-11.7448,1.3925869045293608,43.66146666666666,0.5125572249894517,-13.305466666666666,0.6994694767385414,27.3099,1.2145620802028465,-4.3552333333333335,0.6763676531459959,43.3177,0.5190182334626287,-13.257033333333334,0.6644691582174613,40.19533333333333,1.6889235315891455,-10.1328,0.8257923992545986
1400000.0,41.41406666666666,2.0990667106015364,-11.458866666666667,2.156814849623294,42.75,0.7631479061536286,-13.050533333333334,0.3881375386237323,26.70573333333333,0.5523555155473298,-4.104966666666667,0.30505490580476763,43.677099999999996,0.40269911844949463,-13.252866666666668,0.6364424055290125,39.5052,1.2783550550088447,-9.9216,1.1087638732690863
1440000.0,40.679700000000004,2.167972888821876,-10.845066666666668,1.0828332414960713,43.09373333333334,0.7921412093532021,-12.945066666666667,0.7609475555005237,25.434866666666665,1.1035606140529337,-3.405466666666667,0.5498872874406981,43.53906666666666,0.575045345071926,-13.495033333333334,0.2746014849842504,39.90886666666667,1.3733498227165413,-10.332033333333333,0.6949352216014253
1480000.0,40.76043333333333,1.6759194657129421,-11.048966666666667,0.7158901327873032,42.919266666666665,0.7608466964872448,-13.022933333333333,0.30819087519839977,25.638,0.9848375128246623,-3.5065000000000004,0.30628406205133596,42.760400000000004,0.6511566068670928,-12.871133333333333,0.5715437593823318,40.3125,0.44774034886304587,-10.272633333333333,0.28691691402835745
1520000.0,38.83593333333334,3.58253544605245,-9.884633333333333,1.612130595068388,41.80206666666667,0.8213056488840736,-12.238033333333334,0.776494191149482,25.23176666666667,0.1991229156967015,-3.243733333333333,0.19044133538237504,43.38543333333333,0.8157293600753182,-13.089833333333333,0.6644931719405067,40.294266666666665,0.2838611436757235,-9.9703,0.38109403913819867
1560000.0,38.9401,1.4063267496093041,-9.554933333333333,0.5371274357377606,42.49476666666667,0.4914013453606138,-12.467466666666667,0.40584464460622766,25.239566666666665,1.2136180682387505,-3.3723666666666667,0.21380817466962193,43.203100000000006,0.9827739957216328,-13.1383,0.7698756306486569,39.455733333333335,1.5502882169311467,-10.133066666666666,0.8721217970495224
1600000.0,37.64586666666667,3.143131120749215,-9.2758,1.293239312733726,42.97656666666666,0.7294049462099592,-12.601033333333334,0.8171665694472712,24.307266666666663,0.37395116734081213,-2.7786333333333335,0.24846250331902303,43.8125,0.5824324338496272,-13.217466666666667,0.21868611194027734,39.4974,1.5280473029327355,-9.579166666666667,0.9641141091984676
1640000.0,38.90626666666666,1.8088590406355296,-9.606233333333334,0.5486696658484254,42.578133333333334,0.14999978518503215,-12.6529,0.2227219043261496,24.0547,0.879069443597414,-2.9573,0.4967570902027134,43.66923333333333,0.024158136425550692,-12.9943,0.4140288959319945,39.6276,0.27987286399363803,-9.779133333333334,0.14484778984237953
1680000.0,39.315133333333335,1.199370630326126,-9.7432,0.703292390593462,43.440099999999994,0.9251510183027779,-12.6432,0.5877206876286275,23.77863333333333,0.23650178763712534,-2.8195333333333337,0.044463493140128055,43.55466666666667,0.6979551147618467,-12.9169,0.874429048007899,38.4219,1.1582641523705481,-9.260666666666667,0.660711153900771
1720000.0,37.927099999999996,3.3586369507088243,-8.965633333333335,1.1797222111817491,42.76823333333334,1.2822066300804347,-12.765366666666665,0.9407279781578141,23.505233333333337,0.8978689449035546,-2.7213333333333334,0.45912271658989917,42.9297,0.7832044305288369,-13.2453,0.3659616646590183,39.203133333333334,1.051045765997952,-9.7669,0.3037534526552745
1760000.0,38.3906,0.884887476839098,-9.167966666666667,0.08972537111777408,43.40366666666667,0.14953051713799026,-13.425766666666666,0.17224357430362633,23.289066666666667,0.18084509639160407,-2.486466666666667,0.08876415693034863,44.002633333333335,0.4509088772197274,-13.441699999999999,0.4636237770721718,37.71093333333334,0.34662650728926686,-8.708599999999999,0.2312127303300289
1800000.0,39.015633333333334,0.6921897347468291,-9.5206,0.36182473657836034,43.15886666666666,0.14966550110904683,-12.817966666666669,0.5788821545780187,23.210933333333333,0.7970598234901961,-2.446333333333333,0.20482796575554704,43.5625,0.8628871420991265,-13.154166666666667,0.37001933583110025,37.768233333333335,0.6342904819998134,-8.912233333333333,0.15627937235035919
1840000.0,38.309866666666665,2.4954888597004183,-9.076066666666666,0.8683892381236014,42.75,0.7592406864756415,-12.792966666666667,0.5723784897736428,22.9349,0.701176729980871,-2.3645666666666667,0.17620867048915492,43.1172,0.6862917892558517,-12.974466666666666,0.34503965311572854,38.65103333333334,1.9560595787336232,-9.184899999999999,0.7078225248370293
1880000.0,38.31249999999999,2.034762012292019,-9.438533333333334,0.7008609388142244,42.187533333333334,1.3171561014380766,-12.554699999999999,0.9635853914763682,22.39323333333333,0.7761738006972975,-2.1544333333333334,0.3668337528393785,42.96353333333334,0.4607131163268034,-12.951033333333333,0.4112302707189194,37.773466666666664,2.520273229014841,-8.9487,1.324202041482593
1920000.0,36.8125,1.3593024166829095,-8.513533333333333,0.5017664286188236,42.903633333333325,0.2986883585872657,-12.932566666666666,0.23600867686497332,21.4323,0.6766912491429645,-2.0317666666666665,0.19461953196486262,43.36716666666667,0.5690088126636442,-12.804933333333333,0.45226488796820014,37.166666666666664,0.4014480899056415,-8.598966666666668,0.46403196250066897
1960000.0,36.39843333333334,1.0336544888028214,-8.1021,0.750675660100064,42.671866666666666,0.9772586021906147,-12.9469,0.9756483690346642,23.10936666666667,0.5027158662925034,-2.541933333333333,0.1425420249922419,42.6328,0.1823313650107027,-12.820300000000001,0.3824264635194591,38.5078,1.671830256535236,-9.071333333333333,0.8143254277143889
2000000.0,35.16146666666666,2.5140259151850883,-7.5195333333333325,0.5382474606432328,42.6797,0.5212162762871717,-12.417699999999998,0.3226408839561411,21.54686666666667,0.2661838504158782,-1.9103999999999999,0.08774216014360861,41.90103333333334,0.7880725023047609,-12.107033333333334,0.3786842689570762,34.791666666666664,1.883079559893551,-7.3770999999999995,1.1838462822512053
2040000.0,35.46353333333333,1.986870855278612,-8.235933333333334,0.3583464462716986,43.440133333333335,0.6008987953243212,-13.022933333333333,0.6139490224947199,22.3125,0.9984520252203734,-2.208066666666667,0.17296532086583788,43.1953,1.0740870945443233,-12.862233333333334,0.8259902393827395,36.03126666666666,2.46247801659674,-7.7315,1.1029657504504236
2080000.0,34.763,2.725541931433088,-7.5747,0.784614674006717,42.5573,0.5473100461956355,-12.876033333333334,0.47476373305279135,21.7578,0.5846606366089645,-2.0424333333333333,0.18832869020825146,41.9427,0.3838886383662159,-12.158833333333334,0.9582742555703399,36.58593333333334,2.102423391126427,-7.635699999999999,1.2115261477436903
2120000.0,34.296866666666666,2.162157788465546,-7.543466666666666,0.9403708287455304,42.562466666666666,0.8430489757751658,-13.064066666666667,0.9076862171967186,21.39323333333333,0.6218569413905055,-1.7705666666666666,0.26049125811734175,42.669266666666665,0.5534098260381312,-12.462466666666666,0.2369553309990916,35.328133333333334,2.291101729639161,-7.591633333333334,1.593345859783396
2160000.0,32.8125,1.0068302869236045,-6.9559999999999995,0.6114223472090846,42.38283333333333,0.9570663729450672,-12.7802,0.3586770785353677,21.541666666666668,1.0282588044305219,-1.9565000000000001,0.21991372550767882,42.65886666666666,0.12536414514879857,-12.388,0.6682140824616015,35.7057,2.6863987529776727,-7.527066666666666,1.3579218468756669
2200000.0,31.744766666666667,2.369505985549638,-6.387766666666667,1.1283819221443698,42.0755,0.2648677531649816,-12.326533333333332,0.5950364545321758,21.218733333333333,0.55820344160729,-1.8557333333333332,0.15442159463264488,42.96613333333334,0.34308328758804574,-12.578666666666669,0.2599718232595386,35.8203,1.9591047462212572,-7.3882666666666665,0.9980338014761272
2240000.0,32.1875,2.049373427822919,-6.329933333333334,0.9452571231622056,42.500033333333334,0.8644152949955372,-12.525799999999998,0.4891908080357465,21.34636666666667,0.9041139099816028,-1.6994666666666667,0.22408750275035175,42.5625,1.5909727234200695,-12.354166666666666,0.7027572000500757,35.0,2.6295751798848923,-7.3416999999999994,1.7082952984383777
2280000.0,31.674466666666664,1.1059913873484224,-6.125766666666667,0.3593525319546562,42.66143333333333,0.7677302405286783,-13.356766666666667,0.5819836328359149,20.955699999999997,0.33671027110361007,-1.5453000000000001,0.08667252544299524,42.71093333333334,0.9799911983731747,-12.648966666666666,0.11050798865039332,33.5573,0.44678413132070904,-6.197666666666667,0.5277742088768225
2320000.0,32.5573,3.720380758291638,-6.856233333333333,1.3442343281172704,42.75,0.5533991205871821,-12.789866666666667,0.40525411232405245,21.289066666666667,0.10616510202928685,-1.7060000000000002,0.18377020070366867,43.59896666666666,0.12889883199194996,-12.654966666666667,0.3653586578813878,34.96873333333333,3.4013865509360866,-6.888533333333332,1.7110264839823164
2360000.0,32.453133333333334,3.5745564539145582,-6.6281,1.4565031479540302,41.57033333333333,0.9167702596010038,-12.359900000000001,0.29691483627464615,20.747400000000003,0.5987945613202128,-1.5705666666666669,0.24819326251039853,42.19533333333333,0.8032155432316358,-12.318,0.4773487264743317,34.3281,2.1852954964184286,-6.4604,1.5186309119291188
2400000.0,31.408833333333334,1.5798095588463272,-5.949466666666666,1.0974850776003997,43.1901,0.6989803621466525,-12.7539,0.8508300104407852,20.723933333333335,0.884387895037516,-1.5156333333333334,0.3029156354865529,42.53903333333333,0.6074042384515342,-12.226833333333333,0.3467435395536911,34.59373333333334,1.6337973606159204,-6.651833333333333,1.2463656240802246
2440000.0,30.304699999999997,2.548883372511711,-5.533066666666667,0.8435336837113001,43.03123333333334,0.47429765853195066,-12.8224,0.6701515798683161,21.2396,0.6941593044827676,-1.6645666666666667,0.33427627628787665,43.138000000000005,1.4584155854899532,-12.895533333333333,0.5467135833525837,32.7526,2.4871617974443616,-5.8867,1.5395885619216583
2480000.0,31.3776,3.275767162055325,-6.1026,1.0426102755424327,41.77343333333334,1.6793318717738765,-11.916166666666667,0.8410474950256313,21.385433333333335,0.03681669307377918,-1.8125,0.11861073588283094,42.9349,0.6498424629605732,-12.719266666666664,0.5827185560418985,32.21093333333334,3.695183908218313,-5.514833333333333,1.9355498965984372
2520000.0,30.804699999999997,3.1975695217878632,-5.661700000000001,1.1870001291771906,44.11976666666666,0.6157193426301376,-13.814566666666666,1.1372081144432429,21.23176666666667,0.799277058069075,-1.7244666666666666,0.2149231852442997,43.0625,0.3266696904621973,-12.670833333333334,0.17420505796968738,31.3672,2.5575697618377218,-5.033333333333334,1.5911656782232186
2560000.0,30.95573333333333,2.599964730957368,-5.827100000000001,0.7492318599739334,42.320299999999996,1.3390562646879336,-12.9935,0.925335625597545,20.968733333333333,0.3930446737402205,-1.7403666666666666,0.07866995755828403,42.882799999999996,0.9138430828101718,-12.472133333333334,0.30051445149203004,30.947933333333328,2.3226719417276485,-4.949466666666666,1.3305011219670413
2600000.0,29.10936666666667,2.365260952951187,-5.071866666666667,0.9837841102373811,42.8724,0.8576304137952805,-13.254166666666668,1.031247895620749,20.84633333333333,0.9167891845396567,-1.619,0.3001685637548787,42.958333333333336,0.3546117626676003,-12.666166666666667,0.40907796594564005,31.169233333333334,3.6164890509381564,-5.383300000000001,2.0317703134622938
2640000.0,29.138033333333336,2.3797016148155135,-4.942166666666666,0.7232545533136232,43.041666666666664,0.5459184696963032,-13.2318,0.6412793619008796,20.520833333333332,0.5069371843620154,-1.4958,0.09820919848296626,42.65363333333334,0.6692705498442982,-12.314333333333332,0.5952894272723326,31.739599999999996,3.7539885703963796,-5.581499999999999,2.1486563817108273
2680000.0,29.148433333333333,2.535628222398194,-4.924733333333333,1.0288019742507408,42.997366666666665,0.8750085193236055,-13.291400000000001,0.46729916185102083,20.364566666666665,0.9399934550599576,-1.5450666666666668,0.2421324752187438,42.4922,0.3955056763183024,-12.559099999999999,0.3646464132096552,31.328100000000003,2.8527882582951496,-5.1117,1.842330312403289
2720000.0,30.148433333333333,2.7765874190371806,-5.162233333333333,0.8879460806953439,42.583333333333336,0.6469864209462872,-13.106766666666667,0.3422016981580051,20.263,0.8168108267320329,-1.3187333333333333,0.21197075794133072,42.5078,0.554737938129348,-12.408333333333333,0.1913965574983579,29.825533333333336,3.052110793460087,-4.733333333333333,1.8523779389986506
2760000.0,30.968766666666667,2.993561134984367,-5.396599999999999,0.9228019325221782,42.708333333333336,0.5622398321791939,-13.166166666666667,0.553925442235284,20.843733333333333,0.5868010584706042,-1.5668999999999997,0.2527255164534572,42.1849,0.24071179170673587,-12.005466666666665,0.5025198260323225,30.932299999999998,2.2744598670160494,-4.990366666666667,1.5788641514990733
2800000.0,27.888,2.7830548647125157,-4.354166666666667,0.9944260734491808,42.76043333333333,1.4978877490950007,-13.604966666666668,1.7167835358276502,20.338533333333334,0.7080634450543403,-1.3479,0.21212793309698746,42.4323,0.7011985501030822,-12.378900000000002,0.267526497130035,30.421900000000004,2.588033358878256,-4.8398666666666665,1.5629083622813236
2840000.0,28.92186666666667,2.14388949704866,-4.5927,0.8037342885970876,42.77863333333334,1.5926969628763517,-13.197633333333334,1.2815353456780747,20.23436666666667,0.47737120659806054,-1.3656333333333333,0.05304025724757457,42.612,0.34681396550119825,-12.445333333333332,0.044892637357242214,31.151033333333334,2.908556712338423,-4.9198,1.693320715832257
2880000.0,27.4349,1.943147911680083,-4.161733333333333,0.7636246256433125,42.39846666666666,0.6052764951289247,-12.4094,0.6330883403338484,20.408866666666665,0.19563437552968352,-1.4645666666666666,0.07812427421885092,42.5677,0.3459928419298112,-12.6995,0.10404627175764969,29.8854,2.9262657705683535,-4.5549333333333335,1.545501015492675
2920000.0,28.39323333333333,0.6879603105477,-4.131266666666666,0.3287808726526259,43.3203,0.4489373304445376,-13.151299999999999,0.45112797149663264,19.687466666666666,0.6409538794293674,-1.2109333333333332,0.13755932861456138,42.270833333333336,0.7362214084242739,-12.472666666666667,0.8472618105140558,29.546900000000004,3.278864599217233,-4.1505,1.7084209922225455
2960000.0,27.6172,1.636186794552097,-3.9023333333333334,0.7003916348888115,43.044266666666665,0.7299143343951783,-12.884133333333333,0.8825178726550273,20.085933333333333,0.3030116866099747,-1.4116999999999997,0.06947234461760064,42.4323,1.3281278352126609,-12.306,0.5149745689513865,29.64323333333333,4.045972510479472,-4.6305000000000005,2.29834563110077
3000000.0,26.29426666666667,2.069667423095368,-3.5294000000000003,0.6934445471701395,42.8203,1.071708533137625,-12.537233333333333,0.5311584969563126,20.1224,1.071709525322355,-1.3674666666666668,0.33661840248104213,42.46093333333334,0.8680274278820639,-12.491666666666667,0.11508762844989978,29.1849,3.164483358570032,-4.497133333333333,1.6701186990417445
3040000.0,26.28646666666667,2.197456848773651,-3.3521,0.6860518493525106,42.604166666666664,1.2623102770537664,-13.202866666666667,0.9804477015912452,20.252599999999997,0.5788556008769953,-1.3588666666666667,0.16739259906645285,42.453133333333334,0.5102716422542894,-12.560933333333333,0.4121900640346504,29.203099999999996,3.3512250267626023,-4.1776,1.623091297493767
3080000.0,25.1146,1.5084701322863505,-3.1685000000000003,0.33011938244620936,42.53646666666666,0.9817252036876493,-12.877366666666667,0.8793045332660482,20.414066666666667,1.0220280535397375,-1.4364666666666668,0.304464123046086,42.02603333333334,0.5474686069132698,-11.8573,0.09151987033790336,28.67186666666667,3.509684842768018,-4.0617,1.8186140565459914
3120000.0,25.406266666666667,1.4274584741965546,-3.1,0.3982777171773486,42.45053333333333,1.6016645868873067,-12.545833333333333,1.1086426545806165,19.83073333333333,0.6887282595883196,-1.1885333333333332,0.1586752728758398,42.1302,0.5731210052568879,-11.973199999999999,0.4386168943394679,29.804666666666666,3.288182785200496,-4.429166666666666,1.8124029101965407
3160000.0,26.346333333333334,1.8861129010628055,-3.326033333333333,0.38469330930264717,43.333333333333336,1.1934307585928732,-13.024999999999999,0.8978832552175144,19.835933333333333,0.8552895701976554,-1.2325333333333333,0.27770440319799683,42.13803333333333,0.8490545578596372,-12.372666666666666,0.33367531457324723,29.161466666666666,4.420669343989538,-4.3486666666666665,2.0342160362710304
3200000.0,25.531266666666667,0.6897563353983946,-3.080966666666667,0.1807378457570215,42.416666666666664,0.7286753704878155,-12.990900000000002,0.2973632907180481,19.9349,0.7107175106890213,-1.4463666666666668,0.3818353077207793,42.41406666666666,0.45246784293349507,-12.6901,0.102959409477716,28.7526,4.402684341020449,-4.121333333333333,1.794488646445617
3240000.0,25.630200000000002,2.4416830383978985,-3.0927000000000002,0.7164723907218383,43.656233333333326,1.0193254708656885,-13.8297,0.7279306697756313,19.859333333333332,0.475141700501604,-1.3804666666666667,0.15061651820287034,42.796899999999994,0.25966913563224975,-12.954166666666666,0.5100900203776675,30.1146,5.474811531611536,-4.702833333333333,2.520076774403686
3280000.0,26.8672,2.1170395477332637,-3.3341,0.4983551411058852,42.6224,0.4401147804834545,-12.537766666666665,0.3602798572714764,20.221366666666665,0.4645765838074732,-1.3706000000000003,0.17963865582514996,41.580733333333335,0.2597518602221927,-12.1708,0.4277707797407395,29.273433333333333,4.676422354987007,-4.4177,2.2526578982763157
3320000.0,25.4922,1.7743310983767,-2.9841333333333337,0.5240247852492816,42.53126666666667,0.741751824324609,-12.721366666666668,0.5180177238495052,19.91146666666667,0.785537278272366,-1.2258,0.2076498976643138,41.91406666666666,0.8247355953408456,-12.383833333333333,0.5675726111158717,29.3828,3.32734604552437,-4.2687333333333335,1.3997364283638865
3360000.0,25.45053333333333,1.3229543059212416,-3.142433333333333,0.27866358849974565,43.03643333333334,0.2763336911939782,-12.609900000000001,0.5625878242550224,19.817700000000002,0.4629196906591901,-1.1979,0.17016347042378596,42.018233333333335,0.6570548142194003,-12.238266666666666,0.4470540036977886,28.656233333333333,3.281316457697362,-4.145333333333333,1.5835265839960575
3400000.0,26.156266666666667,1.4347030083687082,-3.237,0.39681008891743996,41.052099999999996,1.4608811747252637,-11.745566666666667,0.9880022986927824,19.48176666666667,0.46248627607265924,-1.1179666666666668,0.16793336641524093,41.257799999999996,0.33702213577152473,-11.937233333333333,0.3166291872557266,30.585933333333333,5.127837908471323,-4.6349,2.3174490472643985
3440000.0,24.51823333333333,0.9619724955643079,-2.850233333333333,0.18572474854531962,42.02086666666667,0.3019817360187345,-12.196599999999998,0.20129045349113484,19.72393333333333,0.27093567912370237,-1.3151,0.12396604373779141,42.29946666666667,0.7642140203430497,-12.4805,0.32812156283914073,28.098966666666666,3.901459796309865,-3.8768,1.8957730314219228
3480000.0,24.848966666666666,1.467863495772758,-2.7715666666666667,0.25997872135148964,41.9349,0.6800334452559432,-12.439300000000001,0.3056383156608471,19.604166666666668,0.7826369500321041,-1.3284,0.1347427425380182,41.916666666666664,0.2707187265205136,-12.4427,0.1752888093024387,29.114566666666665,4.654517987456443,-4.2672,2.181577898372338
3520000.0,24.651033333333334,1.1612105589522599,-2.709633333333333,0.14058146708897615,43.7005,0.689101622887848,-13.303633333333332,0.6543684809714541,19.838499999999996,0.6709702079824407,-1.3020666666666667,0.09927686314320956,42.46093333333334,0.06087738131322239,-12.558866666666667,0.19329966948296193,29.5104,4.086713049220199,-4.332566666666667,1.9224492647546143
3560000.0,26.458333333333332,2.488630714982749,-3.3851666666666667,0.7035947807904458,42.90626666666666,0.8453588442522812,-12.664833333333334,0.5720147803063212,20.460966666666668,0.41059162463720866,-1.4450333333333332,0.24904127279540544,42.0026,0.700837746319835,-12.039333333333332,0.4053579351086584,29.182299999999998,3.605377301568681,-4.315633333333333,2.153277694327624
3600000.0,25.218733333333333,1.2855881593868064,-2.9169,0.16941334067894412,43.3776,0.9784657514019945,-13.077566666666668,0.36921294066643395,19.6172,0.5524285293139727,-1.2924666666666667,0.1611413734030532,41.63803333333333,0.831850516752994,-12.400766666666668,0.3835210583817037,27.013033333333336,3.672148858033339,-3.5814999999999997,1.7649810499454852
3640000.0,24.76823333333333,1.363618735407943,-2.7276000000000002,0.25635823892878234,42.864599999999996,0.7790570753588401,-12.693466666666666,0.5656097260675614,19.4974,0.39284748524925966,-1.2783999999999998,0.1670364231737098,42.453133333333334,0.8698971369586662,-12.427333333333335,0.2096977083539275,28.3151,3.5476313590901745,-3.9974000000000003,2.066001521457975
3680000.0,24.557266666666663,1.2514462815301162,-2.695066666666667,0.22196735996287573,42.53903333333333,0.8455559762008007,-13.0008,0.7585927365853167,19.3828,0.254213964997992,-1.2164,0.18471303870237926,41.96353333333334,0.6317875187821371,-12.478366666666666,0.26892742928571317,27.239599999999996,3.953998105715275,-3.7205666666666666,2.0526876111305605
3720000.0,25.098966666666666,0.24885506268151694,-2.763033333333333,0.05919517622989975,43.28123333333334,1.154661775393799,-13.005733333333334,0.862422658690171,19.752599999999997,0.5794652074686334,-1.1585666666666667,0.1263166479746654,42.546866666666666,0.5627513917254103,-12.844033333333334,0.6821824014799042,27.5026,3.987330551467568,-3.801833333333333,2.185619272019311
3760000.0,24.317700000000002,0.6095343632642877,-2.5609333333333333,0.07646403657203082,43.53126666666666,1.2298618711962008,-13.816133333333333,0.9706487738735474,20.2422,0.44748712458200063,-1.5036333333333334,0.3131757476064979,41.937533333333334,0.4143296701366633,-12.5482,0.3831021882822736,27.898433333333333,3.890423313157008,-3.833833333333333,1.926547345786123
3800000.0,25.130233333333337,0.2594470701746739,-2.9195666666666664,0.09543131328633986,43.390633333333334,0.7287530827074111,-13.737733333333333,0.7877759086327929,19.484366666666663,0.4538305581993753,-1.1249666666666667,0.09607172089410887,42.16146666666666,0.4239659603736552,-12.646366666666667,0.490194140678523,28.625,4.2338885483677995,-4.035166666666666,2.166560783566639
3840000.0,25.21616666666667,0.368363410536691,-2.752333333333333,0.08149847987675729,42.97396666666666,0.9262648661275146,-13.390366666666667,0.4497016813646828,19.705699999999997,0.22974604820685487,-1.3278666666666668,0.09476927537738983,42.169266666666665,1.1841628275799847,-12.4013,0.768235410968973,27.15363333333333,3.1805904004689993,-3.510666666666667,1.615138651083003
3880000.0,25.484400000000004,1.5123253816556803,-2.8684999999999996,0.3767640729510533,43.6172,0.5040624035441118,-13.102833333333335,0.24767896335556874,19.528633333333335,0.6051808838869771,-1.0729333333333333,0.14617173309349366,42.34896666666666,1.1694395533283832,-12.649766666666666,0.4849296054297182,26.8828,3.512854213693854,-3.6565,1.8834292199779281
3920000.0,24.083333333333332,0.2170931801989385,-2.5223666666666666,0.06616415612365621,44.268233333333335,0.42465404219853603,-13.773433333333335,0.4396075320353623,20.585966666666668,0.691133061637836,-1.5255333333333334,0.17935031518108785,42.104166666666664,0.9879050572915498,-12.354433333333333,0.4508832393819443,27.934866666666665,3.9684035693070436,-3.7880000000000003,1.7686282952239192
3960000.0,25.29426666666667,0.5625474518256709,-2.8117,0.36029987510405814,42.7526,0.6663559009018117,-12.510666666666665,0.5543425254800098,20.049500000000002,0.22120492761238456,-1.1859333333333335,0.09458295594638369,40.987,0.5484789391277175,-11.668266666666666,0.4729348463466074,29.8099,4.308613267243496,-4.528933333333334,2.380060792407529
4000000.0,24.97136666666667,0.9050118464540791,-2.682,0.3014923658513871,42.96613333333334,0.7200245196442184,-12.542733333333333,0.10749872350663287,19.736966666666664,0.12541399000466089,-1.2221333333333335,0.10643888178459761,42.263,1.0661430704491155,-12.717466666666667,0.26321864591163635,29.5729,3.408024343614151,-4.2773666666666665,1.44943831956459
}\dataSotaStepsPp

%% file: data_new/sota_steps_lj.tex
\pgfplotstableread[col sep=comma]{
0.0,39.919266666666665,0.11417417493558639,-19.622,0.21644234028180862,39.5469,0.3510219461325273,-18.988266666666664,0.32492955475849716,39.830733333333335,0.23937921565768736,-19.654133333333334,0.2422347942710857,39.8073,0.1504601608400033,-19.09936666666667,0.20014828391858744,39.7604,0.3388455695445943,-19.52253333333333,0.31819482850745645
40000.0,39.84896666666666,0.13520528424913278,-19.224866666666667,0.28101139162358274,39.51820000000001,0.09576401551035092,-19.02253333333333,0.18028533557176127,39.91146666666666,0.12520504072209762,-19.500366666666668,0.2222150659958848,39.90626666666667,0.10541553754335807,-19.275766666666666,0.280148357521907,39.669266666666665,0.08935861582534785,-19.2599,0.08823162698261905
80000.0,39.45053333333333,0.31095475912457304,-18.855733333333337,0.34794764293241376,39.92706666666667,0.051571654574539265,-19.133066666666668,0.16742294413318096,39.447900000000004,0.051163333224748496,-19.203,0.12702419716993568,39.52343333333334,0.2878877597645009,-18.812866666666665,0.2628263854503362,39.59896666666666,0.05196340592720639,-18.74506666666667,0.19269672775864446
120000.0,39.34116666666666,0.20810231991872447,-18.568866666666665,0.3163385352575444,39.859366666666666,0.055614466543237706,-19.0423,0.03547196075775868,39.64843333333334,0.25202500316877674,-19.0307,0.3837876235628242,39.67706666666667,0.07390829152101591,-18.940233333333335,0.12071736504009045,39.0703,0.22918517985826872,-18.372666666666664,0.21456040848415792
160000.0,39.596333333333334,0.3996222994556528,-18.839233333333333,0.33908311993112095,39.744800000000005,0.12376254145203455,-18.959233333333334,0.12165745170582697,39.2448,0.07979289859797445,-18.569,0.0783852452102223,39.953133333333334,0.044674029243944466,-19.057433333333332,0.07549416460156912,39.22656666666666,0.4754655075700974,-18.429566666666663,0.33533579521959134
200000.0,38.96616666666666,0.1399600022228577,-18.17186666666667,0.120789964630989,39.59113333333334,0.3141011443609976,-18.872933333333332,0.22212969684898562,39.28123333333333,0.26583190595227973,-18.60636666666667,0.39496773145269576,39.645833333333336,0.20095124339556106,-18.848566666666667,0.19370392068538114,38.28906666666666,0.27089610964763305,-17.661333333333335,0.2992770659810434
240000.0,38.515633333333334,0.7354676713191084,-17.797666666666668,0.5370020877261302,39.6198,0.3433353268550536,-18.800666666666668,0.2955168278720451,39.604166666666664,0.32215383833745476,-18.740366666666667,0.5063771475447474,39.695299999999996,0.12806579038395263,-18.8018,0.12607714569527062,37.69533333333333,0.5700861708743901,-17.237266666666667,0.3639501919524456
280000.0,38.54423333333333,0.7868027805972434,-17.614833333333333,0.5585012941395523,39.72396666666666,0.12473278995071434,-18.919,0.20810402847294154,39.0703,1.1425851594811933,-18.415733333333332,1.0752147857779653,39.8073,0.10092465836784623,-18.8534,0.09950209378031528,38.46876666666666,0.7661825776028989,-17.794933333333333,0.6169737019427071
320000.0,36.78126666666666,1.5352294168046108,-16.438433333333332,1.0606585889070153,39.66146666666667,0.24942590527493672,-18.903000000000002,0.1745081277954315,38.59373333333333,0.3701605267388109,-17.957166666666666,0.29178476466205183,39.63283333333333,0.11498273300321728,-18.812633333333334,0.20910222911825305,37.0703,0.696753820704752,-16.6026,0.4072241888689828
360000.0,37.578133333333334,1.110248072379423,-17.089199999999998,0.7904904089656414,39.705733333333335,0.13448227475106958,-19.0401,0.10248730002623012,37.5417,1.2811550439609827,-17.081,0.9034928813591538,39.6953,0.1980628856365242,-18.818266666666663,0.1857007329609181,38.11716666666667,0.3904675516465979,-17.494,0.2675564364142016
400000.0,37.833333333333336,1.0262745614871098,-17.2901,0.8005058026023301,39.708333333333336,0.06818725850349393,-18.9004,0.20166844076354645,37.69006666666667,1.416360260981961,-17.106666666666666,1.0709989801841804,39.729166666666664,0.11345840157911803,-18.953633333333332,0.07393611356359538,36.83593333333334,0.20462411609800368,-16.625,0.24296297385952997
440000.0,37.8151,1.1287071926175847,-17.220566666666667,0.6796955315106575,39.539033333333336,0.33150569963258175,-18.830866666666665,0.3553313039341684,37.421866666666666,0.10126928897197283,-17.058966666666667,0.04828252501912297,39.6875,0.20161990642460723,-18.854166666666668,0.2184387073961224,36.875,0.5282999589879468,-16.519499999999997,0.4636362582887588
480000.0,37.3333,1.5150140879432985,-16.82003333333333,1.031750723554656,39.283833333333334,0.3891713789864588,-18.4931,0.19592581929563768,37.66143333333334,0.7837371384732409,-17.153000000000002,0.6137452077205986,39.41146666666666,0.04826954411312366,-18.7419,0.16522064035706993,36.1719,1.269937560144854,-16.19846666666667,0.9226611127000475
520000.0,37.57033333333334,0.3005891585240919,-17.113033333333334,0.19215024214284984,39.458333333333336,0.5504102005676211,-18.586066666666667,0.4420873619043582,36.4948,0.9296901670269884,-16.316666666666666,0.549002716042664,39.330733333333335,0.2867944250194238,-18.651733333333333,0.1873398397446618,36.05466666666667,1.525961424443255,-15.926266666666669,0.9195290618330422
560000.0,38.03126666666667,0.9833704535366571,-17.219933333333334,0.7661078006535528,39.48956666666667,0.18222528791459164,-18.67343333333333,0.3116596576751977,36.21093333333334,1.1916588363379101,-16.045866666666665,0.7946951001624605,39.77343333333334,0.027807233271618187,-18.8142,0.10893175233450833,36.153666666666666,0.2641026231515989,-16.017033333333334,0.08024505523014389
600000.0,37.77343333333334,1.574501305881397,-17.275533333333332,1.2114690595397886,39.4323,0.1832559048616624,-18.561333333333334,0.18772919029519236,35.268233333333335,0.6667532993710806,-15.484399999999999,0.44255513404169977,39.674499999999995,0.08707548449477717,-18.9375,0.18460662682218787,36.580733333333335,0.9161088739275953,-16.358733333333333,0.6994442667019461
640000.0,36.825500000000005,0.8886834344504607,-16.376433333333335,0.6141755684565205,39.645833333333336,0.12209871234192209,-18.760433333333335,0.21972125270189197,35.330733333333335,0.4053252301821603,-15.431133333333335,0.27896095704516705,39.70053333333333,0.21197028617762084,-18.9311,0.3815510538141217,35.117200000000004,0.8713866994624139,-15.376166666666668,0.46451221966943174
680000.0,36.41403333333333,0.6402720064333778,-16.181900000000002,0.527540791471775,39.4792,0.17694136505256988,-18.641133333333332,0.09412255604027919,34.84896666666666,0.5422068629427534,-15.292700000000002,0.4171807601827614,39.67966666666667,0.11284178698021888,-18.7858,0.18242063114315335,35.74476666666666,0.6461439278949798,-15.654166666666667,0.40301502315532695
720000.0,35.328133333333334,0.9063905425121985,-15.390633333333334,0.5794568107767439,39.27863333333333,0.26457148162432215,-18.350266666666666,0.30094617828147047,34.080733333333335,0.8706253742122537,-14.651033333333332,0.514281611138835,39.578133333333334,0.17125741508682882,-18.6918,0.13593596531700813,33.8073,1.7365629905073978,-14.381,1.0682561521782439
760000.0,35.93226666666667,0.934509826355806,-15.889466666666666,0.6913259449942711,39.45053333333333,0.4141987070101608,-18.60466666666667,0.45854983250339165,34.544266666666665,1.8718512624909296,-15.0035,1.1854246946418263,39.23696666666667,0.2533218550031226,-18.319266666666667,0.33311289711180897,34.5052,1.8434896473807492,-14.89076666666667,1.1355890581054795
800000.0,36.2552,1.9865690574455246,-16.182566666666666,1.3590144721656041,39.3568,0.6667120267901792,-18.56103333333333,0.6770964694110355,33.69793333333333,1.5521479575807906,-14.463933333333335,0.9326828161575378,39.5182,0.18056599901421067,-18.562499999999996,0.23483340193990013,33.03903333333333,1.5442049979052501,-13.814466666666666,1.0007735619120957
840000.0,35.72396666666666,1.8159118211582366,-15.735666666666667,1.1380062809824718,39.6406,0.2505422519256992,-18.88423333333333,0.1861803844543121,32.955733333333335,1.7205733701944317,-14.0224,1.1681097237274698,39.5,0.3515415290782398,-18.720933333333335,0.4087785246587958,32.94013333333333,1.2416643946287929,-13.9824,0.8699750034723223
880000.0,35.84636666666666,1.0918087938014724,-15.863433333333333,0.7818709200088949,39.434900000000006,0.24222569365504276,-18.67656666666667,0.0796717153211994,34.291666666666664,0.6115143189020372,-14.850499999999998,0.4828308261354766,39.4609,0.2651650429449553,-18.764466666666667,0.2554991106746871,33.31773333333334,0.9251879857028474,-13.964566666666665,0.6305161183947284
920000.0,35.994800000000005,0.9380472944722279,-15.853033333333334,0.6505846823349666,39.606766666666665,0.03736346997923211,-18.650133333333333,0.09472716376813735,34.84376666666666,2.1818489060631303,-15.381133333333333,1.5879208215637066,39.26303333333333,0.2855730651786945,-18.389333333333337,0.24499253503366544,32.96353333333334,2.217700500267988,-13.7293,1.3900681709901854
960000.0,34.976533333333336,1.5469089012895656,-15.2712,0.9632296749304745,39.42446666666667,0.12778801021831268,-18.6651,0.18947986700438635,34.14843333333334,1.48601188271008,-14.634099999999998,1.0157342795567483,39.515633333333334,0.22432233256831507,-18.457399999999996,0.2348479933914703,32.77346666666667,0.8742960342787544,-13.5085,0.6429286948539991
1000000.0,34.8229,0.7405541753758921,-15.1113,0.4166165223159859,39.70053333333333,0.12391648621371973,-18.824366666666666,0.07735547096934323,32.27343333333334,1.68053560972553,-13.468366666666666,1.1414683886215262,39.5521,0.0321360648908159,-18.657666666666668,0.040707192921590404,31.9141,1.2272814374326138,-13.0771,0.7980364068553931
1040000.0,34.0755,2.275831946051084,-14.737766666666667,1.531700167207089,39.375,0.34661824341293335,-18.666666666666668,0.25035177029309896,33.02863333333333,1.7359938831177444,-13.969166666666666,1.122927202547976,38.953133333333334,0.5880744642947472,-18.1375,0.5205028402099908,31.479133333333333,0.7366466196367316,-12.673466666666668,0.4663509503462918
1080000.0,34.01303333333333,1.2039111161358875,-14.639066666666665,0.8259560332662318,39.46353333333334,0.24294243945611443,-18.714966666666665,0.21819432215853338,32.63543333333333,1.8940321404054605,-13.744166666666667,1.3258937019568682,39.01303333333333,0.47121804106190907,-18.236733333333333,0.43183720955419796,32.0104,2.127202958503646,-13.016933333333332,1.4205147947917407
1120000.0,33.91146666666666,0.46039470264352744,-14.5276,0.48382220563618955,39.36196666666667,0.12173184555498268,-18.4405,0.0192947315779896,32.015633333333334,1.8084933625043298,-13.283733333333332,1.3564512679127922,39.15626666666666,0.2237885509930223,-18.389833333333332,0.27125167075778356,31.013033333333336,0.8470196429573262,-12.4181,0.7425990349217179
1160000.0,33.4401,1.1047392482693217,-14.309600000000001,0.7283754846689084,39.044266666666665,0.3540961953418252,-18.33843333333333,0.3480782894056389,32.184866666666665,2.2531223273986307,-13.168233333333333,1.567816571612324,39.28126666666667,0.613966781584223,-18.381766666666667,0.6887669384890336,31.971366666666665,1.2878057470838606,-12.9116,0.9920983855780976
1200000.0,32.875,2.538243002551175,-13.869133333333332,1.5752275228959427,39.3125,0.3923653739394778,-18.522033333333333,0.23877546402891112,31.182299999999998,1.2280568173609332,-12.555733333333334,0.9913933976423733,39.080733333333335,0.6934100149903302,-18.08413333333333,0.5723431158162228,30.78643333333333,1.300313322071091,-12.2095,1.0024407713176875
1240000.0,34.1849,2.2689882605837046,-14.725766666666665,1.4779574516503802,39.479166666666664,0.03511546034947609,-18.632033333333336,0.20184590381994014,31.864566666666665,1.2482974868017451,-12.643533333333332,1.1397307645034225,39.5026,0.3479023522005363,-18.621333333333336,0.26612846939442997,30.5,1.1214416822406168,-11.842466666666667,0.8303255198347742
1280000.0,33.218733333333326,1.9109816279133147,-14.140866666666668,1.3463323174049147,39.46613333333334,0.3923940224938316,-18.577066666666667,0.4596816386249172,30.442700000000002,1.547107320130054,-11.8423,1.2738804731998996,39.375,0.01684992581586173,-18.364833333333333,0.043929817765260074,31.5703,1.5147413662624627,-12.5122,1.2327239674801496
1320000.0,31.734400000000004,3.299431858163866,-13.288533333333334,2.0901833991834837,38.520833333333336,0.8906815530193087,-17.8677,0.7244376991846854,29.9375,1.8223230942947513,-11.5664,1.348694549060931,39.04163333333333,0.026560664315654733,-18.281499999999998,0.08192122232161031,31.098966666666666,1.1067218570575377,-12.302766666666669,1.0325957787160585
1360000.0,32.34636666666666,2.2348844747672203,-13.414833333333334,1.6567194679714357,39.1146,0.37095962942976135,-18.352999999999998,0.4358070521075426,30.75,2.081412718964374,-12.074599999999998,1.6637930780799237,39.40623333333333,0.1796522628734634,-18.387633333333333,0.20087512428261403,31.364599999999996,1.4473959122046267,-12.4341,1.2771842545224241
1400000.0,31.276033333333334,2.328357412903403,-12.805966666666665,1.6270337229312593,39.3724,0.4838054223204464,-18.49116666666667,0.4276960473150162,30.76823333333333,1.2325092706435195,-11.921066666666666,1.0786819745514526,38.734399999999994,0.572939071804323,-17.744033333333334,0.4693239984867126,31.27863333333333,1.5732017127148359,-12.215633333333335,1.2958459536362936
1440000.0,30.90363333333333,1.8098278856166274,-12.453766666666667,1.3912892590048351,39.1875,0.3161724845713185,-18.437766666666665,0.2240855838488692,30.294300000000003,1.2031160625642072,-11.754833333333332,1.2138610857736378,37.78386666666666,0.8314970007295415,-17.305733333333333,0.6233393208268587,31.9141,1.8693701416965736,-12.793500000000002,1.3742985483511212
1480000.0,32.00783333333333,3.944956968361278,-13.374733333333333,2.8157089815690988,39.59636666666666,0.035115460349475915,-18.79256666666667,0.1444123802018214,30.325533333333336,1.0574404453942332,-11.646866666666668,1.1280180179806032,39.3151,0.3190148690369583,-18.26393333333333,0.1866644999874263,30.6849,1.1464350221447355,-12.0146,0.76570502588573
1520000.0,31.66926666666667,2.803900940158591,-13.137100000000002,1.8216559664217606,38.731766666666665,0.1698804743210824,-17.883466666666667,0.3203708302715602,30.255200000000002,0.9446693813181408,-11.666400000000001,0.9360416052006806,39.10153333333334,0.1504204182357645,-18.276566666666668,0.3347339872528969,31.736966666666664,1.2008217223588566,-12.637233333333333,1.0064094638311427
1560000.0,30.296866666666663,3.531960889112762,-12.266533333333333,2.2867162079181482,39.14843333333334,0.23861157185313167,-18.364833333333333,0.18079399572134358,29.1354,1.1348274083166423,-11.006133333333333,1.1405350099356393,38.971333333333334,0.30594468272694,-18.196233333333335,0.39048036513447704,31.20573333333333,1.354386464623579,-12.265366666666665,1.0779285391692506
1600000.0,30.0755,2.9785871852272523,-12.047766666666666,1.9838422809174012,39.13283333333333,0.3133358616919263,-18.337866666666667,0.14474940030580075,30.083333333333332,0.838812340289663,-11.331499999999998,0.8963677035681283,39.479166666666664,0.1781908402683938,-18.628666666666668,0.1562599827922107,30.885433333333335,1.951612848105097,-12.243133333333333,1.5150172811628995
1640000.0,30.335966666666664,2.7309654218893997,-12.229966666666668,1.8062622111851747,39.51303333333333,0.22576235785051144,-18.680966666666666,0.2736625212840725,29.411466666666666,0.6563501419889312,-11.1888,0.6113283460356369,38.85153333333333,0.8456469804567109,-18.024200000000004,0.7238360633918893,30.41143333333333,1.0382141536739373,-11.863533333333331,0.8694381416881952
1680000.0,29.8151,2.361862067663281,-11.876266666666666,1.5819389080773274,39.239599999999996,0.27849830639820294,-18.4896,0.1916135694568629,28.783833333333334,0.9486317245848825,-10.7035,0.9502240156931417,39.265633333333334,0.5451242814461866,-18.188533333333336,0.520604426497593,30.481800000000003,2.0005235914629953,-11.806633333333332,1.5023171177292172
1720000.0,29.25,2.186396135195999,-11.600266666666668,1.334606844313669,39.09636666666666,0.31309790091208733,-18.19243333333333,0.3274333452102204,29.052066666666665,0.4871994275676259,-10.900133333333335,0.5811202304361998,38.7526,0.5138898001193098,-17.875633333333333,0.6396236723435279,30.997433333333333,1.2248903633477661,-12.137633333333333,0.9961098544950865
1760000.0,29.35676666666667,2.417183493701342,-11.579566666666667,1.5162487424195576,39.05206666666667,0.4542826457418583,-18.23176666666667,0.5647872185365229,29.3151,1.7997968940966653,-10.856766666666667,1.3545793106676658,39.02086666666666,0.4107515821299068,-18.1965,0.4927864513830167,31.4401,1.9649024522012972,-12.228233333333336,1.240681088040848
1800000.0,29.76823333333333,1.1169047865517558,-11.832799999999999,0.7538314267792239,38.79946666666667,1.291096219841451,-18.00766666666667,1.0777031914627,28.79686666666667,0.5872794356654725,-10.694,0.6341811150347093,39.166666666666664,0.503891260315381,-18.2655,0.6104917362258074,30.263,0.7429366908873645,-11.545566666666666,0.6295175736669756
1840000.0,28.9974,2.900282063294304,-11.455733333333333,1.7756003629445696,38.89843333333334,0.7946812540607435,-18.083833333333335,0.7523151925150039,29.601533333333332,1.359927354268929,-10.9492,0.7519572195278134,38.895833333333336,0.8731288157476453,-18.180466666666664,0.7844345791006969,30.88803333333333,1.647220222745647,-11.983199999999998,0.9712637163338631
1880000.0,28.9479,2.28836354774906,-11.324866666666667,1.3810362204599202,38.96093333333334,0.7789602699895692,-18.098066666666664,0.8198142079493117,30.354166666666668,2.72834387658317,-11.568733333333332,2.246254014032152,38.7474,0.5292848256531335,-17.8504,0.39712544617538725,30.4505,1.120563813443928,-11.8112,0.8909363202085024
1920000.0,29.0,1.9848804968225837,-11.416533333333334,1.2204166783885282,39.22656666666666,0.7257281278519905,-18.536166666666666,0.5473526975868079,30.664066666666667,2.275315468725648,-11.736733333333333,2.0041024496988395,38.82033333333333,0.7755107105786636,-17.859366666666663,0.501682106074708,30.557299999999998,1.7453385402265091,-11.764866666666668,1.2352532057121817
1960000.0,29.203100000000003,1.5090956651805298,-11.548566666666668,0.8799659740896553,39.45570000000001,0.27725574235111244,-18.8668,0.33538617542568283,30.5651,1.544832588556658,-11.7104,1.67621329390584,37.9401,0.7163745016865611,-17.213400000000004,0.6299517812235061,30.968733333333333,1.7578591569166038,-12.12876666666667,1.1642198799577723
2000000.0,28.567733333333337,2.0481643575542354,-11.208466666666666,1.17799126293685,39.23696666666667,0.20025266262621388,-18.486700000000003,0.14030345683553294,29.942666666666668,1.2274334885804972,-11.299466666666666,1.2287750874572432,38.065133333333335,0.7879022203858083,-17.354566666666667,0.5244011462823311,31.20573333333333,1.931519533654497,-12.106866666666667,1.3108652672524697
2040000.0,27.47136666666667,2.214867265147557,-10.666033333333333,1.2448824772733451,39.427099999999996,0.5248798021134604,-18.754166666666666,0.49743569321962566,29.460933333333333,0.2604898249239082,-10.7815,0.6131640563503379,37.606766666666665,1.1526888575076208,-16.970433333333332,0.8242950698762073,30.296900000000004,1.2426218974410523,-11.6599,0.7000538026942403
2080000.0,28.57813333333333,1.895840756908542,-11.181899999999999,0.951930631226176,38.6901,0.4079813802940841,-18.043499999999998,0.5287413230178512,29.8802,0.9275698284585719,-11.069400000000002,0.7651570819119426,38.778666666666666,0.6670818856948685,-17.792733333333334,0.7370068942840512,31.17186666666667,1.3260942810457423,-12.1693,1.1275064286586869
2120000.0,26.648433333333333,2.855819294392105,-10.034766666666668,1.5701777931884724,38.236999999999995,1.6769003071142896,-17.78633333333333,1.2672252058029079,28.484400000000004,0.5625914858936273,-10.102733333333333,0.7335668673603579,38.3177,0.7384939313675274,-17.610566666666667,0.6559711595963824,30.3776,0.7259508706976429,-11.653,0.6671604454702033
2160000.0,28.6276,2.1010024527988223,-11.088033333333334,1.1335695283287903,38.697900000000004,0.841679158983199,-18.042833333333334,0.7186687152104387,29.0,0.7895334613977217,-10.366033333333334,0.9052881026992948,38.90886666666666,0.7383352144445563,-17.888299999999997,0.651333570453728,30.630166666666668,0.9980675338984943,-11.812266666666666,0.8934071126249717
2200000.0,27.20573333333333,2.3798646590836947,-10.3397,1.3003000986951687,38.38023333333333,1.3164441812026157,-17.943233333333335,0.9809991176799736,28.325533333333336,0.6654013392098201,-9.978399999999999,0.9987933953859862,38.859366666666666,0.34766754554059637,-17.8659,0.31878934528410285,31.151033333333334,1.5402746602112518,-12.1203,1.1208037027062319
2240000.0,27.130233333333337,1.6764614725334095,-10.323066666666668,0.8631357186960162,39.21616666666666,0.6011691627339373,-18.349966666666663,0.633527780676498,28.8828,0.6780400774782166,-10.201833333333333,0.7342833301178022,38.786433333333335,0.5376846432208708,-17.828233333333333,0.5998854686984468,31.020833333333332,1.0719482584942668,-11.987233333333334,0.7742032133461829
2280000.0,28.567700000000002,1.4560204279702487,-11.2129,0.5997493865496377,38.48956666666667,0.42561482848019294,-17.7897,0.2946243823356561,28.9505,0.23946101979236592,-10.0992,0.4146673445867982,38.106766666666665,0.7329422639077541,-17.23266666666667,0.5663911624389004,30.828133333333337,1.2925927544624756,-12.025766666666668,0.8958773924049106
2320000.0,27.934866666666665,2.831435378429503,-10.8609,1.533442795368209,39.27603333333334,0.52705519213414,-18.528766666666666,0.4408055756856481,29.3828,0.7342745944127437,-10.514566666666667,0.23301715721284458,38.1667,0.48619260243926654,-17.352833333333333,0.44668521603274736,30.28386666666667,0.19663781144248121,-11.690633333333333,0.15294670822072506
2360000.0,27.8125,1.785931056900013,-10.673933333333332,0.9607349594046334,38.583333333333336,0.6425077189333134,-17.901166666666665,0.4518033520115683,28.729166666666668,0.487507034707078,-9.958033333333333,0.36050853280085104,38.7448,0.6162324453212982,-17.699466666666666,0.48035599530163225,30.4245,1.3698201122775209,-11.882533333333333,1.0274020612961396
2400000.0,26.1328,1.802163684759701,-9.855466666666667,0.9458635889433997,38.677099999999996,0.7625606904808729,-17.950366666666667,0.7197007819611961,29.1901,1.6548719044083138,-10.251966666666666,1.3573362549084473,38.40366666666666,0.6220337950804788,-17.712366666666668,0.3845792968368889,30.804699999999997,0.7191579010667031,-11.968766666666667,0.6758965667089076
2440000.0,26.414066666666667,2.584904079888115,-9.9362,1.3239417383958656,37.971333333333334,0.9602506524629699,-17.472,0.8345438514541942,29.125,0.5058845784036783,-10.107266666666668,0.26409392689386557,38.24736666666667,0.7805460581470454,-17.555333333333333,0.5766445024642329,31.065133333333335,1.7631734086268664,-12.039700000000002,1.167405339488674
2480000.0,27.179699999999997,1.4819181646321309,-10.300933333333333,0.7965295404998411,38.611999999999995,0.09763267895535836,-17.886966666666666,0.2790010075648865,29.138033333333336,0.7375445493136136,-10.0522,0.7168918886415161,38.356766666666665,0.23283144594796784,-17.766000000000002,0.18740844876009943,31.031233333333333,1.3181273214509877,-12.146066666666668,0.8768653273007327
2520000.0,27.330699999999997,2.5303029950317546,-10.483600000000001,1.3305377784940946,38.625,0.9120585763352402,-18.029833333333332,0.6031947741447674,29.249966666666666,0.2530073033640646,-10.226933333333333,0.3933137622249641,38.65363333333334,0.30393302259251365,-17.818866666666665,0.2646084066859718,29.77606666666667,0.11238082082316785,-11.429033333333331,0.18488600692198298
2560000.0,26.619833333333336,3.111592289416394,-10.0116,1.6265548930177554,38.94533333333333,0.41123448569182813,-18.189300000000003,0.3525734627941632,28.507833333333334,0.46140297884700415,-9.867199999999999,0.502120191454861,38.502633333333335,0.6790037031481431,-17.559633333333334,0.45676802524792426,30.156233333333333,0.15678159614217801,-11.5871,0.2958056907273195
2600000.0,26.458333333333332,1.2466036240744518,-9.912333333333333,0.5001723791743092,38.416666666666664,0.7063553276424619,-17.93386666666667,0.6463092156407981,28.7526,0.39535837750914904,-9.9814,0.11870588303309433,38.07553333333333,0.41625253018917324,-17.300766666666664,0.34714654286370605,30.781233333333336,2.3074549274519383,-11.911999999999999,1.5630194112678188
2640000.0,26.854200000000002,2.0332691033571204,-10.192466666666668,1.054180750894056,38.296866666666666,1.3921813156178868,-17.7811,1.2394436574528098,28.72396666666667,0.9343357367075761,-9.996733333333333,0.6152737999369785,38.14846666666667,1.405759752668366,-17.5914,1.1608838787751339,30.325533333333336,0.18623992291903746,-11.692333333333332,0.31928074723596367
2680000.0,26.395833333333332,1.741297807830572,-9.891133333333334,0.8967716035250496,38.520833333333336,0.9560018177574513,-17.79086666666667,0.831098150374284,29.34636666666667,0.4790726133780651,-10.287633333333334,0.21316049248290725,37.5625,0.21234737263895423,-17.035433333333334,0.2626253901578359,29.6354,0.4394260195603657,-11.3401,0.4995254548068596
2720000.0,26.093733333333333,1.6373114554727273,-9.757933333333332,0.7693070489444092,37.07553333333333,2.2448241153571216,-16.95896666666667,1.6624755924691217,28.755200000000002,1.2177697894101336,-9.989833333333333,0.8803486064559253,38.57293333333333,0.5172823750675793,-17.736733333333333,0.297087824650482,30.755200000000002,1.2845117697657218,-11.856366666666666,0.7549312080506998
2760000.0,26.16926666666667,2.635695608795184,-9.857566666666669,1.2601330732197389,37.63543333333333,1.5544734463977048,-17.364433333333334,1.15642949729857,28.729166666666668,0.9577087982378691,-10.0548,0.3406778634820089,38.1198,0.1457679205678231,-17.406233333333333,0.21738888860493497,31.072933333333328,1.1322112916275324,-12.100666666666667,0.7338311083318529
2800000.0,25.72136666666667,1.4724123297802452,-9.564933333333334,0.6383454098074353,38.2344,0.7303770167979463,-17.678266666666662,0.6456166578465039,28.656233333333333,0.6486346343581184,-10.0289,0.5913738298797695,38.46613333333334,0.5408647910727987,-17.780633333333334,0.40198824471480266,30.281266666666667,1.3917677424372534,-11.747533333333335,0.8883500861459719
2840000.0,26.166700000000002,1.6878872948156223,-9.801433333333334,0.8154178656097473,38.5729,1.309626086586041,-17.922666666666668,1.104577253472516,28.67186666666667,0.8955307153985418,-9.830066666666667,0.46586526187538635,38.97136666666666,0.2185851067408037,-18.13553333333333,0.3297830229442108,31.546866666666663,1.0373289588596724,-12.3453,0.7226232951314716
2880000.0,25.763033333333336,2.5947473150364537,-9.494,1.2440320360290835,37.84896666666666,1.2792103432282822,-17.432033333333333,0.9253560840142689,29.541666666666668,0.812420442600724,-10.329166666666667,0.7484853965769068,38.83593333333334,0.539758398627468,-17.874733333333335,0.4914672883881031,29.955699999999997,2.123170040293523,-11.537766666666665,1.167710755662073
2920000.0,25.953166666666664,2.4954168928034632,-9.7004,1.3112464019651937,38.15103333333334,1.6005169046974244,-17.680333333333333,1.163542184691021,29.380200000000002,0.2725933601539109,-10.438,0.27922766099845203,38.796866666666666,0.37255636113509205,-17.898966666666666,0.28693183085108553,30.023433333333333,1.733719808837506,-11.5969,1.0995386578015345
2960000.0,25.5677,1.7125604398093524,-9.502366666666667,0.8120607544321345,38.16143333333333,1.3321087151163342,-17.625266666666665,0.8470000091827361,27.500033333333334,1.1257748778310677,-9.449333333333334,0.3796972683014134,38.63023333333334,0.8679873130153226,-17.787233333333337,0.7713289498578992,29.73436666666667,1.0170004468479297,-11.491766666666669,0.6111708235473582
3000000.0,25.3099,1.4501932721767345,-9.343733333333333,0.663457973214749,38.734366666666666,0.8421809874104004,-17.988133333333334,0.8229156997462681,28.73696666666667,0.44692222539895704,-10.097,0.12600574060996805,38.749966666666666,0.9669907077572599,-17.904833333333332,0.9017798413261526,30.0078,1.5428870945946327,-11.593466666666666,0.882976905448583
3040000.0,25.942700000000002,2.2760069698194405,-9.654433333333332,1.1537846400241059,38.53386666666666,0.3953381787224152,-17.8909,0.3546908794993185,28.523433333333333,0.5911713194065564,-10.000266666666667,0.5219091640846661,38.61196666666667,0.9443659118983255,-17.9508,0.8071044170366067,30.351533333333332,1.4109413414061176,-11.756266666666667,1.0215853442349085
3080000.0,24.807299999999998,0.37952921714496,-9.078666666666665,0.14152376321860485,38.8177,0.5123599711140591,-18.10966666666667,0.6618257995038337,29.505200000000002,0.9368319379696658,-10.441533333333334,0.7961471653462621,38.416666666666664,1.3653731732468688,-17.721733333333333,1.2239428971792572,29.6901,1.6386375397465625,-11.393866666666666,1.0125828668421273
3120000.0,24.4974,1.1951012118923936,-8.9112,0.5502949451582005,38.92966666666667,0.8293484283192178,-18.14923333333333,0.7249779092419921,29.046833333333336,0.363314100408387,-10.3108,0.31240730892004825,38.7474,0.7190154703945293,-17.9961,0.8201040096638149,30.468733333333333,0.8857212591379355,-11.768366666666667,0.6325695605140109
3160000.0,25.499966666666666,1.7560147196295244,-9.494766666666665,0.9402991734312837,39.09373333333333,0.4262151751821563,-18.194033333333334,0.4166247338899705,27.8776,0.42198005482091977,-9.689166666666667,0.49023784930265074,37.919266666666665,1.3126975796258469,-17.400133333333333,1.1293067794989207,29.5026,0.5635053327165586,-11.2931,0.19994050781836736
3200000.0,25.75,2.164641942677818,-9.644533333333333,1.1781638661729341,38.4922,0.9667777304013586,-17.820066666666666,0.6400907557179347,27.8125,0.8462198808032496,-9.5866,0.12056301257019093,38.833333333333336,0.8834919895254025,-17.995833333333334,0.7314883518477172,30.010433333333335,0.4710890951354692,-11.584133333333334,0.42219809200053143
3240000.0,25.528633333333335,1.8934453540804628,-9.455966666666667,0.9944492926014664,38.609399999999994,0.5785657669329113,-17.915266666666668,0.5300572065562579,28.789066666666667,0.6871717026252537,-10.230200000000002,0.3354847934954233,38.61983333333333,0.9988609724191944,-17.98646666666667,0.8706619907990825,31.145833333333332,1.109709554593252,-12.179933333333333,0.6533845030982054
3280000.0,25.067700000000002,1.8508041549553536,-9.262233333333333,1.0295495530678562,38.302099999999996,1.158782127925693,-17.756133333333334,0.9305531915061184,28.776033333333334,0.12505962133674056,-10.130600000000001,0.271914116343133,38.77603333333334,0.9798363956407329,-18.1182,0.873417986991338,31.70573333333333,1.346086994051854,-12.459133333333334,0.9326957965429505
3320000.0,24.32813333333333,2.027676809114861,-8.887233333333334,0.9092262583586599,38.21093333333334,0.9075046274751936,-17.656633333333335,0.7777193166917971,29.041666666666668,1.0593084830313702,-10.370600000000001,0.5134236132733545,38.520833333333336,1.0722252323504096,-17.845033333333333,1.0700934455561453,30.33856666666667,0.9988086114077229,-11.557699999999999,0.6409048135253784
3360000.0,25.1172,1.9060809863871646,-9.181633333333332,0.9033315166033394,38.494800000000005,1.0627324404571479,-17.93303333333333,0.9145862464646093,28.388,0.9730470115398672,-9.972366666666666,0.1336030521940108,38.934866666666665,0.45018290603807704,-18.17226666666667,0.5077546804861136,30.20573333333333,0.7069136360892256,-11.549700000000001,0.5732690874856825
3400000.0,24.333333333333332,1.0728566736625258,-8.877733333333333,0.5135406334675204,38.69006666666667,1.0162821699158608,-18.053266666666666,0.7450729106753399,28.041633333333333,1.1022034335316184,-9.729000000000001,0.32328100263805604,38.96093333333334,0.44214999214695766,-18.022133333333333,0.5894433098742877,30.398466666666668,1.396351668535624,-11.682833333333335,0.8652965439020828
3440000.0,25.17186666666667,2.546918059581467,-9.363033333333332,1.306238876401335,38.99216666666667,0.6956798896682943,-18.1781,0.7374161240439483,28.789033333333332,1.3774154210775422,-10.277066666666668,0.7948293499585656,38.47133333333334,0.607193853357853,-17.70183333333333,0.5430972431853771,30.960933333333333,1.9350935159716594,-11.9457,1.0855375565436083
3480000.0,24.8099,2.1295146692771723,-9.138266666666667,1.0120234066243505,38.0781,0.9030243961267039,-17.55613333333333,0.7175824288694806,28.666666666666668,0.17227970151923203,-10.171733333333334,0.1969832198155187,38.958333333333336,1.1625617240483272,-18.148733333333336,1.0563785821801248,30.416633333333333,0.18336037982314718,-11.711466666666666,0.36838890983427885
3520000.0,24.666666666666668,2.124372522783035,-9.171633333333334,1.1088647928800375,37.73696666666667,1.426426879381561,-17.35493333333333,1.0736774976168997,29.302066666666665,1.5628273616181094,-10.592966666666667,0.7973106225862595,39.231766666666665,0.904204336542479,-18.4181,0.7851314794351335,30.5573,0.27435977596336186,-11.861366666666667,0.33000412455334915
3560000.0,25.205699999999997,1.930617036079398,-9.405833333333334,0.9652689066898519,38.40363333333334,0.7546563073493923,-17.685533333333332,0.7441089540890884,28.8099,0.7105240178910217,-10.167566666666668,0.41835235813313554,38.6927,1.2632625248406086,-18.067466666666665,1.1710738357972523,31.0,0.46836484354258007,-12.179599999999999,0.5532156421023054
3600000.0,24.617166666666666,2.624572125551557,-9.064066666666667,1.4144517343793988,37.69533333333333,0.6088207417258007,-17.296333333333333,0.48433896073812704,27.7474,1.011127499378788,-9.764333333333333,0.33727219801749936,39.04163333333334,0.3807396690069963,-18.33203333333333,0.3243379341914168,30.7995,0.9679676165382122,-11.847133333333332,0.6765380715245985
3640000.0,24.270833333333332,1.57921825456634,-8.847533333333333,0.7529347086936261,37.71873333333334,1.3221054025892007,-17.2625,1.0967674594005783,29.1797,0.7317249619905005,-10.399833333333333,0.776758007504411,39.3151,0.22912618066617194,-18.5082,0.1538092324927209,31.015633333333337,1.7680309015650395,-11.980733333333333,1.3179059762449752
3680000.0,23.914066666666667,0.41324451465069534,-8.763933333333334,0.15567704462194223,36.856766666666665,1.9192783441237034,-16.769266666666667,1.3784380540629628,29.041666666666668,0.5237418787490222,-10.157033333333333,0.344798340418796,39.03123333333333,0.8526779201759337,-18.240466666666666,0.7455985038134729,30.619766666666663,1.0287260255070616,-11.858866666666666,0.7751719135492176
3720000.0,23.343766666666667,1.8085255455450133,-8.371733333333333,0.8783091685480434,37.11456666666667,1.3514413844320265,-16.90716666666667,1.0872452110213628,28.6901,1.2567699656924756,-10.230833333333331,0.3436578951353928,39.205733333333335,0.5331634666987412,-18.3751,0.6444710544314615,31.156266666666664,1.0022734900659034,-12.208866666666665,0.8149142177077424
3760000.0,24.64063333333333,2.1100740355626275,-9.137500000000001,1.1823919739240452,37.6823,1.488785283377023,-17.20713333333333,1.1424050838277797,28.192700000000002,0.9706081529982462,-9.863933333333334,0.45542127262080717,38.75,1.7512467102039033,-18.246199999999998,1.403546752576012,31.119766666666663,0.6860334896263368,-12.158999999999999,0.6203485525627238
3800000.0,23.66926666666667,1.1986424886326854,-8.582433333333332,0.7232976396723241,38.1823,1.0402584967208885,-17.660933333333332,0.9405152004206103,27.804666666666662,0.27162356713330765,-9.8389,0.2132210277310059,39.333333333333336,0.25052501316679127,-18.494133333333334,0.407122768816593,30.864566666666665,0.9688953859364243,-12.0005,0.8205144361923189
3840000.0,23.3255,1.0409976080664163,-8.388166666666665,0.5232614026999849,38.1823,1.5917282452311623,-17.828633333333332,1.317449840495728,27.648466666666668,0.6717962654125302,-9.6685,0.08958046662079834,39.58593333333333,0.34836158737087525,-18.8207,0.35250603777335016,30.755233333333337,1.4609803656753528,-12.025533333333334,1.1207519271552562
3880000.0,24.1849,0.34619948006893403,-8.879966666666666,0.16577688486503653,37.354166666666664,1.543073682261767,-17.189466666666664,1.1867400735722304,28.294233333333334,0.18291696355328962,-10.112,0.2127797609423099,39.02606666666666,0.6333032992878592,-18.123566666666665,0.5901740101209324,30.35936666666667,1.5815643423170915,-11.651566666666668,1.1685370264660948
3920000.0,23.825533333333336,2.469953165997732,-8.7181,1.3467899786776951,38.16663333333333,1.234461837761254,-17.567566666666668,0.9634645862141941,27.625,0.46574851583231064,-9.763799999999998,0.3421205343150277,39.08856666666667,0.5870417834835571,-18.278233333333333,0.6878300484535074,30.510433333333335,1.2527108108240932,-11.9182,0.8682304110468988
3960000.0,23.179666666666666,0.7568600766030382,-8.346633333333335,0.40873333061490735,38.22656666666666,1.1741815911045825,-17.6194,1.0569360466303848,28.013033333333336,0.36101062157350683,-9.847533333333333,0.2248161520492292,38.71613333333333,0.8171160681866985,-17.857166666666668,0.6541040403143485,29.01563333333333,0.4431139081044018,-10.860966666666668,0.49320982237673383
4000000.0,23.61196666666667,0.9539533403450906,-8.587266666666666,0.4909344989122502,38.117200000000004,1.2357794490388105,-17.655733333333334,1.0183078523816966,28.713533333333334,0.9057479573381445,-10.430966666666666,0.585403845411202,39.00000000000001,0.6064052110594041,-18.240099999999998,0.6163504252182085,30.002566666666667,1.6018652322284255,-11.444266666666666,1.1587698141659641
}\dataSotaStepsLj

%% file: data_new/sota_steps_pp_grid7.tex
\pgfplotstableread[col sep=comma]{
0.0,39.91146666666666,0.12520504072209762,-10.859366666666666,0.1844265044822881,40.0,0.0,-11.4643,0.23554884843700646,40.0,0.0,-11.187766666666667,0.2920434024060271,39.82293333333333,0.25041008144419524,-11.073966666666669,0.3263160260987633,40.0,0.0,-11.194033333333332,0.18382223901246436
40000.0,38.0964,1.3740561075395232,-9.502066666666666,0.5519119152747316,37.6172,0.17621250428578103,-8.799999999999999,0.39459214386502905,38.67966666666667,0.15794311915651302,-9.400766666666668,0.2238037880724,38.421866666666666,0.41235895635828,-9.195333333333332,0.29011506146508304,38.82033333333333,0.8201462728616684,-9.321600000000002,0.3657512087015803
80000.0,36.6901,1.6358088967439497,-8.614866666666666,1.2403788919340557,36.187533333333334,0.7186143441064571,-8.257800000000001,0.3747221192652851,35.893233333333335,1.1763840198771085,-7.879933333333334,0.7087562125927991,37.14846666666667,0.1486539007971935,-8.241133333333334,0.10954007891584207,37.08336666666667,0.9774947717961905,-8.035433333333334,0.14015397088757628
120000.0,35.11196666666667,2.049553961773687,-7.382300000000001,1.1787917231922973,33.53906666666666,1.4745564967888558,-6.492466666666668,0.49851601333914625,34.37243333333333,0.737043835572589,-6.836200000000001,0.31940495717296963,36.82033333333333,0.06084742302586663,-7.660700000000001,0.1793984578157422,36.80206666666667,1.1629874930062167,-8.319266666666666,0.5836944539359232
160000.0,33.596333333333334,0.8482433979832793,-6.5336,0.6343830125930761,32.28906666666666,1.7320911953153286,-6.238533333333334,0.6649072984676551,30.4974,0.4787401870186658,-5.1479,0.09959882864103718,36.062466666666666,0.27739546179096264,-7.400533333333333,0.2464410995665203,35.1693,1.9018984427846461,-7.438533333333333,0.6506382217143066
200000.0,33.78646666666666,1.1360827943224707,-6.246633333333333,0.2711429430310799,32.14843333333333,1.6191937939055405,-6.174733333333333,0.7903490129191165,31.5,0.6907440046790144,-5.336233333333333,0.48140107556543316,34.66406666666666,0.8805827855586427,-6.5385333333333335,0.36567069277637704,36.22396666666666,0.8147074580622315,-7.494533333333333,0.32339532395437554
240000.0,34.765633333333334,0.8453588442522842,-6.5185,0.7307190476966277,31.388033333333336,2.5417360265945956,-5.403133333333333,1.0907918051682557,30.161466666666666,1.5426816269801813,-4.872366666666667,0.5065120882620234,35.40363333333334,1.134220614440693,-6.8724,0.44635493350770405,36.46616666666666,0.9555021588440054,-7.4664,0.3890010882589743
280000.0,31.9479,1.005183754345444,-5.086166666666667,0.40736160294700746,30.421899999999997,3.971992122348682,-4.876833333333333,1.462294093387358,29.6328,0.3472987282825364,-4.503633333333333,0.08892972256538027,33.64320000000001,0.5841160672332166,-6.0716,0.5221263129422481,34.9323,1.0653748385740416,-6.429433333333333,0.46150407967379403
320000.0,31.562499999999996,1.2264721439967552,-4.956766666666667,0.4528754525866416,30.01823333333333,1.4076021510197962,-4.851066666666667,0.09984712759458296,26.53643333333333,1.0197394972355545,-3.3794333333333335,0.6317325401008106,33.3047,1.3943370635060477,-6.2146,0.41674458204836556,33.81773333333333,2.1320276613798628,-6.425,1.1679246922069364
360000.0,30.677066666666665,3.026665648678244,-4.7297,1.6124913973930732,26.611966666666664,1.7420108999531412,-3.691966666666666,0.7280171258669369,22.2396,1.5541453235353075,-1.9684666666666668,0.7465405027339255,33.390633333333334,1.5049113824039229,-6.4143,0.28070035031447066,33.0677,1.8373777183801925,-5.871366666666667,0.6645376554835366
400000.0,30.942733333333333,2.53139745375729,-4.794,1.407665841029042,25.8021,0.4795578797183917,-3.014066666666667,0.39444874895929843,20.388033333333333,1.2235298452519345,-1.3422,0.47481416013706534,30.109366666666663,2.184464069031324,-5.201833333333333,0.5988817598008996,32.390633333333334,0.9617460521826375,-5.843733333333333,0.1720427143344218
440000.0,29.638,2.0371476349706876,-4.1133,0.8245774230881336,22.325533333333336,0.6101914581142183,-1.8419333333333334,0.2757114957503384,17.513,1.3122876234525214,-0.6473666666666666,0.3532176319998134,29.651066666666665,2.0964474591291067,-4.915633333333333,0.4796449056912371,31.328133333333337,0.6423817885200535,-5.2411666666666665,0.2611413452945016
480000.0,25.862000000000005,0.18821144138087614,-2.8903999999999996,0.21948594488030446,20.79686666666667,0.39099790564940684,-1.3656666666666668,0.1942071974864875,16.52606666666667,0.8246477443261594,-0.33593333333333336,0.1464527303337914,29.54426666666667,2.5031942744332802,-4.811466666666667,1.017740993027641,30.005233333333337,0.6256345916125651,-4.782,0.4436519882370266
520000.0,21.84636666666667,0.660002779455426,-1.5453,0.38771072549862046,17.71353333333333,1.10968173615481,-0.6273333333333333,0.2890286529433064,15.898433333333335,0.6850548850680187,-0.21743333333333334,0.19872906738125207,29.020799999999998,0.7163041113940367,-4.390899999999999,0.26421855852053017,30.59113333333333,0.7284834307579617,-4.689066666666666,0.3767082100984203
560000.0,21.46613333333333,2.2012734440672177,-1.4474,0.5500851812825598,17.502633333333332,0.8389268396125031,-0.6046999999999999,0.3329717205209275,15.565100000000001,0.6643500934497316,-0.14403333333333332,0.19479564220542053,27.0078,0.8105558874418629,-3.5744666666666665,0.23404901670851377,28.51563333333333,2.8272707620994177,-3.9677000000000002,0.8434356446502997
600000.0,21.42186666666667,2.552106309610859,-1.2015666666666667,0.627954628580406,16.04686666666667,0.6907270388671791,-0.13283333333333333,0.06315601493303882,15.278633333333334,0.13077452181352264,-0.040633333333333334,0.07087107229955597,24.6979,1.296261694257761,-2.9468666666666667,0.3956037270917564,26.895866666666667,2.2870264337975827,-3.6208333333333336,1.0371295943880667
640000.0,19.02863333333333,1.6750897315931719,-0.7599,0.36363966596985353,15.197899999999999,0.33755397593076364,0.00936666666666667,0.11992957192545224,14.531266666666667,0.4968838451344093,0.06823333333333333,0.13456533316158695,23.15363333333333,1.8769833320754048,-2.083066666666667,0.6948966845670097,26.270833333333332,2.181373748097489,-3.562233333333333,1.063014456889254
680000.0,17.6875,0.33711233538194146,-0.5159,0.08279859097029772,14.669266666666667,0.6546826525543162,0.06953333333333334,0.08397465226020422,14.8802,0.10717176245012858,-0.025500000000000005,0.051673848963152205,21.536466666666666,2.926454232609073,-1.6468999999999998,0.733105099332058,25.127633333333335,0.937565657552698,-2.962766666666667,0.4681763224351366
720000.0,17.200533333333336,0.8438949078857836,-0.33126666666666665,0.15799629390871447,14.291666666666666,0.11433040812584458,0.12316666666666666,0.032485928577702,14.567700000000002,0.3732521489109829,0.06013333333333334,0.09161238392755038,19.833333333333332,2.3411745062871514,-1.0713333333333332,0.6324083561188046,25.911466666666666,2.7731346960113963,-3.1245,0.9811579315618189
760000.0,18.52863333333333,0.5185377539024731,-0.5768,0.1690317327210091,14.309899999999999,0.44019746326695985,0.13123333333333334,0.09366046242797557,13.940133333333334,0.12779072301575292,0.19113333333333335,0.11826820743077526,18.359366666666663,2.039522944764834,-0.7135333333333334,0.4575560754948209,23.833333333333332,3.0948551935250355,-2.4354333333333336,1.0605073764739006
800000.0,16.7474,1.151653127754475,-0.29350000000000004,0.2042610747711533,13.851566666666669,0.404811959089924,0.21223333333333336,0.0862087518114541,13.966166666666666,0.5129137051092403,0.2617,0.12131284625573116,15.921866666666666,1.033040774714252,-0.11173333333333334,0.3004164923716554,24.1849,2.147283162510246,-2.563566666666667,1.022545778382997
840000.0,15.601566666666665,0.7558655627086669,0.007533333333333336,0.10858840740255021,14.171866666666666,0.17620511406375888,0.1846,0.057826810391028834,13.921866666666668,0.2461620784948184,0.25780000000000003,0.1071584185524715,15.921899999999999,1.0960482106184934,-0.1336,0.2552020506709667,22.6901,1.9887579658336174,-1.9888333333333332,0.8806459914302808
880000.0,16.8828,0.7047964859919966,-0.31303333333333333,0.2046861065686243,13.614600000000001,0.12376254145203502,0.36379999999999996,0.07691298113235936,13.617166666666668,0.29153319231645297,0.3088333333333333,0.10391394944322391,14.526033333333332,0.557283508538418,0.14403333333333335,0.14588921215162629,21.375,2.7319430972600194,-1.6062666666666665,0.9389787939860813
920000.0,15.841133333333334,0.870259717300275,-0.10416666666666667,0.13494093358042086,13.763033333333333,0.3397331927001275,0.29323333333333335,0.10993138263884836,13.289066666666665,0.27625612914266556,0.32526666666666665,0.08443855096393407,15.9375,1.5683672550352061,-0.13906666666666667,0.3106280770024214,21.908866666666665,2.95180914619417,-1.5851666666666666,0.859101330202413
960000.0,15.0755,0.36649759435317814,0.08933333333333333,0.048170553476394914,13.190100000000001,0.40074716052227605,0.3739666666666667,0.06122403304440359,13.361966666666666,0.22101599238265338,0.3005,0.08523395254631023,14.283866666666666,0.5760731801506542,0.22163333333333335,0.062217325204122646,21.52606666666667,4.122472997553236,-1.4707999999999999,1.1533470856598198
1000000.0,15.276033333333332,0.44564121842077825,0.04243333333333333,0.03101637560314802,13.7448,0.21373258993424435,0.2955666666666667,0.08112173294226126,13.929666666666668,0.5476971202731993,0.23646666666666669,0.10258230300051216,14.0,0.12305042326894622,0.31016666666666665,0.0552486098367089,18.34893333333333,1.0832869282368771,-0.46459999999999996,0.4121044527786615
1040000.0,14.984366666666666,0.5413777875835771,0.09996666666666666,0.048211639903888576,13.6432,0.19558866702001593,0.24946666666666664,0.06278398592705699,13.408833333333334,0.12440718986010789,0.3070333333333333,0.09420907009884393,13.666666666666666,0.37957472986956814,0.3711,0.08330934321351158,17.609366666666666,1.3495694729143144,-0.31926666666666664,0.2846289904809807
1080000.0,15.427100000000001,0.2836213320609007,0.03593333333333334,0.04302188073785503,12.846333333333334,0.5319223084457183,0.4328333333333334,0.15477338990350448,13.5573,0.2742858484622685,0.3458666666666666,0.034693835507510866,13.544233333333333,0.6444444239463593,0.3723666666666667,0.1548316361586209,16.968733333333333,0.8924429181870523,-0.3247333333333333,0.20684100710986258
1120000.0,15.109366666666666,0.594190790534114,0.024966666666666675,0.14193863306216373,13.445300000000001,0.5824324338496265,0.35233333333333333,0.10417627155718116,12.968766666666667,0.28579214280467713,0.36043333333333333,0.07042581597365809,13.580733333333335,0.39656434814821956,0.29139999999999994,0.04691929240728167,16.229166666666668,0.5273932519687954,-0.08126666666666667,0.07466861604598161
1160000.0,14.671866666666668,0.5745686227268444,0.19426666666666667,0.09996273750197565,13.0807,0.12310534783942824,0.4226666666666667,0.10547683263267921,13.382800000000001,0.3803941902816081,0.35936666666666667,0.03250336734692098,13.776066666666667,0.40149860384425834,0.2864666666666667,0.11854822741071341,15.528633333333334,0.1313913831099879,0.06953333333333334,0.060784884817051535
1200000.0,14.791699999999999,0.5827582174452801,0.15986666666666668,0.1559640200673077,12.796866666666666,0.45383055819937385,0.4437333333333333,0.10678271811903313,13.278633333333334,0.09292636272279724,0.34793333333333337,0.05204749967310843,13.833333333333334,0.5841451950404875,0.2976333333333333,0.10229474842608274,15.273433333333335,0.5667121805251369,0.13486666666666666,0.10987399854176397
1240000.0,14.729199999999999,0.3558689178147854,0.16846666666666665,0.04900750509417466,13.0052,0.5071653050699213,0.4265666666666667,0.09040060225954741,13.200533333333333,0.48798255660992235,0.3619666666666667,0.06989202783977266,13.726566666666665,0.5328992170716295,0.26513333333333333,0.2074275188000752,16.4505,1.7146206013770704,-0.10959999999999999,0.24725221940358796
1280000.0,14.343733333333333,0.1930585806317755,0.21903333333333333,0.025981702963602853,12.903666666666666,0.3016379986378079,0.4106666666666667,0.08546411852675692,13.093733333333333,0.4887835671905875,0.3679666666666666,0.13600981173756874,13.101566666666665,0.38227602883547646,0.49033333333333334,0.0582106710339455,15.020833333333334,0.6079558445223541,0.11796666666666666,0.08193722529405599
1320000.0,14.513033333333333,0.5436893128330634,0.15156666666666666,0.08899641690664979,12.9818,0.37064833287992405,0.40313333333333334,0.062799009192467,13.138033333333334,0.48318936476522545,0.4205666666666667,0.12883522637677772,13.6693,0.6068350736952064,0.3450333333333333,0.1480362643258589,15.039066666666665,0.16349037348487006,0.10779999999999999,0.05594288754316018
1360000.0,14.346333333333334,0.5875948509720698,0.21223333333333336,0.06814104653013647,13.010433333333333,0.2105381253412841,0.413,0.06612115143179727,12.851566666666665,0.0939854716905164,0.5093666666666666,0.0617428178459289,13.283833333333334,0.15152716661451268,0.4161333333333333,0.05794551080301407,14.927100000000001,0.4440657383766511,0.17316666666666666,0.15269451711032572
1400000.0,14.5573,0.2206352797416283,0.16796666666666668,0.03843967857421402,12.8177,0.2547268864228242,0.4424333333333334,0.0906125205967083,13.138033333333334,0.10488223025003919,0.3825333333333334,0.02359976459392952,13.270866666666668,0.5065171621530269,0.4643,0.0981371828955094,14.843733333333333,0.12805562159554817,0.15623333333333334,0.03892183049252552
1440000.0,14.411466666666668,0.36547299337829203,0.20783333333333331,0.041339354400162336,13.0052,0.7343012233863334,0.43019999999999997,0.15317534614508518,13.580733333333333,0.501053838313697,0.27523333333333333,0.0924770722335482,13.018233333333333,0.04522730247194567,0.4901333333333333,0.05944461474534275,14.789066666666665,0.2234234445074095,0.20753333333333335,0.10131934113923605
1480000.0,14.283866666666668,0.6901215802708645,0.24996666666666667,0.06911986368299317,12.760433333333333,0.39037008366705295,0.4614666666666667,0.11360895895815416,13.213566666666665,0.3907393447754598,0.3708,0.1340130590651523,12.929666666666668,0.22652305744792364,0.4495,0.0457900280264892,14.7552,0.2853651111587866,0.2619666666666667,0.041715251673964775
1520000.0,14.276033333333336,0.752912444371116,0.25886666666666663,0.10819039801305023,13.096333333333334,0.5711512953869777,0.3692666666666667,0.13312035490070212,13.320300000000001,0.21512215134662466,0.38409999999999994,0.019672485015032215,13.468733333333333,0.14377807745117335,0.34663333333333335,0.0885147947457875,14.257800000000001,0.4158836135266695,0.3213333333333333,0.04715593517493023
1560000.0,13.742199999999999,0.4551690967834558,0.39113333333333333,0.061895036598709215,12.835933333333335,0.35499990923316493,0.4166666666666667,0.0703790846456215,13.638033333333334,0.3998763920124428,0.3559666666666667,0.09458122905148195,13.260399999999999,0.3170479143599595,0.3729,0.08342529592395821,14.611966666666666,0.5838566852309642,0.19376666666666664,0.12642323977637795
1600000.0,13.906233333333333,0.15519141585653365,0.3393,0.03084099004031268,12.723966666666668,0.467757560661028,0.5549666666666667,0.12479439980312508,12.966166666666666,0.17517755437143043,0.4812666666666667,0.05747267949981877,13.393233333333333,0.5105465198088108,0.40569999999999995,0.08867728006654241,14.6094,0.13501543121683043,0.2008,0.00984107040248503
1640000.0,14.044266666666667,0.046166173282562825,0.29036666666666666,0.06053193830991664,12.489566666666667,0.18101853189352987,0.5182333333333333,0.025381533094401935,13.757799999999998,0.054495198565255684,0.2703333333333333,0.01888320829614384,13.520833333333334,0.4954423365931585,0.33359999999999995,0.1035341811512829,14.403666666666666,0.2043694258499111,0.225,0.07690201557826687
1680000.0,13.919266666666667,0.38592160113450774,0.2981666666666667,0.08265278109187993,12.773433333333331,0.2860612793713186,0.4291666666666667,0.08479014617801346,12.8958,0.2606395723344148,0.48489999999999994,0.08932278544693956,13.583333333333334,0.18993357318342138,0.3481666666666667,0.07798471360180503,14.481766666666667,1.119959919917773,0.2487,0.18666195827395218
1720000.0,13.973966666666668,0.24970412269101364,0.31143333333333334,0.05797156391043995,12.5755,0.3717801590546041,0.48699999999999993,0.08471225806615398,13.354166666666666,0.9521588114501818,0.28046666666666664,0.1998897418300621,13.388,0.512297576284201,0.35856666666666664,0.16331012079952534,14.148433333333335,0.14617408190996917,0.23096666666666668,0.05944590444728347
1760000.0,13.901066666666665,0.4718648135030015,0.33956666666666663,0.0987824995746829,13.114566666666667,0.11592107468254262,0.38566666666666666,0.10905938239733844,13.247399999999999,0.3110322491318224,0.4291666666666667,0.13211331836302082,13.406233333333333,0.19650113372587877,0.3799333333333333,0.048414345899628654,13.984399999999999,0.3590829430646914,0.2856666666666667,0.07452795150516049
1800000.0,13.955733333333335,0.3334003532624939,0.22293333333333334,0.05125702119926812,13.112,0.6577758888861766,0.3903666666666667,0.14073732822373586,12.893266666666667,0.14774043304239873,0.4989666666666667,0.06130586341361557,13.356766666666667,0.10312310873686632,0.42186666666666667,0.04706147280124394,14.020833333333334,0.11489985011111001,0.28883333333333333,0.033476890868511404
1840000.0,14.065100000000001,0.2286820645933272,0.32003333333333334,0.06968889597505633,12.804700000000002,0.552113128504174,0.4151,0.09968874894724415,13.179700000000002,0.1658681404007414,0.4171666666666667,0.012563527459365141,13.4193,0.2712566804092883,0.3448,0.036200644561480776,14.361966666666667,0.4795504584735813,0.2182333333333333,0.17896711678095753
1880000.0,14.088566666666667,0.3885446549482946,0.27603333333333335,0.050186607332589804,12.64846666666667,0.46964546439013166,0.5260333333333334,0.10496012999653198,13.270800000000001,0.3312366022447795,0.4023666666666667,0.05317733435298246,13.197899999999999,0.2780526209191344,0.45103333333333334,0.04374611093825625,14.510433333333333,0.7467960646804607,0.18306666666666668,0.13839307143864615
1920000.0,13.987,0.2457402286968904,0.3182333333333333,0.027959295810556856,12.9427,0.2969192595078106,0.4388,0.07371734305213845,13.200533333333333,0.1583447785337078,0.3737,0.03743981837562783,13.3151,0.43431487041853256,0.39399999999999996,0.09169889130554779,14.234366666666666,0.22571356971958,0.22736666666666663,0.10003573805840035
1960000.0,13.968733333333333,0.03313973781160933,0.325,0.03945655839020937,13.033833333333334,0.4684951393083553,0.3429666666666667,0.08664149634493214,12.570333333333332,0.46562793682891873,0.5362,0.13850516235866445,12.848966666666668,0.27895323781753983,0.4578333333333333,0.08732629742650387,14.122399999999999,0.6378577270833992,0.29686666666666667,0.13211768323060402
2000000.0,13.632833333333332,0.27651488607708286,0.36199999999999993,0.05836734246705657,13.0755,0.4002095534425266,0.34323333333333333,0.09190318577478998,13.182333333333332,0.49214487252795364,0.3987,0.08579747471031221,12.648433333333335,0.5012500662233264,0.5193,0.10513775091120539,13.781233333333333,0.31915589015749374,0.35259999999999997,0.05686551386092159
2040000.0,13.804699999999999,0.26013730989614,0.3531333333333333,0.007979278719839823,12.908833333333334,0.2603148136818613,0.46613333333333334,0.07094158786557353,13.304666666666668,0.17125741508682832,0.31223333333333336,0.050989301710151785,13.236933333333333,0.25645482166564065,0.3893333333333333,0.09141532815793108,14.468766666666667,0.2579824584906667,0.2258,0.028419711469330578
2080000.0,13.851533333333334,0.15042041823576385,0.3273333333333333,0.04700243492510667,13.174466666666667,0.2061398608280843,0.3966,0.08759942922188475,13.565133333333334,0.2196048775009845,0.2864333333333333,0.06568456608840635,12.768233333333333,0.4557928647484029,0.4471333333333333,0.1450040076074528,13.987,0.1805659990142111,0.269,0.04435139982758906
2120000.0,14.218733333333333,0.5568551836479172,0.2823,0.11474519597787092,12.674466666666667,0.5301630021879014,0.5453,0.09381368059439231,12.848966666666668,0.5422068629427536,0.4593666666666667,0.07217429520881301,12.979166666666666,0.1186886777339026,0.43929999999999997,0.059371092853902116,13.5573,0.39301748052726637,0.3612,0.08650194602820602
2160000.0,14.041666666666666,0.622575105687838,0.3487,0.14151975127168645,12.716133333333334,0.4635170642909374,0.5375,0.06409747784949629,12.968733333333333,0.4028501982402666,0.44193333333333334,0.1294519044106944,12.585933333333335,0.6668266258098047,0.5823,0.1680783349116318,14.718766666666667,0.5111467782241116,0.1539,0.11461608380444112
2200000.0,14.0182,0.17864355571920337,0.2802,0.028064330860839475,12.955733333333333,0.6051762406293086,0.4338666666666667,0.15078373770256376,12.888033333333334,0.03209821732675479,0.4731666666666667,0.034487131255328006,13.265633333333334,0.6866565387602612,0.39606666666666673,0.17170653905881264,14.112,0.28600490671781664,0.2768333333333333,0.0654327814546263
2240000.0,13.809899999999999,0.2378256644407131,0.3010333333333333,0.023209815931100265,12.9193,0.38190446449341164,0.4437666666666667,0.08267624944457885,13.127566666666667,0.0351228637151869,0.3799666666666666,0.058086448983868465,13.109366666666666,0.2797706481308491,0.4573,0.0856215315599217,13.510399999999999,0.26188155083294173,0.4150666666666667,0.08033182571199426
2280000.0,13.859366666666668,0.29709695761185867,0.34636666666666666,0.08900312104390247,12.898433333333335,0.24262896685176571,0.4216,0.08291598157170908,12.5052,0.24889517204370765,0.5351333333333333,0.06292478755537348,12.940100000000001,0.31521060684353,0.43723333333333336,0.07533752200744473,14.273433333333335,0.6472887677759349,0.25053333333333333,0.1975724395984645
2320000.0,13.765633333333334,0.22648511847114586,0.3771,0.06279193154113565,12.520833333333334,0.25230994607602936,0.5724333333333335,0.11686471761067249,12.776033333333332,0.3464116658287102,0.5185,0.04169732205629836,12.744766666666665,0.20294163912043486,0.48750000000000004,0.053038288056836816,13.703133333333334,0.11813465010552826,0.3981666666666667,0.06968913513278491
2360000.0,13.825533333333333,0.2459415829464844,0.3054666666666666,0.07328639406844599,12.401033333333332,0.3014888205039932,0.5812333333333334,0.10231201732391407,12.973933333333333,0.25692141902837884,0.43616666666666665,0.06720418307086415,13.268233333333333,0.08798637520788187,0.3737,0.04558047827743802,13.9375,0.5202796235359087,0.31276666666666664,0.07477951294007974
2400000.0,13.786466666666668,0.45382747333712176,0.37110000000000004,0.0732018214709625,13.065100000000001,0.35689467167032163,0.3820333333333334,0.057777869658046606,12.6406,0.3572172541558801,0.49713333333333337,0.050846457322238525,12.802066666666667,0.06641146152752696,0.5034,0.05896886183967489,13.973966666666668,0.2475887629832093,0.27109999999999995,0.038288379438153296
2440000.0,13.7292,0.42632007537373423,0.34273333333333333,0.05631638798392131,12.822933333333333,0.10371391206369367,0.47866666666666663,0.05188739945518779,13.106766666666667,0.3236251981673999,0.37526666666666664,0.11150438954986883,12.9427,0.1867691801841689,0.4672,0.023193102422918745,14.434899999999999,0.019456618411224886,0.21329999999999996,0.04786007382638128
2480000.0,13.601566666666665,0.2727574298814889,0.39143333333333336,0.06517444966310716,12.445299999999998,0.5413609516764202,0.5171666666666667,0.11995855765869962,12.989566666666667,0.4611536572071783,0.41276666666666667,0.07375555722941983,12.763,0.1250165855663429,0.4525666666666666,0.04786121138830021,13.781266666666667,0.3094483299177568,0.30963333333333337,0.09406587526243983
2520000.0,13.7448,0.30092032168000843,0.35390000000000005,0.07071067811865477,12.674466666666666,0.3057250325955588,0.5492333333333334,0.0559097685760031,12.898433333333331,0.17150623571429952,0.4872666666666667,0.024000601844305667,12.768233333333333,0.4025335624378284,0.4857,0.1263199377242827,14.049466666666667,0.28123453241417967,0.29893333333333333,0.0478351567596702
2560000.0,13.588533333333332,0.048730642880589,0.39136666666666664,0.04302188073785506,12.591133333333332,0.07610485456847613,0.5046666666666667,0.08269116572457359,12.968766666666667,0.6188806688063723,0.41923333333333335,0.11878320120660532,12.742199999999999,0.6142072831436196,0.5070333333333333,0.13640606861703608,14.299466666666666,0.6854173635254819,0.2682,0.12874090259121224
2600000.0,13.721333333333334,0.2033631617465553,0.35000000000000003,0.06743545061760914,12.958366666666665,0.07872891604882062,0.4523333333333333,0.048365437613605386,13.101566666666665,0.3827639858828005,0.38933333333333336,0.0823647308554389,12.47653333333333,0.7639018450618438,0.5294,0.1720797683246542,13.979166666666666,0.12059027968934956,0.2856666666666667,0.04444812207006676
2640000.0,14.000033333333334,0.4009896618229564,0.3544333333333333,0.05626510661344401,12.921866666666666,0.4348054686357518,0.3953333333333333,0.13155608014159675,12.843766666666667,0.3544039346408124,0.4953,0.10914128458104201,12.809899999999999,0.271256680409288,0.4497,0.05233220805584262,14.151033333333332,0.17517755437143043,0.29533333333333334,0.031127943858997337
2680000.0,13.8073,0.16036410654091757,0.32630000000000003,0.07775358512634643,12.283866666666666,0.3556242992572674,0.5861666666666667,0.09818290188328217,12.7917,0.3969479646838694,0.4919,0.07102286016957264,12.9323,0.17819912083584116,0.4336,0.052337430837467225,13.781266666666667,0.13639494940144295,0.3468333333333333,0.0970998569629339
2720000.0,13.763033333333334,0.3899533675824439,0.3333333333333333,0.07260166817795734,12.374966666666666,0.394710056736447,0.5807333333333333,0.06812108500473421,12.5,0.19924597862943189,0.5533666666666667,0.08327249779422309,12.911466666666668,0.45618457034650156,0.4249666666666667,0.09921943134061774,13.744766666666669,0.22101599238265268,0.4015666666666667,0.022533284023614705
2760000.0,13.973966666666668,0.5677988278333167,0.3127333333333333,0.07826895226645678,12.822899999999999,0.4298965534482299,0.46120000000000005,0.08112118506695196,13.075533333333333,0.38464525936966365,0.47079999999999994,0.10386452714955186,12.833333333333334,0.3859213765637876,0.39399999999999996,0.09859036464077003,13.716166666666666,0.10718066782566563,0.3804666666666667,0.03255624193436473
2800000.0,13.783833333333334,0.1636567410432238,0.3484333333333334,0.043396953298077925,12.468766666666667,0.1442006087211692,0.6010333333333334,0.04296047278862543,12.921866666666666,0.05845895046003523,0.5406333333333334,0.03234958354531876,13.007799999999998,0.08861749262984141,0.4198,0.036256585608686324,14.1719,0.30647143423164264,0.25,0.07017962667327321
2840000.0,13.7448,0.4641239561439023,0.3713333333333333,0.06424626232102706,12.911466666666664,0.46790096767965267,0.3692666666666667,0.1336006320676998,13.1875,0.3054201150328296,0.40806666666666663,0.07905079941966994,12.932266666666669,0.23038497250375403,0.5047,0.05858896369340104,13.479166666666666,0.42585830455159035,0.4325666666666667,0.03934991176045451
2880000.0,13.4948,0.18000066666543205,0.4049333333333333,0.06303714954074481,13.0026,0.3182625645595156,0.42106666666666664,0.13468195457777143,12.9427,0.035157455349707455,0.48773333333333335,0.04854038410322781,12.856766666666665,0.4016029908033834,0.4476333333333333,0.05225924692232838,14.125,0.42156331276175674,0.2763,0.11617971710529626
2920000.0,13.632799999999998,0.29073575631490545,0.4002666666666667,0.09944114954193874,12.796866666666666,0.7213397642966564,0.4447666666666666,0.18475252516692578,12.729166666666666,0.4026844615273297,0.47446666666666665,0.09040583806124221,13.125,0.2934488711854254,0.4773666666666667,0.08512615474824539,14.101566666666669,0.18824176535036613,0.25833333333333336,0.046076771684753354
2960000.0,13.210966666666666,0.5268954503082711,0.5109333333333334,0.1093677994455203,12.804666666666668,0.2574639694317552,0.41559999999999997,0.12102917003764009,12.966133333333334,0.3194883548564621,0.3859333333333333,0.05033834412144372,12.6901,0.048298309149147685,0.42863333333333337,0.05524336058648938,13.583333333333334,0.14049145486074557,0.35286666666666666,0.03513140412160543
3000000.0,13.6927,0.37307942854035786,0.3573,0.022386752034778683,12.843733333333333,0.16914633375341717,0.46510000000000007,0.04163275953701205,13.117166666666668,0.27117426541289325,0.37526666666666664,0.058394425152482554,12.953133333333334,0.20250771727407166,0.4492333333333333,0.0647691970684282,13.789066666666665,0.22015634343701249,0.36769999999999997,0.0892052689026831
3040000.0,13.583333333333334,0.18924781519361156,0.3851333333333334,0.07312137550371679,12.875,0.23209025543237827,0.41406666666666664,0.09605891710588639,13.278633333333332,0.14774043304239934,0.35756666666666664,0.04141669014083841,12.927066666666667,0.34241369066606514,0.4791666666666667,0.09648123594196384,13.708333333333334,0.3099095065050801,0.3466,0.04775625613466784
3080000.0,14.080733333333333,0.423447457058002,0.3125,0.0655884644329067,12.703133333333334,0.3317142980873086,0.4281333333333333,0.10429830082774869,12.981766666666667,0.2685649807567793,0.45416666666666666,0.01900040935231543,12.721366666666666,0.27954976102456025,0.482,0.10671297952920254,13.830733333333335,0.4110426444489131,0.32706666666666667,0.1339249126272712
3120000.0,13.911466666666668,0.09910668774384315,0.2841,0.01753795883220164,13.080733333333333,0.08611969706299634,0.35440000000000005,0.051927449388545946,13.151066666666665,0.10135486613324934,0.4320333333333333,0.033641079385510544,12.7474,0.5510827221630764,0.4731666666666667,0.11036972209603299,13.718766666666667,0.3659992744376538,0.3932,0.06373920823689817
3160000.0,13.487,0.27849830639820156,0.4817666666666666,0.11370327856115477,12.557299999999998,0.22788138727563242,0.5101666666666667,0.07289692875713091,12.875,0.2509466477162025,0.5286333333333334,0.014048803824129966,12.973933333333335,0.16636476656899332,0.4567666666666666,0.05144351811021052,13.640633333333334,0.24977424918424976,0.3619666666666667,0.0042695302890234715
3200000.0,13.5677,0.07052153335447778,0.3245,0.03350054725921156,12.585933333333331,0.30076562672988766,0.49816666666666665,0.0692705966161369,12.971366666666668,0.13643177375116405,0.4398333333333333,0.026823414315771864,12.572933333333332,0.47328397629987684,0.5104333333333333,0.16180898890015008,13.9375,0.3591958797091077,0.27190000000000003,0.01746844774634159
3240000.0,13.867166666666668,0.3315119840300728,0.35569999999999996,0.07118403379035687,12.645833333333334,0.19993819600622179,0.48280000000000006,0.08487343518439679,12.528633333333334,0.31204491485824404,0.5198,0.05058774555166497,12.953133333333334,0.20680434123962566,0.4265666666666667,0.07102404913517357,13.789066666666669,0.2993705433442344,0.32580000000000003,0.08602898736278758
3280000.0,13.434899999999999,0.31121440626466307,0.43436666666666673,0.026210600060959224,12.523433333333335,0.5121828926814674,0.5372333333333333,0.13530337106747273,12.8125,0.46116791591205286,0.4963666666666667,0.10089969716946078,12.848966666666668,0.3725198369065586,0.5086,0.08370690931259302,14.039066666666665,0.6003907690459235,0.2979,0.200859104847154
3320000.0,13.570300000000001,0.3764760018912226,0.3919333333333333,0.10841996536103894,12.760433333333333,0.07448652376250518,0.46069999999999994,0.02935518125760199,12.835933333333331,0.30401748122252575,0.4554666666666667,0.0831317161029545,12.640633333333334,0.03981811424744046,0.5364333333333333,0.03959009415946816,13.893233333333333,0.07027291243588933,0.36093333333333333,0.041544059610116216
3360000.0,13.502633333333334,0.3878174943048458,0.3729,0.0887893386993431,12.729199999999999,0.22465014281470386,0.4489666666666667,0.10824578000499094,13.690133333333334,0.18224957856985316,0.23929999999999998,0.08294781491998447,12.3177,0.28307816352849735,0.6114666666666667,0.09308506265179654,14.080733333333333,0.3456948783087319,0.30573333333333336,0.11250103209403113
3400000.0,13.721366666666668,0.11696752635762667,0.3781333333333334,0.0236423819066993,12.828133333333334,0.36246865918157495,0.45649999999999996,0.06502599480207896,12.859366666666666,0.13255895124643732,0.44060000000000005,0.033262892237446835,12.403633333333332,0.30144976732819345,0.6263,0.0433310512219586,13.822899999999999,0.19202553649623483,0.3348666666666667,0.033175727405573036
3440000.0,13.627600000000001,0.19478815843543174,0.37316666666666665,0.03172489383573867,12.257799999999998,0.22572049087311466,0.5666666666666668,0.0809519336122094,12.609366666666666,0.291129825946356,0.5687333333333333,0.054527811456377244,12.869766666666669,0.33549756415741006,0.4942666666666667,0.030805446848821305,13.976566666666665,0.14363695748502756,0.3398333333333334,0.03194759597980557
3480000.0,13.5443,0.10232666644949727,0.42680000000000007,0.03612561418163018,12.625,0.26890003099045323,0.4976666666666667,0.06774483170118752,12.671833333333334,0.42633896008796685,0.5114666666666667,0.06335583810685659,12.736966666666666,0.10488223025003919,0.46563333333333334,0.038942207892665184,13.677100000000001,0.5483560704505789,0.3271,0.1275390397747555
3520000.0,13.776066666666667,0.12842529172852035,0.3221333333333333,0.04277587585959586,12.8099,0.29865057620347324,0.5028333333333334,0.051633150419301575,12.7552,0.14576792056782192,0.4604333333333333,0.051007014768123374,13.190133333333334,0.06878455899077605,0.4062666666666667,0.05196475301159005,14.309899999999999,0.44635208822931094,0.16666666666666666,0.05490685041251431
3560000.0,13.684899999999999,0.1695424430636769,0.40313333333333334,0.06974813418452297,12.421833333333334,0.40600481387403403,0.5937333333333333,0.04458559807331909,12.6198,0.22646911489207508,0.5372333333333333,0.0423396845629356,13.229166666666666,0.32914072302824465,0.32630000000000003,0.04117507336565008,13.791666666666666,0.28170661255205837,0.34686666666666666,0.06946684261014188
3600000.0,13.796866666666666,0.25611124319109635,0.3398333333333334,0.05530655978774631,12.4323,0.4237392201185378,0.5309666666666667,0.03877751352982234,12.9818,0.24577413750569152,0.4195333333333333,0.05678734209506747,12.507799999999998,0.21686149189440407,0.5570333333333334,0.055043639253071,13.781233333333333,0.2267112309132965,0.3377666666666667,0.06612600765877893
3640000.0,13.6875,0.3387695086633388,0.3278666666666667,0.09938602629254388,12.703100000000001,0.3175894939488187,0.45049999999999996,0.07994302137564395,13.059933333333333,0.2568505834570407,0.40650000000000003,0.0442697037110784,12.5625,0.2934488711854254,0.5315,0.06651711559190361,14.0052,0.18592930556172899,0.3390333333333333,0.024009211195335483
3680000.0,13.427100000000001,0.10849307197543348,0.3601666666666667,0.013623835322290457,12.934866666666666,0.13075925291245075,0.42083333333333334,0.03899079663487554,12.875,0.2936561254256413,0.4648333333333334,0.05626866110209324,12.619766666666665,0.34174837071870395,0.49556666666666666,0.09088565465584887,13.260399999999999,0.07798683649608217,0.4817666666666667,0.016268852312180972
3720000.0,13.580733333333333,0.25487346315813736,0.3372333333333333,0.09280464547759568,12.835966666666666,0.31327778230970826,0.4252666666666667,0.07942956348586813,12.981766666666667,0.45115152172585576,0.4437666666666667,0.0636988923678339,12.820300000000001,0.5497534962750728,0.5041333333333333,0.1250035821708944,13.6875,0.31682147444052217,0.3791333333333333,0.08656594146789039
3760000.0,13.4167,0.2767648942092665,0.4445333333333333,0.04870375299251131,12.296866666666666,0.07017788508893312,0.5867000000000001,0.040857394271620745,13.151033333333332,0.053121328631309465,0.41483333333333333,0.057270721625936896,12.7448,0.2961923811759288,0.4737,0.10269336233012659,13.703133333333334,0.28901446715038676,0.34946666666666665,0.09596160111674298
3800000.0,14.0573,0.07368052659963903,0.30833333333333335,0.010673747649672502,12.377600000000001,0.3105746609110282,0.5828333333333333,0.07836735431429481,13.2005,0.2277204865619255,0.3416666666666666,0.030712031663321922,12.325533333333333,0.3353634479519531,0.5531333333333334,0.06119718584670012,14.0495,0.057864381675315935,0.31743333333333335,0.08529073155325195
3840000.0,13.578100000000001,0.22345957128751476,0.363,0.07280238091344723,12.601566666666665,0.398357631058101,0.47136666666666666,0.09168410743174389,13.132833333333332,0.3448766575007491,0.4221333333333333,0.07244199211936554,12.203133333333334,0.15002756043103835,0.6346333333333334,0.05486810447690794,13.552100000000001,0.31697079781372084,0.3786666666666667,0.06716419350285456
3880000.0,13.473966666666668,0.42440667865726306,0.4476333333333334,0.10771519030397815,12.927100000000001,0.45303689327323776,0.37786666666666663,0.10366112525382351,12.669266666666667,0.2619397555842858,0.5117333333333334,0.010263635916293127,12.869766666666665,0.2543870716491346,0.4599,0.07999516652065092,13.919266666666667,0.41635079226803695,0.32759999999999995,0.05237613960574032
3920000.0,13.726566666666665,0.11068060153231747,0.33726666666666666,0.028940436915998513,12.671866666666666,0.2133806822455014,0.5101666666666667,0.09285032160537854,12.596366666666666,0.29728265263138925,0.5434666666666667,0.09306464897526284,13.065066666666667,0.3441176381542929,0.4007666666666667,0.04850046964262879,13.893233333333333,0.2136883140360169,0.3404,0.09345255480723895
3960000.0,13.432299999999998,0.2457402286968904,0.4091,0.06336013467999153,12.539066666666665,0.24342392834084503,0.4559666666666667,0.036229300603547704,12.653633333333332,0.2634013962672857,0.5351666666666667,0.049335067537086524,12.630233333333331,0.307560794351657,0.4932,0.04211658422363648,13.929699999999999,0.6741078697063254,0.3307333333333333,0.12201817715224055
4000000.0,13.929700000000002,0.04816354084436359,0.34896666666666665,0.03695062771982214,12.388033333333333,0.12826525466998298,0.5588666666666667,0.07442474200306119,12.885433333333333,0.24758876298320864,0.49346666666666666,0.06336225655353159,12.8958,0.45868795493232634,0.4486666666666667,0.08556604986143095,13.757800000000001,0.09632708168872728,0.31093333333333334,0.0590981857213532
}\dataSotaStepsPpGridseven

%% file: data_new/sota_steps_lj_T2.tex
\pgfplotstableread[col sep=comma]{
0.0,39.22656666666666,0.40600481387403253,-19.3522,0.2521316058463647,38.88803333333333,0.2182590255229369,-18.912766666666666,0.20668109627045075,39.541666666666664,0.3460006197810769,-19.562766666666665,0.18772072045698393,39.106766666666665,0.33900330709629073,-19.029433333333333,0.2024966885874656,39.7552,0.34619948006893403,-19.72723333333333,0.3364196222312585
40000.0,38.393233333333335,0.13866105757886282,-18.706933333333335,0.13587401844682892,38.84896666666666,0.551022723871006,-18.956400000000002,0.39964281552406244,38.966166666666666,0.26883489770158614,-19.136166666666668,0.18354222644636486,39.046866666666666,0.3815100726790251,-18.978133333333332,0.19515676319877304,38.53646666666666,0.37748059847126264,-18.779033333333334,0.2853102911256846
80000.0,39.10156666666666,0.3039703640525198,-19.146733333333334,0.21145578471370574,38.78123333333334,0.4789356556179782,-18.911066666666667,0.3342127499396488,39.1901,0.6160442570681633,-19.3475,0.3705797979742905,37.580733333333335,0.8300508270916662,-18.020333333333333,0.530950351309382,38.53123333333334,0.5408675397503118,-18.6332,0.333721111109261
120000.0,38.3307,0.57913625915381,-18.534233333333333,0.33247079403894936,38.4818,0.37337823539497483,-18.69946666666667,0.23315597545181835,38.36456666666667,0.15679758359816287,-18.60076666666667,0.09635539539757094,38.481766666666665,0.31908060980817393,-18.566666666666666,0.28085066415366644,37.14843333333334,0.4285439715543257,-17.7736,0.3078675797590041
160000.0,37.6302,1.1249244093123192,-18.176033333333333,0.8488287787036648,38.47653333333333,0.31684942094874546,-18.633966666666666,0.17838341352889842,37.74476666666667,1.3438618736892385,-18.22536666666667,0.8633400617498417,38.4219,0.6239495866387493,-18.594933333333334,0.4046885331818653,37.109399999999994,1.036003407330305,-17.632166666666667,0.6671261016895953
200000.0,37.5625,0.8643608775659992,-17.9573,0.5810626529615077,38.84376666666666,0.1461740819099679,-18.894633333333335,0.08186201126828517,37.864599999999996,0.9754576806128848,-18.240366666666667,0.5601967531343096,38.3776,0.6593180112813543,-18.573033333333335,0.4479080287539188,37.09636666666666,0.5272473191544521,-17.780333333333335,0.2647507800848864
240000.0,36.3099,1.2937767375659035,-17.136566666666667,0.8205353753630762,38.822900000000004,0.08707548449477717,-18.8746,0.04710053078257099,37.7005,1.247547116010721,-18.14493333333333,0.8530105834956302,38.78126666666666,0.459412910378259,-18.814866666666667,0.3250119006368168,36.82293333333333,1.2919379173250645,-17.5137,0.8403424659030401
280000.0,37.25,0.5814347828146053,-17.7026,0.4148725828492415,38.3151,0.7000344896265239,-18.430466666666664,0.4729686764352265,37.388000000000005,1.8861412513382974,-17.994266666666665,1.287628301266411,38.8672,0.07968199294696326,-18.74103333333333,0.04640203539597144,35.0651,0.3795779322703918,-16.4806,0.2141487956227331
320000.0,35.77603333333334,1.2051226115582123,-16.747633333333333,0.7103565364581994,38.0182,0.4478965356716516,-18.237199999999998,0.2910451969482862,37.612,1.2345499260864277,-17.891266666666667,0.8070805591072613,38.59373333333334,0.4032436842517055,-18.5775,0.22723891978855004,34.09113333333333,1.4062857019665516,-15.925533333333334,0.8734413177515443
360000.0,34.265633333333334,0.7034039727559739,-15.861166666666668,0.3697737716796876,37.9505,0.8910486443885482,-18.348399999999998,0.5735344860308452,37.30986666666667,0.9956876094215288,-17.7802,0.512581746326053,38.0625,0.7287152027141109,-18.3108,0.45409776480401254,34.237,0.3268901038575503,-15.863933333333334,0.2290230604594697
400000.0,33.27343333333334,1.140424953934084,-15.3793,0.6487254170037321,38.10413333333333,1.192175963894963,-18.409000000000002,0.759160066564798,36.565133333333335,1.098509554907112,-17.313266666666667,0.7108565084147117,38.114599999999996,0.8032126119527763,-18.2194,0.5023037394512084,34.270833333333336,0.5581923224918891,-15.955433333333332,0.3529636651114232
440000.0,35.34633333333334,1.0631953609546836,-16.6176,0.7112964501528175,38.4505,0.6269475735657628,-18.5194,0.43008032583072964,35.73696666666667,1.531257575400764,-16.868366666666667,0.8879535473335438,38.208333333333336,0.8002850381096853,-18.452866666666665,0.4181451449224576,35.106766666666665,0.2492236657212856,-16.43283333333333,0.07233184330262538
480000.0,34.92446666666667,1.2218243636282418,-16.27463333333333,0.7723736329580978,38.421866666666666,0.5773610385962062,-18.57656666666667,0.4236722186890347,34.919266666666665,1.5111927680551607,-16.3168,0.9091404108643873,38.53906666666666,0.31530047397504607,-18.526933333333332,0.29488366444338004,33.6536,0.705275492272347,-15.551666666666668,0.468744972121183
520000.0,34.56773333333334,0.2906779011590369,-16.171233333333333,0.26897727702457647,37.927099999999996,0.6602216950893607,-18.226699999999997,0.4796627426293044,34.99216666666667,2.3486789284380474,-16.3465,1.3803736981943204,38.3255,0.5333254603585575,-18.435933333333335,0.3485273239726782,33.52343333333334,1.7691829382200395,-15.513399999999999,1.0592118705276428
560000.0,34.140633333333334,0.4499906690390622,-15.816033333333332,0.27266392419158697,38.0469,0.6208193510729725,-18.395799999999998,0.3025235307652396,35.484366666666666,2.497636317712318,-16.6297,1.4881758229456623,38.3177,0.11799807908041052,-18.38566666666667,0.04873413132041631,33.106766666666665,1.5260789174293115,-15.250933333333334,0.8888781331293705
600000.0,34.5052,0.626885784387129,-16.056900000000002,0.3862004401861818,38.625,0.11811260164210606,-18.711466666666666,0.15263093468305183,35.9401,1.2562675856148877,-17.035533333333333,0.8033640035639013,37.5677,0.6293285363517737,-17.8271,0.4191663711065878,33.36456666666667,0.5759573672494247,-15.3423,0.37157039709858475
640000.0,33.21353333333334,1.152759935497799,-15.345433333333332,0.6631238211844163,37.999966666666666,0.30756467431889545,-18.398833333333332,0.1833267453361768,34.66406666666666,1.7852439690106474,-16.188000000000002,1.0525224083125257,37.6146,1.0030571004018969,-17.9978,0.5666842918827609,32.2474,1.247547116010721,-14.726799999999999,0.7565407766053769
680000.0,32.8802,1.4461829851947035,-15.073299999999998,0.8102362165854265,38.04423333333333,0.1079364730858937,-18.320066666666666,0.055261338705786874,34.2005,3.195802704590301,-15.925533333333334,1.8127249273461816,36.96616666666666,1.1349404869370403,-17.574466666666666,0.7508773727373123,33.19533333333333,0.647932760571821,-15.248166666666668,0.2674598703024853
720000.0,32.234366666666666,0.6784369011845448,-14.766766666666667,0.361968887551894,38.015633333333334,0.7173968698627631,-18.163833333333333,0.4797994117915899,33.958333333333336,2.6748536923644157,-15.729433333333333,1.4540893171405327,37.330733333333335,1.0475288073471873,-17.792733333333334,0.6925463033055782,32.52343333333334,0.3270141926923396,-14.937633333333332,0.17767562829180858
760000.0,32.89846666666667,2.4700882053526394,-15.170566666666666,1.5359334498025037,37.6719,0.43673571718679666,-17.913133333333334,0.3203082300257389,32.8594,0.8386054972393152,-15.094,0.49680157675541575,36.625,1.5239077159285825,-17.37073333333333,1.0136927224536803,31.78906666666667,0.8643301465425238,-14.498033333333334,0.5030730187778135
800000.0,33.34113333333333,1.3742535460710628,-15.360533333333334,0.758221745636167,38.45316666666667,0.27739546179096597,-18.504133333333332,0.14751093368138987,32.682300000000005,1.3013482265199696,-15.010933333333334,0.7482223481173382,36.34113333333334,1.4656661405502798,-17.1422,0.8741196752542902,32.09116666666667,0.9553751107403959,-14.572266666666666,0.5923649231869001
840000.0,33.49216666666667,1.5249143589796204,-15.3663,0.9189431139448553,36.955733333333335,0.9507873661106143,-17.575666666666667,0.5264324099276402,33.0599,1.5181487081310587,-15.236066666666666,0.8191846569743731,36.34373333333333,1.4662970875265662,-17.134266666666665,0.9352129181219755,34.1667,0.4053029566468352,-15.820966666666669,0.2418598124717893
880000.0,32.76043333333333,1.8303278218820689,-14.965733333333333,1.122295957203605,38.125,0.7812829577048236,-18.26523333333333,0.47136574146009197,34.07553333333333,2.1336831228860795,-15.869033333333334,1.197902701483816,36.6927,0.28883179649523694,-17.2492,0.17350563871720975,32.2344,0.9863124454248764,-14.697899999999999,0.5272262575656369
920000.0,32.15626666666667,0.3827639858828017,-14.690733333333334,0.17558349074506505,37.2526,0.7854768021187304,-17.8099,0.5134509324171106,33.54946666666667,1.550166497150834,-15.573433333333332,1.0427880396109055,36.16663333333333,0.48167351552224286,-17.068633333333334,0.36014353928522663,31.645833333333332,0.8956758950027002,-14.3254,0.5245083475662394
960000.0,33.57553333333333,1.637510145176377,-15.503633333333333,0.9545271406420147,37.63023333333333,1.005367409899927,-17.915133333333333,0.5437294506972717,31.77863333333333,2.320462661242846,-14.483199999999998,1.3292438477069084,35.70053333333333,1.4527151827602758,-16.681900000000002,0.9475920043281636,32.32553333333333,0.7961241290708947,-14.757566666666667,0.5024588496141305
1000000.0,32.768233333333335,1.4208483248475974,-15.007566666666667,0.8095072753773669,37.039033333333336,0.754068176994333,-17.54116666666667,0.5455823636771586,33.0651,1.659683698781186,-15.290766666666665,0.9915246251214449,35.90883333333334,0.7179050741954379,-16.83513333333333,0.5672012360901738,31.666666666666668,0.6005904946152757,-14.293766666666665,0.3915319172458638
1040000.0,32.9323,0.9349433779646763,-15.145066666666665,0.558930325612136,37.6719,1.2747548783981961,-18.0289,0.8251437369735449,31.72396666666667,4.125580357019145,-14.5698,2.526553981216313,36.606766666666665,1.1148046415802593,-17.2824,0.6496370114661466,31.255200000000002,1.0189081443715473,-14.105866666666666,0.6132335299233259
1080000.0,32.08856666666667,2.20076413598146,-14.631,1.2989053724835646,38.03903333333333,0.18456081803989693,-18.202733333333338,0.17972826402346687,33.8047,3.186663493373593,-15.728533333333333,1.9451070190494804,36.458333333333336,1.5812777983502946,-17.180733333333333,1.0764224707592998,32.3542,1.7917347794804885,-14.668999999999999,1.0344037348476012
1120000.0,31.552066666666665,2.9145674777267074,-14.421733333333334,1.7477087654666283,37.28126666666667,0.5378991375920062,-17.8276,0.38735007422227247,33.00523333333334,4.187613915929796,-15.2965,2.608219393890526,35.13023333333334,2.7463675504604663,-16.431233333333335,1.7238496692641798,31.557266666666667,1.7231237029173374,-14.294933333333333,0.9835283467642858
1160000.0,31.1875,1.000281503711163,-14.0319,0.7243442827827113,37.97136666666666,0.667504893032421,-18.2504,0.5428344376204095,32.640633333333334,3.7920521324944305,-15.053000000000003,2.4359232554960886,36.4427,0.6546788576597441,-17.1311,0.3804677472094925,31.664033333333332,1.794965779308588,-14.3958,1.091475148594781
1200000.0,32.458333333333336,2.161652852692947,-14.7841,1.289182270536895,37.76820000000001,0.5907691991519756,-18.01796666666667,0.5328011031853784,32.0026,4.915045153403985,-14.676033333333335,2.954220283293414,36.8906,0.8925810028600568,-17.4112,0.5577034875271986,31.057266666666667,1.3372149798077448,-13.951433333333332,0.724785008736307
1240000.0,30.0599,3.2836267946687654,-13.463933333333335,1.9425587495763306,37.059866666666665,1.023491935592178,-17.5806,0.6386849144922716,32.21876666666667,4.230218601391132,-14.796633333333332,2.5796888214580376,37.166666666666664,1.7894401681966212,-17.630466666666667,1.2374340045253154,30.812466666666666,1.7490224018640301,-13.815366666666668,1.0578333527650858
1280000.0,29.848966666666666,2.4102791889921966,-13.263166666666669,1.5218651809174455,37.257799999999996,1.2306383059209574,-17.69373333333333,0.7605299219769214,32.268233333333335,4.249034656117655,-14.892466666666666,2.7426588150106372,35.48696666666667,0.7084664863084369,-16.567566666666668,0.40641391325702486,31.34113333333333,1.9700470321514874,-14.128766666666666,1.1980191159669458
1320000.0,29.5104,2.665275666793213,-13.100933333333336,1.6791580714420211,37.830733333333335,0.28630446885944627,-18.163500000000003,0.24796073613914538,32.8594,4.73474665073715,-15.221099999999998,3.007523879650279,36.31773333333333,0.8011711940791576,-17.1647,0.5389902472834418,30.763033333333336,2.200076590080951,-13.861733333333333,1.2644724389597781
1360000.0,29.869766666666667,3.277465062446213,-13.2629,2.01492704251709,36.39066666666667,1.0287868627120451,-17.180733333333336,0.6003668674698466,31.46876666666667,5.155783369080677,-14.484633333333335,3.2062017864687737,35.56773333333333,0.9560205448745447,-16.6108,0.6450364382472262,29.697933333333335,1.7777790195134553,-13.1974,1.0426744554270042
1400000.0,30.28386666666667,1.5011929574693434,-13.6053,0.9532899279163014,36.15886666666667,0.5107897240765725,-17.0426,0.2771373666613726,32.1797,4.214981580821756,-14.834366666666668,2.621796943997677,36.270833333333336,1.5012590279125344,-17.1422,1.0038733618672564,30.625,0.9364930361015332,-13.7228,0.5362805981946392
1440000.0,29.867199999999997,2.878168543825511,-13.283366666666666,1.784330605265989,35.729166666666664,0.8308441343329998,-16.767566666666667,0.4914111742952349,31.531233333333336,2.865158211260866,-14.424733333333334,1.8333467581326657,36.1198,1.0454366264867494,-17.0418,0.6803948461494004,30.614566666666665,1.9999473465291262,-13.619266666666666,1.2042507278617502
1480000.0,30.4505,2.85758609435773,-13.5522,1.760471785251518,37.3776,1.0444293784965395,-17.8768,0.6582633515546803,30.153633333333335,3.4397673994359326,-13.459533333333333,2.064035497649096,36.015633333333334,1.6685192763512084,-16.968366666666668,1.0888175125132555,28.48176666666667,2.132493955182262,-12.4671,1.2208258297835395
1520000.0,30.114566666666665,3.967844491520413,-13.4458,2.3768944963263867,37.51043333333333,0.7342656391857704,-17.793900000000004,0.5140702351495047,29.898433333333333,3.457340968946448,-13.379833333333332,2.134945745342386,36.234366666666666,1.3631265050936723,-17.126933333333337,0.8027446266126953,29.7578,2.795040355105212,-13.175400000000002,1.6377922355007877
1560000.0,28.346333333333334,2.6591503283735,-12.363933333333334,1.65912043431318,36.92966666666667,0.9923173965801242,-17.4785,0.5748598843776346,32.28646666666666,2.245302489396225,-14.807166666666667,1.3526582823791418,36.14843333333334,1.193833439899481,-16.9982,0.6459566755337913,30.14323333333333,3.344123275571973,-13.351166666666666,1.9671714420004736
1600000.0,29.286466666666666,2.4923612597071245,-12.904966666666667,1.5120089711668008,38.03906666666666,0.6232624291224009,-18.204433333333334,0.48647946399502734,32.21353333333333,4.296700731439827,-14.773166666666667,2.716754156873399,35.5417,0.5915595377192968,-16.648433333333333,0.3120397553019312,29.945333333333338,2.9014631726462126,-13.320433333333334,1.6923247606637317
1640000.0,30.192733333333337,2.3176382293667452,-13.4525,1.3898579663644293,36.46093333333334,1.4981310408490804,-17.290233333333333,0.9863909175485261,31.723933333333335,4.696421574868348,-14.441033333333332,2.950507192934048,36.51303333333333,0.7696534472662914,-17.2484,0.41052688909091795,30.23436666666667,1.4914500468410676,-13.483066666666668,0.914264528946021
1680000.0,28.09896666666667,2.6505006776498323,-12.2017,1.5731918149630277,36.05726666666667,0.5999585670879453,-16.9879,0.3698832878985829,30.78906666666667,5.848729891371479,-13.9522,3.595038300213225,36.5,2.143391536793967,-17.351166666666668,1.3519661838308765,29.841166666666666,2.2419061017109714,-13.1056,1.293308411271908
1720000.0,28.3125,2.514851666931207,-12.36,1.4660270211243263,36.580733333333335,0.1918286793527548,-17.283199999999997,0.07413586626368338,30.848966666666666,5.69975666517635,-14.0298,3.4614933087710376,36.096333333333334,1.701787031982034,-16.966533333333334,1.0047608880834402,29.976566666666667,1.6115344537289789,-13.176566666666666,0.9622689240655247
1760000.0,27.395833333333332,2.4139397870608854,-11.848566666666665,1.4455660951874723,36.85153333333333,1.1690755730728244,-17.4061,0.7743057277329165,31.1328,3.5714546177526403,-13.996833333333333,2.252372625082764,36.32033333333333,1.4063302062065244,-17.14816666666667,0.8620344553567578,29.148433333333333,2.79175650124115,-12.804166666666667,1.5628285901602332
1800000.0,26.830766666666666,3.715244534198104,-11.512233333333334,2.169004420363305,37.33856666666667,0.9605520854638196,-17.802466666666664,0.5804885088344894,32.15103333333334,4.541301688821048,-14.682433333333334,2.8232214369797886,35.526066666666665,1.6285979866812508,-16.6263,0.9172687610509805,29.8047,1.9632196633761256,-13.140766666666666,1.1469961648681402
1840000.0,27.164033333333332,1.8064612226363699,-11.656866666666666,1.0958790393510087,35.630199999999995,0.7259939164116087,-16.708466666666666,0.4718826501955278,31.692700000000002,4.150347301933497,-14.464966666666667,2.6150855821984536,36.57293333333333,1.4709562996756773,-17.318233333333335,0.9032258718369144,29.088533333333334,1.8424666968194814,-12.751566666666667,1.0592178949059015
1880000.0,27.6797,1.8445891629303262,-11.9315,1.1546652068889927,37.31506666666667,0.14552400336561722,-17.8004,0.021200157232121864,32.0755,3.9977455371746706,-14.506633333333335,2.51414344901444,36.8151,0.5918341715942635,-17.429933333333334,0.46390688960417664,29.294300000000003,2.993157644807013,-12.761966666666666,1.737633045905326
1920000.0,28.16926666666667,1.0512833184044896,-12.327766666666667,0.6548647510915688,36.828133333333334,0.30167630923815747,-17.494933333333332,0.18817738676283388,31.856800000000003,4.568485668869571,-14.541766666666666,2.808704837148649,35.104166666666664,1.8453600377403017,-16.3867,1.0360360708006264,29.182266666666667,1.4285060198994224,-12.738666666666667,0.9174547121732434
1960000.0,26.4922,3.4326885362156965,-11.373466666666667,2.086479478824452,35.1849,0.33819119838734324,-16.451066666666666,0.18921670703778212,31.078100000000003,3.7728890000458097,-13.9957,2.2960794600071366,34.9974,1.7608850218001169,-16.306533333333334,0.9867392574648196,29.174500000000005,1.8912921632224524,-12.766300000000001,1.1034679243186005
2000000.0,28.156233333333333,2.6063096891113218,-12.251166666666668,1.5225870447658778,36.30466666666667,0.6034244185387999,-17.092933333333335,0.4340848636947481,32.78903333333333,2.91557120799491,-14.929933333333333,1.9319336260740319,33.734366666666666,1.296575492938567,-15.558333333333332,0.8058065331634443,29.89323333333333,1.465090987694015,-13.078533333333333,0.844409980729477
2040000.0,25.8125,2.0283215589907506,-10.968866666666665,1.1918480393443158,36.947900000000004,0.52306063383385,-17.491933333333332,0.2955849605255457,30.2604,4.840990824889744,-13.488433333333333,3.007701303136481,35.7474,0.7524618307041665,-16.794933333333333,0.46279824497890637,28.763,1.343339512806299,-12.590733333333333,0.851205906673324
2080000.0,26.281266666666667,2.5416807593051938,-11.1203,1.527485151046211,37.72396666666666,0.8519353353133979,-18.0569,0.5982635595343139,31.255200000000002,4.581789856231587,-14.1137,2.85971810848552,35.221333333333334,1.4129317188817778,-16.535533333333333,0.7523185598903939,29.567733333333337,1.6070418835722842,-13.0357,1.0653048671624472
2120000.0,25.864566666666665,1.8591775284308432,-11.034266666666666,1.1468147229997043,36.1719,0.5108237334606409,-17.1328,0.23604619604362684,29.877599999999997,5.534956683359559,-13.311100000000001,3.4450865601123386,35.49216666666667,0.9856914809186262,-16.696233333333335,0.6269017165570875,29.085966666666668,1.1170750089208672,-12.692466666666666,0.7774477488694801
2160000.0,26.5703,2.385392885878552,-11.319766666666666,1.4594331007468464,36.737,1.2753017838927392,-17.378500000000003,0.712226373189498,29.82813333333333,5.756673804859501,-13.475266666666668,3.5664983399531938,37.109366666666666,0.555041994407229,-17.65416666666667,0.31455170569487473,28.0104,1.2404606751794538,-12.089166666666666,0.7041213926274041
2200000.0,25.729166666666668,2.693053471846081,-10.9013,1.5598812540275837,36.39586666666667,0.9967578954902848,-17.297666666666668,0.5315379342582766,30.367199999999997,4.154565761504643,-13.6659,2.5597708504213164,35.992200000000004,0.8077247097041601,-17.015766666666668,0.492022366248617,29.302066666666665,1.4906476675891214,-12.694400000000002,0.8762567812386194
2240000.0,26.3021,2.7513798465497277,-11.128900000000002,1.633690309289575,36.43486666666667,1.0219228455329799,-17.249966666666666,0.5641202728339253,28.46613333333333,5.385287855093934,-12.4848,3.20436815092565,36.421866666666666,0.7392138902615105,-17.191666666666666,0.44263212967680343,29.255200000000002,1.3840684544727795,-12.654166666666667,0.8533892442620904
2280000.0,24.3724,0.886005037607951,-10.1772,0.5671370910106303,35.5,1.055728699366775,-16.639966666666666,0.6590005429099095,30.726533333333332,4.542781626810702,-13.760533333333333,2.9536775013908034,37.02863333333334,0.2937776294334829,-17.577233333333336,0.11742845576018644,29.148433333333333,2.212625246674687,-12.512766666666666,1.267495626115618
2320000.0,26.1875,1.9863386938452035,-11.142199999999997,1.0753828248581987,36.53386666666666,0.6325132954246016,-17.342833333333335,0.43319372366439385,30.1875,4.9861046699803655,-13.491933333333334,3.143694415103917,36.02343333333334,1.8627008246689045,-16.913833333333333,1.1403326337326118,29.671899999999997,1.6503619986738263,-12.949333333333334,1.0631418385563092
2360000.0,25.6927,2.7275905081225074,-10.864600000000001,1.5567819564730316,36.83856666666666,0.32914072302824315,-17.474866666666667,0.13133200506942513,28.429666666666662,4.61827841396433,-12.4728,2.874987660263374,36.52603333333334,1.2729764630799552,-17.304566666666663,0.7757379812511158,29.04946666666667,0.7494928122107355,-12.554066666666666,0.5379109116655744
2400000.0,25.197933333333335,2.2517502886026732,-10.582966666666666,1.314019853562173,36.91403333333334,0.45411687433474185,-17.5393,0.2667183283290954,28.9375,4.665577465509138,-12.800633333333332,2.8814564943599077,36.03646666666666,1.241198991119295,-16.957966666666664,0.7827288603909325,28.492166666666666,1.6407428730777893,-12.284500000000001,0.9937336598237307
2440000.0,25.15886666666667,1.9713388789235486,-10.494666666666667,1.129384213729863,36.69533333333333,0.1617557074383753,-17.4314,0.14499041347620167,29.869799999999998,5.215858715494505,-13.349233333333336,3.2211342767554556,37.02603333333334,0.8449307716546319,-17.525233333333336,0.48325838729285325,29.903633333333335,2.206811489205386,-13.052100000000001,1.3451259866644458
2480000.0,24.58073333333333,1.5834061183270571,-10.223033333333333,0.8770744412850915,35.73696666666667,1.1452883081953147,-16.775000000000002,0.5987797146419266,30.1953,5.061748684002399,-13.450533333333334,3.1090832135677693,36.72136666666666,1.017055542020964,-17.3974,0.6728995814136516,28.21613333333333,1.522932391437285,-12.0456,0.8618928278310862
2520000.0,24.04686666666667,1.0144406877136232,-9.911866666666667,0.5801311020412161,36.906233333333326,1.8332839247887631,-17.55506666666667,1.166404119029459,28.6276,3.5768363460838764,-12.576166666666666,2.2274394721792605,37.583333333333336,0.190743533456723,-17.938933333333335,0.14043528363231494,29.348933333333335,2.0281904255326273,-12.666266666666667,1.2696157537704948
2560000.0,24.22393333333333,2.251105722873884,-10.0789,1.24173587368651,37.05466666666667,0.8164749325947205,-17.6267,0.4980049062676665,28.682299999999998,4.6232136449299714,-12.68073333333333,2.723538891874972,37.21616666666666,0.48220002304253484,-17.678900000000002,0.3422107635166762,27.809866666666665,0.6766323341043898,-11.835566666666665,0.5253232422888681
2600000.0,23.947933333333335,0.6197767680562268,-9.865766666666666,0.4094399779644821,37.03123333333333,0.5304604435980325,-17.6935,0.3195827592346004,28.625,3.968131767805434,-12.557566666666666,2.479738583444275,36.018233333333335,1.7458396419933744,-16.999100000000002,1.0655017347084266,28.192733333333337,2.236262072198953,-12.004166666666668,1.3806173144245615
2640000.0,24.333333333333332,2.435251478230168,-10.106733333333333,1.3033547133796264,36.90886666666666,0.5655324708233421,-17.575000000000003,0.41176061815898124,28.8672,6.0461868848611235,-12.811433333333333,3.664049611503037,36.21613333333334,0.6078934628882096,-17.049999999999997,0.33212652207655197,28.4896,0.8810391175575958,-12.132666666666665,0.6855075265005403
2680000.0,24.9922,1.5693920500202188,-10.4543,0.7931260219326226,36.88543333333334,1.2451960818370023,-17.508999999999997,0.7171538235739008,30.333333333333332,5.934671268814213,-13.614466666666667,3.6545156809380663,35.55466666666667,0.8458754097908792,-16.731266666666667,0.5225818107137769,27.718733333333333,1.8922484937825217,-11.6975,1.2155934874235992
2720000.0,24.825533333333336,1.5528109872815248,-10.401833333333334,0.9580835361398415,36.75,0.8424819562855134,-17.488533333333336,0.5280453095037285,28.893266666666666,6.259855918114693,-12.883166666666668,3.810398789569874,36.1432,1.7246210269698854,-17.080433333333332,1.105546032008116,27.807266666666663,0.4283985086601285,-11.874366666666667,0.3473708521003009
2760000.0,24.632833333333334,1.3457391285915044,-10.306399999999998,0.6836693986618583,36.40363333333334,1.3925595004243856,-17.202333333333332,0.8526035785886784,29.72136666666667,6.119621880657514,-13.317933333333334,3.7083832140465516,37.13803333333333,0.6887345610288157,-17.674233333333333,0.4970576179433891,28.20053333333333,1.453008470572541,-12.033066666666668,0.9156002232901045
2800000.0,23.875,2.1840802289903802,-9.9012,1.2569196818678061,35.65363333333334,0.14673028165840923,-16.810666666666666,0.09735270355202726,27.843766666666667,6.193578480078288,-12.185033333333331,3.6134385012370447,37.42706666666667,0.6158904357829325,-17.713266666666666,0.38460217830318577,28.320300000000003,1.5896738931827086,-12.0996,0.9133675382889406
2840000.0,24.419300000000003,2.278628711893771,-10.239566666666667,1.3557126276939704,34.283833333333334,1.1304359699789397,-15.906666666666665,0.6567284640972668,29.158833333333334,7.1294648337414195,-13.019,4.323735630987014,35.26303333333333,1.3864632567155275,-16.456133333333334,0.7309054217582162,26.882833333333334,1.7642018787227527,-11.323966666666665,1.0511289496959395
2880000.0,24.29686666666667,2.056955932656042,-10.161966666666666,1.182647804246425,35.75,0.9668874908695404,-16.813,0.6766272090302015,29.91926666666667,6.50417075233696,-13.472133333333332,3.9669110949901,36.21093333333334,0.6583105059333495,-17.117933333333333,0.36022863418791284,27.90366666666667,1.6341977875670037,-11.760933333333334,0.9230766899643579
2920000.0,25.742166666666666,2.051309588747854,-10.932933333333333,1.2058834944646282,36.66143333333334,0.9893223011514283,-17.387533333333334,0.6323599256401016,29.200533333333336,6.493365225281025,-13.045833333333333,3.953707318735445,38.15366666666666,0.24227263338827326,-18.3749,0.12639258944521475,27.89323333333333,0.9008611337060898,-11.7906,0.45696171247344836
2960000.0,23.726566666666667,1.5146093893219545,-9.812233333333333,0.8599866911114896,34.854166666666664,1.5150644658084862,-16.243233333333333,0.9704371294536405,28.48696666666667,6.6315251098237,-12.665633333333334,4.005534558860056,37.078133333333334,1.1550728529212166,-17.740633333333335,0.8023002652097007,27.04426666666667,1.2507015480210388,-11.3305,0.784710766248728
3000000.0,25.257833333333334,0.9561327883138874,-10.632666666666667,0.5368619085844042,35.9323,0.9176184101611445,-17.001833333333334,0.5752529028368504,28.770833333333332,6.849416301327355,-12.813533333333334,4.106175075284648,36.221333333333334,0.6825901861455544,-17.044633333333334,0.47815691520206544,27.135433333333335,1.484826378926357,-11.321766666666667,0.8905476305185603
3040000.0,24.328133333333337,2.4726275852937416,-10.118066666666666,1.3131356475585032,35.63543333333333,1.103689480283693,-16.772133333333333,0.6683537303620667,28.734400000000004,7.313200149045561,-12.766933333333332,4.354604628308854,36.708333333333336,1.4026061203662608,-17.4529,0.9102738525667246,27.8776,1.7777727695068342,-11.737333333333334,1.038764112887147
3080000.0,24.1797,0.8368487836321836,-10.0306,0.4221075060534538,34.58593333333334,2.026044162620571,-16.063266666666667,1.2683932154063617,28.7891,7.604158457493286,-12.7887,4.548642397463225,36.140633333333334,1.2059842379659125,-16.976033333333334,0.7476620552688826,26.9401,1.401644650639622,-11.214433333333332,0.7196314349875371
3120000.0,24.770833333333332,2.1726907991909226,-10.3716,1.164673982995527,35.611999999999995,1.6416857616486775,-16.727999999999998,1.0615280055969631,27.1849,6.963914483009297,-11.910033333333333,4.069394446624979,36.65623333333334,1.5251182715521532,-17.3681,0.9227482899830631,27.789066666666667,0.6347781467217938,-11.692033333333333,0.3621166877617698
3160000.0,24.684933333333333,2.660766287285592,-10.351833333333333,1.5085167313917627,35.9401,1.3598282930821344,-16.9716,0.8723989263328248,28.479166666666668,6.94665793041677,-12.5298,4.0406501316826064,36.3307,0.3249212314802895,-17.0315,0.2288132134879163,26.804666666666666,0.609617526723838,-11.136466666666665,0.3659672146821657
3200000.0,24.52606666666667,1.8330098969242423,-10.319666666666668,1.0152380093138538,35.40886666666666,0.7850458344722417,-16.617466666666665,0.5460432237673337,28.64063333333333,6.5279871784664385,-12.675933333333333,3.9026123979827787,36.736999999999995,1.1075032821621775,-17.401533333333333,0.6545862068682943,27.791666666666668,0.3637070738321657,-11.698300000000001,0.29164877278443424
3240000.0,24.179699999999997,2.0749429743168037,-10.039066666666669,1.0622322543691765,36.3047,0.5358549803818199,-17.18473333333333,0.3746313773416335,27.596333333333334,6.82659362119124,-12.0918,4.09215635657616,35.682300000000005,0.9682805206481566,-16.790366666666667,0.6265947139543674,27.122366666666665,0.8391688520329044,-11.377366666666667,0.5146157293446136
3280000.0,24.0651,2.633960570446464,-9.955866666666665,1.3993568197171469,35.96356666666666,2.1520765821772114,-17.044933333333333,1.3100620960685625,27.460933333333333,6.682237975575415,-11.983199999999998,3.8814420241279755,36.06513333333333,0.7782231785005878,-16.964299999999998,0.5114001433971914,28.1224,1.2841382506049208,-11.863399999999999,0.8035663673068125
3320000.0,23.6354,1.7011223256035009,-9.695966666666667,0.8534820143909819,35.919266666666665,1.3986093124560854,-16.906633333333335,1.0609978458455467,27.3125,6.974913007342816,-11.949599999999998,3.9888541119808663,36.7109,1.424202604968828,-17.4156,0.8804601864934046,26.9323,0.5841165694163004,-11.148333333333333,0.36368202534020766
3360000.0,24.8047,2.4545628748652306,-10.442966666666669,1.4132495328929782,36.684900000000006,1.602366814018146,-17.474566666666664,0.9644773414768341,27.979166666666668,6.635081427952151,-12.278366666666665,3.8595758578141997,35.390633333333334,2.132460738729373,-16.660933333333332,1.3837109291884464,27.130200000000002,0.892647399592919,-11.307033333333331,0.3176837246207126
3400000.0,24.994766666666663,1.8143838152080412,-10.5047,1.0447070307028667,36.4401,1.7156589190939633,-17.267599999999998,1.0984926429733914,27.20573333333333,5.998111112304901,-11.729333333333331,3.4593332809789934,36.205733333333335,2.3350129511893996,-17.132833333333334,1.478537640899127,27.143266666666666,1.1207481380260635,-11.310166666666667,0.7420675455928671
3440000.0,23.986966666666664,1.6556807267371594,-9.938633333333334,0.8493542972295022,37.484366666666666,1.0085012025553362,-17.8836,0.6802819170510609,27.2396,5.953204618242739,-11.783633333333333,3.369479422040675,36.557300000000005,2.1485354096841554,-17.285666666666668,1.4059430650713505,27.283866666666665,1.5660267352620625,-11.392833333333334,0.8904581941649791
3480000.0,23.828099999999996,2.2341905081408493,-9.836466666666666,1.217127526414367,36.72653333333334,1.0519492076880685,-17.383866666666666,0.6734808798737755,26.598966666666666,5.538451967432377,-11.392299999999999,3.1794835083705024,35.58593333333334,2.7123707248252202,-16.748666666666665,1.6061089675223024,27.380200000000002,2.0150465205547974,-11.440233333333332,1.3475308019567571
3520000.0,24.361966666666664,1.6108166548541625,-10.2078,0.8728196148116745,37.42446666666667,0.680811228527327,-17.935799999999997,0.4858718829760227,26.9375,5.275661706996257,-11.582266666666667,3.0018495968689405,36.58593333333334,2.1146614312671663,-17.342066666666668,1.331905305275951,28.0651,1.6683557913906328,-11.728,0.9755797490039785
3560000.0,24.0026,1.7857616414292254,-9.919666666666666,0.9067807134154445,37.463566666666665,0.8089769108090947,-17.968866666666667,0.5241632082556813,26.781266666666667,6.258359227642834,-11.528,3.61208655581046,37.1901,1.9270293632082176,-17.796466666666667,1.276672874658536,26.89063333333333,1.6556468833124471,-11.234766666666667,1.0011875426490053
3600000.0,23.92186666666667,1.9471073131414436,-9.913533333333334,1.0459012201074358,37.09113333333334,0.4552127439145604,-17.672366666666665,0.20846074503901624,27.6771,6.81705845997133,-12.075099999999999,4.023776421058539,36.82553333333334,2.193171106462563,-17.486333333333334,1.3112703569016155,26.6979,0.9318690501710355,-11.079433333333332,0.6090249821549922
3640000.0,24.067700000000002,1.318714619114639,-9.938433333333334,0.6715018854941678,37.007799999999996,0.9278414627510451,-17.662499999999998,0.619010328400639,27.044266666666662,6.670266502088868,-11.670599999999999,3.846946157668443,36.833333333333336,0.6820390278829768,-17.4414,0.45910889775738395,27.4401,1.4882491480483604,-11.498033333333334,0.792605998519371
3680000.0,25.039066666666667,2.9829966603326183,-10.545566666666666,1.6277359661675954,35.36456666666667,1.0102691797513967,-16.601566666666667,0.6142765627594431,27.247366666666665,6.027439102600338,-11.720699999999999,3.501466585113539,36.11456666666667,1.972353200846362,-17.0393,1.1862044258895679,26.8203,0.9996016139776221,-11.139066666666666,0.582544912908486
3720000.0,24.52606666666667,2.8828642254227335,-10.219033333333334,1.5427422907133759,37.21093333333334,0.4561289096540813,-17.806366666666666,0.2799917895621612,27.302066666666665,7.284043629895569,-11.839566666666668,4.201060223589067,35.82033333333334,1.2595384772032805,-16.750933333333336,0.7269217121228098,27.010433333333335,0.1645345421349436,-11.276299999999999,0.02138660016614751
3760000.0,23.325499999999995,2.692288559323957,-9.535666666666666,1.3896015384114813,36.49216666666667,1.6549952796173006,-17.348033333333333,1.0595790684145383,27.432299999999998,6.124740352047586,-11.882799999999998,3.6098433788739364,36.17706666666667,1.9323210522293894,-16.98006666666667,1.1597425126102585,26.494799999999998,1.485873381774729,-11.0383,0.9000843219758168
3800000.0,23.578166666666664,2.492719045986174,-9.702333333333334,1.2943420911369952,37.4297,0.9463962630244622,-17.9379,0.7037247331165791,26.29686666666667,6.050627232683311,-11.2904,3.463955823428854,36.3672,1.0909189551321705,-17.1771,0.6950858987683942,26.739566666666665,1.9157957203789298,-11.135166666666668,1.1527790835869447
3840000.0,23.23176666666667,1.2150694694360298,-9.475666666666667,0.6216526861698765,36.70053333333333,0.9349720898983515,-17.44673333333333,0.6738185528925588,26.742166666666666,5.42683726451748,-11.473433333333332,3.156498602706627,35.91406666666666,1.4791225717371155,-16.9574,0.9031700209078395,26.48696666666667,1.1600216731691795,-11.004933333333332,0.6695861872994556
3880000.0,23.3177,1.3798667109543583,-9.527733333333334,0.5908983913405832,37.45053333333333,0.6926400812992416,-17.87483333333333,0.4379765011453875,27.424500000000005,6.898743156546705,-11.914700000000002,3.906433845167056,36.77603333333334,1.7072895757766342,-17.5285,1.1688667674290343,27.088533333333334,1.3770537930265785,-11.227866666666666,0.7682506376321605
3920000.0,22.7604,2.0981375932002164,-9.3384,1.1212101438475597,36.9349,1.0079486726350044,-17.594133333333335,0.7516309126751569,27.3229,5.97011836231075,-11.8543,3.439522435842899,37.497366666666665,0.2540755049630377,-17.948033333333335,0.23732905895027323,26.979166666666668,1.536437351653348,-11.249233333333331,0.9451071908636722
3960000.0,23.557299999999998,2.0370986099515815,-9.698566666666666,1.036180503365874,37.375033333333334,0.7697823776164859,-17.9108,0.4754083998696981,26.541700000000002,5.6015094614457865,-11.365366666666667,3.294488797303089,37.32553333333333,1.6792595934584467,-17.83843333333333,1.1628688786311578,26.70573333333333,1.1344544572416972,-11.105500000000001,0.8074009454209644
4000000.0,23.7396,0.9015118117177753,-9.7876,0.4142590815741601,37.26043333333333,1.8517626563778489,-17.9272,1.0833077340565174,26.9141,5.899743191586109,-11.536700000000002,3.4581559623591294,36.42966666666666,2.241543352742083,-17.373833333333334,1.415150781915325,27.526066666666665,0.27187264583910575,-11.444433333333334,0.3154448111617767
}\dataSotaStepsLjTtwo

%% file: data_new/sota_steps_lj_Gseven.tex
\pgfplotstableread[col sep=comma]{
0.0,39.7292,0.0036769552621715262,-15.286066666666665,0.026445709586917367,39.0521,0.29984009071503465,-14.532666666666666,0.2919215914514638,39.8203,0.2541341771584441,-15.508966666666666,0.08764083269547071,38.92446666666667,0.1952086120595651,-14.251199999999999,0.12224314568378342,39.921866666666666,0.11049721967341822,-15.692066666666667,0.2567038414637039
40000.0,36.1797,1.5312710036654755,-12.721233333333332,0.8951749711027945,39.166666666666664,0.5007198174184403,-14.5306,0.22740647894610808,37.0052,1.5600236558035463,-13.251966666666666,0.9903634562904442,38.03646666666666,0.5783179306306245,-13.779433333333335,0.33032082721028805,36.333333333333336,1.2727822917617242,-12.792200000000001,0.755821597115792
80000.0,32.4375,4.317936880810865,-10.7082,2.266574715291777,37.30466666666667,0.7416010622670087,-13.448166666666665,0.6467456601237378,35.01303333333333,2.4105055490959204,-12.146733333333332,1.3981156111789261,35.53126666666666,1.4148660816094545,-12.2435,0.8813074529735158,32.90363333333333,0.8755923721813827,-10.804400000000001,0.4412089376550146
120000.0,34.2708,2.1420203126954704,-11.577833333333333,1.059786029766806,38.40623333333334,0.8004664987090679,-13.919933333333333,0.4718792595098409,34.8073,1.3610376727580564,-11.944166666666668,0.89464694091518,34.3099,3.4634491748352048,-11.5284,1.7731931216499421,30.497366666666665,1.0087849798423627,-9.5822,0.5336775118614862
160000.0,30.4349,5.6055923915318715,-9.651166666666667,2.9042383124591478,36.49736666666667,1.528401006135352,-12.685533333333334,0.9194004000192496,34.09376666666666,4.381426255101057,-11.797266666666667,2.231255706746515,33.94533333333333,3.0996824539871106,-11.368900000000002,1.6558715670808126,29.1146,1.8688429058288087,-8.956366666666666,0.9429184706125029
200000.0,29.658866666666665,5.399912062452702,-9.323933333333333,2.8268456877272627,37.51043333333333,0.6324431797051467,-13.2134,0.44062685649727107,33.9974,2.766097519370326,-11.503900000000002,1.4862518786082883,33.59376666666666,2.614414731615641,-11.0513,1.383462663030701,25.742166666666662,1.4680139334791371,-7.384233333333333,0.7835247723517672
240000.0,27.29686666666667,5.750177556292403,-8.022133333333334,2.9587232216316255,36.76303333333333,1.6460198668977901,-12.741533333333331,0.9147441803890065,32.0651,2.681167232133548,-10.5323,1.250589916799268,32.5443,4.283206537941715,-10.567333333333332,2.10342042925855,23.460933333333333,1.1239239070724003,-6.349366666666666,0.5736923817595005
280000.0,26.755200000000002,6.299008404503046,-7.716799999999999,3.14797482942076,37.437533333333334,1.0947246848205971,-13.242600000000001,0.8041967047930504,28.77863333333333,4.258302646464772,-8.905866666666666,2.097524015489586,30.2005,1.9412444273369271,-9.395166666666666,0.8738446022542502,22.333299999999998,0.6530568479593998,-5.7007666666666665,0.278062035444531
320000.0,23.575533333333336,1.8860126993798914,-6.167866666666668,1.1647330004006164,36.434866666666665,1.6152600127808798,-12.5448,0.9339093888952327,25.7422,2.762461573790061,-7.558466666666667,1.4138263974838718,30.765599999999996,3.4092727318300606,-9.542966666666667,1.5936940156615875,21.54426666666667,1.767183328224764,-5.282,0.9361596587477303
360000.0,22.875,2.810974482274785,-5.822400000000001,1.435425449126495,34.88803333333333,1.2156951655557215,-11.8207,0.7356609409231948,25.567700000000002,4.782310283395116,-7.3763000000000005,2.2582219155787144,27.177066666666665,4.518020882594011,-7.897766666666667,2.135283488334876,20.1198,0.7107175106890213,-4.6796999999999995,0.27154905020394865
400000.0,20.054666666666666,1.1397816233335614,-4.4414,0.7276054196243088,34.49736666666667,1.4361628745453012,-11.5254,0.6880757516436693,25.559933333333333,3.1546837319910357,-7.356133333333333,1.4692080890360255,26.570333333333334,3.2521345455288633,-7.686966666666667,1.531236296446618,19.750033333333334,1.2488816873062427,-4.5193,0.43531384080913404
440000.0,18.973966666666666,0.6998340581086975,-3.9121,0.333357385798885,31.588533333333334,2.392590926636273,-10.129933333333332,1.1550571828076546,23.6901,3.7128906366872707,-6.3944,1.7902631817696528,27.843733333333333,2.637371208019244,-8.230733333333333,1.2609252061702774,19.6849,0.7073490557473494,-4.449366666666667,0.25784473795074586
480000.0,20.77863333333333,2.485219698314193,-4.6686,1.3151048855509586,31.276033333333334,2.2768445333155456,-9.869133333333334,1.0485038748405058,23.3516,2.8756924777636894,-6.1813666666666665,1.5304817571238438,25.9349,3.706847185951965,-7.216033333333333,1.6551082347958062,18.979166666666668,0.7278056944603344,-4.067033333333334,0.3077132467447935
520000.0,18.5,1.5309881580208256,-3.5387,0.8616792249246041,28.927066666666672,2.781384442243747,-8.732033333333334,1.299952012789532,20.033833333333334,3.0123527043750293,-4.636066666666667,1.6070035497437112,24.013,2.752150658424547,-6.359500000000001,1.3118995769493942,19.4375,0.9629415593205362,-4.303633333333333,0.40907166717934523
560000.0,18.4974,0.9281426650395226,-3.5116,0.5082165352157154,26.145833333333332,3.353761116849095,-7.413433333333333,1.5438551536836465,20.270833333333332,2.586460368062027,-4.7503,1.386059279636577,21.992166666666662,1.6313332713526754,-5.4525999999999994,0.7179539585980892,19.335933333333333,0.8595079070155319,-4.178533333333333,0.4406245138689202
600000.0,18.213533333333334,0.6403141017413944,-3.2149666666666668,0.44563230234004464,26.114566666666665,2.9825425767660425,-7.469266666666666,1.5018596390993255,19.79426666666667,2.60162784596277,-4.577866666666666,1.3116251505500935,21.744766666666663,1.3123346456686351,-5.2943,0.5926030543289496,18.638033333333336,0.7092733198296838,-3.8537666666666666,0.36259825641549287
640000.0,18.601533333333332,0.1946626711918747,-3.4773666666666667,0.1929289564119968,25.7578,2.4104863008668316,-7.1881666666666675,1.2114391698398328,20.583333333333332,3.683089703254893,-4.875,1.6767643026575523,22.890599999999996,1.8079347499287692,-5.716266666666667,0.7593708331390022,17.588533333333334,0.8813231353419064,-3.4551,0.30379138675523154
680000.0,17.5495,0.6387474357417551,-2.8482000000000003,0.4210126918118582,23.013033333333336,1.0932747880666702,-5.992566666666666,0.4918662239087192,18.92183333333333,1.913360457995885,-3.9950666666666663,0.9229257295266083,22.679666666666666,1.3192153358046683,-5.638433333333334,0.5806345111647737,18.674466666666667,1.0669115625121994,-3.8868666666666662,0.40123479687362873
720000.0,17.2578,0.41783770373994045,-2.6842333333333332,0.3631313842429792,24.53646666666667,3.0191095048411567,-6.645033333333333,1.3607461123303721,17.950499999999998,1.2377393936797305,-3.579833333333333,0.610477940269389,22.067666666666664,1.9683417070101308,-5.378766666666667,0.9424297403815073,17.992166666666666,0.8046996969194307,-3.6248666666666662,0.33670562150869326
760000.0,17.546866666666663,0.4598773483827577,-2.8607666666666667,0.27812642608872823,23.044300000000003,2.4096041514461795,-5.912366666666666,1.0495206090824938,18.023433333333333,1.1043532174485156,-3.5712333333333333,0.550039430909781,21.645833333333332,2.231105801067165,-5.220566666666667,1.0529280517785735,18.1849,1.9163411126414838,-3.7143,0.814629809586334
800000.0,16.763033333333333,0.3902560213785599,-2.5185333333333335,0.3442879640972398,23.2995,1.024564779146086,-5.923966666666668,0.5629499938320947,23.460933333333333,7.999124427224659,-6.334766666666667,4.2181824334922045,23.171900000000004,2.7021193941053023,-5.967299999999999,1.2741323191359157,18.052066666666665,1.0439933599192843,-3.663933333333333,0.4511495268262568
840000.0,17.6328,0.21718982480770224,-2.7860666666666667,0.35336442316050365,22.914066666666667,2.183449419814641,-5.735433333333333,1.0704611135186346,17.723966666666666,1.113401905073915,-3.414733333333333,0.6604266011871082,21.263033333333336,1.9125484121686671,-5.037133333333333,0.8561319771052955,17.703133333333334,0.4210264626151458,-3.5539,0.14212930732259277
880000.0,17.231766666666665,0.6817195920774326,-2.5277,0.4652233298821831,22.031266666666667,1.9372756426372468,-5.3608,0.9646415327294731,17.807299999999998,0.9046816051333566,-3.4427,0.4535950690502121,21.359366666666663,0.7968911357407745,-5.131900000000001,0.4061183653402869,18.65363333333333,0.24004805537410345,-3.9151000000000002,0.052792486839195754
920000.0,16.888,0.5996463624504036,-2.4330666666666665,0.31812174119701747,20.3073,0.7575361685534677,-4.540366666666666,0.2907119574806346,17.135433333333335,1.4765281853802252,-3.135966666666667,0.7095160103118807,22.479166666666668,0.5621140651346683,-5.502066666666667,0.1955343845863318,17.656266666666667,0.30305379427128504,-3.4939,0.24385935837417977
960000.0,17.0547,0.6621038035434229,-2.4784,0.5432464327233698,20.0052,0.9837193807178961,-4.417433333333333,0.4732993156790132,17.888033333333336,0.7698389283191365,-3.3066333333333335,0.3753844814172382,23.343733333333333,3.10620721173302,-6.025533333333333,1.549845201589142,17.479200000000002,0.7938959293677396,-3.390233333333333,0.29822452764467383
1000000.0,16.604166666666668,0.49987279270719354,-2.2079333333333335,0.377115688114698,19.20573333333333,0.26272274782024907,-4.0027333333333335,0.0669442222218535,16.958333333333332,0.4274014064345393,-2.9983000000000004,0.4293222643500645,22.664066666666667,3.6079796907533837,-5.652333333333334,1.7029973895719537,17.658833333333334,0.2637864077030667,-3.5262999999999995,0.10034364288118441
1040000.0,17.59893333333333,1.0563093654586033,-2.6431,0.6625301854758519,18.856733333333334,0.23011168785806793,-3.842166666666667,0.1585952780577726,17.3229,1.2790249906341418,-3.150766666666667,0.6589173055517328,19.934866666666665,0.9441029969706106,-4.447933333333333,0.4349828527910292,18.0573,0.9017713568305435,-3.6674666666666664,0.47503870251684643
1080000.0,17.08073333333333,0.42745880373304884,-2.3970333333333333,0.2396391361100177,18.33853333333333,0.4038329550142669,-3.596133333333333,0.30926983184411766,17.5521,0.9522531421143566,-3.166,0.5171151000180392,20.833333333333332,0.4436265496513421,-4.816933333333333,0.18620519744506453,16.049466666666664,0.281850555357898,-2.914066666666667,0.10771865617843952
1120000.0,16.966133333333335,0.45807784151701736,-2.3916666666666666,0.36179206121134777,18.895833333333332,0.08659139038546966,-3.7442999999999995,0.16122173137225237,16.9141,1.0185748507923544,-2.936733333333333,0.4367173329384682,22.023433333333333,0.5521442585250911,-5.369933333333333,0.24413521299112553,17.1328,0.8080458444090076,-3.203633333333333,0.3262071564035482
1160000.0,17.1979,0.4683065733754609,-2.3233,0.11038722752202812,19.16146666666667,0.6410879936136349,-3.8871333333333333,0.38255130665339815,17.195333333333334,0.1662141657287033,-2.985666666666667,0.23505662202021796,20.627633333333335,0.9636400515176246,-4.776,0.4757589095329692,17.776033333333334,0.8559870728515048,-3.5044,0.3855321257690466
1200000.0,16.6979,0.19990129230864495,-2.1999999999999997,0.13621123301695798,19.21353333333333,0.41373292781159704,-3.8273333333333337,0.2965718050583292,16.914066666666667,0.3699954263981598,-2.8697999999999997,0.0453334313724431,20.914066666666667,0.46512508234046246,-4.8168,0.24822231970554143,18.046833333333332,0.9125823664500402,-3.619266666666667,0.39427413756871693
1240000.0,16.901033333333334,1.1440114403079877,-2.2621,0.6099486371818533,19.85676666666667,0.26807169853521506,-4.248166666666666,0.13928752835611516,16.53643333333333,0.10603604209051849,-2.7159999999999997,0.1629075197773265,21.91926666666667,0.8604638180784953,-5.348433333333333,0.3339720281034856,17.781233333333333,0.6104132716629139,-3.6200666666666663,0.3587604059659995
1280000.0,17.059900000000003,0.5473960418807083,-2.2382666666666666,0.2979140181700454,19.799500000000002,0.4189854492302407,-4.1116,0.14665396914733225,17.325533333333336,0.6957647798565895,-3.058866666666667,0.2879145397895394,23.41926666666667,0.970232636478946,-6.020066666666668,0.4450234475720231,16.541666666666668,0.6482125028791784,-2.9562333333333335,0.3347429295577264
1320000.0,16.612,0.5860398336859592,-2.1618666666666666,0.4087770731774907,19.283833333333334,0.6595103048643145,-3.9105666666666665,0.3664871105812893,17.276033333333334,0.970960247498435,-3.023566666666667,0.4114148055457194,22.177066666666665,1.2625354525275274,-5.376333333333332,0.6216544931783964,17.2578,0.6077209392476128,-3.076166666666666,0.09615391596578662
1360000.0,17.666666666666668,0.1568236022485848,-2.4878666666666667,0.16442944451107966,19.197933333333335,1.195873193760758,-3.8165666666666667,0.5701350502780508,17.299466666666664,0.17701083457109384,-2.986466666666667,0.07742804545009656,21.471333333333334,0.6890905375122257,-5.0719,0.3453004585381647,17.083333333333332,0.7598795620944738,-3.2022333333333335,0.4725356447460397
1400000.0,17.2474,0.3891596416211057,-2.345966666666667,0.33152299400728685,18.710933333333333,0.5662247629892414,-3.5779,0.25671323819909775,17.156266666666667,0.43024724932944775,-3.0211,0.18268039486126209,21.906233333333333,0.8509827351689088,-5.276566666666667,0.3892259783496243,17.4948,0.6785265556090396,-3.3403666666666667,0.31703671010713513
1440000.0,16.187533333333334,0.08364888257206729,-1.9270666666666667,0.18591977361814488,19.697933333333335,0.925151311347982,-4.103633333333334,0.3657484776303094,16.658833333333334,0.3412598260693581,-2.860033333333334,0.175593571130102,21.679666666666666,0.45378522330380977,-5.204199999999999,0.18100880273253747,17.020866666666667,0.30461436093672434,-3.078933333333333,0.2657862089892718
1480000.0,17.031233333333333,0.8437995746753022,-2.1797,0.37083530036931484,18.976566666666667,0.2929451522422282,-3.7199333333333335,0.18453368858347322,16.520833333333332,0.4916974973384428,-2.7129,0.24939792835279667,21.2526,1.0659243125100397,-4.940733333333334,0.479353288875288,17.0729,0.28329300497306087,-3.119033333333333,0.23112284371928465
1520000.0,16.9531,0.1026564497080756,-2.1814999999999998,0.09022741638031452,18.4349,0.7145094867576407,-3.4906333333333333,0.3573269682268918,16.479166666666668,0.18667637474755347,-2.7754,0.03228663294099683,21.041666666666668,0.4472437540710378,-4.9060999999999995,0.23603910692933916,16.372366666666665,0.7086210662657122,-2.8666666666666667,0.09479712841407989
1560000.0,17.34893333333333,0.5878260589286214,-2.2563999999999997,0.24522760855988443,18.739566666666665,0.6163520981459288,-3.595466666666667,0.4365194179211532,16.718766666666667,0.6518581713498384,-2.750666666666667,0.13402102654268672,21.0078,0.582623880961522,-4.9126,0.18775837309336335,17.406266666666667,0.73295514339025,-3.220833333333333,0.35080525778398614
1600000.0,16.47136666666667,0.5355872187504934,-2.0177,0.22474030939434675,18.263,0.40555370380094713,-3.4403666666666672,0.2243784947112554,16.5703,0.5374484595444166,-2.7854000000000005,0.06603095233802594,21.1901,1.0113110204086566,-4.965266666666667,0.4960833352931297,16.5026,0.17415096898955237,-2.8580666666666663,0.163761744277743
1640000.0,17.018233333333335,0.24491535045580268,-2.129033333333333,0.18927935850365607,18.9375,0.8733894778390678,-3.7123666666666666,0.39617135462097985,17.10676666666667,0.3768733592188353,-2.9390666666666667,0.14742524282571906,20.520799999999998,0.7208406943747461,-4.696633333333334,0.3099191435770876,17.3776,0.2791035769506848,-3.2345333333333333,0.34906329257345625
1680000.0,16.8776,0.08848246530622195,-2.036566666666667,0.1690924861987927,19.322933333333335,0.8323361874200161,-3.8274666666666666,0.5197994249922006,16.848966666666666,0.6698229334848304,-2.7712333333333334,0.32254812905294533,21.60936666666667,0.9699278128923944,-5.153766666666667,0.4021996381013897,15.966133333333332,1.1979688847749295,-2.6249,0.39474818133421047
1720000.0,17.02863333333333,1.0296794333944697,-2.046466666666667,0.3972877238923392,18.864566666666665,0.8046623984973796,-3.6300666666666666,0.42608798256802405,17.041666666666668,0.3227843483745082,-2.950533333333334,0.2819767169269281,20.6328,0.5962543137509925,-4.7217666666666664,0.2990136042984593,17.0677,0.5401002190951845,-3.0394666666666663,0.056149878796742755
1760000.0,17.1276,0.9388260648277712,-2.0357,0.43687300214135455,18.471333333333334,0.2540755049630374,-3.5531,0.1431699922004143,17.406233333333333,0.11484024652630445,-3.0474,0.14686601603729388,21.96353333333333,1.8112202559477846,-5.284633333333333,0.7946237446462041,16.5625,0.40686580097127734,-2.8892,0.2579552803620555
1800000.0,17.0078,0.20642937452471977,-2.126733333333333,0.04120916834341069,19.1901,0.7551934630719908,-3.752366666666667,0.4095586594806117,17.0,0.4627595704034654,-2.9151000000000002,0.16014757777333588,20.1172,0.8801968075379516,-4.5147,0.44444304472001817,16.47136666666667,1.1357140671058992,-2.855333333333333,0.5214864800633752
1840000.0,16.8646,0.4831444504493446,-2.016,0.11240604965925986,19.062466666666666,0.2534805887812496,-3.620533333333333,0.2001260325117372,17.036466666666666,0.04339664605575898,-2.854033333333333,0.09655659250178393,21.367166666666666,0.5390464069900559,-5.077466666666667,0.2059903935192665,16.1771,0.7057092177377308,-2.7403666666666666,0.2505469527963882
1880000.0,17.005233333333337,0.7095023905683624,-2.0388,0.25918975031174885,18.5677,0.5129559500256011,-3.4812666666666665,0.33219767542166134,17.140633333333334,0.465601090586733,-2.9308333333333336,0.21464023129154713,22.0703,0.32592147315981823,-5.283433333333334,0.14991127005295124,16.747366666666665,0.4977995201104776,-2.9840999999999998,0.3485427663859917
1920000.0,17.04426666666667,0.6415506024902141,-2.0710666666666664,0.208419854034004,18.226533333333332,0.4583999006205053,-3.2550666666666666,0.2748804507821939,16.796866666666666,0.36912163607979265,-2.735666666666667,0.14198531222943064,20.91926666666667,1.337923730594868,-4.859133333333333,0.6556223828461692,16.91926666666667,0.30764552473404916,-2.9440000000000004,0.23681085279184327
1960000.0,16.645833333333332,0.3014032219838112,-1.9755333333333331,0.10439202821842081,18.45573333333333,0.5113785377671706,-3.3512999999999997,0.31755629422198506,16.755166666666664,0.30461436093672434,-2.7975333333333334,0.29872964737739416,22.598966666666666,0.8492224927674867,-5.619133333333333,0.42501261419188774,16.085966666666668,0.9898158829914897,-2.752466666666667,0.5675562605494621
2000000.0,16.4349,0.34602821657585536,-1.9282666666666666,0.08175476880416348,18.375,0.45639502626562456,-3.2716333333333334,0.26055564131720926,17.52603333333333,0.3297691953809855,-3.1388,0.2075342381391562,21.453099999999996,1.0735686936568156,-5.170166666666667,0.44072086656093595,16.195333333333334,0.03374614381262258,-2.6747333333333336,0.21378971495893387
2040000.0,16.8724,0.30643825914311995,-1.9926666666666666,0.17612499980285462,18.7526,0.25200132274785214,-3.5008,0.09928870362063699,16.4948,0.39428829554020556,-2.6532666666666667,0.13708465348908394,20.958333333333332,0.4276489396170133,-4.862466666666667,0.12115423595107538,17.1953,0.54580496516613,-3.1307333333333336,0.3598064695854642
2080000.0,17.177066666666665,0.6477074459627243,-2.0466333333333337,0.2028170001640779,18.374966666666666,0.2451619102189865,-3.2428333333333335,0.07116948472172299,17.122400000000003,0.47413597908898175,-2.8806,0.1644349314065192,21.97393333333333,0.06028650116089177,-5.338933333333333,0.02782053598013932,16.549466666666664,0.8554470968771568,-2.9257666666666666,0.5627102352800141
2120000.0,16.375,0.6368579799819326,-1.8002666666666665,0.2166128086907348,18.20573333333333,0.6824172444740507,-3.1711000000000005,0.352941081013059,16.51823333333333,0.8366035314824389,-2.6074333333333333,0.2171742208970075,20.479166666666668,1.5654359570278042,-4.673166666666667,0.7142010656826426,16.3021,0.9654558129022098,-2.7912999999999997,0.5033047254563249
2160000.0,16.848966666666666,0.5593567277586718,-2.0428333333333333,0.23844821007692402,18.578133333333334,0.4088465101830868,-3.3598,0.07427170838661705,17.138033333333333,1.1508293077409675,-2.9237,0.7256912061384421,23.14323333333333,0.2675037735974752,-5.889600000000001,0.0962813931487628,15.114600000000001,0.1844720213654821,-2.2480333333333333,0.22285854008516007
2200000.0,16.528633333333335,0.5697168029429676,-1.8871333333333336,0.24406068006861634,19.434900000000003,0.7174990778158994,-3.7753666666666668,0.4487475484303199,16.708366666666667,0.3545379183607238,-2.708066666666667,0.22215136481437353,20.8255,0.5679005605444198,-4.825933333333333,0.21271554297282166,16.8698,0.7116096261293835,-2.8976666666666664,0.5204851412117567
2240000.0,16.598966666666666,0.5088652233701529,-1.9760333333333335,0.18482579786262401,18.932266666666667,0.6339635022792891,-3.6369666666666665,0.1803811212097567,16.505233333333333,0.3158617454238827,-2.5796666666666668,0.08615804599042914,20.911466666666666,0.8673551457672273,-4.816166666666667,0.3513455880215694,16.773433333333333,0.9783599962295177,-2.9290333333333334,0.5049759224183092
2280000.0,16.59113333333333,0.1023300650943907,-1.8907333333333334,0.026276774704839436,19.15883333333333,0.59548665998679,-3.5355333333333334,0.22350785718229726,16.7396,0.016027476407718486,-2.7722333333333338,0.08413042784206479,21.57813333333333,0.3401596618583031,-5.116666666666667,0.1487191611356414,16.716166666666666,0.35179487710248963,-2.865866666666667,0.1918724634287638
2320000.0,17.145799999999998,0.5614007006289421,-2.0679666666666665,0.15395129821544937,18.6823,0.3375532056826991,-3.356,0.06666678333323124,17.471333333333334,0.6876355930933825,-3.0060000000000002,0.4156833730938329,22.08856666666667,0.9860114817902597,-5.4452,0.4230838923901501,15.736966666666667,1.3622804760971787,-2.6488333333333336,0.6461231685126985
2360000.0,16.82813333333333,0.6542374407574739,-1.9889,0.1765400993164631,19.447933333333335,0.9007064647017674,-3.730966666666667,0.39949902238790147,16.9896,0.3253130184914206,-2.817066666666667,0.3017626477143179,23.0729,0.7164760986941568,-5.834366666666667,0.3503306374777341,16.7552,0.6269475735657644,-3.0289,0.48161119173042477
2400000.0,16.971366666666665,0.9022807594584337,-1.9765666666666666,0.3232435648581765,18.841133333333335,1.015507952810918,-3.4453,0.42511748964256924,17.0677,0.07023337288402517,-2.886833333333333,0.06837525543320165,22.0078,0.8783738156388774,-5.357166666666667,0.3971780485150481,16.487,0.2728164584477992,-2.8767,0.2386086335403646
2440000.0,17.296866666666666,0.20253857794394567,-2.1334666666666666,0.08242533321470746,18.567733333333337,0.6347062120032825,-3.426966666666667,0.05743078926460569,16.5573,0.25646734684945804,-2.686866666666667,0.14605536241059042,21.75,0.297094328454786,-5.2366,0.1248512982177863,16.851566666666667,0.33627604070994366,-3.1187666666666662,0.33337546733708867
2480000.0,17.455699999999997,0.7509859696869621,-2.1572666666666667,0.17850031434768465,18.437466666666666,0.6132684911376274,-3.118233333333333,0.19899551642073637,17.138,0.1842407121132565,-2.8876000000000004,0.19500405123996772,20.83856666666667,1.2515709231024643,-4.854066666666667,0.5411534368570732,16.343766666666667,0.9593857560381482,-2.7965,0.5046489142628434
2520000.0,17.208299999999998,0.46883594145500357,-2.1302333333333334,0.09127213278006739,18.596333333333334,0.9055329419126001,-3.3821666666666665,0.4471437675836363,16.971333333333334,0.161546078737788,-2.8334666666666664,0.0798284132095557,21.096333333333334,0.4814704583068642,-4.951433333333333,0.16425533645990176,16.1146,0.224849149431347,-2.6116,0.2053556102634322
2560000.0,17.052066666666665,0.3857686468453455,-2.0347666666666666,0.04871059661123252,19.372400000000003,0.8703474402022832,-3.6984333333333335,0.6101151603499857,16.637999999999998,0.5319606564399292,-2.6723999999999997,0.275916811134564,21.35676666666667,1.511750998750198,-5.072033333333334,0.7186219234680286,16.41143333333333,0.1936413236430243,-2.8281000000000005,0.18659069287257254
2600000.0,17.3229,0.6388991834919386,-2.1043,0.24440222312136736,18.153666666666666,0.47345032357037015,-3.1745,0.1855961386092574,17.416666666666668,0.11974863487971653,-2.9565,0.054400919109882685,21.184866666666665,0.22975813853315888,-5.053233333333333,0.10312401379999821,16.898433333333333,0.38312593519914934,-2.983233333333333,0.32594400677962404
2640000.0,16.76563333333333,0.6175524503140509,-1.8561333333333334,0.1506874543181644,18.96613333333333,0.42476045275216384,-3.5,0.06996627759142243,17.2526,0.4098982556684034,-2.8511666666666664,0.0628153555182878,22.145866666666667,0.6751151523176537,-5.404166666666666,0.3415936800091919,16.6354,0.4798424810983986,-2.940466666666666,0.3808258248712774
2680000.0,17.416666666666668,0.5663975125494655,-2.1302000000000003,0.15291121606997965,18.333333333333332,0.1329768233774438,-3.1709666666666667,0.2944476448463388,16.6901,0.2665732669767671,-2.7016666666666667,0.04414705224839176,21.60676666666667,0.5872217317353152,-5.159533333333333,0.30424286716737026,16.664066666666667,1.4895454907081185,-2.9544666666666664,0.7809153361422876
2720000.0,16.4349,0.7194391472992464,-1.7354333333333332,0.19499221067063732,18.499966666666666,0.7367286308781243,-3.2883,0.17841014545142883,16.64323333333333,0.14691333802241724,-2.596366666666667,0.09411752700155737,20.307266666666663,0.7068696736708647,-4.630466666666667,0.30516085739964904,16.661466666666666,0.9760698927615565,-2.9358,0.6164042666951617
2760000.0,17.15886666666667,0.15312762288003032,-2.0383,0.09951083693079192,20.398433333333333,1.5677393200259906,-4.063666666666666,0.8914730369201053,17.237000000000002,0.29329815319341235,-2.8263,0.22414180035563788,22.03386666666667,0.880760256192846,-5.366933333333333,0.41805475186338403,16.484366666666663,0.47137673774687744,-2.786733333333333,0.16998371163797507
2800000.0,16.73176666666667,0.20615038739285052,-1.9131666666666665,0.147917371829305,18.953133333333334,0.3039703640525206,-3.476566666666667,0.17377683645667194,17.437466666666666,0.3811909961627578,-2.929933333333333,0.16234400375608452,21.1953,0.1762125042857822,-4.916300000000001,0.0758211492043393,16.8047,0.8020156897883399,-2.9187333333333334,0.47582653235069716
2840000.0,16.9479,0.7375966422555535,-2.0537666666666667,0.2846807139391231,18.627633333333332,0.4126215888142008,-3.3777333333333335,0.2813488976737285,16.6797,0.44612312949080213,-2.5875999999999997,0.28597553508415136,20.729166666666668,0.14552400336561566,-4.748033333333333,0.06160261538459388,17.338533333333334,1.1190613219221823,-3.1635333333333335,0.4819206735091943
2880000.0,16.90103333333333,1.0136157369645666,-1.9878999999999998,0.27286879631060784,18.66143333333333,0.5103411299208496,-3.2685,0.1529731348962949,16.960933333333333,0.21569066327085804,-2.8467333333333333,0.23821875567544112,21.721333333333334,0.8432860223883449,-5.1704333333333325,0.3911991933643127,17.583333333333332,0.6959228517267957,-3.2231666666666663,0.5651040041935249
2920000.0,17.273433333333333,0.4207508948957271,-2.126333333333333,0.132580298519635,19.273433333333333,0.13802457108146535,-3.5355666666666665,0.2696773916285077,16.770833333333332,0.4702139182211533,-2.625433333333333,0.2697340088968307,21.453133333333337,0.781171787907258,-5.125900000000001,0.36641698468639095,16.54686666666667,0.4680473219190092,-2.8714999999999997,0.22838442737337986
2960000.0,17.013033333333336,0.19780194359229433,-1.9483333333333333,0.08321447523651698,19.1979,0.6241199884637576,-3.4464666666666663,0.13613650828814772,17.054666666666666,0.28734450094307085,-2.9042999999999997,0.20039687289642683,20.763,0.7944538249640449,-4.7208000000000006,0.37941065702832666,16.5547,1.3468134416713642,-2.8256666666666668,0.8370893792713469
3000000.0,16.770833333333332,0.6584069882847701,-1.9343666666666666,0.19859732011171644,18.916666666666668,0.8451919913381954,-3.4514333333333336,0.24320206048103735,17.153666666666666,0.6999700057065926,-2.8442666666666665,0.23058268702475374,19.651033333333334,0.357504371379645,-4.3237,0.24895643527867817,16.4505,0.9528340289193427,-2.7698,0.4909169040343453
3040000.0,16.520833333333332,0.36880970100521526,-1.8359333333333332,0.08753910872036309,19.114566666666665,0.7179224857570322,-3.413933333333334,0.09549720182055357,16.898433333333333,0.16729776514413458,-2.7975666666666665,0.09514810677161277,21.710933333333333,0.21338068224550114,-5.158233333333333,0.06743937194909896,16.34893333333333,0.40644020743797177,-2.703666666666667,0.3859192863223545
3080000.0,17.151033333333334,0.3131372684445106,-2.0665333333333336,0.19246700727368535,19.023433333333333,0.5554312218647988,-3.4578,0.037846267979815364,16.52866666666667,0.16014296390690666,-2.6823,0.2298712828229455,21.10673333333333,0.9934411317346503,-4.891533333333333,0.4461677885678835,16.111966666666664,0.6508048572515588,-2.5805666666666665,0.06016368413660264
3120000.0,17.40886666666667,0.5646417763109253,-2.0638,0.22169343697998817,19.062466666666666,1.0536317710134262,-3.437633333333333,0.20965664522950117,16.505233333333337,0.35652532713523766,-2.682533333333333,0.21805397395038595,21.398433333333333,1.2020313482693465,-5.043866666666667,0.5986735634346614,17.104166666666668,0.9978433889588529,-2.9623666666666666,0.7406230590223041
3160000.0,17.127633333333335,0.8997903768224149,-2.0320666666666667,0.24089491946120872,19.093766666666667,0.6219909341961691,-3.4636666666666667,0.24548488525546525,17.104166666666668,0.9380593241131161,-2.8577,0.4059884481115197,21.1224,0.3263169114015792,-4.9526,0.22287935450971383,16.919300000000003,1.7218791149981079,-2.9452,0.7202799779715293
3200000.0,16.9479,0.4084481933693259,-2.0313666666666665,0.18368949041490884,18.9401,0.6658679949259214,-3.4226666666666667,0.267676375913569,17.6146,0.1119929759702212,-2.947266666666667,0.13499297101042781,20.98436666666667,2.2011908918179315,-4.935266666666667,1.0004268966573997,16.84636666666667,0.7281985092602762,-2.8234333333333335,0.6636128104717556
3240000.0,17.203133333333337,0.954312682277437,-2.0016666666666665,0.3095170467393068,18.510433333333335,0.45573922611754886,-3.2960999999999996,0.038114126864807876,16.908833333333334,0.4239659603736552,-2.677466666666666,0.09557336914061829,20.773433333333333,1.040254751918436,-4.798866666666666,0.45687193197024284,16.9974,1.395804112330953,-2.9738333333333333,0.8275295375728624
3280000.0,16.505233333333333,0.6461692674696176,-1.790333333333333,0.28823867810471854,18.729200000000002,0.38232972680658783,-3.296466666666667,0.1329273569368706,17.2578,0.356363194882225,-2.9227666666666665,0.24688320225474134,21.838533333333334,0.6124245277764621,-5.286566666666666,0.26406393586065896,16.28386666666667,0.19892915545210857,-2.69,0.46541195372128846
3320000.0,16.645799999999998,0.5316590135290356,-1.7903666666666667,0.11333517645560101,18.591166666666666,0.5847258863965428,-3.2902333333333336,0.06344584742562398,17.291666666666668,0.5660384282204008,-2.9186,0.24730735263365425,20.807299999999998,0.0036769552621698513,-4.824633333333334,0.0564636953173356,16.20573333333333,0.597818463712485,-2.6924333333333337,0.2754653799582726
3360000.0,16.658833333333334,0.6305636226600939,-1.7769666666666668,0.18678965591155078,19.1953,0.45315165231961746,-3.500633333333333,0.15842567833389756,16.9974,0.46641680930258084,-2.7941333333333334,0.24868080388767885,21.1979,0.781253945056705,-4.957966666666667,0.37989158394936257,16.875,1.238311918163864,-2.9168000000000003,0.7312798096488101
3400000.0,16.54686666666667,0.45542097985148783,-1.8237333333333334,0.10192495060364541,18.770833333333332,0.8174552559960427,-3.3300666666666667,0.12741727599592698,17.335933333333333,0.35407661443378186,-2.966666666666667,0.032013157017423674,20.2604,0.3656823029169811,-4.583466666666666,0.19101173320563916,16.5677,0.6276633704993994,-2.7428666666666666,0.6739620381660149
3440000.0,16.6849,0.09228531844231758,-1.8683666666666667,0.05059013957504195,18.604166666666668,0.3300029932524183,-3.2509333333333337,0.1694758258736495,16.927066666666665,0.5979271378874037,-2.7618666666666662,0.2698783717817262,22.052066666666665,0.519503720444382,-5.380600000000001,0.18482512455471695,16.2422,0.20282333856503537,-2.6003666666666665,0.3977629305112393
3480000.0,16.914033333333332,0.7458566543839966,-1.9560000000000002,0.23242897122920517,18.533866666666665,0.2393701503715295,-3.1347666666666663,0.11713127488231116,16.98176666666667,1.06183879609331,-2.8379,0.5239112138521184,20.778633333333335,0.769653447266293,-4.728133333333333,0.3614178222254989,16.4922,0.7768814109416354,-2.803266666666667,0.602848269098913
3520000.0,17.567733333333333,0.6621148809349894,-2.1984,0.23484799339147008,18.531233333333333,0.48340595316519996,-3.2532666666666668,0.055757949109422,16.5703,0.6457895993794474,-2.7169333333333334,0.2763049321954921,21.395833333333332,0.11236228113660807,-5.053133333333333,0.08555708945233906,16.494833333333332,0.3737176860798664,-2.7757666666666663,0.41505447299788917
3560000.0,17.1901,0.5150899791945741,-2.0216333333333334,0.18738438806071572,19.098966666666666,0.18591342668624655,-3.4019666666666666,0.18046263386702033,17.34636666666667,0.20503323849778268,-2.9653666666666667,0.17718081787321724,21.036466666666666,0.41485564825477467,-4.935666666666666,0.24678590361327846,17.01823333333333,1.3480910264353658,-3.0131333333333337,0.8462708799327924
3600000.0,17.210933333333333,0.7094422237849177,-2.1806,0.2987231605796019,18.416666666666668,0.27515506335317425,-3.098433333333333,0.34200791738723757,17.385433333333335,0.24378538831430382,-2.9304666666666663,0.07096146058875126,21.84893333333333,0.7258854745910144,-5.3008,0.32720590866710636,16.466166666666666,0.14953051713799356,-2.685266666666667,0.42670412335585517
3640000.0,17.39063333333333,0.45879335459684056,-2.1229,0.21450600613191853,18.968733333333333,0.6520575911851824,-3.320933333333333,0.08950353189803302,17.169233333333334,0.4679924667010028,-2.8021,0.17264172921593038,21.9271,0.24889517204370765,-5.3247333333333335,0.12312999454054857,17.265633333333334,1.5081847020691985,-3.1229333333333336,0.8316448454051096
3680000.0,16.513033333333333,0.5521340195600658,-1.6798000000000002,0.10423716547693856,18.539066666666667,0.6165985854310375,-3.0696666666666665,0.06675559069388037,17.075533333333333,0.6223863234001491,-2.8376333333333332,0.21269935485458236,21.096366666666665,1.1146233932987812,-4.966133333333333,0.4831248377892496,15.971366666666668,0.29539194677956665,-2.5787666666666667,0.45998321950069243
3720000.0,16.953133333333334,0.2548043606813307,-1.8300999999999998,0.12438110789022586,18.458299999999998,0.7587624705180575,-3.1186333333333334,0.504788842542657,16.8698,0.7134559411764675,-2.7523666666666666,0.14716922987575293,20.716133333333335,0.9468534217196574,-4.780866666666667,0.4831327652818515,16.544300000000003,0.36677487191282104,-2.8825333333333334,0.40303765194940394
3760000.0,16.825533333333336,0.525313172201962,-1.920333333333333,0.0952180771819207,18.723966666666666,0.09053516198447815,-3.3339999999999996,0.26030401456758206,16.91146666666667,0.5731472081605418,-2.7808333333333333,0.1689288673442825,20.502633333333335,0.6790327548964197,-4.605833333333333,0.32067878355485596,16.2578,0.8052891157838906,-2.6900999999999997,0.6198922056831063
3800000.0,17.14323333333333,0.5785265498565204,-1.9580666666666666,0.21520163155102298,18.927066666666665,0.2841476885158294,-3.3017000000000003,0.16176946559842512,17.2604,0.6059638328041257,-2.863766666666667,0.15039869975797276,20.924466666666664,0.2621646344485756,-4.801566666666666,0.1594577129593368,17.289066666666667,1.453003109272502,-3.120433333333333,0.877244133003401
3840000.0,16.684900000000003,0.39399939932272443,-2.014466666666667,0.20472622911803193,19.0521,0.36982529208623166,-3.3889333333333336,0.3724055704321418,17.013,0.35065801953850434,-2.7177000000000002,0.25767865000163803,20.601566666666667,1.1217982508256803,-4.7016,0.5776975160064307,16.398433333333333,0.891925622210482,-2.7678,0.543726867707185
3880000.0,17.16926666666667,0.47960222638719724,-2.0256333333333334,0.11793798747175954,19.276033333333334,0.7670103968584063,-3.4465000000000003,0.3608869906217181,16.497366666666668,0.8285770385962242,-2.5787666666666667,0.22662228977358378,20.2682,1.0501750742931706,-4.522033333333334,0.466620576295369,16.908833333333334,1.2030470213402116,-2.9858,0.641262338412811
3920000.0,17.42183333333333,0.24364915121725506,-2.0471666666666666,0.02816195226818549,19.427066666666665,0.2501192426734811,-3.5217333333333336,0.3615623167434472,17.265600000000003,0.2801691036975112,-2.8598999999999997,0.15676889572446007,20.497366666666665,0.684438496547028,-4.622,0.2760488362590937,17.0078,1.8228054549695274,-3.0242333333333336,1.006261670187675
3960000.0,16.901033333333334,0.26949944135171855,-1.9567666666666668,0.1593509822861332,18.502599999999997,0.5106414267043623,-3.0836,0.2368549907995748,17.070333333333334,0.25131529112960044,-2.7504000000000004,0.17661693010580834,20.684900000000003,0.7423096029734939,-4.7297,0.33170723035029975,16.104166666666668,0.8429658527419066,-2.6198,0.5972240283176824
4000000.0,16.9974,0.5741693884792776,-1.9282666666666666,0.12740754381990985,18.416666666666668,0.13475390738511245,-3.1346333333333334,0.2209074818913012,16.695333333333334,0.09202015479713685,-2.6735666666666673,0.07435865039717243,20.791666666666668,0.336854838502019,-4.757933333333334,0.150713023841855,17.125,0.5966802158610598,-2.9987,0.5046159794008377
}\dataSotaStepsLjGseven

%% file: data_new/ablation_gnnmixer.tex
\pgfplotstableread[col sep=comma]{
0.0,39.919266666666665,0.11417417493558639,-19.622,0.21644234028180862,39.91406666666666,0.12152808545992944,-19.5733,0.1301116699864648,45.0,0.0,-17.275533333333332,0.07913256107458043,45.0,0.0,-16.843233333333334,0.26263588652141345
40000.0,39.84896666666666,0.13520528424913278,-19.224866666666667,0.28101139162358274,39.770833333333336,0.210238631612325,-19.34436666666667,0.28961139864000995,45.0,0.0,-16.750766666666664,0.20717724027722872,45.0,0.0,-16.837999999999997,0.6338533321413297
80000.0,39.45053333333333,0.31095475912457304,-18.855733333333337,0.34794764293241376,39.7578,0.34252252480676254,-19.0207,0.6018882343647084,45.0,0.0,-17.26406666666667,0.1547226622774513,44.830733333333335,0.23937921565768738,-17.0805,1.065635832105259
120000.0,39.34116666666666,0.20810231991872447,-18.568866666666665,0.3163385352575444,39.61193333333333,0.20889169974468344,-18.8182,0.13112447012921227,44.7552,0.27252659809029023,-15.842166666666666,0.6211464150173361,45.0,0.0,-16.1086,1.0509932540221174
160000.0,39.596333333333334,0.3996222994556528,-18.839233333333333,0.33908311993112095,39.203133333333334,0.6775820606309526,-18.446766666666665,0.5577843390495005,44.8099,0.26884199820712684,-15.440900000000001,0.6754534230179509,44.71356666666666,0.2041398268072385,-16.160966666666667,0.6628879861803367
200000.0,38.96616666666666,0.1399600022228577,-18.17186666666667,0.120789964630989,39.080733333333335,0.6639044174845921,-18.325266666666668,0.6799767463344277,44.5026,0.703429825924376,-14.334133333333334,0.2043081713708218,44.91146666666666,0.12520504072209762,-16.35653333333333,0.034977548354463824
240000.0,38.515633333333334,0.7354676713191084,-17.797666666666668,0.5370020877261302,38.33856666666666,0.8160475571653613,-18.0608,0.6594834544298035,44.143233333333335,0.6125671736407525,-14.347666666666667,0.9103915872975881,45.0,0.0,-16.281233333333333,0.16959037577514696
280000.0,38.54423333333333,0.7868027805972434,-17.614833333333333,0.5585012941395523,39.1198,1.0728796515297818,-18.614433333333334,1.2841110189110947,44.24743333333333,0.4310552349241967,-14.0849,0.8442401830442959,44.416666666666664,0.5374060806834586,-14.783366666666666,0.6840551456002818
320000.0,36.78126666666666,1.5352294168046108,-16.438433333333332,1.0606585889070153,39.286433333333335,0.14720272491439085,-18.502866666666666,0.1532515434035022,43.3724,0.9511565626471119,-13.569033333333332,0.2580365004327014,44.734366666666666,0.14838080139364893,-15.203666666666669,0.8946616728623668
360000.0,37.578133333333334,1.110248072379423,-17.089199999999998,0.7904904089656414,39.541666666666664,0.3610106215735051,-18.874200000000002,0.6168135915061096,43.49216666666667,0.3090346402733677,-12.973966666666668,0.1729404971530834,44.41146666666666,0.6634327161731537,-15.091166666666666,1.560964066068004
400000.0,37.833333333333336,1.0262745614871098,-17.2901,0.8005058026023301,39.291666666666664,0.39876096710797726,-18.863166666666668,0.5121137004828341,42.46093333333334,0.9465946028909205,-12.809899999999999,0.6853226733931012,43.46873333333334,1.2698397414197153,-14.268766666666666,2.3963668783296845
440000.0,37.8151,1.1287071926175847,-17.220566666666667,0.6796955315106575,38.768233333333335,0.9222695135130273,-18.318966666666665,0.8012090044981325,42.36456666666667,0.5646202519294606,-12.751566666666667,0.29117257578434247,43.72136666666666,0.9222257219478441,-14.449733333333333,2.0730393023020297
480000.0,37.3333,1.5150140879432985,-16.82003333333333,1.031750723554656,38.1432,0.5782093219587516,-17.472533333333335,0.5426610871949045,42.05466666666667,0.7020453752350241,-12.368733333333333,0.0064095414985954075,43.9453,0.5374484595444173,-13.954666666666668,0.623063834789199
520000.0,37.57033333333334,0.3005891585240919,-17.113033333333334,0.19215024214284984,39.21353333333333,0.172631482901844,-18.2426,0.22017231130790943,40.3958,1.021421078693798,-11.5362,0.2905575445013713,43.12756666666667,0.7497279165729976,-13.063833333333335,0.3752069858387803
560000.0,38.03126666666667,0.9833704535366571,-17.219933333333334,0.7661078006535528,38.9349,0.1684717384805709,-18.243866666666666,0.27322755270205706,41.71093333333334,0.17361262114898734,-11.624733333333332,0.18816188656461622,44.0573,0.6699336136265042,-13.768266666666667,0.9499534386250492
600000.0,37.77343333333334,1.574501305881397,-17.275533333333332,1.2114690595397886,38.59896666666666,0.5186241692101241,-17.7672,0.5176964425864506,41.7448,0.3467262897445193,-12.220033333333333,0.5238247692586605,43.1927,1.2036629123914477,-13.295066666666665,1.1692369544089662
640000.0,36.825500000000005,0.8886834344504607,-16.376433333333335,0.6141755684565205,38.296866666666666,0.7206054414325656,-17.5413,0.5424177172622585,42.08593333333334,0.6519826037215698,-12.380166666666668,0.18179443213573812,43.888000000000005,0.8131720646128141,-13.915899999999999,1.1747535996965486
680000.0,36.41403333333333,0.6402720064333778,-16.181900000000002,0.527540791471775,37.6276,1.1602524667789618,-17.167333333333332,1.0005132860464956,38.200500000000005,0.8245400657336212,-10.136700000000001,0.4127995881780893,44.205733333333335,0.8261554023843739,-13.968766666666667,0.9308291118973209
720000.0,35.328133333333334,0.9063905425121985,-15.390633333333334,0.5794568107767439,37.51043333333333,0.7395805312622962,-16.83526666666667,0.5208796683389448,39.171866666666666,0.7095448133995631,-10.814566666666666,0.06174713668575187,44.01043333333333,0.41954464469099056,-13.352633333333335,0.47961145616935497
760000.0,35.93226666666667,0.934509826355806,-15.889466666666666,0.6913259449942711,36.33853333333334,1.8131985905820946,-16.167833333333334,1.15414185821713,38.97396666666666,1.3651839249305915,-11.009633333333333,0.6285705069618064,42.9401,1.564468295193822,-12.946599999999998,1.9358419994066323
800000.0,36.2552,1.9865690574455246,-16.182566666666666,1.3590144721656041,37.32033333333333,1.1691102666367923,-16.7482,0.920654781482542,38.25783333333333,1.4299367383054244,-9.934366666666667,0.7250918301996119,41.947900000000004,2.000513665703555,-12.087266666666666,1.2296670994848793
840000.0,35.72396666666666,1.8159118211582366,-15.735666666666667,1.1380062809824718,36.01043333333334,0.6571715267383046,-15.936599999999999,0.4840154198645601,36.8776,0.5946076409420498,-9.2315,0.24105287110231005,42.6042,1.694854857109205,-12.438,1.608876782106075
880000.0,35.84636666666666,1.0918087938014724,-15.863433333333333,0.7818709200088949,36.424499999999995,1.358577655736567,-16.339299999999998,1.030080504944476,37.52863333333334,0.16649913580022851,-9.655466666666667,0.47810510931755923,42.2448,1.326524370928281,-12.200000000000001,1.1799509933326333
920000.0,35.994800000000005,0.9380472944722279,-15.853033333333334,0.6505846823349666,36.111999999999995,1.3073031706532345,-15.924766666666665,0.9014050895993182,35.802099999999996,0.3242691783071591,-8.750533333333333,0.8562684794437371,41.0677,1.505704315815913,-11.665333333333335,1.1071817987826371
960000.0,34.976533333333336,1.5469089012895656,-15.2712,0.9632296749304745,35.731766666666665,1.4325868940098847,-15.594699999999998,0.8019130002687316,34.763000000000005,1.964042083731066,-8.663033333333333,0.8671135886119853,41.0599,1.1220816221053922,-11.190633333333333,0.7376052256382735
1000000.0,34.8229,0.7405541753758921,-15.1113,0.4166165223159859,35.4427,0.8868829159853414,-15.5867,0.4659836263217841,37.3099,0.8718320136356565,-9.240900000000002,0.35234461350596313,40.97656666666666,1.1150435576942976,-11.342966666666667,0.8239401124407577
1040000.0,34.0755,2.275831946051084,-14.737766666666667,1.531700167207089,34.59896666666666,1.4155169664197687,-14.932666666666668,0.9149369535049338,34.82293333333333,0.8596802868250227,-8.1224,0.4221536292236117,40.69793333333333,1.8263135078317974,-11.328899999999999,1.0801711623627066
1080000.0,34.01303333333333,1.2039111161358875,-14.639066666666665,0.8259560332662318,34.859366666666666,1.8230417755925281,-15.217166666666666,1.2245191881804964,33.427099999999996,0.6611055336832894,-7.665366666666666,0.4262922341409575,39.91143333333334,1.5724557064950646,-10.780700000000001,0.8238062393548615
1120000.0,33.91146666666666,0.46039470264352744,-14.5276,0.48382220563618955,35.841166666666666,1.1273326227082334,-15.758833333333333,0.8006905408597139,30.921900000000004,0.9576158206713169,-6.065366666666667,0.40283621430496463,40.01043333333333,2.0206062000190843,-10.9565,0.9383914748120846
1160000.0,33.4401,1.1047392482693217,-14.309600000000001,0.7283754846689084,34.59896666666666,1.1314067860657187,-15.016533333333333,0.7453074101037473,32.32033333333333,0.39544056724395715,-6.848466666666667,0.14144118526401317,40.828100000000006,1.7892010954613227,-10.980466666666667,1.1677497715216598
1200000.0,32.875,2.538243002551175,-13.869133333333332,1.5752275228959427,34.09373333333333,2.0285976113780886,-14.714333333333334,1.3032943088786797,30.0651,2.2246779916802937,-5.694533333333333,1.3054754800029333,40.25256666666667,2.3330774273954615,-10.983833333333335,0.9961678316874568
1240000.0,34.1849,2.2689882605837046,-14.725766666666665,1.4779574516503802,33.5703,1.9453809978167944,-14.361333333333334,1.1956710593730295,30.2266,0.6654000500951791,-5.466133333333334,0.40642327962633035,41.145833333333336,2.576101015531462,-11.644800000000002,1.1557467283103162
1280000.0,33.218733333333326,1.9109816279133147,-14.140866666666668,1.3463323174049147,33.55466666666667,1.2153721085421618,-14.281233333333333,0.8011899580554466,29.388,0.6762177805017163,-4.9911666666666665,0.34333744657730286,40.49476666666666,2.14259818341087,-11.176066666666665,0.9714923445686824
1320000.0,31.734400000000004,3.299431858163866,-13.288533333333334,2.0901833991834837,31.218766666666667,1.8020353258345285,-12.870800000000001,1.2539739258320588,28.143199999999997,1.0630179396416606,-4.792433333333333,0.6300294825552917,39.41406666666666,2.2404203330823638,-10.221633333333335,1.2384159514835429
1360000.0,32.34636666666666,2.2348844747672203,-13.414833333333334,1.6567194679714357,35.447900000000004,2.9255141850052047,-15.484233333333334,2.1009485482091703,27.3099,1.2145620802028465,-4.3552333333333335,0.6763676531459959,39.80726666666667,1.0954965916068495,-10.882566666666667,0.7154171758879957
1400000.0,31.276033333333334,2.328357412903403,-12.805966666666665,1.6270337229312593,32.84636666666666,1.3152092161410003,-13.685566666666668,0.9238956374444509,26.70573333333333,0.5523555155473298,-4.104966666666667,0.30505490580476763,38.90103333333334,1.899743832438703,-10.358600000000001,1.3698463076807805
1440000.0,30.90363333333333,1.8098278856166274,-12.453766666666667,1.3912892590048351,31.695333333333334,1.306772822218494,-13.172933333333333,0.8522020782785157,25.434866666666665,1.1035606140529337,-3.405466666666667,0.5498872874406981,38.80206666666667,3.1477660833606342,-10.470066666666668,1.7906046619942537
1480000.0,32.00783333333333,3.944956968361278,-13.374733333333333,2.8157089815690988,32.2448,1.7539616928542086,-13.5099,1.239705975893746,25.638,0.9848375128246623,-3.5065000000000004,0.30628406205133596,39.278666666666666,1.5756739009776795,-10.233333333333334,0.6012148830123514
1520000.0,31.66926666666667,2.803900940158591,-13.137100000000002,1.8216559664217606,31.835933333333333,2.1376666755028864,-13.301533333333333,1.3053175509771127,25.23176666666667,0.1991229156967015,-3.243733333333333,0.19044133538237504,38.60156666666667,3.8123103181958076,-10.2948,1.8169656701948627
1560000.0,30.296866666666663,3.531960889112762,-12.266533333333333,2.2867162079181482,31.645866666666667,1.807114772841565,-13.017433333333335,0.9080757322798333,25.239566666666665,1.2136180682387505,-3.3723666666666667,0.21380817466962193,38.572900000000004,3.999237278115248,-10.664033333333334,1.9117277793195937
1600000.0,30.0755,2.9785871852272523,-12.047766666666666,1.9838422809174012,30.658833333333334,1.45780264172563,-12.441933333333333,0.967418387025777,24.307266666666663,0.37395116734081213,-2.7786333333333335,0.24846250331902303,37.947900000000004,3.7035150609477294,-9.7375,1.6487261587864335
1640000.0,30.335966666666664,2.7309654218893997,-12.229966666666668,1.8062622111851747,32.237,1.2355997005503025,-13.407966666666667,0.717675480670446,24.0547,0.879069443597414,-2.9573,0.4967570902027134,37.359366666666666,4.6158688790832105,-9.446100000000001,1.6602118920989173
1680000.0,29.8151,2.361862067663281,-11.876266666666666,1.5819389080773274,31.26563333333333,1.1678798292442407,-12.975900000000001,0.7419870124649528,23.77863333333333,0.23650178763712534,-2.8195333333333337,0.044463493140128055,37.63283333333333,3.993116497451861,-9.751033333333334,1.8843989763906743
1720000.0,29.25,2.186396135195999,-11.600266666666668,1.334606844313669,32.481766666666665,2.0040714580296006,-13.463000000000001,1.1803017777952662,23.505233333333337,0.8978689449035546,-2.7213333333333334,0.45912271658989917,37.00523333333334,4.296392899734483,-9.555466666666668,1.7464732946394825
1760000.0,29.35676666666667,2.417183493701342,-11.579566666666667,1.5162487424195576,32.1797,2.2013290848939424,-13.4005,1.4154656995726407,23.289066666666667,0.18084509639160407,-2.486466666666667,0.08876415693034863,37.09893333333333,4.214457989139555,-9.0284,1.7617055883432962
1800000.0,29.76823333333333,1.1169047865517558,-11.832799999999999,0.7538314267792239,32.1432,2.3166012993176035,-13.476433333333334,1.6331805887762556,23.210933333333333,0.7970598234901961,-2.446333333333333,0.20482796575554704,35.3776,4.46559818688008,-8.7276,2.2599604436066283
1840000.0,28.9974,2.900282063294304,-11.455733333333333,1.7756003629445696,33.0755,2.2617247415781323,-13.954299999999998,1.455490810688958,22.9349,0.701176729980871,-2.3645666666666667,0.17620867048915492,35.4141,4.686557995658079,-8.509133333333333,2.2892524578026676
1880000.0,28.9479,2.28836354774906,-11.324866666666667,1.3810362204599202,31.54426666666667,0.933659600116065,-13.166133333333335,0.622897243175434,22.39323333333333,0.7761738006972975,-2.1544333333333334,0.3668337528393785,35.263,3.9980828305576663,-8.1987,1.8934508725252597
1920000.0,29.0,1.9848804968225837,-11.416533333333334,1.2204166783885282,31.895799999999998,1.020726695382592,-13.105833333333335,0.8149962631134504,21.4323,0.6766912491429645,-2.0317666666666665,0.19461953196486262,34.02863333333334,4.066206292793758,-7.701066666666667,1.514790382271495
1960000.0,29.203100000000003,1.5090956651805298,-11.548566666666668,0.8799659740896553,31.398433333333333,1.92843112803013,-12.839166666666666,1.1605960575880347,23.10936666666667,0.5027158662925034,-2.541933333333333,0.1425420249922419,33.604166666666664,4.542811846814213,-7.771866666666667,2.332373206747916
2000000.0,28.567733333333337,2.0481643575542354,-11.208466666666666,1.17799126293685,30.468733333333333,1.270858393195543,-12.3496,0.7000231758068208,21.54686666666667,0.2661838504158782,-1.9103999999999999,0.08774216014360861,34.466166666666666,2.5899935525445286,-7.751833333333334,1.2640345731910272
2040000.0,27.47136666666667,2.214867265147557,-10.666033333333333,1.2448824772733451,30.927099999999996,1.1176025799302123,-12.663400000000001,0.6923179038563138,22.3125,0.9984520252203734,-2.208066666666667,0.17296532086583788,32.61716666666667,2.912383995210032,-7.3275999999999994,1.5303134471952686
2080000.0,28.57813333333333,1.895840756908542,-11.181899999999999,0.951930631226176,29.854166666666668,1.301280052700245,-11.982433333333333,0.732031430715618,21.7578,0.5846606366089645,-2.0424333333333333,0.18832869020825146,32.53903333333333,2.867648265662224,-6.7711,1.4984939795229961
2120000.0,26.648433333333333,2.855819294392105,-10.034766666666668,1.5701777931884724,31.01823333333333,1.0927836209525748,-12.637866666666667,0.6892071064314096,21.39323333333333,0.6218569413905055,-1.7705666666666666,0.26049125811734175,32.25,3.2748992411981166,-7.0359333333333325,1.371233919585649
2160000.0,28.6276,2.1010024527988223,-11.088033333333334,1.1335695283287903,30.08073333333333,1.2127251992470878,-12.0539,0.6069598394182818,21.541666666666668,1.0282588044305219,-1.9565000000000001,0.21991372550767882,31.921899999999997,3.14784463509028,-6.8263,1.6926823939140698
2200000.0,27.20573333333333,2.3798646590836947,-10.3397,1.3003000986951687,29.651033333333334,0.8677153616762164,-11.867433333333333,0.5485628516607934,21.218733333333333,0.55820344160729,-1.8557333333333332,0.15442159463264488,31.528633333333335,1.9427572044791293,-6.134633333333333,1.0314744344330058
2240000.0,27.130233333333337,1.6764614725334095,-10.323066666666668,0.8631357186960162,30.281233333333333,1.2458105054764772,-12.064433333333334,0.7693615809372226,21.34636666666667,0.9041139099816028,-1.6994666666666667,0.22408750275035175,31.86196666666667,1.7709060098027671,-6.4497333333333335,1.0613501035107231
2280000.0,28.567700000000002,1.4560204279702487,-11.2129,0.5997493865496377,29.3776,1.7191606265849617,-11.585533333333332,1.0755451217354337,20.955699999999997,0.33671027110361007,-1.5453000000000001,0.08667252544299524,28.838533333333334,2.671521297355165,-5.005199999999999,1.2856151082912282
2320000.0,27.934866666666665,2.831435378429503,-10.8609,1.533442795368209,29.45573333333333,1.686318124066618,-11.545833333333334,1.0255515665024142,21.289066666666667,0.10616510202928685,-1.7060000000000002,0.18377020070366867,30.526,3.3458452125982574,-5.842433333333333,1.6373492527727724
2360000.0,27.8125,1.785931056900013,-10.673933333333332,0.9607349594046334,29.4974,2.189336010453094,-11.798933333333332,1.2433788972884416,20.747400000000003,0.5987945613202128,-1.5705666666666669,0.24819326251039853,29.257833333333334,4.2071408789765465,-4.9883,1.5722533383650361
2400000.0,26.1328,1.802163684759701,-9.855466666666667,0.9458635889433997,30.3828,1.6396706152964584,-12.113533333333331,0.8689076526817389,20.723933333333335,0.884387895037516,-1.5156333333333334,0.3029156354865529,29.73176666666667,3.1583535587742895,-5.171633333333333,1.5820742867795081
2440000.0,26.414066666666667,2.584904079888115,-9.9362,1.3239417383958656,28.343766666666667,1.6166140795570927,-11.101700000000001,1.0084677915861602,21.2396,0.6941593044827676,-1.6645666666666667,0.33427627628787665,29.375,3.827073478085659,-5.182300000000001,1.7151386552307268
2480000.0,27.179699999999997,1.4819181646321309,-10.300933333333333,0.7965295404998411,29.406266666666667,1.463465226402352,-11.469133333333332,0.7176635717536609,21.385433333333335,0.03681669307377918,-1.8125,0.11861073588283094,27.585933333333333,3.3616979760564782,-4.316166666666667,1.829729313921112
2520000.0,27.330699999999997,2.5303029950317546,-10.483600000000001,1.3305377784940946,29.244799999999998,2.1308292392086856,-11.477200000000002,1.167917850992383,21.23176666666667,0.799277058069075,-1.7244666666666666,0.2149231852442997,27.1953,1.9236479459610063,-4.209133333333334,1.182938742661212
2560000.0,26.619833333333336,3.111592289416394,-10.0116,1.6265548930177554,29.335933333333333,1.1505068805627474,-11.5586,0.7252167308237364,20.968733333333333,0.3930446737402205,-1.7403666666666666,0.07866995755828403,27.864566666666665,3.9164112589064093,-4.4944999999999995,1.7631373986920778
2600000.0,26.458333333333332,1.2466036240744518,-9.912333333333333,0.5001723791743092,28.403633333333335,1.0948448789161354,-10.982933333333333,0.6140710997560102,20.84633333333333,0.9167891845396567,-1.619,0.3001685637548787,27.03386666666667,2.207539303286707,-3.960133333333333,1.1827182936871408
2640000.0,26.854200000000002,2.0332691033571204,-10.192466666666668,1.054180750894056,28.83073333333333,1.043927536863656,-11.317966666666669,0.6008889877136668,20.520833333333332,0.5069371843620154,-1.4958,0.09820919848296626,27.77863333333333,2.0663910477502134,-4.379966666666667,0.9695558169709584
2680000.0,26.395833333333332,1.741297807830572,-9.891133333333334,0.8967716035250496,28.5703,1.5242733285077195,-11.199333333333334,0.9832900736247103,20.364566666666665,0.9399934550599576,-1.5450666666666668,0.2421324752187438,27.226566666666667,2.807528951864176,-4.308599999999999,1.4045222200686847
2720000.0,26.093733333333333,1.6373114554727273,-9.757933333333332,0.7693070489444092,28.807299999999998,1.5163356510570698,-11.326033333333333,0.7286285488657591,20.263,0.8168108267320329,-1.3187333333333333,0.21197075794133072,26.526033333333334,3.229153817264758,-4.130966666666667,1.5159371168429854
2760000.0,26.16926666666667,2.635695608795184,-9.857566666666669,1.2601330732197389,28.695333333333334,2.120664188932851,-11.411466666666668,1.311586114426177,20.843733333333333,0.5868010584706042,-1.5668999999999997,0.2527255164534572,25.474000000000004,2.168819038094234,-3.6005333333333334,1.093803139915446
2800000.0,25.72136666666667,1.4724123297802452,-9.564933333333334,0.6383454098074353,27.89323333333333,1.9026072608105153,-10.762766666666666,1.08702987488334,20.338533333333334,0.7080634450543403,-1.3479,0.21212793309698746,26.0,2.1920667067100545,-3.6583333333333337,1.2712631024125398
2840000.0,26.166700000000002,1.6878872948156223,-9.801433333333334,0.8154178656097473,28.010433333333335,1.004587491903462,-10.9185,0.6764138723198001,20.23436666666667,0.47737120659806054,-1.3656333333333333,0.05304025724757457,25.302066666666665,3.352171258884141,-3.3541666666666665,1.4284905701084936
2880000.0,25.763033333333336,2.5947473150364537,-9.494,1.2440320360290835,27.72136666666667,0.35493858936754386,-10.807933333333333,0.18053864467814698,20.408866666666665,0.19563437552968352,-1.4645666666666666,0.07812427421885092,24.817733333333337,1.840409302905803,-3.1401000000000003,0.941428556326324
2920000.0,25.953166666666664,2.4954168928034632,-9.7004,1.3112464019651937,27.4479,1.7844872204641873,-10.626033333333334,0.9994603621732859,19.687466666666666,0.6409538794293674,-1.2109333333333332,0.13755932861456138,24.35676666666667,2.975137897457678,-3.0325333333333333,1.2691247622760427
2960000.0,25.5677,1.7125604398093524,-9.502366666666667,0.8120607544321345,27.479166666666668,0.32415317709722086,-10.677866666666667,0.21783648811793638,20.085933333333333,0.3030116866099747,-1.4116999999999997,0.06947234461760064,24.963533333333334,2.6327049743984268,-3.308866666666667,1.2902870102767403
3000000.0,25.3099,1.4501932721767345,-9.343733333333333,0.663457973214749,27.90106666666667,0.7092227874386318,-10.987366666666667,0.37062110331472226,20.1224,1.071709525322355,-1.3674666666666668,0.33661840248104213,24.23176666666667,1.1622884275810754,-2.9479333333333333,0.5010925019949466
3040000.0,25.942700000000002,2.2760069698194405,-9.654433333333332,1.1537846400241059,27.76823333333333,1.3348688832823832,-10.735433333333333,0.8061021330382949,20.252599999999997,0.5788556008769953,-1.3588666666666667,0.16739259906645285,23.565066666666667,1.2281626991386585,-2.5781333333333336,0.6970400339212153
3080000.0,24.807299999999998,0.37952921714496,-9.078666666666665,0.14152376321860485,27.049466666666664,0.7101962185825059,-10.3651,0.4406255628838014,20.414066666666667,1.0220280535397375,-1.4364666666666668,0.304464123046086,24.255200000000002,1.9504720095402541,-2.8557,0.8681367672588616
3120000.0,24.4974,1.1951012118923936,-8.9112,0.5502949451582005,27.487000000000005,0.3848714330786321,-10.5772,0.31525156092661377,19.83073333333333,0.6887282595883196,-1.1885333333333332,0.1586752728758398,23.2422,2.2249089254768757,-2.4523333333333333,0.9385207379464534
3160000.0,25.499966666666666,1.7560147196295244,-9.494766666666665,0.9402991734312837,27.59633333333333,0.6648432914771895,-10.733866666666666,0.4093047628465728,19.835933333333333,0.8552895701976554,-1.2325333333333333,0.27770440319799683,22.4974,0.8802312991481268,-2.3521,0.5542787445560822
3200000.0,25.75,2.164641942677818,-9.644533333333333,1.1781638661729341,27.33073333333333,1.1583920474328009,-10.570333333333332,0.6120885466081591,19.9349,0.7107175106890213,-1.4463666666666668,0.3818353077207793,22.713533333333334,1.3364629695165111,-2.363,0.6581971943625
3240000.0,25.528633333333335,1.8934453540804628,-9.455966666666667,0.9944492926014664,27.825499999999995,0.9131167176215761,-10.758566666666667,0.4862858510062663,19.859333333333332,0.475141700501604,-1.3804666666666667,0.15061651820287034,21.98436666666667,0.45411687433474185,-2.1544333333333334,0.33897009045374804
3280000.0,25.067700000000002,1.8508041549553536,-9.262233333333333,1.0295495530678562,27.789066666666667,0.34801810617009804,-10.815766666666667,0.235374514810381,20.221366666666665,0.4645765838074732,-1.3706000000000003,0.17963865582514996,21.9922,0.9063337611866096,-2.110666666666667,0.5253913610083651
3320000.0,24.32813333333333,2.027676809114861,-8.887233333333334,0.9092262583586599,26.682299999999998,0.22335126296187852,-10.053633333333332,0.18891102550024852,19.91146666666667,0.785537278272366,-1.2258,0.2076498976643138,21.359400000000004,0.8534054409638283,-2.0492000000000004,0.32490727908127875
3360000.0,25.1172,1.9060809863871646,-9.181633333333332,0.9033315166033394,27.140599999999996,0.6184360004613794,-10.475566666666667,0.37684976080955923,19.817700000000002,0.4629196906591901,-1.1979,0.17016347042378596,21.5,0.7421638812733116,-1.8838333333333332,0.3625665946492528
3400000.0,24.333333333333332,1.0728566736625258,-8.877733333333333,0.5135406334675204,27.002566666666667,0.3940470389292228,-10.406933333333333,0.205656677229038,19.48176666666667,0.46248627607265924,-1.1179666666666668,0.16793336641524093,21.9896,0.292326746410017,-2.1164,0.2088244717460096
3440000.0,25.17186666666667,2.546918059581467,-9.363033333333332,1.306238876401335,26.72133333333333,0.19375588306480554,-10.165766666666668,0.26724403745557285,19.72393333333333,0.27093567912370237,-1.3151,0.12396604373779141,21.901033333333334,0.6030519177944432,-1.9859333333333333,0.24802005743263766
3480000.0,24.8099,2.1295146692771723,-9.138266666666667,1.0120234066243505,25.72136666666667,1.4425182640861864,-9.6027,0.775233659399969,19.604166666666668,0.7826369500321041,-1.3284,0.1347427425380182,21.781266666666667,1.0107344436376715,-2.023166666666667,0.36743033202076403
3520000.0,24.666666666666668,2.124372522783035,-9.171633333333334,1.1088647928800375,25.875,0.8046702471613241,-9.8181,0.3469738414731963,19.838499999999996,0.6709702079824407,-1.3020666666666667,0.09927686314320956,21.583333333333332,0.4265732475854001,-1.8799666666666666,0.3430861537411396
3560000.0,25.205699999999997,1.930617036079398,-9.405833333333334,0.9652689066898519,27.502566666666667,0.4679924667010028,-10.724466666666666,0.1669044104337037,20.460966666666668,0.41059162463720866,-1.4450333333333332,0.24904127279540544,21.898433333333333,0.411974807751909,-1.9637666666666667,0.20268419332109305
3600000.0,24.617166666666666,2.624572125551557,-9.064066666666667,1.4144517343793988,27.59636666666667,1.2321585675372932,-10.747533333333331,0.6964672393986349,19.6172,0.5524285293139727,-1.2924666666666667,0.1611413734030532,22.39323333333333,0.7545817089988048,-2.1021,0.30310672487865836
3640000.0,24.270833333333332,1.57921825456634,-8.847533333333333,0.7529347086936261,25.84636666666667,1.1766641954648278,-9.705066666666667,0.5098670044454948,19.4974,0.39284748524925966,-1.2783999999999998,0.1670364231737098,21.328133333333334,0.4630163232063817,-1.8369666666666669,0.12227666262301336
3680000.0,23.914066666666667,0.41324451465069534,-8.763933333333334,0.15567704462194223,27.445300000000003,0.8582835118226773,-10.536200000000001,0.21881951466905333,19.3828,0.254213964997992,-1.2164,0.18471303870237926,21.265599999999996,0.5276075498575309,-1.8075333333333334,0.1537481562671747
3720000.0,23.343766666666667,1.8085255455450133,-8.371733333333333,0.8783091685480434,26.942700000000002,0.4729775117980424,-10.339933333333333,0.25878435467049066,19.752599999999997,0.5794652074686334,-1.1585666666666667,0.1263166479746654,21.682299999999998,1.1079091749778058,-1.9005,0.307192588886299
3760000.0,24.64063333333333,2.1100740355626275,-9.137500000000001,1.1823919739240452,26.908866666666665,0.6626086946473174,-10.3857,0.4186874490595585,20.2422,0.44748712458200063,-1.5036333333333334,0.3131757476064979,21.127599999999997,0.2934719066622909,-1.7283666666666668,0.12407944587597453
3800000.0,23.66926666666667,1.1986424886326854,-8.582433333333332,0.7232976396723241,26.95313333333333,0.7486210849881862,-10.326166666666667,0.4847532591145957,19.484366666666663,0.4538305581993753,-1.1249666666666667,0.09607172089410887,21.682299999999998,0.9632490885885474,-1.9062333333333334,0.37483747885657787
3840000.0,23.3255,1.0409976080664163,-8.388166666666665,0.5232614026999849,27.15363333333333,0.8306144325471098,-10.507033333333334,0.39250992627221826,19.705699999999997,0.22974604820685487,-1.3278666666666668,0.09476927537738983,20.958333333333332,0.5570430523956141,-1.612,0.1789609082080963
3880000.0,24.1849,0.34619948006893403,-8.879966666666666,0.16577688486503653,26.791666666666668,1.1810469771445253,-10.263699999999998,0.7604385620592016,19.528633333333335,0.6051808838869771,-1.0729333333333333,0.14617173309349366,21.04426666666667,0.3807962651544191,-1.7981666666666667,0.2516640662117304
3920000.0,23.825533333333336,2.469953165997732,-8.7181,1.3467899786776951,27.429666666666666,0.858844950435694,-10.607433333333333,0.6281151557548097,20.585966666666668,0.691133061637836,-1.5255333333333334,0.17935031518108785,20.593733333333333,0.7579592000159961,-1.5854,0.29202466790780984
3960000.0,23.179666666666666,0.7568600766030382,-8.346633333333335,0.40873333061490735,26.59113333333333,1.3466939007147185,-10.147533333333334,0.6993495374671781,20.049500000000002,0.22120492761238456,-1.1859333333333335,0.09458295594638369,21.166666666666668,1.0133655422512757,-1.814066666666667,0.5533411866912573
4000000.0,23.61196666666667,0.9539533403450906,-8.587266666666666,0.4909344989122502,26.557299999999998,0.8370134885412543,-10.1413,0.44748598488295305,19.736966666666664,0.12541399000466089,-1.2221333333333335,0.10643888178459761,21.382800000000003,0.8116201574628367,-1.7853999999999999,0.2779257574725068
}\dataAblationGnnmixer

%% file: 1_intro.tex
\section{Introduction}
\label{sec: intro}
\IEEEPARstart{C}{ooperation} in multi-agent reinforcement learning (MARL) can be significantly improved by inter-agent communication: by sharing local observations and intentions, agents mitigate partial observability and coordinate toward shared objectives~\cite{dial,tarmac}.
When agents are deployed to physical environments, for example, to conduct safety-critical tasks such as search-and-rescue~\cite{sar} or firefighting~\cite{firefight}, they communicate over wireless channels that are both (1)~\textit{stochastic}: transmissions fail due to path loss, fading, contention, and interference; and (2)~\textit{dynamic}: link quality varies with mobility and obstacles.
However, most MARL communication approaches assume idealized channels that deliver messages reliably and synchronously.
Value decomposition methods~\cite{vdn, qmix} offer high sample efficiency under the Centralized Training with Decentralized Execution (CTDE) paradigm, making them attractive for the added complexity that realistic wireless channels introduce; yet two gaps persist.
On the communication side, methods such as SchedNet~\cite{schednet}, VBC~\cite{vbc}, and RMADDPG~\cite{rmaddpg} address bandwidth constraints but still assume perfect network reliability, abstracting away the MAC/PHY stochasticity of real wireless channels.
On the value decomposition side, existing mixers, from VDN's additive decomposition~\cite{vdn} to QMIX's monotonic mixing network~\cite{qmix}, treat agent utilities as an unstructured collection, without leveraging the \emph{relational structure} of the realized communication topology for credit assignment.
To the best of our knowledge, no prior work addresses both: learning communication under realistic wireless stochasticity \emph{and} exploiting the resulting communication structure for topology-aware credit assignment.

We show that the realized communication graph (who successfully communicated with whom at a given step) provides a \emph{graph-structured relational inductive bias} for value decomposition.
Since value decomposition serves as an implicit credit-assignment mechanism~\cite{implicitVD}, the communication graph is a natural structural prior: it encodes which agents shared information and thus which utilities should be coupled during credit assignment.
Concretely, we design a GNN-based mixer that propagates individual utilities along the edges of this directed communication graph: multi-hop propagation reshapes the mixing of \emph{every} agent's utility, not only those that directly communicated. Hence, structurally different topologies induce different global $Q_{\text{tot}}$ values and correspondingly different credit assignment.
This relational inductive bias is architecturally unattainable by QMIX-style mixers, regardless of their state input.
Because the mixer is used exclusively during centralized training, communication actions can be treated as fully differentiable graph edges without conflicting with decentralized execution.

Posing this investigation under realistic channels requires addressing non-differentiable channel stochasticity, joint game/channel state influence, and variable-size unordered message reception.
Following~\cite{LAUREL_ACML}, we defer communication to the end of each step, yielding an augmented MDP that separates channel stochasticity from the agent computation graph, and employ a stochastic receptive field encoder that is permutation invariant and provably asymptotically lossless.

We refer to the complete framework as \coolname{} (\textit{Communication-enhanced vaLue decOmposition oVer rEalistic wiReless channels}).
Our main contributions are as follows:

\begin{enumerate}
\item \textbf{Communication-graph-conditioned GNN mixer.}
We introduce a GNN-based centralized mixer whose weights are generated by a Permutation-Equivariant Hypernetwork (PEHypernet), explicitly conditioning $Q_{\text{tot}}$ on the realized inter-agent communication graph.
We prove that this mixer is permutation invariant (Theorem~\ref{theorem: PI}), monotonic and therefore IGM-consistent (Theorem~\ref{theorem: monotonicity}), and represents a strictly larger monotone function class than QMIX-style graph-agnostic mixers (Theorem~\ref{thm: exp power}).

\item \textbf{Behavioral evidence of communication-aware coordination.}
Through positive signaling and positive listening analysis, we demonstrate that agents learn genuine communication strategies---not merely incidental message exchange---and that the communication-graph-conditioned mixer exploits these strategies for differentiated credit assignment.

\item \textbf{End-to-end CTDE framework under realistic wireless channels.}
We provide a complete CTDE framework that operates under realistic $p$-CSMA wireless channels without idealized communication assumptions, integrating the augmented MDP and message encoder of~\cite{LAUREL_ACML} with the proposed GNN mixer.
\end{enumerate}

\parag{Relation to~\cite{LAUREL_ACML}}
The present paper extends the preliminary investigation of~\cite{LAUREL_ACML}, which introduced realistic wireless channels into cooperative MARL via an augmented MDP and a message encoder with provable injectiveness.
The key new contributions here are:
(i)~the communication-graph-conditioned GNN mixer and its PEHypernet weight generation (Sections~\ref{sec: method enhanced mixer}--\ref{sec: method complete});
(ii)~theoretical guarantees on permutation invariance, monotonicity, and strict expressive-power improvement over QMIX-style mixers (Theorems~\ref{theorem: PI},~\ref{theorem: monotonicity},~\ref{thm: exp power});
(iii)~extended experimental evaluation with more complex environments (larger field, more agents and obstacles), behavioral analysis (positive signaling/listening, speaker consistency), and a mixer ablation isolating the GNN contribution.

The remainder of this paper is organized as follows.
Section~\ref{sec: prelim} introduces preliminaries.
Section~\ref{sec: related} surveys related work.
Sections~\ref{sec: method mdp}--\ref{sec: method how} formulate the augmented MDP and describe the decentralized agent architecture.
Section~\ref{sec: method enhanced mixer} introduces the communication-enhanced GNN mixer with its theoretical properties.
Section~\ref{sec: method complete} presents the training algorithm.
Section~\ref{sec: exp} reports experimental results, and Section~\ref{sec: conclusion} concludes.

%% file: 2_related.tex
\section{Preliminaries}
\label{sec: prelim}
\parag{Dec-POMDP}
We consider cooperative multi-agent reinforcement learning (MARL) problems formulated as a Decentralized Partially Observable Markov Decision Process (Dec-POMDP), represented by $\langle \mathcal{S}, \bm{\mathcal{A}}, \mathcal{T}, \bm{\Omega}, \mathcal{O}, \mathcal{R}, \gamma \rangle$.
At each step $t$, agent $i \in \{1,\dots,N\}$ selects an action $a_i$ according to a local policy $\pi_{\theta_i}(a_i \mid \tau_i)$ based on its action--observation history $\tau_i$.
The joint action $\bm{a}$ induces a state transition $s_{t+1} \sim \mathcal{T}(\cdot \mid s,\bm{a})$, a joint observation $\bm{o} \sim \mathcal{O}(\cdot \mid s_{t+1}, \bm{a})$, and a shared team reward $r_t = \mathcal{R}(s,\bm{a})$.
The objective is to maximize the discounted return $J = \mathbb{E}_{\bm{\pi}_\theta}\left[\sum_{t\ge 1}\gamma^t r_t\right]$.

\parag{Wireless communication environment}
In many MARL applications with mobile agents (\eg, search-and-rescue), inter-agent communication forms a mobile ad-hoc network (MANET)~\cite{manet_overview}.
Realistic wireless links are both \emph{dynamic} (topology and link quality vary with mobility and obstacles) and \emph{stochastic} (contention, fading, interference, and noise cause packet loss and variable latency).
Under common MAC protocols such as $p$-CSMA~\cite{gaiyi}, agents contend for channel access; interference can prevent correct decoding when SINR falls below a threshold; and path loss with log-normal fading (Appendix~\ref{appendix: pathloss}) produces location-dependent RSS variations.
These effects lead to message dropouts and out-of-order receptions, which complicate learning communication-aware policies.

\section{Related Work}
\label{sec: related}
We organize related work into four areas: learned communication, communication under constraints, value decomposition, and graph-structured factorization.

\parag{Learning to communicate in MARL}
Early neural communication approaches such as CommNet~\cite{nips16_commnet} and BiCNet~\cite{bicnet} learn message content end-to-end, but typically assume an \emph{idealized} communication layer in which messages are delivered reliably and synchronously. TarMAC~\cite{tarmac} improves message aggregation via targeted attention, while MAGIC~\cite{magic} further introduces structured communication scheduling together with attention-based aggregation. Because these methods assume reliable delivery, their learned communication strategies may degrade under packet loss and asynchronous reception; in particular, the mixer receives no signal about which messages were actually delivered. Recent surveys~\cite{comm_madrl_survey, robust_comm_survey} emphasize that bridging the gap between idealized assumptions and non-ideal channels remains a key open challenge. In contrast, \coolname{} explicitly conditions the value mixer on the realized communication graph, enabling differentiated credit assignment based on which messages succeeded.

\parag{Communication under constraints and uncertainty}
A growing line of work incorporates communication constraints. These methods can be grouped by mechanism.
\emph{Budget-based selection} methods enforce bandwidth limits: VBC~\cite{vbc} reduces redundancy by transmitting only messages with high embedding variance, and SchedNet~\cite{schednet} schedules a fixed number of messages based on learned importance scores.
\emph{Event-triggered gating} methods learn when to communicate: ETCNet~\cite{etcnet} formulates communication as a penalty-constrained MDP and learns gating policies that control bandwidth occupancy.
\emph{Partner and intention selection} methods determine whom to communicate with: I2C~\cite{i2c} infers which agents should exchange information based on intentions, R-MADDPG~\cite{rmaddpg} limits communication in partially observable settings using recurrent actor--critic learning, and ECOM~\cite{ecom} selects cooperative partners via sequential Q-learning conditioned on task similarity and instantaneous channel-state features.
A complementary thread addresses robustness to corrupted or adversarial messages: MAGI~\cite{magi} regularizes message encoders via adversarial training, while~\cite{lossy_comm_prior} imposes information-theoretic bottlenecks that force compact, noise-resilient representations.
While the above methods significantly improve communication efficiency, partner selection or message robustness, they typically abstract the wireless channel as a reliable selection/budget mechanism, rather than explicitly modeling the MAC/PHY stochasticity (\eg, contention-driven interference, path fading, SINR-threshold decoding) that governs packet \emph{delivery}. In contrast, \coolname{} targets this physical delivery stochasticity directly and integrates it into the learning formulation.

\parag{Value decomposition under CTDE}
Value decomposition is a widely used CTDE paradigm for cooperative MARL~\cite{ctde}. VDN~\cite{vdn} decomposes $Q_{\text{tot}}$ as a sum of per-agent utilities, and QMIX~\cite{qmix} generalizes to monotonic mixing to satisfy the IGM property.
Several extensions enlarge the factorization class beyond QMIX's monotonic constraint: WQMIX~\cite{wqmix} applies weighted projections, QTRAN~\cite{qtran} learns factorization-error corrections, and QPLEX~\cite{qplex} introduces dueling-style decomposition.
Others introduce auxiliary structure: RODE~\cite{rode} employs role-based decomposition, CDS~\cite{cds} promotes diversity shaping, and FOP~\cite{fop} derives richer IGM-consistent factorization from maximum-entropy objectives.
More recent mixer designs include XMIX~\cite{xmix} (topology-aware encoders with temporal credit signals) and HyperMARL~\cite{hypermarl} (gradient-variance reduction in hypernetwork-based weight generation).
Closest to our setting, NDQ~\cite{ndq} couples communicated messages with value decomposition, but assumes an idealized channel without contention-induced interference or fading-driven loss.
Our work is orthogonal to these advances: rather than enlarging the factorization class, we introduce a \emph{graph-structured relational inductive bias} by replacing the flat hypernetwork with a GNN mixer that propagates utilities along the realized communication graph, while retaining the monotonic class with clean IGM guarantees (Theorem~\ref{theorem: monotonicity}). This architectural change enables topology-dependent credit assignment (Theorem~\ref{thm: exp power}). Indeed, recent studies show that QMIX remains a strong baseline across diverse cooperative benchmarks~\cite{sota_qmix1, sota_qmix2}.

\parag{Graph-structured factorization and GNNs in MARL}
Graphs provide a natural abstraction of inter-agent interactions.
Graph-based MARL mixers fall into two categories: those that \emph{infer} an interaction graph from latent states or spatial proximity, and those that \emph{exploit} a realized communication graph whose topology is determined by physical delivery outcomes.
Prior work addresses the former; \coolname{} addresses the latter.

\textit{Coordination graphs}
DCG~\cite{dcg} and DICG~\cite{dicg} model $Q_{\text{tot}}$ via pairwise payoff terms on a static or dynamically inferred interaction graph; DMCG~\cite{dmcg} composes meta coordination graphs from base graphs, and DDFG~\cite{ddfg} generalizes to higher-order factor graphs. Unlike such coordination-graph methods, \coolname{} does not factorize payoffs over pre-specified pairwise terms; instead, its graph is induced by realized communication outcomes and used to condition value mixing.

\textit{GNN-based mixers (graph inference)}
At the individual-agent level, DGN~\cite{dgn} uses graph attention to aggregate neighboring observations for Q-value estimation, while DPN~\cite{pi_pe_marl} addresses ordering sensitivity through permutation-invariant and permutation-equivariant architectures. In value decomposition, VGN~\cite{vgn} incorporates graph attention into the Q-value mixer and proves IGM guarantees, SVMIX~\cite{svmix} improves robustness to volatile topologies via stochastic edge sampling before a monotonic mixer, and GCM~\cite{gcm} learns an interaction graph from hidden states for interpretable credit assignment.
These methods construct graphs from spatial proximity or learned latent interactions under idealized communication.
\coolname{} addresses a complementary problem: \emph{graph exploitation}, where the graph is induced by realized wireless delivery outcomes and reflects both agents' communication decisions and the stochastic channel realization.

The above gaps motivate the design of \coolname{}, whose technical framework begins with an augmented MDP that couples game and wireless dynamics.

%% file: 4_method.tex
\section{Problem Setup: Augmented MDP via Causal State Alignment}
\label{sec: method mdp}

\begin{figure}
 \centering
 \resizebox{\columnwidth}{!}{\input{./diagrams/env_2026.tex}}
\vspace{-.2cm}
\caption{Two computational interfaces for agent–environment interaction under realistic wireless channels. (a) The instantaneous-oracle interface embeds non-differentiable channel arbitration inside the agent's forward pass, blocking gradient flow. (b) The proposed causal state alignment interface buffers messages for consumption at the next step, yielding an end-to-end differentiable decision interface.}
\label{fig: 2env}
\end{figure}

In this section, we formulate the augmented MDP that couples game and wireless dynamics via causal state alignment.

Under a realistic wireless channel, message delivery outcomes are stochastic and governed by non-differentiable arbitration events (contention resolution, SINR-threshold decoding).
To enable end-to-end learning, we require a well-defined communication graph at each step, specifying which agents successfully received whose messages, that is resolved \emph{before} the value mixer processes it.
Following~\cite{LAUREL_ACML}, we achieve this by enforcing \emph{causal state alignment}, which shifts message delivery to the end of each step so that:
(i)~the communication graph for step $t$ is fully resolved before the mixer processes it; and
(ii)~all non-differentiable channel operations are isolated outside the within-step agent computation graph, enabling end-to-end gradient flow.

The standard ``communicate-then-act within the same step'' paradigm (Figure~\ref{fig: 2env}(a)) places wireless arbitration inside the agent's forward pass.
Causal alignment (Figure~\ref{fig: 2env}(b)) buffers messages generated at step $t$ for consumption at step $t{+}1$, yielding a well-posed decision interface and end-to-end trainability.
The resulting one-step staleness is standard in discrete-time coordination settings~\cite{vbc,ndq} and is bounded by a single decision epoch.

\begin{remark}[\textbf{End-to-end Differentiability}]
\label{prop: differentiability}
Under causal state alignment, the gradient of the TD loss $\nabla_\theta \mathcal{L}$ (Eq.~\ref{eq: loss}) is well-defined.
We do not backpropagate through the wireless medium; it is part of the environment transition.
The wireless channel induces a stochastic reception mask determined by non-differentiable MAC/PHY events; during backpropagation this mask is treated as a fixed constant, while message \emph{values} on realized edges are outputs of the sender's message head $M(h_j^t;\theta_M)$ and retain their gradient paths to $\theta_M$.
This is the standard ``stochastic mask, deterministic value'' treatment: the mask is sampled from the environment and held constant during the backward pass, while the continuous message content remains fully differentiable.
All within-step agent computations are implemented as differentiable neural modules (Section~\ref{sec: method how}), so non-differentiable channel arbitration does not obstruct gradient flow (see Appendix~\ref{appendix: prop diff}).
\end{remark}

\parag{Augmented MDP Formulation}
Causal state alignment couples the game and channel dynamics into a single system with augmented action and observation spaces:

\begin{definition}[\textbf{Augmented Dec-POMDP}]
\label{def: augmented mdp}
Relative to the base Dec-POMDP of Section~\ref{sec: prelim}, causal alignment augments the state with wireless components, $\mathcal{S} = \mathcal{S}^{\mathcal{T}} \times \mathcal{S}^{\mathcal{C}}$; the joint action with a communication action, $\bm{\mathcal{A}} = \bm{\mathcal{A}}^{\mathcal{T}} \times \bm{\mathcal{A}}^{\mathcal{C}}$; and each agent's observation with local wireless measurements (\eg, RSS, ACK statistics). The remaining elements of the tuple are inherited from the base Dec-POMDP. Each agent's learning objective is defined over the augmented action $a_i = (a_i^T, a_i^C)$; the complete formal tuple is given in Appendix~\ref{appendix: aug mdp def}.
\end{definition}

\noindent
The augmented state is Markov by construction, since the formulation defines $\mathcal{S}$ to include all quantities that influence the next-step transition (Definition~\ref{def: augmented mdp}). The primary communication action is a binary transmit decision $a_{i}^{C}\in \mathcal{A}_{i}^{C}=\{0,1\}$; the formulation accommodates richer actions (variable power, contention parameters), but the binary action already captures the key communication--coordination tradeoff (Section~\ref{sec: exp behavioral}).
Under this formulation, agents can learn explicit communication policies on ``when to transmit.''

\parag{Observation Augmentation}
\phantomsection\label{sec: method what}
The wireless state $\mathcal{S}^{\mathcal{C}}$ is not directly observable to any single agent. We therefore augment each agent's observation with local wireless measurements:
$o_{i}^{t} = o_i^{\mathcal{T},t} \| o_i^{\mathcal{C},t}$,
where $o_i^{\mathcal{C}, t}$ incorporates RSS values, ACK-derived statistics, and other lightweight link indicators.
RSS provides a physically grounded proxy of link quality under path loss and shadowing, enabling the agent to modulate its communication decisions based on channel conditions, while ACK-related feedback provides empirical estimates of delivery reliability.
Each transmitted message encodes the sender's latent state $\bm{h}_i^t$, summarizing both game and wireless observations at step $t$.
For notational simplicity, we present the method using the current
observation $o_i^t$ as input, i.e., a history window of size one. More generally, $o_i^t$ can be replaced by a longer local action-observation history without changing the architecture or the subsequent derivations.

\section{Decentralized Agent Execution}
\label{sec: method how}

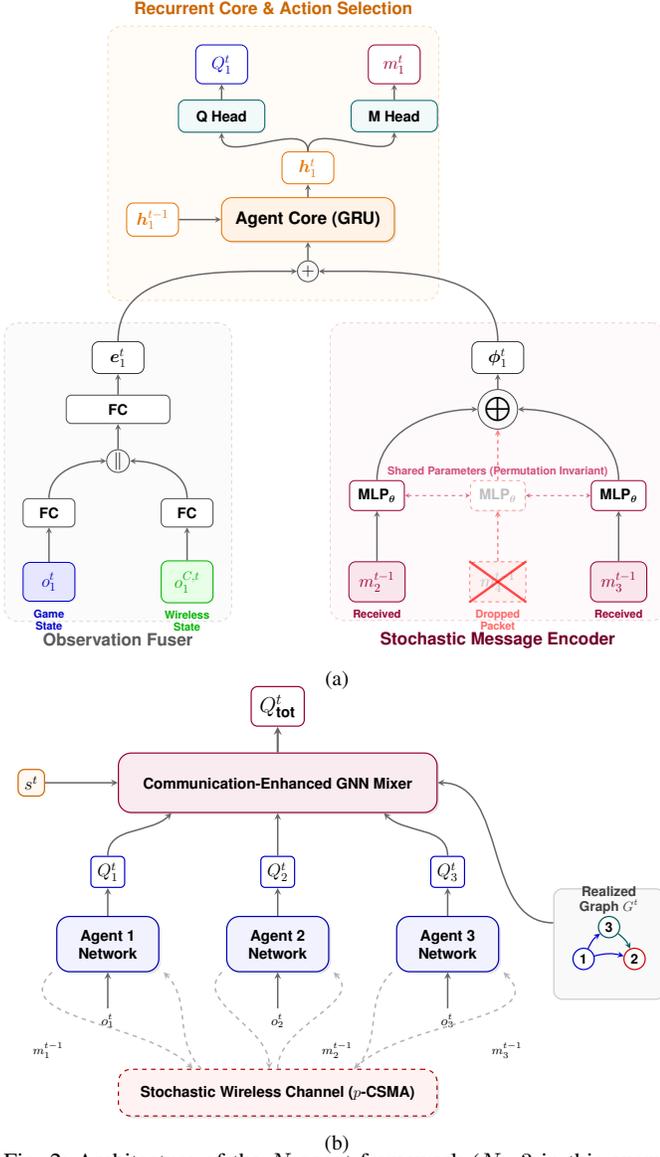
\begin{figure}
 \centering
 \begin{subfigure}[t]{\columnwidth}
   \centering
   \resizebox{\columnwidth}{!}{\input{./diagrams/overall_arch_2026.tex}}
   \caption{}
   \label{fig: overall_archi a}
 \end{subfigure}

 \vspace{2mm}

 \begin{subfigure}[t]{\columnwidth}
   \centering
   \resizebox{\columnwidth}{!}{\input{./diagrams/overall_arch_2026_b.tex}}
   \caption{}
   \label{fig: overall_archi b}
 \end{subfigure}
\caption{Architecture of the $N$-agent framework ($N{=}3$ in this example). (a)~Decentralized per-agent network: an Observation Fuser, a Stochastic Message Encoder, and a Recurrent Core that outputs both a local Q-value $Q_i^t$ and an outgoing message $m_i^t$ (agent~1 shown for brevity). (b)~Overall CTDE architecture: all agents' Q-values feed into the centralized Communication-Enhanced GNN Mixer, which combines them with the global state $s^t$ and the realized communication graph $G^t$ to produce the joint value $Q_{\text{tot}}^t$. The mixer is used only during centralized training.}
\label{fig: overall_archi}
\end{figure}

Following the CTDE paradigm, each agent executes a fully decentralized network, while the centralized mixing network is utilized exclusively during training (Figure~\ref{fig: overall_archi}).
As shown in Figure~\ref{fig: overall_archi a}, the per-agent network consists of three modules:
(i)~an observation fuser that embeds game and wireless inputs,
(ii)~a stochastic receptive field encoder that aggregates received messages, and
(iii)~a recurrent action-value generator that produces individual Q-values and message content.
During centralized training (Figure~\ref{fig: overall_archi b}), the per-agent Q-values are fed into the communication-enhanced GNN mixer, which conditions $Q_{\text{tot}}$ on the realized inter-agent communication graph.

\subsection{Observation Fuser}
The observation fuser independently embeds game and wireless observations via fully connected layers:
\begin{equation}
\label{eq: obs fuser}
e_i^t = \operatorname{FC}\!\Big(\operatorname{FC}(o_i^{\mathcal{T},t}) \,\|\, \operatorname{FC}(o_i^{\mathcal{C},t})\Big),
\end{equation}
where $\|$ denotes concatenation and $o_i^{\mathcal{C},t}$ is the wireless observation defined in Section~\ref{sec: method what}.
Separate embedding allows each modality-specific subnetwork to learn representations tailored to game and wireless observations before fusion, avoiding interference between heterogeneous input domains.

\subsection{Stochastic Receptive Field Encoder}

We model the set of received messages as a variable-size, unordered multiset $c_i=\{\bm{m}_{ij}\!:\,j\in\mathcal{N}_i\}$ and encode it via elementwise embedding and additive pooling:
\begin{align}
\label{eq: how gnn}
\textstyle
\Phi(c_i) = \sum_{j\in\mathcal{N}_i}\func[MLP]{\bm{m}_{ij}}.
\end{align}
This encoder is permutation invariant by commutativity of addition, and we show below that, with suitable parameters, it is also injective. Achieving these two properties \emph{jointly} is not trivial: concatenation-based encoders are injective but depend on the order in which neighbors are listed, so they are not permutation invariant; naive average pooling is permutation invariant but collapses distinct multisets that share the same mean. Their joint satisfaction matters for two reasons. First, permutation invariance eliminates any dependence on arbitrary neighbor indexing and removes the combinatorial burden of learning to recognize that different orderings of the same multiset encode the same information, which is critical under stochastic reception where the cardinality and identity of $\mathcal{N}_i$ vary across steps. Second, injectiveness guarantees that the aggregated embedding retains all information present in the multiset of received messages, so no coordinator-relevant content is destroyed at the aggregation stage before it reaches the recurrent action-value generator or the GNN mixer. The neighbor set $\mathcal{N}_i$ also defines the edges of the communication graph consumed by the GNN mixer (Section~\ref{sec: method enhanced mixer}).

\begin{theorem}
\label{thm: gin emb}
Assuming the message space is finite (e.g., under finite-precision quantization), there exists a set of parameters for the encoder of Equation~\ref{eq: how gnn} such that $\Phi$ is permutation invariant and injective.
\end{theorem}

\begin{proof}
The proof follows directly from Lemma~5 of Xu et al.~\cite{gin}; see Appendix~\ref{appendix: proof} for the full derivation and a quantitative finite-width bound.
\end{proof}

\begin{remark}[\textbf{Asymptotically Lossless Aggregation}]
\label{remark: lossless}
With a finite-width MLP, the encoder approximates the injective target mapping; as model capacity increases, the multiset reconstruction error is bounded and driven toward zero (Appendix~\ref{appendix: proof}, Eq.~\ref{eq: approx inject}).
Equation~\ref{eq: how gnn} therefore functions as an \emph{asymptotically lossless} aggregation mechanism for the stochastic receptive field.
\end{remark}

\subsection{Memory-Aided $Q$ Generator}
Partial observability under stochastic communication requires each agent to maintain memory across steps; we therefore use a GRU to encode the action--observation--message history.
The $Q$ generator combines the fused observation $e_i^{t}$ and encoded message $\phi_i^{t}$, and updates a GRU hidden state $h_i^{t}$.
The GRU hidden state $h_i^t$ serves as a learned, fixed-dimensional summary of the agent's past observations and actions $(o_i^{1:t}, a_i^{1:t-1})$, i.e., a recurrent approximation to the action--observation history used in the Dec-POMDP preliminaries.
Two output heads map $h_i^{t}$ to:
\begin{itemize}
\item \emph{Action-value head:} per-action utilities $Q_i^{t}(a_i^T, a_i^C)$ over the augmented action space $\mathcal{A}_i^T \times \mathcal{A}_i^C$, where $a_i^T$ is the game action and $a_i^C \in \{0,1\}$ is the transmit/silent decision.
The agent selects $(a_i^{T,t}, a_i^{C,t}) = \arg\max_{(a_i^T, a_i^C)} Q_i^t(a_i^T, a_i^C)$ with $\epsilon$-greedy exploration during training.
\item \emph{Message head:} a continuous message vector $m_i^{t} = M(h_i^t;\theta_M)$ encoding the content to be broadcast. The message is transmitted to the wireless channel only when $a_i^{C,t} = 1$; otherwise the agent remains silent.
\end{itemize}

\noindent
Given per-agent utilities $Q_i$, the centralized mixer aggregates them into $Q_{\text{tot}}$ using the realized communication graph, as described next.

\section{Centralized Value Decomposition: Communication-Enhanced Mixer}
\label{sec: method enhanced mixer}
Communication directly affects collaboration: different coordination patterns not only yield different individual action-values $Q_i$ (via the agent network), but should also induce different mixing strategies. Therefore, the mixer must capture the inductive bias from the dynamic communication structure.
As in VDN and QMIX, the mixing network can be viewed as having two stages. First, an aggregation stage $f(\cdot)$ maps the individual action-values to a hidden representation $v=f(Q_1,\ldots,Q_N)$. Second, a non-negative MLP maps $v$ to the scalar joint value $Q_{\mathrm{tot}}$. The novelty of our construction lies in the first stage, so in this section we focus on $f(\cdot)$ and use ``mixer'' as shorthand for this graph-conditioned aggregation stage.
We omit the time superscript $t$ for readability; all quantities refer to a single time step.

\subsection{GNN Mixer Architecture}
\begin{figure}[t]
    \centering
    \resizebox{\columnwidth}{!}{\input{./diagrams/gnnmixer_2026.tex}}
\caption{Communication-enhanced GNN mixer ($N{=}3$ for illustration). Dashed edges denote the realized communication graph $G^t$, shown in the left panel. PEHypernet (bottom) generates node-specific weights from each agent's conditioning input $\hat{s}_i$; these weights parameterize the $L$-layer GNN (middle) that embeds each $Q_i$, and sum pooling (top) yields the team embedding $\bm{v}$ that is mapped to $Q_{\text{tot}}$.}
\label{fig: method gnnmixer}
\end{figure}
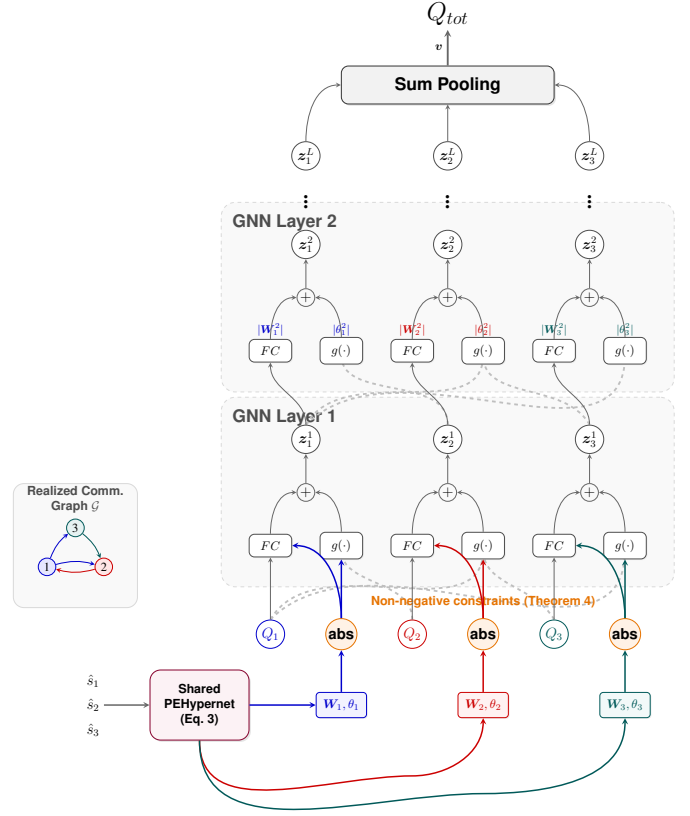

\parag{Communication graph construction}
At each step, we construct a directed graph $G^t=(V,E^t)$ from the realized message-delivery outcomes. The node set $V$ contains one node per agent, and a
directed edge $(j \rightarrow i)\in E^t$ is present if and only if agent
$i$ successfully receives the message sent by agent $j$ ($j \in \mathcal{N}_i^t$). Under causal state alignment (Section~\ref{sec: method mdp}), the edges in $G^t$ reflect messages transmitted at step $t{-}1$ and delivered before agents act at step $t$; $G^t$ is therefore fully resolved when the mixer processes step-$t$ utilities. An example of the resulting graph is visualized in the left panel of Figure~\ref{fig: method gnnmixer}.

\parag{Node features and message passing}
Node $i$'s feature is the scalar individual action-value $Q_i$ evaluated at the executed augmented action; we write $\bm{z}_i^0 = Q_i$. As shown in Figure~\ref{fig: method gnnmixer}, the mixer applies $L$ layers of GNN message passing to these node features. In each layer, every node combines a \emph{self-embedding} that linearly transforms its own previous embedding with a \emph{neighbor aggregation} term that pools information along the incoming edges of $G^t$; the two contributions are summed and passed through a nonlinearity. Stacking $L$ layers integrates information from multi-hop paths in $G^t$, so the communication reshapes the mixing of \emph{every} agent's utility, not only those that received a message directly. After the final layer, a permutation-invariant sum pooling yields the team embedding $\bm{v}$, which a second-stage MLP with non-negative weights maps to $Q_{\text{tot}}$.

\parag{Node-specific, state-conditioned weights}
Unlike classical GNNs that share a single weight matrix across nodes, our mixer uses \emph{node-specific} self-embedding weights and neighbor-aggregation parameters, generated at every forward pass by a Permutation-Equivariant Hypernetwork (PEHypernet, shown in the lower portion of Figure~\ref{fig: method gnnmixer}). This lets the mixer modulate each agent's contribution and each neighbor edge's influence based on the current global state, while still preserving permutation invariance over agent indexing (formalized in Section~\ref{sec: method mixer math}).

To preserve permutation invariance at the mixer level, the node-specific parameters must transform equivariantly under any relabeling of the agents.
We enforce this using a \textit{\textbf{Permutation-Equivariant Hypernetwork (PEHypernet)}}, which generates per-agent parameters from $\hat{s}_i = s \,\|\, o_i$, where $s$ is the global state and $o_i = o_i^{\mathcal{T}} \| o_i^{\mathcal{C}}$ is agent~$i$'s local observation (Section~\ref{sec: method what}).
The same functions $\psi_A^l$ and $\psi_B^l$ are applied independently to each $\hat{s}_i$, which yields permutation equivariance by construction (Lemma~\ref{lemma: PE}).
The dependence on $s$ is consistent with CTDE because PEHypernet is used only during centralized training.

In layer $l$, there are two \textit{PEHypernet}s shared across the nodes (Figure~\ref{fig: method gnnmixer}). For node $i$, they generate self-embedding weight $\bm{W}_i^l$ and neighbor aggregation parameters $\xi^l_i$, respectively. To preserve monotonicity (see Section~\ref{sec: method mixer math}), function \textit{\textbf{abs}}$(\cdot)$ is applied to the generated weights and parameters.

Let the functions approximated by the two \textit{PEHypernet}s be $\psi^{l}_{A}(\cdot)$ and $\psi^{l}_{B}(\cdot)$.
The hidden embedding of $Q_i$ after $l$ layers $\bm{z}_{i}^{l}$ is then:

\begin{align}
	\label{eq: gnnHyper}
    \bm{z}_{i}^{l} = \sigma\paren{\bm{W}^{l}_{i} \cdot\bm{z}_i^{l-1}  + g\paren{ \set {\bm{z}_j^{l-1} \given j\in \mathcal{N}_i}; \xi^l_i}}
\end{align}
\noindent
where $\bm{W}_i^l = \textit{\textbf{abs}}\paren{\bm{\psi}^{l}_{A}\paren{\hat{s}_i}}$, $\xi_i^l = \textit{\textbf{abs}}\paren{\bm{\psi}^{l}_{B}\paren{\hat{s}_i}}$, $\mathcal{N}_i$ is the neighbor set defined in Section~\ref{sec: method how}, equivalently the set of source nodes with incoming edges to $i$ in $G^t$, $g(\cdot)$ is the neighbor feature aggregation function parameterized by $\xi_i^{l}$.
In the first layer, $\bm{z}_i^0=Q_i$.
All nodes share the same $\bm{\psi}_{A}^{l}$ and $\bm{\psi}_{B}^{l}$ but with different inputs $\hat{s}_i$; when $L>1$, multi-hop information is integrated.
We use linear aggregation with non-negative weights for $g(\cdot)$ because it guarantees the non-negative Jacobian required by Theorem~\ref{theorem: monotonicity}, preserving the IGM guarantee while still permitting state-dependent neighbor-influence modulation via PEHypernet parameters.

Finally, a permutation-invariant pooling function is applied after the final layer. We use sum pooling:
\begin{align}
    \label{eq: pooling}
\bm{v} = \sum_i{\bm{z}_i^L}
\end{align}
The architecture specification, including the GNN hidden dimension and the number of message-passing rounds, is detailed in Appendix~\ref{appendix: training hyperparameters}.

\subsection{Mathematical Properties}
\label{sec: method mixer math}
This subsection establishes two key mathematical properties of the proposed mixer: permutation invariance and monotonicity.
Permutation invariance ensures that $Q_{\text{tot}}$ is independent of arbitrary agent indexing---relabeling agents does not change the joint value. Monotonicity guarantees that the optimal joint action decomposes into per-agent greedy actions, enabling decentralized execution (the IGM condition).

\begin{theorem}
    \label{theorem: PI}
    \normalfont
    The proposed communication-enhanced mixer is \textbf{\textit{permutation invariant}} w.r.t. $\set{Q_i \given i=1, 2, \ldots, N}$, provided (i)~the neighbor aggregation $g(\cdot)$ is permutation invariant, (ii)~the final pooling function $\mathrm{pool}(\cdot)$ is permutation invariant, and (iii)~the conditioning inputs $(\hat{s}_1,\ldots,\hat{s}_N)$ are permutation-equivariant with the agent indexing, i.e., relabeling the agents permutes the $\hat{s}_i$'s in the same way.
\end{theorem}
\noindent
Condition~(iii) is satisfied by construction in our implementation: $\hat{s}_i$ is formed by concatenating a shared global-state vector with agent~$i$'s local observation, so the mapping $(s,o_1,\ldots,o_N)\mapsto(\hat{s}_1,\ldots,\hat{s}_N)$ is trivially permutation-equivariant.
The proof of Theorem~\ref{theorem: PI} relies on the following lemma.

\begin{lemma}[\textbf{Permutation Equivariance of PEHypernet}]
    \label{lemma: PE}
    The weights and parameters generated by the proposed \textit{PEHypernet} are \textbf{\textit{permutation equivariant}} with respect to $\set{\hat{s}_i \given i=1,2,\ldots,N}$.
    This property holds by construction: because \textit{PEHypernet} applies the same function $\psi_A(\cdot)$ independently to each agent's input $\hat{s}_i$, permuting the inputs permutes the outputs in the same way.

Formally, define $\Psi_A(\cdot)$, the mapping realized by \textit{PEHypernet{\_}A}, as
\begin{align}
    \label{equation: PEHypernet}
(\bm{W}_1^l,\ldots,\bm{W}_N^l)&=\Psi_A^l(\hat{s}_1,\ldots,\hat{s}_N) \nonumber\\
&\coloneqq(\psi_A^l(\hat{s}_1),\ldots,\psi_A^l(\hat{s}_N)).
\end{align}
Similarly, define
\begin{align}
\label{equation: PEHypernet B}
(\xi_1^l,\ldots,\xi_N^l)&=\Psi_B^l(\hat{s}_1,\ldots,\hat{s}_N) \nonumber\\
&\coloneqq(\psi_B^l(\hat{s}_1),\ldots,\psi_B^l(\hat{s}_N)).
\end{align}
\end{lemma}

\begin{proof}
See Appendix~\ref{appendix: proof pi}.
\end{proof}

\begin{proof}[\textit{Proof of Theorem~\ref{theorem: PI}}]
See Appendix~\ref{appendix: proof pi}.
\end{proof}

Unlike traditional GNNs that share a single weight matrix across nodes, the PEHypernet generates node-specific weights while preserving permutation invariance. Combined with the permutation-invariant encoder (Section~\ref{sec: method how}), the proposed framework consistently guarantees permutation invariance throughout the pipeline.
This design reduces the hypothesis space and discourages spurious index-dependent correlations, which can improve sample efficiency in practice: as the number of agents increases, the number of possible permutations grows combinatorially, and without permutation invariance the model must independently learn that each permutation produces the same $Q_{\text{tot}}$.

Note that standard QMIX does not guarantee permutation invariance, since its hypernetwork generates a single concatenated weight vector with an implicit agent ordering.

We next establish \textbf{\textit{monotonicity}}, which ensures the IGM condition (Section~\ref{sec: related}).

\begin{theorem}
    \label{theorem: monotonicity}
    \normalfont
    Suppose the activation function $\sigma(\cdot)$ satisfies $\sigma'(\cdot) \geq 0$ elementwise (\eg, ReLU or ELU with $\alpha > 0$).
    The proposed mixer is \textbf{\textit{monotonic}} with regard to $\set{Q_i \given i=1, 2, \ldots, N}$, if the Jacobian matrix $\frac{\partial g}{\partial \bm{z}_j^{l-1}}$ is elementwise non-negative.
\end{theorem}
\begin{proof}
See Appendix~\ref{appendix: proof mono}.
\end{proof}

\begin{remark}[\textbf{Monotonicity Implies the IGM Condition}]
\label{remark: igm}
Theorem~\ref{theorem: monotonicity} establishes that $\frac{\partial f}{\partial Q_i} \geq 0$ for all $i$, where $f(\cdot)$ is the first aggregation stage.
The full mixer is $Q_\text{tot} = \text{MLP}_2(f(Q_1,\ldots,Q_N))$, where $\text{MLP}_2$ is implemented with non-negative weights (following QMIX).
By the chain rule, $\frac{\partial Q_\text{tot}}{\partial Q_i} = \left(\nabla_{\bm{v}} Q_\text{tot}\right)^{\!\top} \frac{\partial f}{\partial Q_i} \geq 0$, where $\nabla_{\bm{v}} Q_\text{tot} \in \mathbb{R}^{d_v}$ is elementwise non-negative (non-negative weights and non-negative-derivative activations in $\mathrm{MLP}_2$) and $\frac{\partial f}{\partial Q_i} \in \mathbb{R}^{d_v}$ is elementwise non-negative by Theorem~\ref{theorem: monotonicity}.
This directly implies the Individual-Global-Max (IGM) condition \cite{qmix}: since $Q_\text{tot}$ is monotonically non-decreasing in each $Q_i$, we have
\begin{align}
  &\arg\max_{\bm{a}}\, Q_\text{tot}(\bm{a}) \nonumber\\
  &\quad= \bigl(\arg\max_{a_1} Q_1(a_1),\; \ldots,\; \arg\max_{a_N} Q_N(a_N)\bigr),
\end{align}
where each $a_i = (a_i^T, a_i^C)$ is the augmented action (Definition~\ref{def: augmented mdp}). Thus greedy action selection can be performed independently per agent without loss of joint optimality.
\end{remark}

\subsection{Theoretical Analysis}
\parag{Expressive Power}
We now analyze the expressive power of the proposed communication-enhanced mixer and compare its first aggregation stage $f(\cdot)$ with the corresponding monotonic aggregation stage in QMIX.

\begin{definition}[\textbf{Expressive Power}]
    \label{def: exp power}
    Let $\mathcal{F}_A$ and $\mathcal{F}_B$ denote the function classes realizable by model architectures $A$ and $B$, respectively, over all possible parameter settings.
    Model $A$ has \emph{strictly higher expressive power} than model $B$ if $\mathcal{F}_B \subsetneq \mathcal{F}_A$, i.e., every function representable by $B$ is also representable by $A$, but not vice versa.
\end{definition}

\begin{theorem}[\textbf{Expressive power of communication-conditioned mixing}]
\label{thm: exp power}
Let $G$ denote the realized \emph{communication structure} (i.e., the
collection of neighbor sets $\{\mathcal{N}_i\}_{i=1}^N$) at a step, and let
$\bm Q=(Q_1,\ldots,Q_N)$ be the vector of individual action-values.
Here $s$ denotes the global state as in standard QMIX.
Consider the class $\mathcal{F}_{\mathrm{agn}}$ of monotone mixers of the form
$Q_{\mathrm{tot}} = F(\bm Q, s)$ whose functional form does \emph{not} consume the realized communication graph $G$ as an input (this includes
standard QMIX-style mixers, whose hypernetwork is conditioned on the flat state $s$ regardless of any communication actions that might be absorbed into the individual utilities $Q_i$), and the class $\mathcal{F}_{\mathrm{cond}}$ of
mixers representable by the proposed GNN-based mixing architecture that
explicitly conditions on $G$.
Then $\mathcal{F}_{\mathrm{agn}} \subsetneq \mathcal{F}_{\mathrm{cond}}$.

\end{theorem}

\begin{proof}
\textbf{Containment.}
Any graph-agnostic mixer $F(\bm Q,s)$ can be realized by the proposed
architecture by ignoring $G$ (e.g., setting the neighbor-aggregation term to
zero and using only self-terms), reducing to a standard monotone mixing network.
Hence $\mathcal{F}_{\mathrm{agn}}\subseteq \mathcal{F}_{\mathrm{cond}}$.

\smallskip
\textbf{Strictness.}
We exhibit a monotone function that depends on $G$, which therefore cannot lie in $\mathcal{F}_{\mathrm{agn}}$, but is realizable by the proposed architecture.
Take $N=2$ and define
\begin{align}
\label{eq:witness}
Q_{\mathrm{tot}}^\star(\bm Q,G) \;:=\; Q_2 \;+\; \mathbf{1}\{(1\!\to\!2)\in G\}\,Q_1,
\end{align}
where $\mathbf{1}\{\cdot\}$ is the indicator function. $Q_{\mathrm{tot}}^\star$ is
monotone in each $Q_i$.

First, $Q_{\mathrm{tot}}^\star \notin \mathcal{F}_{\mathrm{agn}}$.
Fix $\bm Q=(1,0)$ and consider two delivery relations $G$ and $G'$ that differ
only by the presence/absence of edge $(1\!\to\!2)$.
Then $Q_{\mathrm{tot}}^\star(\bm Q,G)=1$ and $Q_{\mathrm{tot}}^\star(\bm Q,G')=0$,
whereas any graph-agnostic $F(\bm Q,s)$ must output the same value for both
$G$ and $G'$ because it does not take $G$ as input.

Second, $Q_{\mathrm{tot}}^\star \in \mathcal{F}_{\mathrm{cond}}$.
Instantiate the proposed GNN-based mixer with one aggregation round and sum
pooling: let node~2 aggregate its incoming neighbor values by
$g(\{Q_j:j\in\mathcal{N}_2\})=\sum_{j\in\mathcal{N}_2} Q_j$ and let all other
aggregations be zero; then the pooled representation equals
$Q_2+\mathbf{1}\{1\in\mathcal{N}_2\}Q_1$, and the final monotone readout can be
chosen as the identity. This realizes~\eqref{eq:witness}.

Therefore, $\mathcal{F}_{\mathrm{agn}} \subsetneq \mathcal{F}_{\mathrm{cond}}$.
\end{proof}

\begin{remark}[\textbf{Representability vs.\ Learnability}]
\label{remark: learnability}
Theorem~\ref{thm: exp power} is a \emph{representability} result: it shows that $\mathcal{F}_{\mathrm{cond}}$ strictly contains $\mathcal{F}_{\mathrm{agn}}$.
It does not, by itself, imply superior optimization behavior or guaranteed empirical gains under gradient-based training. We therefore interpret Theorem~\ref{thm: exp power} as architectural justification for graph-conditioned mixing rather than a standalone guarantee of practical advantage.
The ablation study in Section~\ref{sec: exp ablation} provides empirical evidence that the added capacity is consistently realized: the GNN mixer outperforms the standard QMIX hypernetwork mixer in all evaluated environments.
\end{remark}

\begin{remark}[\textbf{Practical Significance}]
\label{remark: practical significance}
Functions in $\mathcal{F}_{\mathrm{cond}} \setminus \mathcal{F}_{\mathrm{agn}}$ correspond to realistic coordination patterns:
(i)~the mixer can upweight the joint contribution of agents who successfully received a teammate's broadcast relative to isolated agents, reflecting that informed agents can coordinate more effectively;
(ii)~under partial communication failure, the mixer can assign differentiated credit based on which links were realized, even when individual $Q_i$ values are similar.
Neither pattern can be expressed by a mixer limited to $(\bm{Q},s)$, since it depends on \emph{which} edges of $G$ were realized.
\end{remark}

\begin{remark}[\textbf{Design Alternatives}]
\label{remark: design alternatives}
Attention-based aggregation could model richer patterns but softmax normalization introduces negative cross-partial derivatives, breaking monotonicity (Theorem~\ref{theorem: monotonicity}).
Mean-pool satisfies the Jacobian condition but loses state-conditioned scaling.
We adopt linear aggregation with non-negative weights, trading flexibility for a clean IGM guarantee.
\end{remark}

\parag{Comparison with GNNs}
A vanilla GNN is not a suitable drop-in replacement for a CTDE value mixer. In particular, its fixed, globally shared weights do not naturally incorporate the global-state conditioning used in standard CTDE mixers, and a standard GNN does not enforce the non-negative Jacobian needed for the IGM condition. Our design is tailored to these two requirements. While the proposed mixer retains the core GNN principles of parameter sharing and neighbor aggregation, it differs in three key ways:
(1)~PEHypernet generates \emph{state-conditioned}, agent-specific parameters $(\bm{W}_i^l, \xi_i^l)$ at every forward pass, allowing the mixer to modulate neighbor influence based on the current global state;
(2)~parameter sharing occurs at the hypernetwork level, so each agent receives distinct parameters while permutation equivariance is preserved;
and (3)~the generated parameters are constrained to be non-negative, which is what ensures the monotonicity required for the IGM condition.

\section{Training Algorithm}
\label{sec: method complete}
This section describes the end-to-end training procedure that combines the decentralized agent networks (Section~\ref{sec: method how}) with the communication-enhanced mixer (Section~\ref{sec: method enhanced mixer}).

\coolname{} trains via off-policy TD learning over the augmented action space $\mathcal{A}_i = \mathcal{A}_i^T \times \mathcal{A}_i^C$ (Section~\ref{sec: method mdp}).
Each agent's network produces (i)~augmented action-values $Q_i^t(a_i^T, a_i^C)$ and (ii)~a message vector $m_i^t = M(h_i^t;\theta_M)$, transmitted only when $a_i^{C,t}=1$.

\parag{Why maximize over augmented actions}
The shared reward $\mathcal{R}$ depends only on the game action $\bm{a}^T$ (see Appendix~\ref{appendix: aug mdp def}); the communication action $\bm{a}^C$ generates no immediate reward but shapes the information available to teammates at the next step by altering the received-message sets $\{\mathcal{N}_j^{t+1}\}$ and thereby the communication graph that conditions the mixer.
The action-value $Q_i(a_i^T, a_i^C)$ captures this long-term effect through the discounted future return, so the $\arg\max$ over $(a_i^T, a_i^C)$ jointly optimizes game execution and communication strategy within a single TD learning loop; no auxiliary communication objective is required.
By the monotonicity guarantee (Remark~\ref{remark: igm}), the optimal joint augmented action decomposes into per-agent $\arg\max$ operations, enabling fully decentralized execution.

Algorithm~\ref{algo: qmix-wink} summarizes the training loop.
Lines~5--12 are executed by every agent at each step: the agent encodes received messages (line~6), fuses observations (line~7), selects an augmented action via the $Q$ head (lines~8--9), computes the message content (line~10), and interacts with both the game and wireless environments (lines~11--12). In lines~8--9, we write $Q_i(\cdot,a_i^T,a_i^C;\theta_Q)$, where $\cdot$ stands for the non-action arguments of $Q_i$: the fused observation $e_i^t$, the encoded received messages $\phi_i^t$, and the previous hidden state $h_i^{t-1}$.
After each episode is stored in the replay buffer~$\mathcal{D}$, a trajectory is sampled for training (line~16).
Because the $Q$ generator uses a GRU~\cite{gru}, gradient descent is performed over the full sampled episode rather than individual transitions (following~\cite{rmaddpg}).

\begin{algorithm}[tb]
\caption{Training algorithm of \coolname}
\label{algo: qmix-wink}
\begin{algorithmic}[1]
\STATE Initialize all neural networks and replay buffer $\mathcal{D}$
\FOR{episode$=1$ to $M$}
    \STATE{Initialize observation $\bm{o}$; initialize received messages $\bm{c}\gets\bm{0}$}
    \FOR{$t=1$ to end of episode}
        \STATE Receive messages $c_{i}^{t-1}$; get wireless measurements ${o}_{i}^{\mathcal{C},t}$
        \STATE $\phi_{i}^{t}\gets\Phi_i\!\paren{c_i^{t-1}}$ \COMMENT{encode received messages}
        \STATE $e_i^t\gets\text{Fuse}\!\paren{o_i^{\mathcal{T},t},\, o_i^{\mathcal{C},t}}$ \COMMENT{observation fusion}
        \STATE $Q^*_i \gets \arg\max_{(a_i^T,a_i^C)} Q_i(\cdot,a_i^T,a_i^C;\theta_Q)$
        \STATE $(a_i^{T,t},a_i^{C,t})\gets\epsilon\text{-greedy on } Q^*_i$
        \STATE $m_i^t\gets M(h_i^t;\theta_M)$ \COMMENT{message content}
        \STATE Execute game action $\bm{a}^{\mathcal{T},t}$ in the game environment
        \STATE If $a_i^{C,t}\!=\!1$: transmit $m_i^t$ via the wireless channel
        \STATE Observe shared reward $r^t$; transit to $\bm{s}^{t+1}$
        \STATE Construct $G^t$ from realized received-message sets $\{\mathcal{N}_i^t\}_{i=1}^N$
    \ENDFOR
    \STATE Sample trajectory from $\mathcal{D}$; compute TD loss (Eq.~\ref{eq: loss})
    \STATE Update online parameters $\theta$; periodically copy to target $\theta'$
\ENDFOR
\end{algorithmic}
\end{algorithm}

\parag{TD loss}
The mixing network $Q_\text{tot}$ (Figure~\ref{fig: method gnnmixer}) combines per-agent utilities into a joint value, conditioned on the global state $s^t$ and the realized communication graph $G^t$.
Training minimizes the TD loss over sampled episodes:
\begin{align}
	\label{eq: loss}
	\mathcal{L}\!\paren{\theta} = \sum_{t}\!\left[\!\paren{y_t^{\text{tot}}-Q_{\text{tot}}\!\paren{s^t,\,\{Q_i(\cdot,a_i^t;\theta)\}_{i=1}^{N},\,G^t;\theta}}^{\!2}\right],
\end{align}

where $a_i^t = (a_i^{T,t}, a_i^{C,t})$ is the executed augmented action of agent~$i$.
Because $G^t$ serves as the GNN's edge set rather than a flat input vector, the mixer inherits the graph-structured inductive bias discussed in Section~\ref{sec: intro}; Theorem~\ref{thm: exp power} formalizes the resulting expressive-power gain.

\parag{TD target}
Each agent selects $a_i^{*,t+1}=\arg\max_{a_i} Q_i(\cdot,a_i;\theta')$ under the target network, and the corresponding utilities form the TD target:

\begin{align}
	\label{eq: target}
y_t^{tot}=r_t+\gamma\,Q_{tot}\!\Big(s^{t+1},\{Q_i(\cdot,a_i^{*,t+1};\theta')\}_{i=1}^N,G^{t+1};\theta'\Big),
\end{align}
where $\theta'$ denotes target-network parameters (periodically copied from~$\theta$ to stabilize training). By the IGM guarantee established in Remark~\ref{remark: igm}, the joint $\arg\max$ over the exponentially sized joint action space $\prod_i(\mathcal{A}_i^T\times\mathcal{A}_i^C)$ decomposes into $N$ independent per-agent $\arg\max$ operations, which makes both target computation (above) and decentralized execution at deployment tractable.

%% file: diagrams/env_2026.tex
\usetikzlibrary{arrows.meta, positioning, fit, backgrounds, shadows, shapes.geometric, calc}

\begin{tikzpicture}[>=stealth]
    \tikzset{
        agentbox/.style={draw=blue!70!black, fill=blue!5, thick, rounded corners=3mm, inner sep=4mm, align=center, drop shadow={opacity=0.2, shadow xshift=1.5pt, shadow yshift=-1.5pt}, font=\sffamily\bfseries},
        wirelessbox/.style={draw=red!70!black, fill=red!5, thick, dashed, rounded corners=3mm, inner sep=4mm, align=center, drop shadow={opacity=0.2, shadow xshift=1.5pt, shadow yshift=-1.5pt}, font=\sffamily\bfseries},
        signal/.style={->, thick, draw=gray!80!black},
        gradfail/.style={->, red!80, dashed, line width=1.5pt},
        gradpass/.style={->, green!60!black, line width=1.5pt},
        stepbox/.style={draw=gray!50, dashed, thick, rounded corners=4mm, fill=gray!3},
        title/.style={font=\sffamily\Large\bfseries, anchor=west, text=black}
    }

    \node[title] at (-1, 2.5) {(a) Instantaneous Oracle (Non-Differentiable Baseline)};

    \node[agentbox] (encA) at (2, 0) {Agent $i$\\Network\\(msg gen.)};
    \node[wirelessbox] (wireA) at (6, 0) {Stochastic\\Wireless\\Medium};
    \node[agentbox] (actA) at (10, 0) {Agent $i$\\Network\\($Q$ head)};

    \node at (0, 0) (obsA) {$o_i^t$};
    \node at (12, 0) (actionA) {$a_i^t$};

    \draw[signal] (obsA) -- (encA);
    \draw[signal] (encA) -- node[above, align=center, font=\sffamily\small] {Latent msg\\$m_i^t$} (wireA);
    \draw[signal] (wireA) -- node[above, font=\sffamily\small] {Received $c_i^t$} (actA);
    \draw[signal] (actA) -- (actionA);

    \draw[gradpass] (12, -1.4) -- (10, -1.4) node[midway, below, font=\sffamily\small] {Gradient Flows};
    \draw[gradfail] (10, -1.4) -- (6, -1.4) node[midway, below, font=\sffamily\small] {Gradient Blocked};
    \node[text=red!80, font=\Huge] at (6, -1.4) {$\times$};

    \begin{scope}[on background layer]
        \draw[stepbox] (-0.5, 1.5) rectangle (12.5, -2.4);
        \node[anchor=south east, font=\sffamily\bfseries, text=black!70] at (12.4, -2.4) {Single Step $t$ Computation Graph};
    \end{scope}

    \node[title] at (-1, -3.5) {(b) \coolname{}'s Causal State Alignment (End-to-End Differentiable)};

    \node[agentbox, minimum height=2.5cm, minimum width=2.5cm] (agentT) at (2, -6.5) {Agent $i$\\Network};
    \node[wirelessbox] (wireB) at (6, -7.1) {Stochastic\\Wireless\\Medium};
    
    \node[agentbox, minimum height=2.5cm, minimum width=2.5cm] (agentT1) at (10, -6.5) {Agent $i$\\Network};

    \coordinate (inTopT)  at ([yshift=6mm]agentT.west);
    \coordinate (inBotT)  at ([yshift=-6mm]agentT.west);
    \coordinate (outTopT) at ([yshift=6mm]agentT.east);
    \coordinate (outBotT) at ([yshift=-6mm]agentT.east);

    \coordinate (inTopT1)  at ([yshift=6mm]agentT1.west);
    \coordinate (inBotT1)  at ([yshift=-6mm]agentT1.west);
    \coordinate (outTopT1) at ([yshift=6mm]agentT1.east);
    \coordinate (outBotT1) at ([yshift=-6mm]agentT1.east);

    \node at (0, -5.9) (obsT) {$o_i^t$};
    \node at (0, -7.1) (msgPrevT) {$c_i^{t-1}$};
    \node at (4, -5.9) (actionT) {$a_i^t$};

    \node at (8, -5.9) (obsT1) {$o_i^{t+1}$};
    \node at (12, -5.9) (actionT1) {$a_i^{t+1}$};
    \coordinate (msgNext) at (12, -7.1);

    \draw[signal] (obsT) -- (inTopT);
    \draw[signal] (msgPrevT) -- (inBotT);
    \draw[signal] (outTopT) -- (actionT);
    \draw[signal] (outBotT) -- node[above, font=\sffamily\small] {$m_i^t$} (wireB);

    \draw[signal] (wireB) -- node[above, font=\sffamily\small] {$c_i^t$} (inBotT1);
    \draw[signal] (obsT1) -- (inTopT1);
    \draw[signal] (outTopT1) -- (actionT1);
    \draw[signal] (outBotT1) -- node[above, font=\sffamily\small] {$m_i^{t+1}$} (msgNext);

    \draw[gradpass] (4, -8.6) -- (0, -8.6) node[midway, below, font=\sffamily\small] {Gradient Flow (TD loss)};
    
    \draw[gradpass] (12, -8.6) -- (8, -8.6) node[midway, below, font=\sffamily\small] {Gradient Flow (TD loss)};

    \begin{scope}[on background layer]
        \draw[stepbox] (-0.5, -4.5) rectangle (4.5, -9.4);
        \node[anchor=south east, font=\sffamily\bfseries, text=black!70] at (4.4, -10.1) {Step $t$};

        \draw[stepbox] (7.5, -4.5) rectangle (12.5, -9.4);
        \node[anchor=south east, font=\sffamily\bfseries, text=black!70] at (12.4, -10.1) {Step $t+1$};
    \end{scope}

\end{tikzpicture}

%% file: diagrams/overall_arch_2026.tex
\begin{tikzpicture}[>=stealth]
    \tikzset{
        gamenode/.style={rectangle, draw=blue!80!black, fill=blue!10, thick, rounded corners=2mm, inner sep=3mm, minimum width=1.5cm, align=center, drop shadow={opacity=0.15, shadow xshift=1pt, shadow yshift=-1pt}, font=\sffamily\Large\bfseries},
        wirenode/.style={rectangle, draw=green!70!black, fill=green!10, thick, rounded corners=2mm, inner sep=3mm, minimum width=1.5cm, align=center, drop shadow={opacity=0.15, shadow xshift=1pt, shadow yshift=-1pt}, font=\sffamily\Large\bfseries},
        msgnode/.style={rectangle, draw=purple!80!black, fill=purple!10, thick, rounded corners=2mm, inner sep=3mm, minimum width=1.5cm, align=center, drop shadow={opacity=0.15, shadow xshift=1pt, shadow yshift=-1pt}, font=\sffamily\Large\bfseries},
        droppednode/.style={rectangle, draw=red!50, fill=red!5, dashed, thick, rounded corners=2mm, inner sep=3mm, minimum width=1.5cm, align=center, font=\sffamily\Large\bfseries, text=gray!50},
        latnode/.style={rectangle, draw=black!80, fill=white, thick, rounded corners=2mm, inner sep=2mm, minimum width=1.5cm, align=center, font=\sffamily\Large\bfseries},
        fcblock/.style={rectangle, draw=black!70, fill=white, thick, rounded corners=1.5mm, inner sep=2.5mm, minimum width=1.5cm, font=\sffamily\large\bfseries, drop shadow={opacity=0.1, shadow xshift=1pt, shadow yshift=-1pt}},
        grublock/.style={rectangle, draw=orange!90!black, fill=orange!10, thick, rounded corners=3mm, inner sep=4mm, minimum width=4cm, font=\sffamily\Large\bfseries, drop shadow={opacity=0.2, shadow xshift=1.5pt, shadow yshift=-1.5pt}},
        headblock/.style={rectangle, draw=teal!80!black, fill=teal!5, thick, rounded corners=2mm, inner sep=3mm, minimum width=2.5cm, font=\sffamily\large\bfseries, drop shadow={opacity=0.15, shadow xshift=1pt, shadow yshift=-1pt}},
        opnode/.style={circle, draw=black!70, fill=gray!5, thick, inner sep=1pt, minimum size=6mm, font=\sffamily\Large\bfseries},
        signal/.style={->, thick, draw=gray!80!black, line width=1.2pt},
        dropsignal/.style={->, dashed, draw=red!50, line width=1.2pt},
        boxtitle/.style={font=\sffamily\Large\bfseries}
    }

    \node[gamenode, text=blue!80!black] (ogame) at (-2, 0) {$o_1^t$};
    \node[font=\sffamily\small\bfseries, text=blue!80!black, align=center, below=1mm of ogame] {Game\\State};
    
    \node[wirenode, text=green!70!black] (owire) at (2, 0) {$o_1^{C,t}$};
    \node[font=\sffamily\small\bfseries, text=green!70!black, align=center, below=1mm of owire] {Wireless\\State};

    \node[fcblock] (fc1) at (-2, 2) {FC};
    \node[fcblock] (fc2) at (2, 2) {FC};
    
    \draw[signal] (ogame) -- (fc1);
    \draw[signal] (owire) -- (fc2);

    \node[opnode] (concat) at (0, 3.5) {$\Vert$};
    \node[fcblock, minimum width=3cm] (fc3) at (0, 5) {FC};
    \node[latnode] (embed) at (0, 6.5) {$\bm{e}_1^t$};
    
    \draw[signal] (fc1.north) to[out=90, in=180] (concat.west);
    \draw[signal] (fc2.north) to[out=90, in=0] (concat.east);
    \draw[signal] (concat) -- (fc3);
    \draw[signal] (fc3) -- (embed);

    \begin{scope}[on background layer]
        \node[fit=(ogame) (owire) (fc3) (embed) (fc1) (fc2) (concat), fill=gray!3, rounded corners=4mm, draw=gray!30, dashed, thick, inner sep=15pt] (fuserbox) {};
    \end{scope}
    \node[boxtitle, text=gray!70!black, anchor=north] at ([yshift=-2mm]fuserbox.south) {Observation Fuser};

    \node[msgnode, text=purple!80!black] (m2) at (7.5, 0) {$m_2^{t-1}$};
    \node[droppednode] (m4) at (11, 0) {$m_4^{t-1}$};
    \node[msgnode, text=purple!80!black] (m3) at (14.5, 0) {$m_3^{t-1}$};
    
    \draw[red, line width=2pt, opacity=0.7] (m4.south west) -- (m4.north east);
    \draw[red, line width=2pt, opacity=0.7] (m4.north west) -- (m4.south east);
    \node[font=\sffamily\small\bfseries, text=red!60, align=center, below=1mm of m4] {Dropped\\Packet};
    \node[font=\sffamily\small\bfseries, text=purple!80!black, align=center, below=1mm of m2] {Received};
    \node[font=\sffamily\small\bfseries, text=purple!80!black, align=center, below=1mm of m3] {Received};

    \node[fcblock, draw=purple!80!black] (mlp2) at (7.5, 2.5) {$\text{MLP}_{\bm{\theta}}$};
    \node[fcblock, draw=red!40, text=gray!50, dashed] (mlp4) at (11, 2.5) {$\text{MLP}_{\bm{\theta}}$};
    \node[fcblock, draw=purple!80!black] (mlp3) at (14.5, 2.5) {$\text{MLP}_{\bm{\theta}}$};
    
    \draw[<->, dashed, draw=purple!60, thick] (mlp2.east) -- (mlp4.west);
    \draw[<->, dashed, draw=purple!60, thick] (mlp4.east) -- (mlp3.west);
    \node[font=\sffamily\small\bfseries, text=purple!70] at (11, 3.2) {Shared Parameters (Permutation Invariant)};

    \draw[signal] (m2) -- (mlp2);
    \draw[dropsignal] (m4) -- (mlp4);
    \draw[signal] (m3) -- (mlp3);

    \node[opnode, minimum size=8mm, font=\sffamily\huge\bfseries] (sum) at (11, 5) {$\bigoplus$};
    \node[latnode] (phi) at (11, 6.5) {$\bm{\phi}_1^t$};
    
    \draw[signal] (mlp2.north) to[out=90, in=180] (sum.west);
    \draw[dropsignal] (mlp4.north) -- (sum.south);
    \draw[signal] (mlp3.north) to[out=90, in=0] (sum.east);
    \draw[signal] (sum) -- (phi);

    \begin{scope}[on background layer]
        \node[fit=(m2) (m3) (m4) (phi) (sum) (mlp2) (mlp3) (mlp4), fill=purple!2, rounded corners=4mm, draw=purple!20, dashed, thick, inner sep=15pt] (encoderbox) {};
    \end{scope}
    \node[boxtitle, text=purple!60!black, anchor=north] at ([yshift=-2mm]encoderbox.south) {Stochastic Message Encoder};

    \node[opnode] (add) at (5.5, 9) {$+$};
    \draw[signal] (embed.north) to[out=90, in=180] (add.west);
    \draw[signal] (phi.north) to[out=90, in=0] (add.east);

    \node[grublock] (gru) at (5.5, 10.5) {Agent Core (GRU)};
    \node[latnode, draw=orange!90!black, text=orange!90!black] (hprev) at (1, 10.5) {$\bm{h}_1^{t-1}$};
    \node[latnode, draw=orange!90!black, text=orange!90!black] (hnext) at (5.5, 12) {$\bm{h}_1^t$};
    
    \draw[signal] (add) -- (gru);
    \draw[signal] (hprev) -- (gru);
    \draw[signal] (gru) -- (hnext);

    \node[headblock] (qhead) at (3, 13.5) {Q Head};
    \node[headblock] (mhead) at (8, 13.5) {M Head};
    
    \node[gamenode, text=blue!80!black, fill=white] (qout) at (3, 15) {$Q_1^t$};
    \node[msgnode, text=purple!80!black, fill=white] (mout) at (8, 15) {$m_1^t$};
    
    \draw[signal] (hnext.north) to[out=90, in=270] (qhead.south);
    \draw[signal] (hnext.north) to[out=90, in=270] (mhead.south);
    
    \draw[signal] (qhead) -- (qout);
    \draw[signal] (mhead) -- (mout);

    \begin{scope}[on background layer]
        \node[fit=(add) (gru) (hprev) (qout) (mout), fill=orange!3, rounded corners=4mm, draw=orange!30, dashed, thick, inner sep=15pt] (corebox) {};
    \end{scope}
    \node[boxtitle, text=orange!80!black, anchor=south] at ([yshift=2mm]corebox.north) {Recurrent Core \& Action Selection};

\end{tikzpicture}

%% file: diagrams/overall_arch_2026_b.tex
\begin{tikzpicture}[>=stealth]
    \tikzset{
        agentbox/.style={rectangle, draw=blue!70!black, fill=blue!5, thick, rounded corners=3mm, inner sep=3mm, minimum width=2.4cm, minimum height=1.3cm, align=center, drop shadow={opacity=0.2, shadow xshift=1.5pt, shadow yshift=-1.5pt}, font=\sffamily\bfseries},
        mixerbox/.style={rectangle, draw=purple!80!black, fill=purple!8, thick, rounded corners=3mm, inner sep=4mm, minimum width=7.5cm, minimum height=1.4cm, align=center, drop shadow={opacity=0.2, shadow xshift=1.5pt, shadow yshift=-1.5pt}, font=\sffamily\bfseries},
        wirelessbox/.style={rectangle, draw=red!70!black, fill=red!5, thick, dashed, rounded corners=3mm, inner sep=3mm, minimum width=7.5cm, minimum height=1.1cm, align=center, drop shadow={opacity=0.2, shadow xshift=1.5pt, shadow yshift=-1.5pt}, font=\sffamily\bfseries},
        graphnode/.style={circle, draw=black!70, fill=white, thick, inner sep=1pt, minimum size=5mm, font=\sffamily\small\bfseries},
        qnode/.style={rectangle, draw=blue!80!black, fill=white, thick, rounded corners=1mm, inner sep=1.5mm, font=\sffamily\large\bfseries},
        qtotnode/.style={rectangle, draw=purple!80!black, fill=white, thick, rounded corners=1.5mm, inner sep=2mm, font=\sffamily\Large\bfseries},
        statenode/.style={rectangle, draw=orange!80!black, fill=orange!5, thick, rounded corners=1.5mm, inner sep=1.5mm, font=\sffamily\large\bfseries},
        signal/.style={->, thick, draw=gray!80!black, line width=1.0pt},
        msgedge/.style={->, thick, draw=gray!60, dashed, line width=1.0pt},
        title/.style={font=\sffamily\Large\bfseries, anchor=west, text=black}
    }

    \node[agentbox] (a1) at (0, 0)  {Agent~1\\Network};
    \node[agentbox] (a2) at (4, 0)  {Agent~2\\Network};
    \node[agentbox] (a3) at (8, 0)  {Agent~3\\Network};

    \node[font=\sffamily\small] (o1) at (0, -1.8) {$o_1^t$};
    \node[font=\sffamily\small] (o2) at (4, -1.8) {$o_2^t$};
    \node[font=\sffamily\small] (o3) at (8, -1.8) {$o_3^t$};
    \draw[signal] (o1) -- (a1.south);
    \draw[signal] (o2) -- (a2.south);
    \draw[signal] (o3) -- (a3.south);

    \node[qnode] (q1) at (0, 1.7) {$Q_1^t$};
    \node[qnode] (q2) at (4, 1.7) {$Q_2^t$};
    \node[qnode] (q3) at (8, 1.7) {$Q_3^t$};
    \draw[signal] (a1.north) -- (q1.south);
    \draw[signal] (a2.north) -- (q2.south);
    \draw[signal] (a3.north) -- (q3.south);

    \node[wirelessbox] (wire) at (4, -3.5) {Stochastic Wireless Channel ($p$-CSMA)};

    \node[font=\sffamily\footnotesize] at (-1.4, -2.5) {$m_1^{t{-}1}$};
    \node[font=\sffamily\footnotesize] at (5.4, -2.5) {$m_2^{t{-}1}$};
    \node[font=\sffamily\footnotesize] at (9.4, -2.5) {$m_3^{t{-}1}$};

    \draw[msgedge] ([xshift=-1mm]a1.south west) to[out=-135, in=135] ([xshift=-20mm]wire.north);
    \draw[msgedge] ([xshift=-1mm]a2.south west) to[out=-135, in=90] ([xshift=-2mm]wire.north);
    \draw[msgedge] ([xshift=-1mm]a3.south west) to[out=-135, in=45] ([xshift=18mm]wire.north);

    \draw[msgedge] ([xshift=-18mm]wire.north) to[out=135, in=-45] ([xshift=1mm]a1.south east);
    \draw[msgedge] ([xshift=0mm]wire.north)  to[out=90, in=-45] ([xshift=1mm]a2.south east);
    \draw[msgedge] ([xshift=18mm]wire.north) to[out=45, in=-45] ([xshift=1mm]a3.south east);

    \begin{scope}[shift={(11.8, 0)}]
        \node[fill=gray!5, draw=gray!40, thick, rounded corners=2mm, minimum width=2.6cm, minimum height=2.6cm] (ginset) at (0, 0) {};
        \node[font=\sffamily\small\bfseries, text=black!75, align=center] at (0, 1.05) {Realized\\Graph $G^t$};
        \node[graphnode, draw=blue!80!black] (gn1) at (-0.6, -0.35) {1};
        \node[graphnode, draw=red!80!black]  (gn2) at (0.6, -0.35) {2};
        \node[graphnode, draw=teal!70!black] (gn3) at (0, 0.4) {3};
        \draw[->, thick, blue!80!black] (gn1) to[bend left=15] (gn2);
        \draw[->, thick, teal!70!black] (gn3) to[bend left=15] (gn2);
        \draw[->, thick, blue!80!black] (gn1) to[bend left=15] (gn3);
    \end{scope}

    \node[mixerbox] (mix) at (4, 3.8) {Communication-Enhanced GNN Mixer};

    \draw[signal] (q1.north) to[out=90, in=-135] ([xshift=-25mm]mix.south);
    \draw[signal] (q2.north) -- ([xshift=0mm]mix.south);
    \draw[signal] (q3.north) to[out=90, in=-45] ([xshift=25mm]mix.south);

    \node[statenode] (state) at (-1.8, 3.8) {$s^t$};
    \draw[signal] (state) -- (mix.west);

    \draw[signal] ([yshift=5mm]ginset.west) to[out=180, in=0] (mix.east);

    \node[qtotnode] (qtot) at (4, 5.6) {$Q_{\text{tot}}^t$};
    \draw[signal, line width=1.4pt] (mix.north) -- (qtot.south);

\end{tikzpicture}

%% file: diagrams/gnnmixer_2026.tex
\begin{tikzpicture}[>=stealth]
    \tikzset{
        peblock/.style={draw=purple!70!black, fill=purple!5, thick, rounded corners=3mm, inner sep=4mm, align=center, drop shadow={opacity=0.2, shadow xshift=1.5pt, shadow yshift=-1.5pt}, font=\sffamily\bfseries},
        fcblock/.style={draw=black!70, fill=white, thick, rounded corners=1.5mm, inner sep=2mm, minimum width=1.2cm, font=\sffamily\bfseries, drop shadow={opacity=0.1, shadow xshift=1pt, shadow yshift=-1pt}},
        gblock/.style={draw=black!70, fill=white, thick, rounded corners=1.5mm, inner sep=2mm, minimum width=1.2cm, font=\sffamily\bfseries, drop shadow={opacity=0.1, shadow xshift=1pt, shadow yshift=-1pt}},
        poolblock/.style={draw=black!80, fill=gray!10, thick, rounded corners=2mm, inner sep=3mm, minimum width=6cm, font=\sffamily\Large\bfseries, drop shadow={opacity=0.2, shadow xshift=1.5pt, shadow yshift=-1.5pt}},
        qnode/.style={circle, thick, inner sep=1pt, minimum size=8mm, font=\sffamily\large\bfseries, fill=white},
        znode/.style={circle, draw=black!70, fill=white, thick, inner sep=1pt, minimum size=8mm, font=\sffamily\large\bfseries},
        addnode/.style={circle, draw=black!60, fill=gray!5, thick, inner sep=0pt, minimum size=5mm, font=\sffamily\large\bfseries},
        absnode/.style={circle, draw=orange!90!black, fill=orange!10, thick, inner sep=1pt, minimum size=6mm, font=\sffamily\large\bfseries},
        weightmat/.style={rectangle, thick, rounded corners=1mm, inner sep=2mm, font=\sffamily\bfseries, align=center, drop shadow={opacity=0.15, shadow xshift=1pt, shadow yshift=-1pt}},
        signal/.style={->, thick, draw=gray!80!black},
        weightedge/.style={->, line width=1.2pt},
        msgarrow/.style={->, line width=1.5pt, draw=gray!50, dashed},
        title/.style={font=\sffamily\Large\bfseries, anchor=west, text=black}
    }

    \begin{scope}[shift={(-0.5, 0)}, on background layer]
        \node[fill=gray!5, draw=gray!30, thick, rounded corners=3mm, minimum width=3.5cm, minimum height=3.5cm] (insetbg) at (0,0) {};
        \node[font=\sffamily\bfseries, text=black!70, align=center] at (0, 1.3) {Realized Comm.\\Graph $\mathcal{G}$};
        
        \node[circle, draw=blue!80!black, fill=blue!10, thick, inner sep=2pt] (n1) at (-0.8, -0.6) {1};
        \node[circle, draw=red!80!black, fill=red!10, thick, inner sep=2pt] (n2) at (0.8, -0.6) {2};
        \node[circle, draw=teal!70!black, fill=teal!10, thick, inner sep=2pt] (n3) at (0, 0.5) {3};
        
        \draw[->, thick, blue!80!black] (n1) to[bend left=15] (n2);
        \draw[->, thick, red!80!black]  (n2) to[bend left=15] (n1);
        \draw[->, thick, blue!80!black] (n1) to[bend left=15] (n3);
        \draw[->, thick, teal!70!black] (n3) to[bend left=15] (n2);
    \end{scope}

    \node[align=center, font=\sffamily\large] (states) at (0, -4.5) {$\hat{s}_1$ \\[2mm] $\hat{s}_2$ \\[2mm] $\hat{s}_3$};
    \node[peblock] (pehyp) at (3, -4.5) {Shared\\PEHypernet\\(Eq.~\ref{eq: gnnHyper})};
    \draw[signal, line width=1.2pt] (states) -- (pehyp);

    \node[weightmat, draw=blue!80!black, fill=blue!5, text=blue!80!black] (w1) at (7, -4.5) {$\bm{W}_1, \theta_1$};
    \node[weightmat, draw=red!80!black, fill=red!5, text=red!80!black] (w2) at (11, -4.5) {$\bm{W}_2, \theta_2$};
    \node[weightmat, draw=teal!70!black, fill=teal!5, text=teal!70!black] (w3) at (15, -4.5) {$\bm{W}_3, \theta_3$};

    \draw[weightedge, blue!80!black]  (pehyp.east) -- (w1.west);
    \draw[weightedge, red!80!black]   (pehyp.south) to[out=270, in=180] (9, -6.5) to[out=0, in=270] (w2.south);
    \draw[weightedge, teal!70!black] (pehyp.south) to[out=270, in=180] (11, -7.0) to[out=0, in=270] (w3.south);

    \node[absnode] (abs1) at (7, -2.5) {$\textbf{abs}$};
    \node[absnode] (abs2) at (11, -2.5) {$\textbf{abs}$};
    \node[absnode] (abs3) at (15, -2.5) {$\textbf{abs}$};

    \draw[weightedge, blue!80!black]  (w1) -- (abs1);
    \draw[weightedge, red!80!black]   (w2) -- (abs2);
    \draw[weightedge, teal!70!black] (w3) -- (abs3);

    \node[font=\sffamily\bfseries, text=orange!90!black, align=center] at (11, -1.6) {Non-negative constraints (Theorem~\ref{theorem: monotonicity})};

    \node[qnode, draw=blue!80!black, text=blue!80!black] (q1) at (5, -2.5) {$Q_1$};
    \node[qnode, draw=red!80!black, text=red!80!black] (q2) at (9, -2.5) {$Q_2$};
    \node[qnode, draw=teal!70!black, text=teal!70!black] (q3) at (13, -2.5) {$Q_3$};

    \node[fcblock] (fc1) at (5, 0) {$FC$};
    \node[gblock]  (g1)  at (7, 0) {$g(\cdot)$};
    \node[addnode] (add1) at (6, 1.5) {$+$};
    \node[znode]   (z1)   at (6, 3) {$\bm{z}_1^1$};
    
    \draw[signal] (q1.north) to[out=90, in=270] (fc1.south);
    \draw[weightedge, blue!80!black] (abs1.north) to[out=90, in=0] (fc1.east);
    \draw[weightedge, blue!80!black] (abs1.north) -- (g1.south);
    \draw[signal] (fc1.north) to[out=90, in=180] (add1.west);
    \draw[signal] (g1.north) to[out=90, in=0] (add1.east);
    \draw[signal] (add1) -- (z1);

    \node[fcblock] (fc2) at (9, 0) {$FC$};
    \node[gblock]  (g2)  at (11, 0) {$g(\cdot)$};
    \node[addnode] (add2) at (10, 1.5) {$+$};
    \node[znode]   (z2)   at (10, 3) {$\bm{z}_2^1$};
    
    \draw[signal] (q2.north) to[out=90, in=270] (fc2.south);
    \draw[weightedge, red!80!black] (abs2.north) to[out=90, in=0] (fc2.east);
    \draw[weightedge, red!80!black] (abs2.north) -- (g2.south);
    \draw[signal] (fc2.north) to[out=90, in=180] (add2.west);
    \draw[signal] (g2.north) to[out=90, in=0] (add2.east);
    \draw[signal] (add2) -- (z2);

    \node[fcblock] (fc3) at (13, 0) {$FC$};
    \node[gblock]  (g3)  at (15, 0) {$g(\cdot)$};
    \node[addnode] (add3) at (14, 1.5) {$+$};
    \node[znode]   (z3)   at (14, 3) {$\bm{z}_3^1$};
    
    \draw[signal] (q3.north) to[out=90, in=270] (fc3.south);
    \draw[weightedge, teal!70!black] (abs3.north) to[out=90, in=0] (fc3.east);
    \draw[weightedge, teal!70!black] (abs3.north) -- (g3.south);
    \draw[signal] (fc3.north) to[out=90, in=180] (add3.west);
    \draw[signal] (g3.north) to[out=90, in=0] (add3.east);
    \draw[signal] (add3) -- (z3);

    \begin{scope}[on background layer]
        \node[fit=(fc1) (g3) (add1) (add3) (z1) (z3), fill=gray!5, rounded corners=4mm, draw=gray!30, dashed, thick, inner sep=22pt] (layer1box) {};
        \node[anchor=north west, font=\sffamily\Large\bfseries, text=black!70, yshift=-6pt, xshift=6pt] at (layer1box.north west) {GNN Layer 1};
    \end{scope}

    \node[fcblock] (fc1_2) at (5, 5.5) {$FC$};
    \node[gblock]  (g1_2)  at (7, 5.5) {$g(\cdot)$};
    \node[addnode] (add1_2) at (6, 7.0) {$+$};
    \node[znode]   (z1_2)   at (6, 8.5) {$\bm{z}_1^2$};
    
    \node[font=\sffamily\small\bfseries, text=blue!80!black] at (5, 6.1) {$|\bm{W}_1^2|$};
    \node[font=\sffamily\small\bfseries, text=blue!80!black] at (7, 6.1) {$|\theta_1^2|$};

    \draw[signal] (z1.north) to[out=90, in=270] (fc1_2.south);
    \draw[signal] (fc1_2.north) to[out=90, in=180] (add1_2.west);
    \draw[signal] (g1_2.north) to[out=90, in=0] (add1_2.east);
    \draw[signal] (add1_2) -- (z1_2);

    \node[fcblock] (fc2_2) at (9, 5.5) {$FC$};
    \node[gblock]  (g2_2)  at (11, 5.5) {$g(\cdot)$};
    \node[addnode] (add2_2) at (10, 7.0) {$+$};
    \node[znode]   (z2_2)   at (10, 8.5) {$\bm{z}_2^2$};
    
    \node[font=\sffamily\small\bfseries, text=red!80!black] at (9, 6.1) {$|\bm{W}_2^2|$};
    \node[font=\sffamily\small\bfseries, text=red!80!black] at (11, 6.1) {$|\theta_2^2|$};

    \draw[signal] (z2.north) to[out=90, in=270] (fc2_2.south);
    \draw[signal] (fc2_2.north) to[out=90, in=180] (add2_2.west);
    \draw[signal] (g2_2.north) to[out=90, in=0] (add2_2.east);
    \draw[signal] (add2_2) -- (z2_2);

    \node[fcblock] (fc3_2) at (13, 5.5) {$FC$};
    \node[gblock]  (g3_2)  at (15, 5.5) {$g(\cdot)$};
    \node[addnode] (add3_2) at (14, 7.0) {$+$};
    \node[znode]   (z3_2)   at (14, 8.5) {$\bm{z}_3^2$};
    
    \node[font=\sffamily\small\bfseries, text=teal!70!black] at (13, 6.1) {$|\bm{W}_3^2|$};
    \node[font=\sffamily\small\bfseries, text=teal!70!black] at (15, 6.1) {$|\theta_3^2|$};

    \draw[signal] (z3.north) to[out=90, in=270] (fc3_2.south);
    \draw[signal] (fc3_2.north) to[out=90, in=180] (add3_2.west);
    \draw[signal] (g3_2.north) to[out=90, in=0] (add3_2.east);
    \draw[signal] (add3_2) -- (z3_2);

    \begin{scope}[on background layer]
        \node[fit=(fc1_2) (g3_2) (add1_2) (add3_2) (z1_2) (z3_2), fill=gray!5, rounded corners=4mm, draw=gray!30, dashed, thick, inner sep=22pt] (layer2box) {};
        \node[anchor=north west, font=\sffamily\Large\bfseries, text=black!70, yshift=-6pt, xshift=6pt] at (layer2box.north west) {GNN Layer 2};
    \end{scope}

    \begin{scope}[on background layer]
        \draw[msgarrow] (q2.north) to[out=90, in=270] (7, 0); 
        
        \draw[msgarrow] (q1.north) to[out=60, in=270] (11, 0);
        \draw[msgarrow] (q3.north) to[out=120, in=270] (11, 0);
        
        \draw[msgarrow] (q1.north) to[out=45, in=270] (15, 0);

        \draw[msgarrow] (z2.north) to[out=90, in=270] (7, 5.5); 
        
        \draw[msgarrow] (z1.north) to[out=60, in=270] (11, 5.5);
        \draw[msgarrow] (z3.north) to[out=120, in=270] (11, 5.5);
        
        \draw[msgarrow] (z1.north) to[out=45, in=270] (15, 5.5);
    \end{scope}

    \node[font=\Huge] at (6, 9.8) {$\vdots$};
    \node[font=\Huge] at (10, 9.8) {$\vdots$};
    \node[font=\Huge] at (14, 9.8) {$\vdots$};

    \node[znode] (zL1)   at (6, 11) {$\bm{z}_1^L$};
    \node[znode] (zL2)   at (10, 11) {$\bm{z}_2^L$};
    \node[znode] (zL3)   at (14, 11) {$\bm{z}_3^L$};

    \node[poolblock] (pool) at (10, 13) {Sum Pooling};
    
    \draw[signal] (zL1.north) to[out=90, in=180] (pool.west);
    \draw[signal] (zL2.north) -- (pool.south);
    \draw[signal] (zL3.north) to[out=90, in=0] (pool.east);

    \node[font=\sffamily\huge\bfseries, text=black] (qtot) at (10, 15) {$Q_{tot}$};
    \draw[signal, line width=1.5pt] (pool) -- node[left, font=\sffamily\bfseries] {$\bm{v}$} (qtot);

\end{tikzpicture}

%% file: 6_exp.tex
\section{Experiments}
\label{sec: exp}

We evaluate \coolname{} on two cooperative MARL benchmarks: Predator-Prey (PP), widely used in the communication literature~\cite{schednet,ic3net,i2c}, and Lumberjacks (LJ)~\cite{lj_paper,lumber_liuyan}, a coordination benchmark requiring multi-agent proximity.
These environments permit controlled variation of wireless conditions while keeping the game structure interpretable for the behavioral analysis of Section~\ref{sec: exp behavioral}.
We evaluate across settings of increasing complexity (3--5 agents, $7\times7$ to $10\times10$ grids, added obstacles with signal attenuation).

\subsection{Setup}

\parag{Wireless Environment}
\label{sec: env wifi setup}
Agents communicate via single-hop broadcast with log-distance path loss, interference, and obstacle attenuation, using slotted $p$-CSMA~\cite{gaiyi} for medium access (see Appendices~\ref{appendix: pathloss}--\ref{appendix: wireless parameters} for details).
Even with 3--5 agents, log-distance path loss, obstacle shadowing, and contention under $p$-CSMA~\cite{gaiyi} produce non-trivial packet loss rates (see Appendix~\ref{appendix: wireless parameters}), ensuring that the realized communication graph varies stochastically from step to step.

\parag{MARL Algorithms and Hyperparameters}
We use a controlled $2 \times 2$ design isolating (i)~communication presence and (ii)~communication-awareness in the mixer, with all other factors held fixed.
All experiments use 5 random seeds~\cite{smac,qmix_journal}.
\subparag{No communication.}{1} VDN~\cite{vdn}, QMIX~\cite{qmix}, graph-agnostic mixers without message exchange.
\subparag{Communication + graph-agnostic mixing.}{2} TarMAC+VDN, TarMAC+QMIX, communication-enhanced agents with standard graph-agnostic mixers. TarMAC~\cite{tarmac} is an attention-based message aggregation architecture; because the original TarMAC is formulated in an actor--critic setting, we adapt its communication module to a value-decomposition framework for a controlled comparison.
\subparag{Communication + communication-aware mixing.}{3} \coolname{} (Section~\ref{sec: method complete}).
The mixer ablation in Section~\ref{sec: exp ablation} further isolates the mixer's contribution by replacing only the GNN mixer with a standard QMIX hypernetwork while keeping all other components identical.
Agents execute binary transmit/silent decisions throughout all experiments; investigating richer communication actions (e.g., power control) is left for future work.
See Appendix~\ref{appendix: training hyperparameters} for training hyperparameters. Code will be released upon publication.

\parag{Game Environment}

\subparag{Predator-Prey.}{}
Predators search for a stationary prey on a grid, receiving reward upon arrival; the episode ends when all predators reach the prey.
Obstacle variants (\ppw{g}) add barriers that block movement and attenuate wireless signals ($\delta_0{=}4.5$).
We evaluate with 3 predators on $7{\times}7$ and 4 predators on $10{\times}10$ ($k{=}1$ barrier, $\ell{=}9$).

\subparag{Lumberjacks.}{}\cite{lj_paper,lumber_liuyan}
Multiple lumberjack agents navigate a grid to cooperatively fell trees, each requiring at least $k{=}2$ adjacent agents for a successful chop.
Agents receive $r_1{=}0.05$ for observing a tree and $r_2{=}0.5$ for chopping one, with a step penalty $r_3{=}{-}0.1$ encouraging efficient coordination.
Trees act as wireless obstacles with signal attenuation $\delta{=}4.5$, creating location-dependent link degradation.
The episode ends when all trees are felled or a step limit is reached.
We denote {\lj{g}{p}{q}} for $p$ lumberjacks and $q$ trees on a $g{\times}g$ grid, and evaluate with 4 agents / 3 trees on $7{\times}7$ and 5 agents / 3 trees on $10{\times}10$.

\subsection{Performance Comparison}
\label{sec: exp sota}

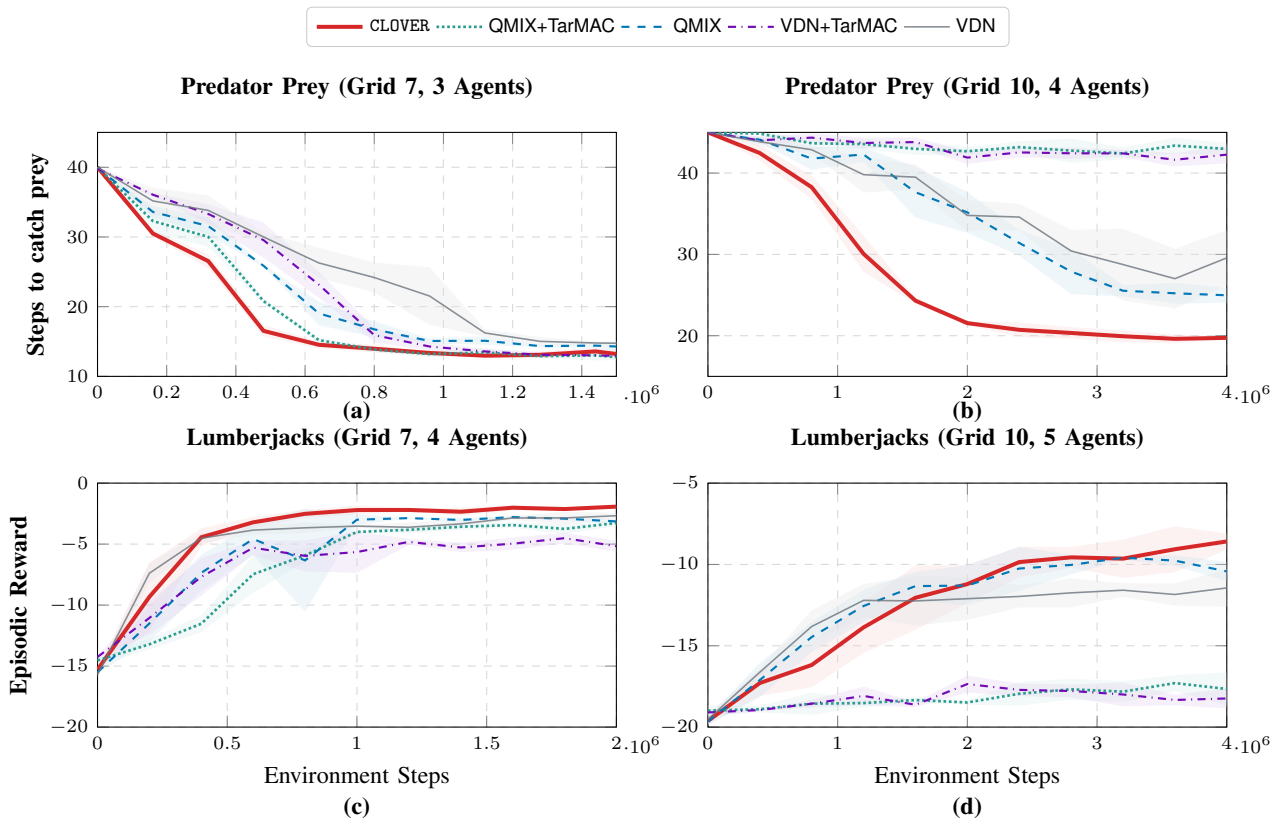
\begin{figure*}[t]
    \centering
    \resizebox{\textwidth}{!}{\input{./plots_new/sota_steps_2026.tex}}
    \vspace{-.2cm}
    \caption{Performance comparison. (a)~PP with obstacles, $7\times7$ grid, 3 agents. (b)~PP with obstacles, $10\times10$ grid, 4 agents. (c)~LJ with tree shadowing, $7\times7$ grid, 4 agents, 3 trees. (d)~LJ with tree shadowing, $10\times10$ grid, 5 agents, 3 trees. Subfigures (a)--(b) show number of steps to catch the prey (lower is better); subfigures (c)--(d) show episode return (higher is better). Shaded regions: $\pm 1$ standard deviation over 5 seeds.}
    \label{fig: sota steps}
\end{figure*}

The wireless observation $o_i^{\mathcal{C}}$ consists of RSS measurements throughout all experiments.
Figure~\ref{fig: sota steps} shows the learning curves (steps to catch prey for PP, lower is better; episode return for LJ, higher is better). All methods train for $4\times10^{6}$ environment steps; the x-axis in subfigures~(a) and~(c) is truncated after convergence for readability.

\subsubsection{PP}
In the smaller grid (Figure~\ref{fig: sota steps}a), all communication-enhanced methods converge to similar terminal performance (${\sim}13$ steps), but \coolname{} reaches this level faster than TarMAC+QMIX.
As environment complexity increases (Figure~\ref{fig: sota steps}b), \coolname{}'s advantage extends to both convergence speed and terminal performance: \coolname{} converges to $19.7 \pm 0.1$ steps, compared with $25.0 \pm 0.9$ for QMIX and $43.0 \pm 0.7$ for TarMAC+QMIX, while exhibiting the lowest variance across seeds.

\subsubsection{LJ}

In Figure~\ref{fig: sota steps}c, \coolname{} achieves the highest episode return and the fastest convergence. TarMAC+VDN fails to converge to a good optimum, while VDN, QMIX, and TarMAC+QMIX reach intermediate returns within a similar range. Compared with PP, LJ requires stronger multi-agent coordination because chopping a tree requires at least $k{=}2$ adjacent agents, making credit assignment more difficult.

In Figure~\ref{fig: sota steps}d, the larger grid ($10{\times}10$) and additional agent (5 lumberjacks) increase coordination difficulty; we fix the training budget at $4\times 10^{6}$ steps. Again, \coolname{} achieves the largest episode return compared with the baseline algorithms. We also observe that up to roughly $2\times 10^{6}$ steps, \coolname{} has a convergence speed similar to those of the QMIX and VDN baselines. However, the baselines converge to poorer local optima, whereas \coolname{} continues to improve and ultimately outperforms them.

TarMAC-aided VDN's failure in Figure~\ref{fig: sota steps}d is driven by the interaction of credit assignment difficulty with wireless unreliability. Tree-shadowing attenuation ($\delta{=}4.5$) degrades messages between agents on opposite sides of a tree, producing inconsistent communication patterns that destabilize joint value estimates. \coolname{} mitigates this by conditioning the mixer on the realized communication graph, explicitly accounting for which agents successfully exchanged messages.

\subsubsection{Overall Performance}
Across all settings, \coolname{} achieves the best terminal performance with the lowest variance. In three of four settings it also converges fastest; in LJ $10{\times}10$ (Figure~\ref{fig: sota steps}d) it initially matches the baseline convergence rate but surpasses all baselines after ${\sim}2\times10^{6}$ steps. Existing communication designs (TarMAC+VDN, TarMAC+QMIX) often converge to poorer local optima under realistic wireless conditions, plausibly because channel stochasticity makes learning harder.
\coolname{}'s advantage over the best-performing baseline grows as grid size and agent count increase (PP $7\times7$/3 $\to$ $10\times10$/4; LJ $7\times7$/4 $\to$ $10\times10$/5), consistent with the expressive-power analysis of Theorem~\ref{thm: exp power}: as the environment grows (larger grid, more agents), the space of possible communication topologies increases, and conditioning the mixer on the realized graph yields a correspondingly larger advantage.

\begin{figure}[t]
    \centering
    \resizebox{\columnwidth}{!}{\input{./plots/traj_comm}}
    \caption{Average communication action per agent (left axis) and prey-finding distribution (right axis, orange bars) over 2000 post-convergence trajectories in PP~$10{\times}10$, 3 agents.}
    \label{fig: exp when traj}
\end{figure}
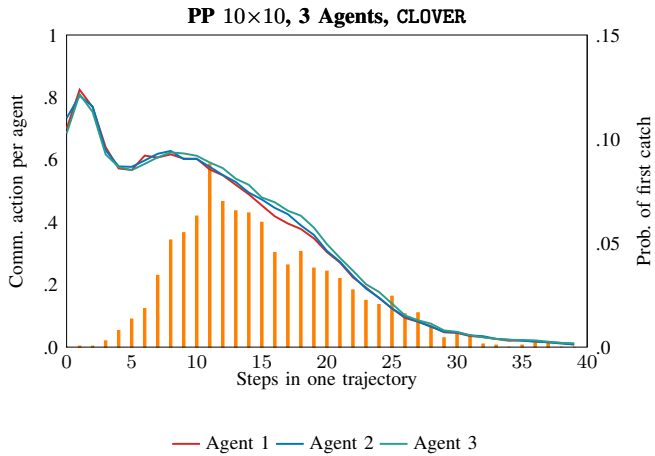

\subsection{Behavior Analysis}
\label{sec: exp behavioral}
The behavioral analyses below use the PP~$10{\times}10$, 3-agent configuration, which provides clean single-prey-finding events for analysis.

\begin{figure}[!htb]
    \centering
    \begin{subfigure}[t]{0.45\columnwidth}\centering\input{./diagrams/traj_demo}
    \caption{}\label{fig: traj_before_catch}\end{subfigure}
    \hfill
    \begin{subfigure}[t]{0.45\columnwidth}\centering\input{./diagrams/traj_demo_2}
    \caption{}\label{fig: traj_after_catch}\end{subfigure}

    \vspace{4mm}

    \begin{subfigure}[t]{0.45\columnwidth}\centering\input{./diagrams/traj_demo_3}
    \caption{}\label{fig: agent_before_catch}\end{subfigure}
    \hfill
    \begin{subfigure}[t]{0.45\columnwidth}\centering\input{./diagrams/traj_demo_4}
    \caption{}\label{fig: agent_after_catch}\end{subfigure}

\caption{Sample trajectory illustrating positive listening in PP~$10{\times}10$, 3 agents. (a)~Steps 1--9: Agent~1 finds the prey at step~9 and broadcasts. (b)~Steps 10--15: Agent~2, after receiving the message, reverses direction toward the prey. (c)~Agent positions at step~9. (d)~Agent positions at step~10.}
\label{fig: trajectory positive listening}
\end{figure}
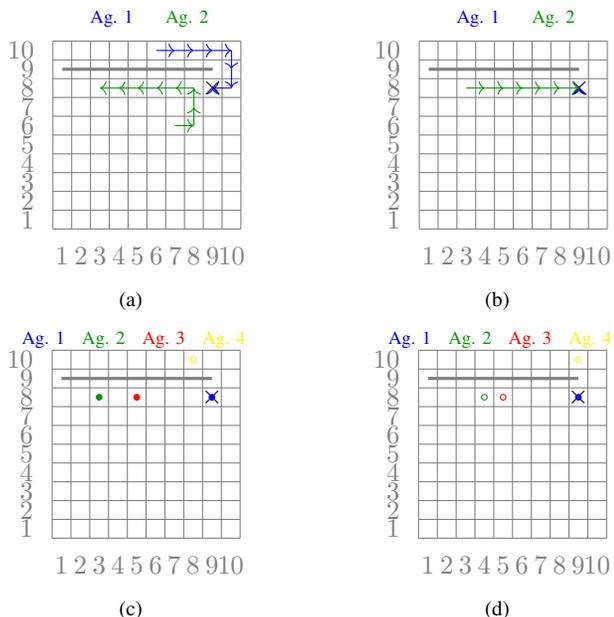

\begin{table}[!htb]
  \vspace{4mm}
  \centering
  \caption{Reward gain from positive listening. $r_\text{ref}$: step penalty, included for scale.}
  \label{tab: exp reward gain positive listening}
  \resizebox{0.8\columnwidth}{!}{\input{tables/positive_listening_reward}}
\end{table}

\begin{figure}[!htb]
    \centering
    \includegraphics[width=0.85\columnwidth]{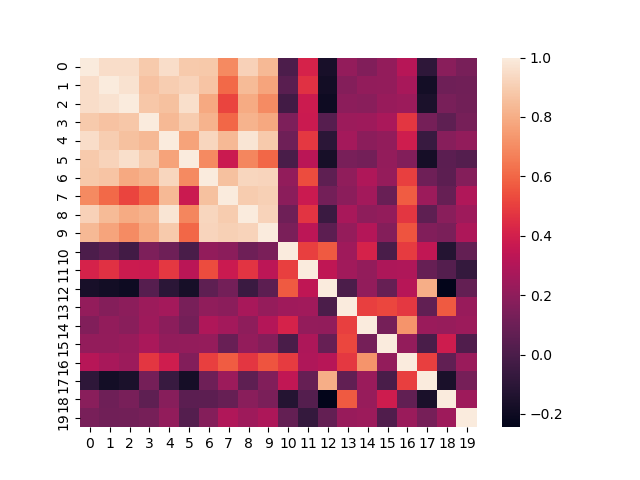}
    \caption{Cosine similarity matrix of messages sampled from 2000 evaluation episodes in PP~$10{\times}10$. Messages~0--9 correspond to prey-finding events; their high intra-group similarity (dark diagonal block) confirms semantically consistent prey-finding representations.}
    \label{fig: cosine sim}
\end{figure}

\subsubsection{Learned Communication Strategy}
Figure~\ref{fig: exp when traj} shows the average communication action per agent and the prey-finding distribution across 2000 post-convergence trajectories. Once training converges, we freeze the model and record two metrics: (1) the binary communication action $a_i^{C,t}$ of each agent $i$ at each step $t$, and (2) the first step $t'$ at which at least one predator catches the prey. The left axis records the average $a_i^{C,t}$ over the 2000 trajectories; the right axis records the probability distribution of $t'$.

We observe the following: (1) The three agent curves nearly overlap because the agents are homogeneous. (2) At initial steps, agents are more likely to communicate, helping them infer others' initial positions and probe both the wireless and game environments. (3) Agents are also more likely to communicate around prey-catching events, as reflected by the similar shapes of the curves and the prey-finding distribution after step~10. After the prey-catching agent informs the others of the prey's position, the remaining agents can approach the prey more directly.

\subsubsection{Positive Signaling and Positive Listening}
Positive signaling means messages reflect meaningful sender observations; positive listening means messages influence receiver behavior~\cite{toru}. Figure~\ref{fig: trajectory positive listening} illustrates positive listening in PP: Agent~1 finds the prey and broadcasts (step 9); Agent~2, which is blind by design, receives the message and immediately reverses direction to approach the prey (step 10).

Table~\ref{tab: exp reward gain positive listening} quantifies positive listening via the reward gain when communication is enabled~\cite{pspl}. The gains of 1.44 (PP) and 1.00 (LJ) are substantial: in LJ the gain represents ${\sim}12\%$ of the absolute episode reward magnitude ($|{-}8.45|$).

For positive signaling, we measure speaker consistency~\cite{jaques2019social}, defined as intra-group cosine similarity of messages associated with the same event type. We load a converged model and roll it out for 2000 episodes. We randomly sample 10 messages from episodes in which an agent first observes the prey (messages~0--9), and another 10 messages from episodes in which the prey is not observed (messages~10--19). Since we use parameter sharing during training, all agents share the same message encoder architecture and weights. We compute pairwise cosine similarity of these messages in the encoder-induced latent space. A Mann-Whitney $U$ test confirms that prey-finding messages are significantly more consistent ($p < 0.01$). Figure~\ref{fig: cosine sim} visualizes the resulting $20\times20$ cosine-similarity matrix: messages~0--9 (prey-finding) exhibit visibly higher mutual similarity than the others, confirming that the encoder learns semantically consistent prey-finding representations.

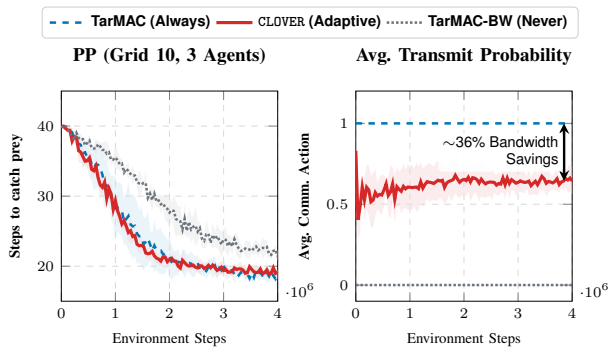
\begin{figure}[!htb]
    \centering
    \resizebox{\columnwidth}{!}{\input{./plots/ablation_when_2026}}
    \caption{Bandwidth adaptation in PP~$10{\times}10$, 3 agents. Left: steps to catch prey. Right: average transmit probability. Under reduced bandwidth (``-BW''), \coolname{} maintains performance by reducing communication frequency, while TarMAC degrades.}
    \label{fig: exp when conv}
\end{figure}

\subsubsection{Adapting to Bandwidth Limit Change}
Figure~\ref{fig: exp when conv} compares \coolname{} (learned transmit policy) against TarMAC with always-communicate in the standard bandwidth setting, and TarMAC under reduced bandwidth (``TarMAC-BW'', with fewer $p$-CSMA time slots~\cite{gaiyi}). TarMAC+QMIX is highly sensitive to bandwidth reduction, while \coolname{} adapts by reducing communication frequency: the learned transmit probability converges to ${\sim}0.65$ in the standard setting, achieving comparable terminal performance to always-communicate with ${\sim}35\%$ fewer transmissions. Under reduced bandwidth, \coolname{} shows only slightly slower convergence.

\begin{table}[!htb]
  \vspace{6mm}
  \centering
  \caption{Benefits of proposed message encoder}
  \label{tab: exp ablation enc}
  \resizebox{0.8\columnwidth}{!}{\input{tables/ablation_mlp_gin}}
\end{table}

\subsection{Ablation Study}
\label{sec: exp ablation}

\subsubsection{Message Encoder Ablation}
Table~\ref{tab: exp ablation enc} compares three encoders (results from~\cite{LAUREL_ACML}, using the same wireless environment and hyperparameters but without the GNN mixer):
(i)~MLP with fixed-order concatenation,
(ii)~average-then-MLP, and
(iii)~the proposed sum-of-MLPs (Eq.~\ref{eq: how gnn}).
The proposed encoder yields substantially better performance, consistent with Theorem~\ref{thm: gin emb}.

\subsubsection{Communication-Enhanced Mixer Ablation}

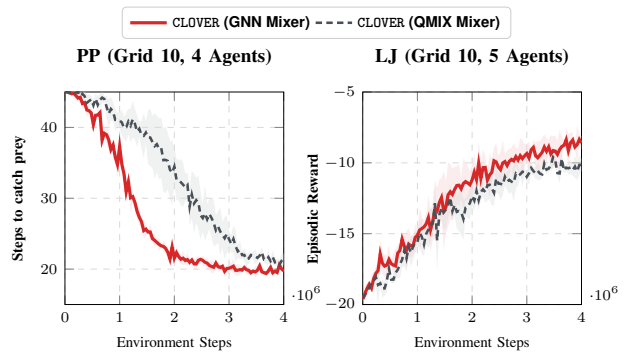
\begin{figure}[t]
    \centering
    \resizebox{\columnwidth}{!}{\input{./plots_new/ablation_gnnHyper_2026.tex}}
  \caption{Mixer ablation comparing the proposed GNN-based mixer with a standard QMIX hypernetwork while holding all other components fixed. (a)~Predator-Prey, $10{\times}10$, 4 agents. (b)~Lumberjacks, $10{\times}10$, 5 agents, 3 trees. This ablation isolates the effect of communication-graph-conditioned mixing during centralized training.}
  \label{fig: ablation gnnhyper}
  \end{figure}

As shown in Figure~\ref{fig: ablation gnnhyper}, the proposed GNN mixer consistently outperforms the QMIX hypernetwork in both the harder PP ($10{\times}10$, 4 agents) and LJ ($10{\times}10$, 5 agents) settings: the GNN mixer converges to $19.7 \pm 0.1$ steps vs.\ $21.4 \pm 0.8$ in PP and a return of $-8.6 \pm 0.5$ vs.\ $-10.1 \pm 0.4$ in LJ. Since the only difference is whether the mixer conditions on the realized communication graph, this is consistent with the prediction of Theorem~\ref{thm: exp power}: structurally different communication topologies induce meaningfully different credit assignment.
The GNN mixer additionally preserves permutation invariance (unlike standard QMIX, Section~\ref{sec: method mixer math}), which may further contribute to sample efficiency.

Together with Table~\ref{tab: exp reward gain positive listening} (agent-level communication ablation), this provides complementary evidence that both message exchange and graph-conditioned mixing contribute to \coolname{}'s performance.

%% file: plots_new/sota_steps_2026.tex
\definecolor{laurelRed}{HTML}{D62828}
\definecolor{qmixBlue}{HTML}{0077B6}
\definecolor{tarmacGreen}{HTML}{2A9D8F}
\definecolor{vdnPurple}{HTML}{7209B7}
\definecolor{grayBase}{HTML}{6C757D}

\pgfplotsset{
    hero curve/.style={color=laurelRed, solid, line width=1.5pt},
    base1 curve/.style={color=qmixBlue, dashed, line width=0.8pt},
    base2 curve/.style={color=tarmacGreen, densely dotted, line width=1pt},
    base3 curve/.style={color=vdnPurple, dashdotted, line width=0.8pt},
    base4 curve/.style={color=grayBase, solid, line width=0.6pt, opacity=0.8},
}

\newcommand{\addlearningcurve}[7]{%
    \addplot[name path=upper, draw=none, forget plot] table[x index=#2, y expr=\thisrowno{#3}+\thisrowno{#4}] {#1};
    \addplot[name path=lower, draw=none, forget plot] table[x index=#2, y expr=\thisrowno{#3}-\thisrowno{#4}] {#1};
    \addplot[#6!20, opacity=0.3, forget plot] fill between[of=upper and lower];
    \addplot[#5] table[x index=#2, y index=#3] {#1};
    \addlegendentry{#7}
}

\newcommand{\addlearningcurvenolegend}[6]{%
    \addplot[name path=upper, draw=none, forget plot] table[x index=#2, y expr=\thisrowno{#3}+\thisrowno{#4}] {#1};
    \addplot[name path=lower, draw=none, forget plot] table[x index=#2, y expr=\thisrowno{#3}-\thisrowno{#4}] {#1};
    \addplot[#6!20, opacity=0.3, forget plot] fill between[of=upper and lower];
    \addplot[#5] table[x index=#2, y index=#3] {#1};
}

\pgfplotstableread[col sep=comma]{
ep,steps_qmix_avg,steps_qmix_var,reward_qmix_avg,reward_qmix_var,steps_qmix_tarmac_avg,steps_qmix_tarmac_var,reward_qmix_tarmac_avg,reward_qmix_tarmac_var,steps_gnnHyer_avg,steps_gnnHyer_var,reward_gnnHyer_avg,reward_gnnHyer_var,steps_vdn_tarmac_avg,steps_vdn_tarmac_var,reward_vdn_tarmac_avg,reward_vdn_tarmac_var,steps_vdn_avg,steps_vdn_var,reward_vdn_avg,reward_vdn_var
0.0,39.911,0.125,-10.859,0.184,40.0,0.0,-11.464,0.235,40.0,0.0,-11.187,0.292,39.822,0.250,-11.073,0.326,40.0,0.0,-11.194,0.183
160000.0,33.596,0.848,-6.533,0.634,32.289,1.732,-6.238,0.664,30.497,0.478,-5.147,0.099,36.062,0.277,-7.400,0.246,35.169,1.901,-7.438,0.650
320000.0,31.562,1.226,-4.956,0.452,30.018,1.407,-4.851,0.099,26.536,1.019,-3.379,0.631,33.304,1.394,-6.214,0.416,33.817,2.132,-6.425,1.167
480000.0,25.862,0.188,-2.890,0.219,20.796,0.390,-1.365,0.194,16.526,0.824,-0.335,0.146,29.544,2.503,-4.811,1.017,30.005,0.625,-4.782,0.443
640000.0,19.028,1.675,-0.759,0.363,15.197,0.337,0.009,0.119,14.531,0.496,0.068,0.134,23.153,1.876,-2.083,0.694,26.270,2.181,-3.562,1.063
800000.0,16.747,1.151,-0.293,0.204,13.851,0.404,0.212,0.086,13.966,0.512,0.261,0.121,15.921,1.033,-0.111,0.300,24.184,2.147,-2.563,1.022
960000.0,15.075,0.366,0.089,0.048,13.190,0.400,0.373,0.061,13.361,0.221,0.300,0.085,14.283,0.576,0.221,0.062,21.526,4.122,-1.470,1.153
1120000.0,15.109,0.594,0.024,0.141,13.445,0.582,0.352,0.104,12.968,0.285,0.360,0.070,13.580,0.396,0.291,0.046,16.229,0.527,-0.081,0.074
1280000.0,14.343,0.193,0.219,0.025,12.903,0.301,0.410,0.085,13.093,0.488,0.367,0.136,13.101,0.382,0.490,0.058,15.020,0.607,0.117,0.081
1440000.0,14.411,0.365,0.207,0.041,13.005,0.734,0.430,0.153,13.580,0.501,0.275,0.092,13.018,0.045,0.490,0.059,14.789,0.223,0.207,0.101
1500000.0,14.283,0.690,0.249,0.069,12.760,0.390,0.461,0.113,13.213,0.390,0.370,0.134,12.929,0.226,0.449,0.045,14.755,0.285,0.261,0.041
}\dataPPseven

\pgfplotstableread[col sep=comma]{
ep,steps_qmix_avg,steps_qmix_var,reward_qmix_avg,reward_qmix_var,steps_qmix_tarmac_avg,steps_qmix_tarmac_var,reward_qmix_tarmac_avg,reward_qmix_tarmac_var,steps_gnnHyer_avg,steps_gnnHyer_var,reward_gnnHyer_avg,reward_gnnHyer_var,steps_vdn_tarmac_avg,steps_vdn_tarmac_var,reward_vdn_tarmac_avg,reward_vdn_tarmac_var,steps_vdn_avg,steps_vdn_var,reward_vdn_avg,reward_vdn_var
0.0,45.0,0.0,-17.321,0.331,45.0,0.0,-17.268,0.624,45.0,0.0,-17.275,0.079,44.973,0.036,-16.293,0.468,45.0,0.0,-17.460,0.313
400000.0,44.101,0.327,-14.029,0.765,44.854,0.179,-15.341,0.184,42.460,0.946,-12.809,0.685,44.028,0.487,-14.225,0.641,43.856,0.567,-13.247,1.056
800000.0,41.783,1.401,-12.310,0.414,43.664,0.663,-13.678,0.506,38.257,1.429,-9.934,0.725,44.356,0.554,-13.702,0.526,42.867,0.499,-12.268,0.467
1200000.0,42.283,0.899,-11.690,0.214,43.562,0.581,-13.001,0.569,30.065,2.224,-5.694,1.305,43.697,0.659,-13.142,1.092,39.804,2.188,-10.251,0.989
1600000.0,37.645,3.143,-9.275,1.293,42.976,0.729,-12.601,0.817,24.307,0.373,-2.778,0.248,43.812,0.582,-13.217,0.218,39.497,1.528,-9.579,0.964
2000000.0,35.161,2.514,-7.519,0.538,42.679,0.521,-12.417,0.322,21.546,0.266,-1.910,0.087,41.901,0.788,-12.107,0.378,34.791,1.883,-7.377,1.183
2400000.0,31.408,1.579,-5.949,1.097,43.190,0.698,-12.753,0.850,20.723,0.884,-1.515,0.302,42.539,0.607,-12.226,0.346,34.593,1.633,-6.651,1.246
2800000.0,27.888,2.783,-4.354,0.994,42.760,1.497,-13.604,1.716,20.338,0.708,-1.347,0.212,42.432,0.701,-12.378,0.267,30.421,2.588,-4.839,1.562
3200000.0,25.531,0.689,-3.080,0.180,42.416,0.728,-12.990,0.297,19.934,0.710,-1.446,0.381,42.414,0.452,-12.690,0.102,28.752,4.402,-4.121,1.794
3600000.0,25.218,1.285,-2.916,0.169,43.377,0.978,-13.077,0.369,19.617,0.552,-1.292,0.161,41.638,0.831,-12.400,0.383,27.013,3.672,-3.581,1.764
4000000.0,24.971,0.905,-2.682,0.301,42.966,0.720,-12.542,0.107,19.736,0.125,-1.222,0.106,42.263,1.066,-12.717,0.263,29.572,3.408,-4.277,1.449
}\dataPPten

\pgfplotstableread[col sep=comma]{
ep,steps_gnnHyper_avg,steps_gnnHyper_var,reward_gnnHyper_avg,reward_gnnHyper_var,steps_qmix_tarmac_avg,steps_qmix_tarmac_var,reward_qmix_tarmac_avg,reward_qmix_tarmac_var,steps_qmix_avg,steps_qmix_var,reward_qmix_avg,reward_qmix_var,steps_vdn_tarmac_avg,steps_vdn_tarmac_var,reward_vdn_tarmac_avg,reward_vdn_tarmac_var,steps_vdn_avg,steps_vdn_var,reward_vdn_avg,reward_vdn_var
0.0,39.729,0.003,-15.286,0.026,39.052,0.299,-14.532,0.291,39.820,0.254,-15.508,0.087,38.924,0.195,-14.251,0.122,39.921,0.110,-15.692,0.256
200000.0,29.658,5.399,-9.323,2.826,37.510,0.632,-13.213,0.440,33.997,2.766,-11.503,1.486,33.593,2.614,-11.051,1.383,25.742,1.468,-7.384,0.783
400000.0,20.054,1.139,-4.441,0.727,34.497,1.436,-11.525,0.688,25.559,3.154,-7.356,1.469,26.570,3.252,-7.686,1.531,19.750,1.248,-4.519,0.435
600000.0,18.213,0.640,-3.214,0.445,26.114,2.982,-7.469,1.501,19.794,2.601,-4.577,1.311,21.744,1.312,-5.294,0.592,18.638,0.709,-3.853,0.362
800000.0,16.763,0.390,-2.518,0.344,23.299,1.024,-5.923,0.562,23.460,7.999,-6.334,4.218,23.171,2.702,-5.967,1.274,18.052,1.043,-3.663,0.451
1000000.0,16.604,0.499,-2.207,0.377,19.205,0.262,-4.002,0.066,16.958,0.427,-2.998,0.429,22.664,3.607,-5.652,1.702,17.658,0.263,-3.526,0.100
1200000.0,16.697,0.199,-2.199,0.136,19.213,0.413,-3.827,0.296,16.914,0.369,-2.869,0.045,20.914,0.465,-4.816,0.248,18.046,0.912,-3.619,0.394
1400000.0,17.247,0.389,-2.345,0.331,18.710,0.566,-3.577,0.256,17.156,0.430,-3.021,0.182,21.906,0.850,-5.276,0.389,17.494,0.678,-3.340,0.317
1600000.0,16.471,0.535,-2.017,0.224,18.263,0.405,-3.440,0.224,16.570,0.537,-2.785,0.066,21.190,1.011,-4.965,0.496,16.502,0.174,-2.858,0.163
1800000.0,17.007,0.206,-2.126,0.041,19.190,0.755,-3.752,0.409,17.0,0.462,-2.915,0.160,20.117,0.880,-4.514,0.444,16.471,1.135,-2.855,0.521
2000000.0,16.434,0.346,-1.928,0.081,18.375,0.456,-3.271,0.260,17.526,0.329,-3.138,0.207,21.453,1.073,-5.170,0.440,16.195,0.033,-2.674,0.213
}\dataLJseven

\pgfplotstableread[col sep=comma]{
ep,steps_gnnHyper_avg,steps_gnnHyper_var,reward_gnnHyper_avg,reward_gnnHyper_var,steps_qmix_tarmac_avg,steps_qmix_tarmac_var,reward_qmix_tarmac_avg,reward_qmix_tarmac_var,steps_qmix_avg,steps_qmix_var,reward_qmix_avg,reward_qmix_var,steps_vdn_tarmac_avg,steps_vdn_tarmac_var,reward_vdn_tarmac_avg,reward_vdn_tarmac_var,steps_vdn_avg,steps_vdn_var,reward_vdn_avg,reward_vdn_var
0.0,39.919,0.114,-19.622,0.216,39.546,0.351,-18.988,0.324,39.830,0.239,-19.654,0.242,39.807,0.150,-19.099,0.200,39.760,0.338,-19.522,0.318
400000.0,37.833,1.026,-17.290,0.800,39.708,0.068,-18.900,0.201,37.690,1.416,-17.106,1.070,39.729,0.113,-18.953,0.073,36.835,0.204,-16.625,0.242
800000.0,36.255,1.986,-16.182,1.359,39.356,0.666,-18.561,0.677,33.697,1.552,-14.463,0.932,39.518,0.180,-18.562,0.234,33.039,1.544,-13.814,1.000
1200000.0,32.875,2.538,-13.869,1.575,39.312,0.392,-18.522,0.238,31.182,1.228,-12.555,0.991,39.080,0.693,-18.084,0.572,30.786,1.300,-12.209,1.002
1600000.0,30.075,2.978,-12.047,1.983,39.132,0.313,-18.337,0.144,30.083,0.838,-11.331,0.896,39.479,0.178,-18.628,0.156,30.885,1.951,-12.243,1.515
2000000.0,28.567,2.048,-11.208,1.177,39.236,0.200,-18.486,0.140,29.942,1.227,-11.299,1.228,38.065,0.787,-17.354,0.524,31.205,1.931,-12.106,1.310
2400000.0,26.132,1.802,-9.855,0.945,38.677,0.762,-17.950,0.719,29.190,1.654,-10.251,1.357,38.403,0.622,-17.712,0.384,30.804,0.719,-11.968,0.675
2800000.0,25.721,1.472,-9.564,0.638,38.234,0.730,-17.678,0.645,28.656,0.648,-10.028,0.591,38.466,0.540,-17.780,0.401,30.281,1.391,-11.747,0.888
3200000.0,25.750,2.164,-9.644,1.178,38.492,0.966,-17.820,0.640,27.812,0.846,-9.586,0.120,38.833,0.883,-17.995,0.731,30.010,0.471,-11.584,0.422
3600000.0,24.617,2.624,-9.064,1.414,37.695,0.608,-17.296,0.484,27.747,1.011,-9.764,0.337,39.041,0.380,-18.332,0.324,30.799,0.967,-11.847,0.676
4000000.0,23.611,0.953,-8.587,0.490,38.117,1.235,-17.655,1.018,28.713,0.905,-10.430,0.585,39.000,0.606,-18.240,0.616,30.002,1.601,-11.444,1.158
}\dataLJten

\begin{tikzpicture}[every node/.style={transform shape}]

\begin{groupplot}[
    group style={
        group size=2 by 2,
        horizontal sep=1.2cm,
        vertical sep=1.4cm,
    },
    scale only axis,
    height=3.2cm,
    width=6.8cm,
    tick label style={font=\scriptsize},
    title style={font=\small\bfseries, yshift=1mm},
    grid=major,
    grid style={dashed, gray!30},
    every x tick scale label/.style={at={(xticklabel* cs:1.05,0cm)}, anchor=near xticklabel},
    legend style={
        at={(1.08, 1.35)},
        anchor=south,
        legend columns=5,
        draw=black!20,
        fill=white,
        rounded corners=2pt,
        font=\scriptsize\sffamily
    }
]

\nextgroupplot[
    title={Predator Prey (Grid 7, 3 Agents)},
    ymin=10, ymax=45, xmin=0, xmax=1500000,
    ylabel={Steps to catch prey},
    ylabel style={font=\small\bfseries},
]
\addlearningcurve{\dataPPseven}{0}{9}{10}{hero curve}{laurelRed}{\coolname{}}
\addlearningcurve{\dataPPseven}{0}{5}{6}{base2 curve}{tarmacGreen}{QMIX+TarMAC}
\addlearningcurve{\dataPPseven}{0}{1}{2}{base1 curve}{qmixBlue}{QMIX}
\addlearningcurve{\dataPPseven}{0}{13}{14}{base3 curve}{vdnPurple}{VDN+TarMAC}
\addlearningcurve{\dataPPseven}{0}{17}{18}{base4 curve}{grayBase}{VDN}

\nextgroupplot[
    title={Predator Prey (Grid 10, 4 Agents)},
    ymin=15, ymax=45, xmin=0, xmax=4000000,
]
\addlearningcurvenolegend{\dataPPten}{0}{9}{10}{hero curve}{laurelRed}
\addlearningcurvenolegend{\dataPPten}{0}{5}{6}{base2 curve}{tarmacGreen}
\addlearningcurvenolegend{\dataPPten}{0}{1}{2}{base1 curve}{qmixBlue}
\addlearningcurvenolegend{\dataPPten}{0}{13}{14}{base3 curve}{vdnPurple}
\addlearningcurvenolegend{\dataPPten}{0}{17}{18}{base4 curve}{grayBase}

\nextgroupplot[
    title={Lumberjacks (Grid 7, 4 Agents)},
    ymin=-20, ymax=0, xmin=0, xmax=2000000,
    xlabel={Environment Steps},
    ylabel={Episodic Reward},
    xlabel style={font=\small},
    ylabel style={font=\small\bfseries},
]
\addlearningcurvenolegend{\dataLJseven}{0}{3}{4}{hero curve}{laurelRed}
\addlearningcurvenolegend{\dataLJseven}{0}{7}{8}{base2 curve}{tarmacGreen}
\addlearningcurvenolegend{\dataLJseven}{0}{11}{12}{base1 curve}{qmixBlue}
\addlearningcurvenolegend{\dataLJseven}{0}{15}{16}{base3 curve}{vdnPurple}
\addlearningcurvenolegend{\dataLJseven}{0}{19}{20}{base4 curve}{grayBase}

\nextgroupplot[
    title={Lumberjacks (Grid 10, 5 Agents)},
    ymin=-20, ymax=-5, xmin=0, xmax=4000000,
    xlabel={Environment Steps},
    xlabel style={font=\small},
]
\addlearningcurvenolegend{\dataLJten}{0}{3}{4}{hero curve}{laurelRed}
\addlearningcurvenolegend{\dataLJten}{0}{7}{8}{base2 curve}{tarmacGreen}
\addlearningcurvenolegend{\dataLJten}{0}{11}{12}{base1 curve}{qmixBlue}
\addlearningcurvenolegend{\dataLJten}{0}{15}{16}{base3 curve}{vdnPurple}
\addlearningcurvenolegend{\dataLJten}{0}{19}{20}{base4 curve}{grayBase}

\end{groupplot}

\node[anchor=north, font=\small\bfseries] at ($(group c1r1.south)+(0,-2mm)$) {(a)};
\node[anchor=north, font=\small\bfseries] at ($(group c2r1.south)+(0,-2mm)$) {(b)};
\node[anchor=north, font=\small\bfseries] at ($(group c1r2.south)+(0,-8mm)$) {(c)};
\node[anchor=north, font=\small\bfseries] at ($(group c2r2.south)+(0,-8mm)$) {(d)};

\end{tikzpicture}

%% file: plots/traj_comm.tex
\begin{tikzpicture}[
    every node/.style={transform shape}
]
\definecolor{agentRed}{HTML}{D62828}
\definecolor{agentBlue}{HTML}{0077B6}
\definecolor{agentTeal}{HTML}{2A9D8F}
\def\colA{agentRed}
\def\colB{agentBlue}
\def\colC{agentTeal}
\def\datafilecnt{./data/term_bincount.csv}
\pgfplotsset{
    scale only axis,
    xmin=0, xmax=40,
    title={PP~$10{\times}10$, 3 Agents, {\coolname}},
    height=4.5cm,
    width=7.5cm,
    title style={at={(axis description cs:0.5, 0.95)},font=\small\bfseries},
    x label style={at={(axis description cs:0.5,-.05)}},
    xlabel={Steps in one trajectory},
    xlabel style={font=\footnotesize},
    tick label style={font=\footnotesize},
    ytick style={draw=none},
    xtick style={draw=none},
}

\begin{axis}[
    axis y line*=right,
    axis x line=none,
    ymin=0, ymax=0.15,
    ylabel={Prob. of first catch},
    ylabel style={font=\footnotesize},
    ybar=\pgflinewidth,
    bar width=1pt,
    tick label style={font=\footnotesize},
    ytick={0, 0.05, 0.10, 0.15},
    yticklabels={.0, .05, .10, .15},
]
\addplot[color=orange,fill=orange] table[x=step, y=term_cnt, col sep=comma] {\datafilecnt};
\end{axis}

\begin{axis}[
    ymin=0, ymax=1,
    ylabel={Comm. action per agent},
    ylabel style={font=\footnotesize},
    yticklabels={.0, .2, .4, .6, .8, 1},
    ytick={0, 0.2, 0.4, 0.6, 0.8, 1},
    legend cell align=left,
    legend style={at={(0.5,-0.25)},anchor=north,draw=none,font=\footnotesize},
    legend columns=3,
    tick label style={font=\footnotesize},
]

\errorband[thick,color=\colA]{\dataTrajComm}{0}{1}{7}{Agent 1}
\errorband[thick,color=\colB]{\dataTrajComm}{0}{3}{7}{Agent 2}
\errorband[thick,color=\colC]{\dataTrajComm}{0}{5}{7}{Agent 3}
\end{axis}

\end{tikzpicture}

%% file: diagrams/traj_demo.tex
\begin{tikzpicture}
\def\colA{red}
\def\colB{blue}
\def\colC{green!60!black}
\def\colD{yellow}
\def\gridsize{10}
\def\scale{0.25}
\def\preyCol{black}

\def\preyPos{9, 8}
\def\predatorInfo{
{\colB/{6/10/0, 7/10/0, 8/10/0, 9/10/0, 10/10/0, 10/9/0, 10/8/1, 9/8/1, 9/8/1, 9/8/1, 9/8/1, 9/8/1, 9/8/1, 9/8/1, 9/8/1}},
{\colC/{7/6/1, 8/6/1, 8/7/0, 8/8/1, 7/8/1, 6/8/1, 5/8/1, 4/8/0, 3/8/1, 4/8/0, 5/8/0, 6/8/0, 7/8/0, 8/8/0, 9/8/0}}}
\def\wallInfo{
{gray/1/9/h/8}}

\foreach \i in {0,...,\gridsize} {
    \draw[thin,gray] ({(\i+0.5)*\scale}, {0.5*\scale}) --
                      ({(\i+0.5)*\scale}, {(\gridsize+0.5)*\scale});
    \ifthenelse{\i > 0}{
        \node[below,gray] at ({(\i+0.)*\scale}, {0.*\scale}) {$\i$};
    }{}
}
\foreach \i in {0,...,\gridsize} {
    \draw [thin,gray] ({0.5*\scale}, {(\i+0.5)*\scale}) --
                      ({(\gridsize+0.5)*\scale}, {(\i+0.5)*\scale});
    \ifthenelse{\i > 0} {
        \node[left,gray] (y-grid label \i) at ({0.*\scale},{\i*\scale}) {$\i$};
    }{}
}
\node[\preyCol] at ({{\preyPos}[0]*\scale}, {{\preyPos}[1]*\scale}) {$\times$};
\foreach \col/\pInfo [count=\ip] in \predatorInfo {
    \foreach \x/\y/\iscomm [count=\istep, remember=\x as \xp, remember=\y as \yp] in \pInfo {
        \ifthenelse{\iscomm = 1} {
        }{}
        \ifthenelse{\istep > 1 \and \istep <10} {
            \draw [\col, ->] ({\xp*\scale}, {\yp*\scale}) -- ({\x*\scale}, {\y*\scale});
        }{
        }
    }
}
\foreach \col/\x/\y/\hv/\len [count=\i] in \wallInfo {
    \ifthenelse{\equal{\hv}{h}} {
        \draw[\col, very thick] ({\x*\scale}, {\y*\scale}) --++ ({\len*\scale}, 0);
    }{
        \draw[\col, very thick] ({\x*\scale}, {\y*\scale}) --++ (0, {\len*\scale});
    }
}

\node [\colB, above right=.2 cm and 0.5 cm of y-grid label 9] (legend 2) {\scriptsize Ag. 1};
\node [\colC, above right=.2 cm and 1.5 cm of y-grid label 9] (legend 3) {\scriptsize Ag. 2};

\end{tikzpicture}

%% file: diagrams/traj_demo_2.tex
\begin{tikzpicture}
\def\colA{red}
\def\colB{blue}
\def\colC{green!60!black}
\def\colD{yellow}
\def\gridsize{10}
\def\scale{0.25}
\def\preyCol{black}

\def\preyPos{9, 8}
\def\predatorInfo{
{\colB/{6/10/0, 7/10/0, 8/10/0, 9/10/0, 10/10/0, 10/9/0, 10/8/1, 9/8/1, 9/8/1, 9/8/1, 9/8/1, 9/8/1, 9/8/1, 9/8/1, 9/8/1}},
{\colC/{7/6/1, 8/6/1, 8/7/0, 8/8/1, 7/8/1, 6/8/1, 5/8/1, 4/8/0, 3/8/1, 4/8/0, 5/8/0, 6/8/0, 7/8/0, 8/8/0, 9/8/0}}}
\def\wallInfo{
{gray/1/9/h/8}}

\foreach \i in {0,...,\gridsize} {
    \draw[thin,gray] ({(\i+0.5)*\scale}, {0.5*\scale}) --
                      ({(\i+0.5)*\scale}, {(\gridsize+0.5)*\scale});
    \ifthenelse{\i > 0}{
        \node[below,gray] at ({(\i+0.)*\scale}, {0.*\scale}) {$\i$};
    }{}
}
\foreach \i in {0,...,\gridsize} {
    \draw [thin,gray] ({0.5*\scale}, {(\i+0.5)*\scale}) --
                      ({(\gridsize+0.5)*\scale}, {(\i+0.5)*\scale});
    \ifthenelse{\i > 0} {
        \node[left,gray] (y-grid label \i) at ({0.*\scale},{\i*\scale}) {$\i$};
    }{}
}
\node[\preyCol] at ({{\preyPos}[0]*\scale}, {{\preyPos}[1]*\scale}) {$\times$};
\foreach \col/\pInfo [count=\ip] in \predatorInfo {
    \foreach \x/\y/\iscomm [count=\istep, remember=\x as \xp, remember=\y as \yp] in \pInfo {
        \ifthenelse{\iscomm = 1} {
        }{}
        \ifthenelse{\istep > 9 } {
            \draw [\col, ->] ({\xp*\scale}, {\yp*\scale}) -- ({\x*\scale}, {\y*\scale});
        }{
        }
    }
}
\foreach \col/\x/\y/\hv/\len [count=\i] in \wallInfo {
    \ifthenelse{\equal{\hv}{h}} {
        \draw[\col, very thick] ({\x*\scale}, {\y*\scale}) --++ ({\len*\scale}, 0);
    }{
        \draw[\col, very thick] ({\x*\scale}, {\y*\scale}) --++ (0, {\len*\scale});
    }
}

\node [\colB, above right=.2 cm and 0.5 cm of y-grid label 9] (legend 2) {\scriptsize Ag. 1};
\node [\colC, above right=.2 cm and 1.5 cm of y-grid label 9] (legend 3) {\scriptsize Ag. 2};

\end{tikzpicture}

%% file: diagrams/traj_demo_3.tex
\begin{tikzpicture}
\def\colA{red}
\def\colB{blue}
\def\colC{green!60!black}
\def\colD{yellow}
\def\gridsize{10}
\def\scale{0.25}
\def\preyCol{black}

\def\preyPos{9, 8}
\def\predatorInfo{
{\colA/{8/5/1, 9/5/1, 9/6/0, 9/7/1, 8/7/1, 7/7/0, 7/8/1, 6/8/1, 5/8/1, 5/8/0, 4/8/0, 5/8/0, 6/8/0, 7/8/0, 8/8/1}},
{\colB/{6/10/0, 7/10/0, 8/10/0, 9/10/0, 10/10/0, 10/9/0, 10/8/1, 9/8/1, 9/8/1, 9/8/1, 9/8/1, 9/8/1, 9/8/1, 9/8/1, 9/8/1}},
{\colC/{7/6/1, 8/6/1, 8/7/0, 8/8/1, 7/8/1, 6/8/1, 5/8/1, 4/8/0, 3/8/1, 4/8/0, 5/8/0, 6/8/0, 7/8/0, 8/8/0, 9/8/0}},
{\colD/{1/10/1, 1/10/1, 2/10/1, 3/10/1, 4/10/0, 5/10/0, 6/10/0, 7/10/0, 8/10/0, 9/10/0, 10/10/0, 10/9/1, 10/8/1, 9/8/1, 9/8/0}}}
\def\wallInfo{
{gray/1/9/h/8}}

\foreach \i in {0,...,\gridsize} {
    \draw[thin,gray] ({(\i+0.5)*\scale}, {0.5*\scale}) --
                      ({(\i+0.5)*\scale}, {(\gridsize+0.5)*\scale});
    \ifthenelse{\i > 0}{
        \node[below,gray] at ({(\i+0.)*\scale}, {0.*\scale}) {$\i$};
    }{}
}
\foreach \i in {0,...,\gridsize} {
    \draw [thin,gray] ({0.5*\scale}, {(\i+0.5)*\scale}) --
                      ({(\gridsize+0.5)*\scale}, {(\i+0.5)*\scale});
    \ifthenelse{\i > 0} {
        \node[left,gray] (y-grid label \i) at ({0.*\scale},{\i*\scale}) {$\i$};
    }{}
}
\node[\preyCol] at ({{\preyPos}[0]*\scale}, {{\preyPos}[1]*\scale}) {$\times$};
\foreach \col/\pInfo [count=\ip] in \predatorInfo {
    \foreach \x/\y/\iscomm [count=\istep, remember=\x as \xp, remember=\y as \yp] in \pInfo {
        \ifthenelse{\istep=9 \and \iscomm>0} {
            \draw[\col, fill=\col] ({\x*\scale}, {\y*\scale}) circle [radius=0.15*\scale];
        }{}
        \ifthenelse{\istep=9 \and \iscomm=0} {
            \draw[\col] ({\x*\scale}, {\y*\scale}) circle [radius=0.15*\scale];
        }{
        }
    }
}
\foreach \col/\x/\y/\hv/\len [count=\i] in \wallInfo {
    \ifthenelse{\equal{\hv}{h}} {
        \draw[\col, very thick] ({\x*\scale}, {\y*\scale}) --++ ({\len*\scale}, 0);
    }{
        \draw[\col, very thick] ({\x*\scale}, {\y*\scale}) --++ (0, {\len*\scale});
    }
}

\node [\colB, above right=1.8 cm and -0.4 cm of y-grid label 2] (legend 1) {\scriptsize Ag. 1};
\node [\colC, above right=1.8 cm and 0.4 cm of y-grid label 2] (legend 2) {\scriptsize Ag. 2};
\node [\colA, above right=1.8 cm and 1.2 cm of y-grid label 2] (legend 3) {\scriptsize Ag. 3};
\node [\colD, above right=1.8 cm and 2.0 cm of y-grid label 2] (legend 4) {\scriptsize Ag. 4};

\end{tikzpicture}

%% file: diagrams/traj_demo_4.tex
\begin{tikzpicture}
\def\colA{red}
\def\colB{blue}
\def\colC{green!60!black}
\def\colD{yellow}
\def\gridsize{10}
\def\scale{0.25}
\def\preyCol{black}

\def\preyPos{9, 8}
\def\predatorInfo{
{\colA/{8/5/1, 9/5/1, 9/6/0, 9/7/1, 8/7/1, 7/7/0, 7/8/1, 6/8/1, 5/8/1, 5/8/0, 4/8/0, 5/8/0, 6/8/0, 7/8/0, 8/8/1}},
{\colB/{6/10/0, 7/10/0, 8/10/0, 9/10/0, 10/10/0, 10/9/0, 10/8/1, 9/8/1, 9/8/1, 9/8/1, 9/8/1, 9/8/1, 9/8/1, 9/8/1, 9/8/1}},
{\colC/{7/6/1, 8/6/1, 8/7/0, 8/8/1, 7/8/1, 6/8/1, 5/8/1, 4/8/0, 3/8/1, 4/8/0, 5/8/0, 6/8/0, 7/8/0, 8/8/0, 9/8/0}},
{\colD/{1/10/1, 1/10/1, 2/10/1, 3/10/1, 4/10/0, 5/10/0, 6/10/0, 7/10/0, 8/10/0, 9/10/0, 10/10/0, 10/9/1, 10/8/1, 9/8/1, 9/8/0}}}
\def\wallInfo{
{gray/1/9/h/8}}

\foreach \i in {0,...,\gridsize} {
    \draw[thin,gray] ({(\i+0.5)*\scale}, {0.5*\scale}) --
                      ({(\i+0.5)*\scale}, {(\gridsize+0.5)*\scale});
    \ifthenelse{\i > 0}{
        \node[below,gray] at ({(\i+0.)*\scale}, {0.*\scale}) {$\i$};
    }{}
}
\foreach \i in {0,...,\gridsize} {
    \draw [thin,gray] ({0.5*\scale}, {(\i+0.5)*\scale}) --
                      ({(\gridsize+0.5)*\scale}, {(\i+0.5)*\scale});
    \ifthenelse{\i > 0} {
        \node[left,gray] (y-grid label \i) at ({0.*\scale},{\i*\scale}) {$\i$};
    }{}
}
\node[\preyCol] at ({{\preyPos}[0]*\scale}, {{\preyPos}[1]*\scale}) {$\times$};
\foreach \col/\pInfo [count=\ip] in \predatorInfo {
    \foreach \x/\y/\iscomm [count=\istep, remember=\x as \xp, remember=\y as \yp] in \pInfo {
        \ifthenelse{\istep=10 \and \iscomm>0} {
            \draw[\col, fill=\col] ({\x*\scale}, {\y*\scale}) circle [radius=0.15*\scale];
        }{}
        \ifthenelse{\istep=10 \and \iscomm=0} {
            \draw[\col] ({\x*\scale}, {\y*\scale}) circle [radius=0.15*\scale];
        }{
        }
    }
}
\foreach \col/\x/\y/\hv/\len [count=\i] in \wallInfo {
    \ifthenelse{\equal{\hv}{h}} {
        \draw[\col, very thick] ({\x*\scale}, {\y*\scale}) --++ ({\len*\scale}, 0);
    }{
        \draw[\col, very thick] ({\x*\scale}, {\y*\scale}) --++ (0, {\len*\scale});
    }
}

\node [\colB, above right=1.8 cm and -0.4 cm of y-grid label 2] (legend 1) {\scriptsize Ag. 1};
\node [\colC, above right=1.8 cm and 0.4 cm of y-grid label 2] (legend 2) {\scriptsize Ag. 2};
\node [\colA, above right=1.8 cm and 1.2 cm of y-grid label 2] (legend 3) {\scriptsize Ag. 3};
\node [\colD, above right=1.8 cm and 2.0 cm of y-grid label 2] (legend 4) {\scriptsize Ag. 4};

\end{tikzpicture}

%% file: tables/positive_listening_reward.tex
\begin{tabular}{lccccc}
    \toprule
     & $r_\text{{no comm.}}$ & $r_\text{{\coolname}}$ & $\bar{r}\uparrow$ & Cohen's $d$ & $r_\text{{ref}}$ \\
    \midrule
    \ppw{10} & -2.55 \std{0.27} & -1.11\std{0.29} & 1.44$^{**}$ & 6.40 & -0.1 \\
    \midrule
    \lj{10}{5}{3} & -9.45\std{0.73} & -8.45\std{0.79} & 1.00$^{*}$ & 1.63 &-0.1\\
    \bottomrule
    \multicolumn{6}{l}{\scriptsize Welch's $t$-test ($n{=}5$ seeds): $^{*}\!p{<}0.05$, $^{**}\!p{<}0.01$. $\pm$: std.}
\end{tabular}

%% file: plots/ablation_when_2026.tex
\definecolor{laurelRed}{HTML}{D62828}
\definecolor{alwaysBlue}{HTML}{0077B6}
\definecolor{neverGray}{HTML}{6C757D}

\pgfplotsset{
    policy curve/.style={color=laurelRed, solid, line width=1.5pt},
    always curve/.style={color=alwaysBlue, dashed, line width=1.2pt},
    never curve/.style={color=neverGray, densely dotted, line width=1.2pt},
}

\newcommand{\addadaptivecurve}[7]{%
    \addplot[name path=upper, draw=none, forget plot] table[x index=#2, y expr=\thisrowno{#3}+\thisrowno{#4}] {#1};
    \addplot[name path=lower, draw=none, forget plot] table[x index=#2, y expr=\thisrowno{#3}-\thisrowno{#4}] {#1};
    \addplot[#6!20, opacity=0.4, forget plot] fill between[of=upper and lower];
    \addplot[#5] table[x index=#2, y index=#3] {#1};
    \addlegendentry{#7}
}

\newcommand{\addadaptivecurvenolegend}[6]{%
    \addplot[name path=upper, draw=none, forget plot] table[x index=#2, y expr=\thisrowno{#3}+\thisrowno{#4}] {#1};
    \addplot[name path=lower, draw=none, forget plot] table[x index=#2, y expr=\thisrowno{#3}-\thisrowno{#4}] {#1};
    \addplot[#6!20, opacity=0.4, forget plot] fill between[of=upper and lower];
    \addplot[#5] table[x index=#2, y index=#3] {#1};
}

\pgfplotstableread[col sep=comma]{
ep,steps_always_avg,steps_always_var,comm_always_avg,comm_always_var,steps_never_avg,steps_never_var,comm_never_avg,comm_never_var,steps_policy_avg,steps_policy_var,comm_policy_avg,comm_policy_var
0.0,40.0,0.0,1.0,0.0,40.0,0.0,0.0,0.0,40.0,0.0,0.83026,0.3393800088396487
40000.0,39.923825,0.13193897026655788,1.0,0.0,39.72136666666666,0.22966947748642808,0.0,0.0,40.0,0.0,0.40408,0.11682471313895874
80000.0,39.6289,0.3644928051416103,1.0,0.0,39.95053333333333,0.06995643088539019,0.0,0.0,39.676579999999994,0.175523416101669,0.48768,0.16649699576869248
120000.0,39.552749999999996,0.5422518257230686,1.0,0.0,39.57293333333333,0.46291218989734606,0.0,0.0,39.40468,0.48468282164731236,0.60634,0.016512492240724888
160000.0,39.005874999999996,0.6447425081961006,1.0,0.0,38.98696666666667,0.4200473015691875,0.0,0.0,39.39378,0.7518042415416412,0.52684,0.13187482853069424
200000.0,39.08985,0.7152378013639925,1.0,0.0,38.375033333333334,1.1324150309061696,0.0,0.0,37.86406,1.2450487662738359,0.60476,0.13920061206761988
240000.0,38.209,0.8356188245845112,1.0,0.0,38.080733333333335,0.21206634706043426,0.0,0.0,38.51404,0.7030079959715952,0.5876,0.15300709787457575
280000.0,37.794925,0.5731010442103563,1.0,0.0,39.0026,0.4272813826976308,0.0,0.0,38.38438,1.01504800162357,0.51512,0.15857225986912085
320000.0,37.480475,1.5833983601339885,1.0,0.0,38.27863333333334,0.447281017805236,0.0,0.0,36.493739999999995,1.728187180371386,0.5253599999999999,0.16815592288111653
360000.0,37.322275,1.5442587985422005,1.0,0.0,38.08076666666667,0.31180478223116176,0.0,0.0,36.26094,1.493811539117301,0.5720000000000001,0.14895591294070876
400000.0,36.12695000000001,2.001828176317836,1.0,0.0,38.140633333333334,0.21001302076035383,0.0,0.0,35.62656,1.4824707816344995,0.55018,0.07913258241710555
440000.0,36.25585,1.7603226472723692,1.0,0.0,37.2578,0.47664251453963435,0.0,0.0,34.981260000000006,0.6348252912416156,0.5608199999999999,0.10885203535074571
480000.0,35.513675,1.4098407044325967,1.0,0.0,37.5807,0.33903121783497486,0.0,0.0,35.04376,1.2182393420013993,0.55696,0.12484980736869401
520000.0,35.695325,3.0543664395214605,1.0,0.0,37.916666666666664,0.7608654078671786,0.0,0.0,35.4625,1.2633156248539008,0.58714,0.09041905993760385
560000.0,34.537125,2.3560783532970637,1.0,0.0,36.91143333333333,0.3963811998008427,0.0,0.0,33.818740000000005,1.3857626399928675,0.62532,0.04051317810293336
600000.0,33.982425,3.1200279793416916,1.0,0.0,36.802099999999996,0.22100381595498977,0.0,0.0,33.6172,0.5336223870865976,0.55454,0.1149687018279323
640000.0,34.541,3.567308786045862,1.0,0.0,36.830733333333335,0.9653237464993579,0.0,0.0,34.28438,1.6130342406781069,0.5748599999999999,0.166910737821148
680000.0,34.46095,2.958975454866091,1.0,0.0,37.393233333333335,0.666623713616276,0.0,0.0,32.63438,1.2771804060507665,0.6257999999999999,0.10338458298992166
720000.0,34.072275,3.499279331787476,1.0,0.0,37.080733333333335,0.4316997207421959,0.0,0.0,31.584339999999997,2.1728534774346855,0.53026,0.14774887613785764
760000.0,33.04295,3.0217285587060916,1.0,0.0,36.60413333333333,1.1274108429888765,0.0,0.0,31.246879999999997,0.9983059839548197,0.60828,0.12339905023945685
800000.0,31.703125000000004,3.5451892815299733,1.0,0.0,36.60156666666666,1.1883622969822354,0.0,0.0,31.99842,0.5914790322572734,0.59202,0.10749994232556594
840000.0,31.52345,3.08958234920191,1.0,0.0,36.0599,1.775417455135551,0.0,0.0,29.66406,1.296725021891688,0.58002,0.13804583876379614
880000.0,31.248075,2.4573858461533873,1.0,0.0,36.2474,0.9066790538369507,0.0,0.0,31.07654,0.91202837148852,0.6104,0.13645954711928368
920000.0,29.773425000000003,3.0877125986521166,1.0,0.0,35.1901,0.7672080856369173,0.0,0.0,30.35,1.441572993642708,0.59512,0.12054077152565434
960000.0,29.425775,3.4427807273881106,1.0,0.0,35.21353333333334,0.40150717165312305,0.0,0.0,28.13906,1.248387456841825,0.6060000000000001,0.14068914670293514
1000000.0,28.24415,4.533859602204284,1.0,0.0,35.51303333333333,1.9077924316398311,0.0,0.0,29.34686,0.8813294380650176,0.6051200000000001,0.13175708557796809
1040000.0,28.341799999999996,3.5158832588696676,1.0,0.0,35.1328,1.1842630225869053,0.0,0.0,27.89064,1.7342202185420392,0.6025,0.1414705340344766
1080000.0,28.334000000000003,3.6829558244703393,1.0,0.0,34.55206666666667,0.9444554139232254,0.0,0.0,27.095319999999997,0.9668569085443821,0.6044999999999999,0.1158973856478221
1120000.0,27.9043,3.656403553356767,1.0,0.0,34.645833333333336,0.8034853009787359,0.0,0.0,26.021859999999997,1.0134242657446089,0.6062399999999999,0.15076268238526402
1160000.0,26.2539,3.227680220374999,1.0,0.0,33.393233333333335,0.44436525766785834,0.0,0.0,26.48594,1.1315795413491703,0.59628,0.13911433283454294
1200000.0,26.740225,3.6129679941669837,1.0,0.0,33.890633333333334,0.6308626598203075,0.0,0.0,25.70624,1.160170223027638,0.63832,0.11792813743971368
1240000.0,27.164099999999998,4.845275470497007,1.0,0.0,33.60156666666666,1.1638656575777493,0.0,0.0,24.80624,1.1980525507672866,0.5997199999999999,0.11871246606822722
1280000.0,25.951175,4.167155158123465,1.0,0.0,33.1745,0.305431268864209,0.0,0.0,25.0625,1.8954536480747826,0.62296,0.11250442835728731
1320000.0,25.68945,4.7001136467643,1.0,0.0,32.95053333333333,1.4769764821719036,0.0,0.0,24.234379999999998,1.0161886132013087,0.63964,0.1050384805678376
1360000.0,25.755875000000003,4.215259705744712,1.0,0.0,32.9427,0.22212217358921874,0.0,0.0,24.3625,0.6291870691614699,0.6359,0.12044253401518917
1400000.0,24.650375,4.4780468450960855,1.0,0.0,32.015633333333334,0.780979412162845,0.0,0.0,23.118740000000003,0.7444272337844716,0.63294,0.11174597263436384
1440000.0,24.2637,3.594743277342626,1.0,0.0,31.807333333333332,1.8031426017434724,0.0,0.0,22.91874,1.2033304277711918,0.6281399999999999,0.10298669040220682
1480000.0,24.265625,4.359374350394217,1.0,0.0,31.666666666666668,1.2804392848636836,0.0,0.0,22.96876,1.0235134422175407,0.6377599999999999,0.11726942653564908
1520000.0,23.1543,3.746503628051092,1.0,0.0,31.877633333333335,0.11417417493558639,0.0,0.0,22.85624,0.2658325232171563,0.64828,0.12389116836966226
1560000.0,23.466825,3.479229318667426,1.0,0.0,31.86196666666667,1.8263516388934777,0.0,0.0,21.895319999999998,0.9060112877884021,0.64324,0.10221214409256858
1600000.0,23.9824,3.8791811971084824,1.0,0.0,29.763033333333336,0.38083501531357244,0.0,0.0,21.775,0.6363373287808913,0.64796,0.12567546459034876
1640000.0,23.1504,2.308075017195065,1.0,0.0,30.362,1.3835580893719877,0.0,0.0,21.83592,1.0956938977652477,0.6042799999999999,0.10163879967807572
1680000.0,22.24415,2.2857120919967158,1.0,0.0,29.4375,1.4270884508910666,0.0,0.0,22.13126,1.0530865502891957,0.6311,0.1151733476113289
1720000.0,22.955075,3.7306216274067516,1.0,0.0,30.3203,1.0405229870919086,0.0,0.0,21.473419999999997,0.9637699256565335,0.64118,0.08477717617377922
1760000.0,22.289025,2.7858625113014814,1.0,0.0,29.223966666666666,0.7590399213626536,0.0,0.0,21.1547,0.8189381710483402,0.6457200000000001,0.12431584613395028
1800000.0,22.125,3.1247512949033234,1.0,0.0,29.3776,0.9589163084788305,0.0,0.0,21.0125,0.763851236825601,0.65196,0.12015324548259194
1840000.0,22.214824999999998,3.306399944331447,1.0,0.0,28.843733333333333,1.3346747077679846,0.0,0.0,20.72498,0.7102854676818323,0.67884,0.11064152204303772
1880000.0,21.720699999999997,2.824769803895532,1.0,0.0,28.97136666666667,0.7855000374849692,0.0,0.0,21.329700000000003,1.161752918653532,0.6549600000000001,0.10817098686801374
1920000.0,21.625025,2.7002973469370004,1.0,0.0,28.42186666666667,1.8884391461969032,0.0,0.0,21.318760000000005,0.44221856406080523,0.63558,0.0968481987442203
1960000.0,21.742199999999997,2.0110233203521037,1.0,0.0,28.17446666666667,1.9058636280932826,0.0,0.0,20.635939999999998,0.42074230402943824,0.6481,0.0982220341878542
2000000.0,21.2969,2.8461090061696517,1.0,0.0,27.65886666666667,2.4486036270132048,0.0,0.0,20.9625,0.9410568569432988,0.6540000000000001,0.0970197917952827
2040000.0,20.578125,2.4540059029829164,1.0,0.0,27.992199999999997,1.184184929251622,0.0,0.0,21.13906,0.5512060580218627,0.6461,0.09159181186110471
2080000.0,20.302725,1.8355772870884526,1.0,0.0,27.880200000000002,2.3022206033885344,0.0,0.0,20.72656,0.7885135346967731,0.66686,0.09033410430175308
2120000.0,20.392575,1.736777905166633,1.0,0.0,26.718733333333333,2.612932193702869,0.0,0.0,20.7172,0.3743053993732928,0.6382199999999999,0.12368298832094898
2160000.0,20.623025,1.861096781974275,1.0,0.0,26.562466666666666,1.9749742383692561,0.0,0.0,20.00626,0.7571136548762014,0.664,0.11331739495770277
2200000.0,20.527350000000002,0.9359868388497782,1.0,0.0,26.406266666666667,2.192964424902714,0.0,0.0,19.73594,0.27197353253579715,0.6607800000000001,0.085031697619182
2240000.0,20.61915,1.4549222358944132,1.0,0.0,25.2474,0.9441391881850194,0.0,0.0,20.84686,1.8627629785885262,0.65176,0.09869114651274448
2280000.0,20.304675,2.1405652937658792,1.0,0.0,26.7422,2.1240175579939704,0.0,0.0,20.521859999999997,1.134369986556414,0.6300399999999999,0.113639334739341
2320000.0,20.41015,1.3111909862792674,1.0,0.0,25.059933333333333,1.068735755720541,0.0,0.0,20.16718,0.41053800506165067,0.6548,0.10213569405452726
2360000.0,20.4707,1.3268707981563241,1.0,0.0,26.309866666666665,1.9325469452397666,0.0,0.0,20.0172,0.4727464648202033,0.63714,0.09751571360555178
2400000.0,20.351575,1.7652797092458177,1.0,0.0,25.2396,1.8386919970457267,0.0,0.0,20.071859999999997,0.7283598660003168,0.62586,0.08554085807378838
2440000.0,20.011725,1.612465229663884,1.0,0.0,25.6354,1.504782624390203,0.0,0.0,20.15626,0.5493564074442022,0.63462,0.08798105250563894
2480000.0,20.00195,1.4107071143579033,1.0,0.0,25.773433333333333,2.75068768654111,0.0,0.0,19.851580000000002,0.4104066685618059,0.63706,0.0889039616665084
2520000.0,20.527325,1.04634481738813,1.0,0.0,24.927066666666665,1.6208466354209938,0.0,0.0,19.75624,0.7341053292273529,0.6364,0.10936260786941757
2560000.0,19.52735,0.7955166890141272,1.0,0.0,25.4245,0.917242054567204,0.0,0.0,20.06406,1.1362443656185934,0.6346,0.059411413044969705
2600000.0,19.64065,1.01673184886675,1.0,0.0,23.916666666666668,1.279853745116561,0.0,0.0,20.0344,0.803107830867064,0.61816,0.06323421858456069
2640000.0,19.556625,0.9802247736488812,1.0,0.0,23.60676666666667,1.0590335919548952,0.0,0.0,19.665640000000003,0.6832510127325097,0.6279399999999999,0.07150277197423888
2680000.0,19.3125,1.0727959102271045,1.0,0.0,24.552066666666665,1.985416114795306,0.0,0.0,19.80312,0.7093772745161775,0.62428,0.055857439969980716
2720000.0,19.882825,1.1632867173981658,1.0,0.0,24.632800000000003,0.7631479061536279,0.0,0.0,20.13124,0.6540832748205692,0.6615400000000001,0.08586208942251519
2760000.0,20.015625,0.9905697688073263,1.0,0.0,23.34113333333333,0.47324032184739045,0.0,0.0,19.784380000000002,0.6505676671953498,0.6213200000000001,0.08871395380660246
2800000.0,19.75,1.536078754491448,1.0,0.0,24.450533333333336,1.8549857741293387,0.0,0.0,20.1578,0.6119636688562491,0.6618999999999999,0.06948677572027646
2840000.0,20.082025,1.3594888513242769,1.0,0.0,23.846333333333334,1.0918472736096791,0.0,0.0,19.81094,0.8068658069344612,0.62866,0.09257398338626245
2880000.0,19.009775,0.845909783531909,1.0,0.0,24.286466666666666,1.707356116599255,0.0,0.0,19.39844,0.9357448575332922,0.6514800000000001,0.06450427582726588
2920000.0,19.730475000000002,0.8169975133836086,1.0,0.0,23.726566666666667,1.5916394322277343,0.0,0.0,19.6172,0.8196571551569594,0.64144,0.058428369821517355
2960000.0,19.40625,0.7616349798295765,1.0,0.0,23.1016,1.9182511566528508,0.0,0.0,19.86564,0.9729456564474706,0.6301599999999999,0.07711216765206387
3000000.0,20.228525,0.9934323061361542,1.0,0.0,23.82813333333333,1.5186423790858592,0.0,0.0,19.3953,1.0685760094630614,0.63498,0.06691610867347263
3040000.0,19.1836,0.9276884363836815,1.0,0.0,23.763033333333336,1.7321259506424156,0.0,0.0,19.56252,0.39518997659353655,0.6444799999999999,0.06875089526689818
3080000.0,18.99805,0.8589027578835692,1.0,0.0,22.895833333333332,1.4761081720373879,0.0,0.0,19.140620000000002,0.404559174410864,0.6386000000000001,0.06718258702967608
3120000.0,19.36135,0.7498258147730039,1.0,0.0,23.458333333333332,1.008557310661566,0.0,0.0,19.16252,0.4961859062891649,0.63464,0.0670067638376903
3160000.0,19.4375,0.7982543830133337,1.0,0.0,22.822933333333335,1.2186390177388695,0.0,0.0,19.19842,0.5446334708774327,0.6365999999999999,0.035215223980545676
3200000.0,18.814475,1.303094548325255,1.0,0.0,23.843766666666667,0.891066740236418,0.0,0.0,19.7531,0.6695935393953559,0.64258,0.08982479390457848
3240000.0,19.148425,0.7139194680599488,1.0,0.0,23.078100000000003,1.60111185742908,0.0,0.0,19.53594,0.5598677793908132,0.6147599999999999,0.05875173529352133
3280000.0,18.6973,0.6307464585711124,1.0,0.0,22.533866666666665,1.3705162295848805,0.0,0.0,19.62186,0.874277572856585,0.62958,0.06617060979014776
3320000.0,19.1289,0.5416591225115658,1.0,0.0,22.166666666666668,1.0073758693865074,0.0,0.0,19.6125,1.0664698082927624,0.64206,0.08439781039813771
3360000.0,19.066375,0.4624179352869007,1.0,0.0,22.244833333333332,0.10980243875049184,0.0,0.0,19.445320000000002,0.44431236489658854,0.62032,0.06623749391394576
3400000.0,18.912125,1.1607237276264328,1.0,0.0,22.51823333333333,1.0998921290542798,0.0,0.0,18.996879999999997,0.8250147208383611,0.64578,0.062497276740670855
3440000.0,18.847649999999998,0.6908853975153914,1.0,0.0,22.343733333333333,1.246824300194519,0.0,0.0,19.06408,0.6686016374493852,0.67256,0.08635229238416316
3480000.0,18.269550000000002,0.6090331374399918,1.0,0.0,22.3828,1.1588802296469938,0.0,0.0,19.92344,0.4245860035375642,0.65838,0.06769831312521753
3520000.0,19.001949999999997,1.0187827503938212,1.0,0.0,22.208333333333332,0.5049103506784369,0.0,0.0,19.559379999999997,0.6058005361503072,0.62936,0.08219015999497747
3560000.0,18.3965,0.9119337832320943,1.0,0.0,22.541666666666668,1.0017346066809423,0.0,0.0,19.198439999999998,0.6943498991142711,0.64086,0.06941321488016526
3600000.0,19.427725,0.9582413562746074,1.0,0.0,22.8047,0.9407265312866795,0.0,0.0,19.08906,0.9294583317179961,0.6373599999999999,0.051682863697748
3640000.0,19.0332,0.80317399111276,1.0,0.0,22.26823333333333,1.1133143381613104,0.0,0.0,19.14686,0.756566638439735,0.63184,0.04038804773692335
3680000.0,19.06445,0.662677329097654,1.0,0.0,22.539033333333332,0.5889443682461777,0.0,0.0,19.604680000000002,0.7969448799007365,0.64028,0.05881728997497248
3720000.0,18.652324999999998,0.6000383086728709,1.0,0.0,22.611966666666664,1.502924283595891,0.0,0.0,18.93438,0.48079827953103244,0.65746,0.06915873336029225
3760000.0,18.490225000000002,0.5177302018184761,1.0,0.0,22.580766666666666,1.0913167378090662,0.0,0.0,19.065620000000003,0.5375108944012201,0.6215200000000001,0.04939159442658234
3800000.0,19.083975,1.238190555558795,1.0,0.0,22.742199999999997,1.066993836283353,0.0,0.0,19.4875,1.0339974584108036,0.63748,0.038263998745557165
3840000.0,18.554675,0.8856092320402944,1.0,0.0,21.76046666666667,0.5621140651346683,0.0,0.0,18.88438,0.5038152335926332,0.64302,0.05724269036304987
3880000.0,19.290975,0.8745753608895006,1.0,0.0,21.60936666666667,0.6590755158216363,0.0,0.0,19.004700000000003,0.9543455307172563,0.65826,0.05182407162699587
3920000.0,18.615225,0.5775553626060449,1.0,0.0,22.16926666666667,0.33356648845803205,0.0,0.0,19.76248,0.11998450566635702,0.6498999999999999,0.08104435822436994
3960000.0,18.195325,0.5059765378700875,1.0,0.0,22.341166666666666,0.683229396973585,0.0,0.0,18.97344,0.5880135529050335,0.6619999999999999,0.08458186566871177
4000000.0,18.949225,0.6228959318176677,1.0,0.0,21.385433333333335,0.8007652187473904,0.0,0.0,19.1922,0.21409000910831832,0.64378,0.08471395162545543
}\dataBwLimitSteps

\begin{tikzpicture}[every node/.style={transform shape}]

\begin{groupplot}[
    group style={
        group size=2 by 1,
        horizontal sep=1.3cm,
    },
    scale only axis,
    height=3.5cm,
    width=3.6cm,
    tick label style={font=\scriptsize},
    title style={font=\small\bfseries, yshift=1mm},
    grid=major,
    grid style={dashed, gray!30},
    every x tick scale label/.style={at={(xticklabel* cs:1.0, -12pt)}, anchor=north west, xshift=0pt},
    legend style={
        at={(1.15, 1.25)},
        anchor=south,
        legend columns=3,
        draw=black!20,
        fill=white,
        rounded corners=2pt,
        font=\scriptsize\sffamily\bfseries
    }
]

\nextgroupplot[
    title={PP (Grid 10, 3 Agents)},
    ymin=15, ymax=45, xmin=0, xmax=4000000,
    ylabel={Steps to catch prey},
    xlabel={Environment Steps},
    ylabel style={font=\scriptsize\bfseries},
    xlabel style={font=\scriptsize},
]
\addadaptivecurve{\dataBwLimitSteps}{0}{1}{2}{always curve}{alwaysBlue}{TarMAC (Always)}
\addadaptivecurve{\dataBwLimitSteps}{0}{9}{10}{policy curve}{laurelRed}{\coolname{} (Adaptive)}
\addadaptivecurve{\dataBwLimitSteps}{0}{5}{6}{never curve}{neverGray}{TarMAC-BW (Never)}

\nextgroupplot[
    title={Avg. Transmit Probability},
    ymin=-0.1, ymax=1.2, xmin=0, xmax=4000000,
    xlabel={Environment Steps},
    ylabel={Avg. Comm. Action},
    xlabel style={font=\scriptsize},
    ylabel style={font=\scriptsize\bfseries},
]
\addadaptivecurvenolegend{\dataBwLimitSteps}{0}{3}{4}{always curve}{alwaysBlue}
\addadaptivecurvenolegend{\dataBwLimitSteps}{0}{11}{12}{policy curve}{laurelRed}
\addadaptivecurvenolegend{\dataBwLimitSteps}{0}{7}{8}{never curve}{neverGray}

\draw[stealth-stealth, solid, line width=1.2pt, color=black] (axis cs: 3850000, 1.0) -- (axis cs: 3850000, 0.645) 
    node[midway, left, align=right, fill=white, fill opacity=0.9, text opacity=1, rounded corners=2pt, inner sep=2pt, font=\scriptsize\sffamily, text=black] {$\sim$36\% Bandwidth\\Savings};

\end{groupplot}

\end{tikzpicture}

%% file: tables/ablation_mlp_gin.tex
\begin{tabular}{lccc}
    \toprule
    Enc. & MLP & Avg & Eq. \ref{eq: how gnn} \\
    \midrule
    \ppw{10} & 24.03\std{5.12}$^{\dagger}$ & 21.24\std{0.98}$^{**}$ & \textbf{18.71}\std{0.58}\\
    \midrule
    \lj{10}{5}{3} & 29.00\std{3.60}$^{*}$ & 38.15\std{1.27}$^{**}$ & \textbf{22.58}\std{1.25}\\
    \bottomrule
    \multicolumn{4}{l}{\scriptsize Welch's $t$-test vs.\ proposed ($n{=}5$ seeds): $^{\dagger}\!p{<}0.1$, $^{*}\!p{<}0.05$, $^{**}\!p{<}0.01$. $\pm$: std.}
\end{tabular}

%% file: plots_new/ablation_gnnHyper_2026.tex
\definecolor{laurelRed}{HTML}{D62828}
\definecolor{ablatedGray}{HTML}{495057}

\pgfplotsset{
    hero curve/.style={color=laurelRed, solid, line width=1.5pt},
    ablation curve/.style={color=ablatedGray, densely dashed, line width=1.2pt},
}

\newcommand{\addablationcurve}[7]{%
    \addplot[name path=upper, draw=none, forget plot] table[x index=#2, y expr=\thisrowno{#3}+\thisrowno{#4}] {#1};
    \addplot[name path=lower, draw=none, forget plot] table[x index=#2, y expr=\thisrowno{#3}-\thisrowno{#4}] {#1};
    \addplot[#6!20, opacity=0.4, forget plot] fill between[of=upper and lower];
    \addplot[#5] table[x index=#2, y index=#3] {#1};
    \addlegendentry{#7}
}

\newcommand{\addablationcurvenolegend}[6]{%
    \addplot[name path=upper, draw=none, forget plot] table[x index=#2, y expr=\thisrowno{#3}+\thisrowno{#4}] {#1};
    \addplot[name path=lower, draw=none, forget plot] table[x index=#2, y expr=\thisrowno{#3}-\thisrowno{#4}] {#1};
    \addplot[#6!20, opacity=0.4, forget plot] fill between[of=upper and lower];
    \addplot[#5] table[x index=#2, y index=#3] {#1};
}

\pgfplotstableread[col sep=comma]{
ep,steps_gnnHyper_lj_avg,steps_gnnHyper_lj_var,reward_gnnHyper_lj_avg,reward_gnnHyper_lj_var,steps_Hyper_lj_avg,steps_Hyper_lj_var,reward_Hyper_lj_avg,reward_Hyper_lj_var,steps_gnnHyper_pp_avg,steps_gnnHyper_pp_var,reward_gnnHyper_pp_avg,reward_gnnHyper_pp_var,steps_Hyper_pp_avg,steps_Hyper_pp_var,reward_Hyper_pp_avg,reward_Hyper_pp_var
0.0,39.919,0.114,-19.622,0.216,39.914,0.121,-19.573,0.130,45.0,0.0,-17.275,0.079,45.0,0.0,-16.843,0.262
40000.0,39.848,0.135,-19.224,0.281,39.770,0.210,-19.344,0.289,45.0,0.0,-16.750,0.207,45.0,0.0,-16.837,0.633
80000.0,39.450,0.310,-18.855,0.347,39.757,0.342,-19.020,0.601,45.0,0.0,-17.264,0.154,44.830,0.239,-17.080,1.065
120000.0,39.341,0.208,-18.568,0.316,39.611,0.208,-18.818,0.131,44.755,0.272,-15.842,0.621,45.0,0.0,-16.108,1.050
160000.0,39.596,0.399,-18.839,0.339,39.203,0.677,-18.446,0.557,44.809,0.268,-15.440,0.675,44.713,0.204,-16.160,0.662
200000.0,38.966,0.139,-18.171,0.120,39.080,0.663,-18.325,0.679,44.502,0.703,-14.334,0.204,44.911,0.125,-16.356,0.034
240000.0,38.515,0.735,-17.797,0.537,38.338,0.816,-18.060,0.659,44.143,0.612,-14.347,0.910,45.0,0.0,-16.281,0.169
280000.0,38.544,0.786,-17.614,0.558,39.119,1.072,-18.614,1.284,44.247,0.431,-14.084,0.844,44.416,0.537,-14.783,0.684
320000.0,36.781,1.535,-16.438,1.060,39.286,0.147,-18.502,0.153,43.372,0.951,-13.569,0.258,44.734,0.148,-15.203,0.894
360000.0,37.578,1.110,-17.089,0.790,39.541,0.361,-18.874,0.616,43.492,0.309,-12.973,0.172,44.411,0.663,-15.091,1.560
400000.0,37.833,1.026,-17.290,0.800,39.291,0.398,-18.863,0.512,42.460,0.946,-12.809,0.685,43.468,1.269,-14.268,2.396
440000.0,37.815,1.128,-17.220,0.680,38.768,0.922,-18.318,0.801,42.364,0.564,-12.751,0.291,43.721,0.922,-14.449,2.073
480000.0,37.333,1.515,-16.820,1.032,38.143,0.578,-17.472,0.543,42.054,0.702,-12.368,0.006,43.945,0.537,-13.954,0.623
520000.0,37.570,0.301,-17.113,0.192,39.213,0.173,-18.242,0.220,40.395,1.021,-11.536,0.291,43.127,0.750,-13.063,0.375
560000.0,38.031,0.983,-17.219,0.766,38.934,0.168,-18.243,0.273,41.710,0.174,-11.624,0.188,44.057,0.670,-13.768,0.950
600000.0,37.773,1.575,-17.275,1.211,38.598,0.519,-17.767,0.518,41.744,0.347,-12.220,0.524,43.192,1.204,-13.295,1.169
640000.0,36.825,0.889,-16.376,0.614,38.296,0.721,-17.541,0.542,42.085,0.652,-12.380,0.182,43.888,0.813,-13.915,1.175
680000.0,36.414,0.640,-16.182,0.528,37.627,1.160,-17.167,1.001,38.200,0.825,-10.136,0.413,44.205,0.826,-13.968,0.931
720000.0,35.328,0.906,-15.390,0.579,37.510,0.740,-16.835,0.521,39.171,0.709,-10.814,0.062,44.010,0.420,-13.352,0.480
760000.0,35.932,0.934,-15.889,0.691,36.338,1.813,-16.167,1.154,38.973,1.365,-11.009,0.629,42.940,1.564,-12.946,1.936
800000.0,36.255,1.987,-16.182,1.359,37.320,1.169,-16.748,0.921,38.257,1.430,-9.934,0.725,41.947,2.000,-12.087,1.230
840000.0,35.723,1.815,-15.735,1.138,36.010,0.657,-15.936,0.484,36.877,0.594,-9.231,0.241,42.604,1.694,-12.438,1.609
880000.0,35.846,1.091,-15.863,0.782,36.424,1.358,-16.339,1.030,37.528,0.166,-9.655,0.478,42.244,1.326,-12.200,1.180
920000.0,35.994,0.938,-15.853,0.650,36.112,1.307,-15.924,0.901,35.802,0.324,-8.750,0.856,41.067,1.505,-11.665,1.107
960000.0,34.976,1.546,-15.271,0.963,35.731,1.432,-15.594,0.802,34.763,1.964,-8.663,0.867,41.059,1.122,-11.190,0.737
1000000.0,34.822,0.740,-15.111,0.416,35.442,0.886,-15.586,0.466,37.309,0.871,-9.240,0.352,40.976,1.115,-11.342,0.823
1040000.0,34.075,2.275,-14.737,1.531,34.598,1.415,-14.932,0.914,34.822,0.859,-8.122,0.422,40.697,1.826,-11.328,1.080
1080000.0,34.013,1.203,-14.639,0.826,34.859,1.823,-15.217,1.224,33.427,0.661,-7.665,0.426,39.911,1.572,-10.780,0.823
1120000.0,33.911,0.460,-14.527,0.483,35.841,1.127,-15.758,0.800,30.921,0.957,-6.065,0.403,40.010,2.020,-10.956,0.938
1160000.0,33.440,1.104,-14.309,0.728,34.598,1.131,-15.016,0.745,32.320,0.395,-6.848,0.141,40.828,1.789,-10.980,1.167
1200000.0,32.875,2.538,-13.869,1.575,34.093,2.028,-14.714,1.303,30.065,2.224,-5.694,1.305,40.252,2.333,-10.983,0.996
1240000.0,34.184,2.268,-14.725,1.477,33.570,1.945,-14.361,1.195,30.226,0.665,-5.466,0.406,41.145,2.576,-11.644,1.155
1280000.0,33.218,1.910,-14.140,1.346,33.554,1.215,-14.281,0.801,29.388,0.676,-4.991,0.343,40.494,2.142,-11.176,0.971
1320000.0,31.734,3.299,-13.288,2.090,31.218,1.802,-12.870,1.254,28.143,1.063,-4.792,0.630,39.414,2.240,-10.221,1.238
1360000.0,32.346,2.234,-13.414,1.657,35.447,2.925,-15.484,2.101,27.309,1.214,-4.355,0.676,39.807,1.095,-10.882,0.715
1400000.0,31.276,2.328,-12.805,1.627,32.846,1.315,-13.685,0.923,26.705,0.552,-4.104,0.305,38.901,1.900,-10.358,1.370
1440000.0,30.903,1.809,-12.453,1.391,31.695,1.306,-13.172,0.852,25.434,1.103,-3.405,0.550,38.802,3.148,-10.470,1.791
1480000.0,32.008,3.944,-13.374,2.815,32.244,1.753,-13.509,1.240,25.638,0.984,-3.506,0.306,39.278,1.575,-10.233,0.601
1520000.0,31.669,2.803,-13.137,1.821,31.835,2.137,-13.301,1.305,25.231,0.199,-3.243,0.190,38.601,3.812,-10.294,1.817
1560000.0,30.296,3.531,-12.266,2.286,31.645,1.807,-13.017,0.908,25.239,1.213,-3.372,0.214,38.572,3.999,-10.664,1.911
1600000.0,30.075,2.978,-12.047,1.983,30.658,1.457,-12.441,0.967,24.307,0.373,-2.778,0.248,37.947,3.703,-9.737,1.648
1640000.0,30.335,2.730,-12.229,1.806,32.237,1.235,-13.407,0.717,24.054,0.879,-2.957,0.496,37.359,4.615,-9.446,1.660
1680000.0,29.815,2.361,-11.876,1.582,31.265,1.167,-12.975,0.742,23.778,0.236,-2.819,0.044,37.632,3.993,-9.751,1.884
1720000.0,29.250,2.186,-11.600,1.334,32.481,2.004,-13.463,1.180,23.505,0.897,-2.721,0.459,37.005,4.296,-9.555,1.746
1760000.0,29.356,2.417,-11.579,1.516,32.179,2.201,-13.400,1.415,23.289,0.180,-2.486,0.088,37.098,4.214,-9.028,1.761
1800000.0,29.768,1.116,-11.832,0.753,32.143,2.316,-13.476,1.633,23.210,0.797,-2.446,0.204,35.377,4.465,-8.727,2.259
1840000.0,28.997,2.900,-11.455,1.775,33.075,2.261,-13.954,1.455,22.934,0.701,-2.364,0.176,35.414,4.686,-8.509,2.289
1880000.0,28.947,2.288,-11.324,1.381,31.544,0.933,-13.166,0.622,22.393,0.776,-2.154,0.366,35.263,3.998,-8.198,1.893
1920000.0,29.000,1.984,-11.416,1.220,31.895,1.020,-13.105,0.814,21.432,0.676,-2.031,0.194,34.028,4.066,-7.701,1.514
1960000.0,29.203,1.509,-11.548,0.879,31.398,1.928,-12.839,1.160,23.109,0.502,-2.541,0.142,33.604,4.542,-7.771,2.332
2000000.0,28.567,2.048,-11.208,1.177,30.468,1.270,-12.349,0.700,21.546,0.266,-1.910,0.087,34.466,2.590,-7.751,1.264
2040000.0,27.471,2.214,-10.666,1.244,30.927,1.117,-12.663,0.692,22.312,0.998,-2.208,0.172,32.617,2.912,-7.327,1.530
2080000.0,28.578,1.895,-11.181,0.951,29.854,1.301,-11.982,0.732,21.757,0.584,-2.042,0.188,32.539,2.867,-6.771,1.498
2120000.0,26.648,2.855,-10.034,1.570,31.018,1.092,-12.637,0.689,21.393,0.621,-1.770,0.260,32.250,3.274,-7.035,1.371
2160000.0,28.627,2.101,-11.088,1.134,30.080,1.212,-12.053,0.606,21.541,1.028,-1.956,0.220,31.921,3.147,-6.826,1.692
2200000.0,27.205,2.379,-10.339,1.300,29.651,0.867,-11.867,0.548,21.218,0.558,-1.855,0.154,31.528,1.942,-6.134,1.031
2240000.0,27.130,1.676,-10.323,0.863,30.281,1.245,-12.064,0.769,21.346,0.904,-1.699,0.224,31.861,1.770,-6.449,1.061
2280000.0,28.567,1.456,-11.212,0.599,29.377,1.719,-11.585,1.075,20.955,0.336,-1.545,0.086,28.838,2.671,-5.005,1.285
2320000.0,27.934,2.831,-10.860,1.533,29.455,1.686,-11.545,1.025,21.289,0.106,-1.706,0.183,30.526,3.345,-5.842,1.637
2360000.0,27.812,1.785,-10.673,0.960,29.497,2.189,-11.798,1.243,20.747,0.598,-1.570,0.248,29.257,4.207,-4.988,1.572
2400000.0,26.132,1.802,-9.855,0.945,30.382,1.639,-12.113,0.868,20.723,0.884,-1.515,0.302,29.731,3.158,-5.171,1.582
2440000.0,26.414,2.584,-9.936,1.323,28.343,1.616,-11.101,1.008,21.239,0.694,-1.664,0.334,29.375,3.827,-5.182,1.715
2480000.0,27.179,1.481,-10.300,0.796,29.406,1.463,-11.469,0.717,21.385,0.036,-1.812,0.118,27.585,3.361,-4.316,1.829
2520000.0,27.330,2.530,-10.483,1.330,29.244,2.130,-11.477,1.167,21.231,0.799,-1.724,0.214,27.195,1.923,-4.209,1.182
2560000.0,26.619,3.111,-10.011,1.626,29.335,1.150,-11.558,0.725,20.968,0.393,-1.740,0.078,27.864,3.916,-4.494,1.763
2600000.0,26.458,1.246,-9.912,0.500,28.403,1.094,-10.982,0.614,20.846,0.916,-1.619,0.300,27.033,2.207,-3.960,1.182
2640000.0,26.854,2.033,-10.192,1.054,28.830,1.043,-11.317,0.601,20.520,0.506,-1.495,0.098,27.778,2.066,-4.379,0.969
2680000.0,26.395,1.741,-9.891,0.896,28.570,1.524,-11.199,0.983,20.364,0.939,-1.545,0.242,27.226,2.807,-4.308,1.404
2720000.0,26.093,1.637,-9.757,0.769,28.807,1.516,-11.326,0.728,20.263,0.816,-1.318,0.211,26.526,3.229,-4.130,1.515
2760000.0,26.169,2.635,-9.857,1.260,28.695,2.120,-11.411,1.311,20.843,0.586,-1.566,0.252,25.474,2.168,-3.600,1.093
2800000.0,25.721,1.472,-9.564,0.638,27.893,1.902,-10.762,1.087,20.338,0.708,-1.347,0.212,26.000,2.192,-3.658,1.271
2840000.0,26.166,1.687,-9.801,0.815,28.010,1.004,-10.918,0.676,20.234,0.477,-1.365,0.053,25.302,3.352,-3.354,1.428
2880000.0,25.763,2.594,-9.494,1.244,27.721,0.354,-10.807,0.181,20.408,0.195,-1.464,0.078,24.817,1.840,-3.140,0.941
2920000.0,25.953,2.495,-9.700,1.311,27.447,1.784,-10.626,0.999,19.687,0.641,-1.210,0.137,24.356,2.975,-3.032,1.269
2960000.0,25.567,1.712,-9.502,0.812,27.479,0.324,-10.677,0.217,20.085,0.303,-1.411,0.069,24.963,2.632,-3.309,1.290
3000000.0,25.309,1.450,-9.343,0.663,27.901,0.709,-10.987,0.370,20.122,1.071,-1.367,0.336,24.231,1.162,-2.947,0.501
3040000.0,25.942,2.276,-9.654,1.153,27.768,1.334,-10.735,0.806,20.252,0.578,-1.358,0.167,23.565,1.228,-2.578,0.697
3080000.0,24.807,0.380,-9.078,0.141,27.049,0.710,-10.365,0.440,20.414,1.022,-1.436,0.304,24.255,1.950,-2.855,0.868
3120000.0,24.497,1.195,-8.911,0.550,27.487,0.384,-10.577,0.315,19.830,0.688,-1.188,0.158,23.242,2.224,-2.452,0.938
3160000.0,25.500,1.756,-9.494,0.940,27.596,0.664,-10.733,0.409,19.835,0.855,-1.232,0.277,22.497,0.880,-2.352,0.554
3200000.0,25.750,2.164,-9.644,1.178,27.330,1.158,-10.570,0.612,19.934,0.711,-1.446,0.381,22.713,1.336,-2.363,0.658
3240000.0,25.528,1.893,-9.455,0.994,27.825,0.913,-10.758,0.486,19.859,0.475,-1.380,0.151,21.984,0.454,-2.154,0.339
3280000.0,25.067,1.850,-9.262,1.030,27.789,0.348,-10.815,0.235,20.221,0.464,-1.370,0.179,21.992,0.906,-2.110,0.525
3320000.0,24.328,2.027,-8.887,0.909,26.682,0.223,-10.053,0.188,19.911,0.785,-1.225,0.207,21.359,0.853,-2.049,0.324
3360000.0,25.117,1.906,-9.181,0.903,27.140,0.618,-10.475,0.376,19.817,0.462,-1.197,0.170,21.500,0.742,-1.883,0.362
3400000.0,24.333,1.072,-8.877,0.513,27.002,0.394,-10.406,0.205,19.481,0.462,-1.117,0.167,21.989,0.292,-2.116,0.208
3440000.0,25.171,2.546,-9.363,1.306,26.721,0.193,-10.165,0.267,19.723,0.270,-1.315,0.123,21.901,0.603,-1.985,0.248
3480000.0,24.809,2.129,-9.138,1.012,25.721,1.442,-9.602,0.775,19.604,0.782,-1.328,0.134,21.781,1.010,-2.023,0.367
3520000.0,24.666,2.124,-9.171,1.108,25.875,0.804,-9.818,0.346,19.838,0.670,-1.302,0.099,21.583,0.426,-1.879,0.343
3560000.0,25.205,1.930,-9.405,0.965,27.502,0.467,-10.724,0.166,20.460,0.410,-1.445,0.249,21.898,0.411,-1.963,0.202
3600000.0,24.617,2.624,-9.064,1.414,27.596,1.232,-10.747,0.696,19.617,0.552,-1.292,0.161,22.393,0.754,-2.102,0.303
3640000.0,24.270,1.579,-8.847,0.752,25.846,1.176,-9.705,0.509,19.497,0.392,-1.278,0.167,21.328,0.463,-1.836,0.122
3680000.0,23.914,0.413,-8.763,0.155,27.445,0.858,-10.536,0.218,19.382,0.254,-1.216,0.184,21.265,0.527,-1.807,0.153
3720000.0,23.343,1.808,-8.371,0.878,26.942,0.472,-10.339,0.258,19.752,0.579,-1.158,0.126,21.682,1.107,-1.900,0.307
3760000.0,24.640,2.110,-9.137,1.182,26.908,0.662,-10.385,0.418,20.242,0.447,-1.503,0.313,21.127,0.293,-1.728,0.124
3800000.0,23.669,1.198,-8.582,0.723,26.953,0.748,-10.326,0.484,19.484,0.453,-1.124,0.096,21.682,0.963,-1.906,0.374
3840000.0,23.325,1.040,-8.388,0.523,27.153,0.830,-10.507,0.392,19.705,0.229,-1.327,0.094,20.958,0.557,-1.612,0.178
3880000.0,24.184,0.346,-8.879,0.165,26.791,1.181,-10.263,0.760,19.528,0.605,-1.072,0.146,21.044,0.380,-1.798,0.251
3920000.0,23.825,2.470,-8.718,1.346,27.429,0.858,-10.607,0.628,20.585,0.691,-1.525,0.179,20.593,0.757,-1.585,0.292
3960000.0,23.179,0.756,-8.346,0.408,26.591,1.346,-10.147,0.699,20.049,0.221,-1.185,0.094,21.166,1.013,-1.814,0.553
4000000.0,23.611,0.953,-8.587,0.490,26.557,0.837,-10.141,0.447,19.736,0.125,-1.222,0.106,21.382,0.811,-1.785,0.277
}\dataAblationGnnmixer

\begin{tikzpicture}[every node/.style={transform shape}]

\begin{groupplot}[
    group style={
        group size=2 by 1,
        horizontal sep=1.3cm,
    },
    scale only axis,
    height=3.5cm,
    width=3.6cm,
    tick label style={font=\scriptsize},
    title style={font=\small\bfseries, yshift=1mm},
    grid=major,
    grid style={dashed, gray!30},
    every x tick scale label/.style={at={(xticklabel* cs:1.0, -12pt)}, anchor=north west, xshift=0pt},
    legend style={
        at={(1.15, 1.25)},
        anchor=south,
        legend columns=2,
        draw=black!20,
        fill=white,
        rounded corners=2pt,
        font=\scriptsize\sffamily\bfseries
    }
]

\nextgroupplot[
    title={PP (Grid 10, 4 Agents)},
    ymin=15, ymax=45, xmin=0, xmax=4000000,
    ylabel={Steps to catch prey},
    xlabel={Environment Steps},
    ylabel style={font=\scriptsize\bfseries},
    xlabel style={font=\scriptsize},
]
\addablationcurve{\dataAblationGnnmixer}{0}{9}{10}{hero curve}{laurelRed}{\coolname{} (GNN Mixer)}
\addablationcurve{\dataAblationGnnmixer}{0}{13}{14}{ablation curve}{ablatedGray}{\coolname{} (QMIX Mixer)}

\nextgroupplot[
    title={LJ (Grid 10, 5 Agents)},
    ymin=-20, ymax=-5, xmin=0, xmax=4000000,
    xlabel={Environment Steps},
    ylabel={Episodic Reward},
    xlabel style={font=\scriptsize},
    ylabel style={font=\scriptsize\bfseries, xshift=-5pt, yshift=-5pt},
]
\addablationcurvenolegend{\dataAblationGnnmixer}{0}{3}{4}{hero curve}{laurelRed}
\addablationcurvenolegend{\dataAblationGnnmixer}{0}{7}{8}{ablation curve}{ablatedGray}

\end{groupplot}

\end{tikzpicture}

%% file: 7_appendix.tex
\section*{Appendix}

\section{Log Distance Path Loss with Fading}
\label{appendix: pathloss}
We summarize the wireless channel model here for completeness.
The received power (dBm) under log-distance path loss with fading is
$P_r = P_t - K_\text{ref} - 10\eta\log_{10}({d}/{d_{0}})+\psi$,
where $P_t$ is transmit power, $K_\text{ref}$ is the reference-distance loss, $\eta$ is the path-loss exponent, and $\psi\sim\mathcal{N}(0,\sigma_\psi^2)$ models log-normal fading.

\section{Proofs}
\label{appendix: proof}

\begin{proof}[Proof of Theorem \ref{thm: gin emb} (\cite{LAUREL_ACML})]
We reproduce the core argument here for completeness and extend it with a quantitative finite-width bound.
We use Lemma~5 of Xu et al.~\cite{gin}.

\textbf{Permutation invariance.}
Vector addition is commutative, so for any permutation $\rho$, the sum $\sum_{k=1}^{|\mathcal{N}_i|}\func[MLP]{\bm{m}_{i\rho(k)}}$ equals $\sum_{j\in\mathcal{N}_i}\func[MLP]{\bm{m}_{ij}}$ regardless of the ordering. Therefore $\Phi$ is permutation invariant.

\textbf{Injectiveness.}
We use the following result from \cite{gin}.

\begin{lemma}[Xu et al.\ \cite{gin}, Lemma 5]
\label{lemma: gin}
Assume the message space $\mathcal{M}$ is countable.
There exists a function $\varphi : \mathcal{M}\rightarrow \mathbb{R}^{d'}$ such that $h(c_i) \coloneqq \sum_{j\in\mathcal{N}_i} \varphi\paren{\bm{m}_{ij}}$ is unique for each finite multiset $c_i$ over $\mathcal{M}$ of cardinality at most $N-1$.
\end{lemma}

\noindent The countability assumption is satisfied in our setting, since messages are neural network outputs expressed in finite-precision floating-point arithmetic, making $\mathcal{M}$ finite.

Since $\mathcal{M}$ is finite, the function $\varphi : \mathcal{M} \rightarrow \mathbb{R}^{d'}$ from Lemma~\ref{lemma: gin} is defined on a finite domain.
Any function on a finite domain can be exactly realized by a sufficiently wide MLP: one can construct a network whose hidden units each activate on exactly one element of $\mathcal{M}$ (an indicator construction analogous to a lookup table), achieving $\func[MLP]{\bm{m}} = \varphi(\bm{m})$ for every $\bm{m} \in \mathcal{M}$.
For this parameter setting, $\Phi(c_i) = \sum_{j\in\mathcal{N}_i} \varphi(\bm{m}_{ij})$ is unique per $c_i$ by Lemma~\ref{lemma: gin}, so $\Phi$ is injective.

\medskip
\noindent\textbf{Quantitative bound for finite-width networks (new contribution).}
We now derive an explicit reconstruction bound.
Under the indicator construction introduced above, set $K\coloneqq|\mathcal{M}|$ and $\varphi(\bm{m}_k)\coloneqq e_k\in\mathbb{R}^{K}$ (the $k$-th standard basis vector).
Then $\Phi(c_i)=\sum_{j\in\mathcal{N}_i}\varphi(\bm{m}_{ij})=\bm{n}(c_i)$ is the exact multiplicity vector of $c_i$.
For a width-$D$ MLP define the worst-case per-element approximation error
\begin{align}
  \label{eq: approx error}
  \varepsilon_D \;\coloneqq\; \max_{\bm{m}\in\mathcal{M}}\,\bigl\|\mathrm{MLP}_D(\bm{m})-\varphi(\bm{m})\bigr\|_2,
\end{align}
and the approximate encoder $\Phi_D(c_i)\coloneqq\sum_{j\in\mathcal{N}_i}\mathrm{MLP}_D(\bm{m}_{ij})$.

\textit{Claim.} For any two multisets $c_i,c_i'$ over $\mathcal{M}$ with $|c_i|,|c_i'|\le N-1$,
\begin{align}
  \label{eq: approx inject}
  &\bigl\|\bm{n}(c_i)-\bm{n}(c_i')\bigr\|_2 \nonumber\\
  &\quad\le
  \bigl\|\Phi_D(c_i)-\Phi_D(c_i')\bigr\|_2
  + 2(N\!-\!1)\varepsilon_D.
\end{align}

\textit{Proof of Claim.}
By the triangle inequality,
\begin{align}
\label{eq: triangle ineq}
&\|\Phi(c_i)-\Phi(c_i')\|_2 \nonumber\\
&\quad\le \|\Phi(c_i)-\Phi_D(c_i)\|_2
   + \|\Phi_D(c_i)-\Phi_D(c_i')\|_2 \nonumber\\
&\qquad + \|\Phi_D(c_i')-\Phi(c_i')\|_2.
\end{align}
For the first term, linearity of the sum and the triangle inequality give
\begin{align}
\label{eq: first term bound}
&\|\Phi(c_i)-\Phi_D(c_i)\|_2 \nonumber\\
&= \Bigl\|\sum_{j\in\mathcal{N}_i}\bigl[\varphi(\bm{m}_{ij})-\mathrm{MLP}_D(\bm{m}_{ij})\bigr]\Bigr\|_2 \nonumber\\
&\le \sum_{j\in\mathcal{N}_i}\bigl\|\varphi(\bm{m}_{ij})-\mathrm{MLP}_D(\bm{m}_{ij})\bigr\|_2 \nonumber\\
&\le |\mathcal{N}_i|\,\varepsilon_D
\le (N-1)\varepsilon_D,
\end{align}
and the identical bound holds for the third term.
Substituting and using $\Phi(c_i)=\bm{n}(c_i)$ yields~\eqref{eq: approx inject}.\hfill$\square$

Three consequences follow.
(i)~If $\Phi_D(c_i)=\Phi_D(c_i')$, then $\|\bm{n}(c_i)-\bm{n}(c_i')\|_2\le 2(N-1)\varepsilon_D$: two distinct multisets can share an approximate embedding only if their multiplicity vectors differ by at most $2(N-1)\varepsilon_D$, so the reconstruction error is explicitly controlled by the MLP approximation quality.
(ii)~Exact injectiveness ($\|\bm{n}(c_i)-\bm{n}(c_i')\|_2=0$ whenever $\Phi_D(c_i)=\Phi_D(c_i')$) is recovered when $\varepsilon_D=0$.
(iii)~$\varepsilon_D=0$ is achieved by a two-layer MLP of width $D=K$: assigning one neuron per message type (each unit activates on exactly one $\bm{m}_k$ and outputs $e_k$) exactly realises the lookup table $\varphi$ with zero approximation error.
For $D<K$, the universal approximation theorem~\cite{cybenko1989approximation} guarantees $\varepsilon_D\to 0$ as $D\to\infty$, so the bound in~\eqref{eq: approx inject} tightens monotonically with network width and is consistent with Remark~\ref{remark: lossless} in the main text.
\end{proof}

\phantomsection
\label{appendix: prop diff}
\begin{proof}[Proof of Remark~\ref{prop: differentiability}]
Fix a step $t$ and an agent $i$.
Under causal alignment, the step-$t$ agent network takes as inputs observations $(o_i^{\mathcal{T},t}, o_i^{\mathcal{C},t})$ and the \emph{buffered} received-message set $\bm{c}_i^{t-1}$, and outputs $(a_i^t,m_i^t,Q_i^t)$.
The encoded message $\phi_i^t=\Phi(\bm{c}_i^{t-1})$ is computed from data realized at the end of step $t{-}1$; hence $\bm{c}_i^{t-1}$ is treated as a fixed input when differentiating the step-$t$ loss with respect to parameters $\theta$.
Crucially, the stochastic wireless channel operation that produced $\bm{c}_i^{t-1}$ from other agents' previous-step transmissions,
$\{m_j^{t-1},a_j^{\mathcal{C},t-1}\}_{j\neq i}\mapsto \bm{c}_i^{t-1}$,
is part of the environment and does not appear as an operator inside the step-$t$ agent network.
All components within the step-$t$ agent computation (observation fusion, message encoding $\Phi$, recurrent state update, and output heads) are differentiable neural modules; in particular, $\Phi$ is implemented as a sum of MLP outputs (Equation~\ref{eq: how gnn}).
Thus $\nabla_\theta \mathcal{L}$ exists and can be computed by standard back-propagation through time.
\end{proof}

\phantomsection
\label{appendix: proof pi}
\begin{proof}[Proof of Lemma~\ref{lemma: PE}]
Permutation equivariance holds by the definition of \textit{PEHypernet}: since $\psi_A^l(\cdot)$ is applied independently to each agent's input $\hat{s}_i$ (with no cross-agent coupling), permuting the inputs is equivalent to permuting the outputs.
Let $\rho$ be a permutation and let $\bm P$ denote its permutation matrix, using the convention $P_{i,\rho(i)}=1$. Since \textit{PEHypernet{\_}A} applies the same function $\psi_A^l(\cdot)$ independently to each input, we have
\begin{align}
\label{eq: PEHyper}
&\Psi_A^l\!\paren{\rho\paren{\hat{s}_1,\hat{s}_2,\ldots,\hat{s}_N}} \nonumber\\
&= \paren{\psi_A^l(\hat{s}_{\rho^{-1}(1)}),\ldots,\psi_A^l(\hat{s}_{\rho^{-1}(N)})} \nonumber\\
&= \rho\!\paren{\Psi_A^l\!\paren{\hat{s}_1,\hat{s}_2,\ldots,\hat{s}_N}}.
\end{align}
Therefore, $\Psi_A^l$ is permutation equivariant. The same argument applies to $\Psi_B^l$, so $(\xi_1^l,\ldots,\xi_N^l)$ is also permutation equivariant.
\end{proof}

\begin{proof}[Proof of Theorem~\ref{theorem: PI}]
By Equation~\ref{eq: gnnHyper}, when $l=1$,
\begin{equation}
    \label{eq: h1}
    \bm{z}_i^1 = \sigma\paren{\bm{W}^{1}_{i} \cdot{Q}_i  + g\paren{ \set {{Q}_j \given j\in \mathcal{N}_i}; \xi^1_i}}
\end{equation}
Let permutation $\rho(\cdot)$ be applied on the indices of agents. Consequently, $(Q_1, Q_2, \ldots, Q_N)$
and $(\hat{s}_1, \hat{s}_2, \ldots, \hat{s}_N)$ are both permuted with $\rho$.
We use $'$ to denote the corresponding variables after permutation.
By Lemma~\ref{lemma: PE}, ${{\bm{W}'}}^{l}_{i} = {\bm{W}}^{l}_{\rho^{-1}(i)}$, and ${\xi'}^{l}_{i} = {\xi}^{l}_{\rho^{-1}(i)}$.
Thus,
\begin{align}
    \label{eq: hprime1}
    \begin{split}
    {{\bm{z}'}_i^1}
    &= \sigma\paren{{\bm{W}'}^{1}_{i} \cdot{{Q}'}_i  + g\paren{ \set {{{Q}'}_j \given j\in \mathcal{N}_i}; {\xi'}^1_i}}\\
    &= \sigma\paren{{\bm{W}}^{1}_{\rho^{-1}(i)} \cdot{{Q}}_{\rho^{-1}(i)}  + g\paren{ \set {{{Q}'}_j \given j\in \mathcal{N}'_i}; {\xi}^1_{\rho^{-1}(i)}}}
    \end{split}
\end{align}
Let the original adjacency matrix be $\bm{A}$, and let $\bm{P}$ denote the permutation matrix for $\rho$. Then $\bm{A}' = \bm{P}^{T} \bm{A} \bm{P}$ \cite{perm_matrix}.
By definition, $\mathcal{N}'_i$ is the set of column indices of nonzero elements in the $i$th row of $\bm{A}'$, giving
\begin{align}
    \begin{split}
    \label{eq: adj}
    &\set{Q'_j \given j \in \mathcal{N}'_i}
    = \set{Q'_{j} \given j=\rho(k), k\in\mathcal{N}_{\rho^{-1}(i)}}\\
    &= \set{Q_{k} \given k\in\mathcal{N}_{\rho^{-1}(i)}}.
    \end{split}
\end{align}
Combining Equations~\ref{eq: hprime1} and~\ref{eq: adj},
\begin{align}
    \label{eq: hprime2}
    {{\bm{z}'}_i^1}
    &= \sigma\biggl({\bm{W}}^{1}_{\rho^{-1}(i)} \cdot{{Q}}_{\rho^{-1}(i)} \nonumber\\
    &\quad + g\paren{ \set{Q_{k} \given k\in\mathcal{N}_{\rho^{-1}(i)}}; {\xi}^1_{\rho^{-1}(i)}}\biggr).
\end{align}
Comparing Equations~\ref{eq: h1} and~\ref{eq: hprime2}, we conclude ${{\bm{z}'}_i^1} = \bm{z}_{\rho^{-1}(i)}^{1}$ if and only if $g(\cdot)$ is permutation invariant.
Thus $(\bm{z}_{1}^{1}, \ldots, \bm{z}_{N}^{1})$ is permutation equivariant with $(Q_1, \ldots, Q_N)$.
The result extends to all layers $L \ge 1$ by the identical argument applied inductively with $\bm{z}^0 = Q$.
Applying a permutation-invariant pooling function $\mathrm{pool}(\cdot)$ to the final layer then yields the same $Q_{\text{tot}}$ regardless of agent ordering.
\end{proof}

\phantomsection
\label{appendix: proof mono}
\begin{proof}[Proof of Theorem~\ref{theorem: monotonicity}]
In Equation~\ref{eq: gnnHyper}, the $\textit{\textbf{absolute}}(\cdot)$ operation ensures $\bm{W}_i^l \ge 0$ and $\xi_i^l \ge 0$ elementwise for all $i,l$.
By the chain rule, the Jacobian of $\bm{z}_i^l$ with respect to $\bm{z}_j^{l-1}$ decomposes into three cases:
\begin{itemize}
  \item \textbf{Self-connection} ($j = i$): $\frac{\partial \bm{z}_i^l}{\partial \bm{z}_i^{l-1}} = \mathrm{diag}(\sigma'(\mathrm{pre}_i^l))\bm{W}_i^l \ge 0$, since $\sigma'(\cdot)\ge 0$ and $\bm{W}_i^l\ge 0$.
  \item \textbf{Incoming neighbor} ($j \in \mathcal{N}_i,\; j \ne i$):
  \begin{align}
  \label{eq: jacobian neighbor}
  \frac{\partial \bm{z}_i^l}{\partial \bm{z}_j^{l-1}} = \mathrm{diag}(\sigma'(\mathrm{pre}_i^l))\,\frac{\partial g}{\partial \bm{z}_j^{l-1}} \ge 0,
  \end{align}
  since $\sigma'(\cdot)\ge 0$ and $\frac{\partial g}{\partial \bm{z}_j^{l-1}}\ge 0$ (theorem hypothesis). For the linear-aggregation example, $\frac{\partial g}{\partial \bm{z}_j^{l-1}} = (\xi_i^l / |\mathcal{N}_i|)\,\bm{I}_d \ge 0$ elementwise, since $\xi_i^l \ge 0$ by construction.
  \item \textbf{Non-neighbor} ($j \notin \mathcal{N}_i \cup \{i\}$): $\frac{\partial \bm{z}_i^l}{\partial \bm{z}_j^{l-1}} = 0$.
\end{itemize}
Here $\mathrm{pre}_i^l := \bm{W}_i^l \bm{z}_i^{l-1} + g(\{\bm{z}_j^{l-1}\mid j\in\mathcal{N}_i\};\xi_i^l)$.
Since $\bm{z}_i^0 = Q_i$, repeated application of the chain rule across the $L$ GNN layers and the final pooling implies $\frac{\partial f}{\partial Q_i} \geq 0$ for all $i$.

A simple example of $g(\cdot)$ satisfying the theorem condition is linear aggregation:
\begin{align}
\label{eq: linear agg}
&g\paren{\set{\bm{z}_j^{l-1} \given j\in\mathcal{N}_i}; \xi_i^l} \nonumber\\
&\quad=
\xi_i^l \cdot \frac{1}{|\mathcal{N}_i|}\sum_{j\in\mathcal{N}_i}\bm{z}_j^{l-1},
\end{align}
where $\xi_i^l = \textit{\textbf{abs}}\paren{\bm{\psi}_B^l(\hat{s}_i)} \ge 0$ by construction.
\end{proof}

\phantomsection
\label{appendix: proof exp power}
\begin{proof}[Proof of Theorem~\ref{thm: exp power}]
\textbf{Containment.}
Any graph-agnostic mixer $F(\bm Q,s)$ can be realized by the proposed
architecture by setting the neighbor-aggregation term to zero,
reducing to a standard monotone mixing network.
Hence $\mathcal{F}_{\mathrm{agn}}\subseteq \mathcal{F}_{\mathrm{cond}}$.

\smallskip
\textbf{Strictness.}
Take $N=2$ and define
\begin{align}
\label{eq:witness}
Q_{\mathrm{tot}}^\star(\bm Q,G) \;:=\; Q_2 \;+\; \mathbf{1}\{(1\!\to\!2)\in G\}\,Q_1,
\end{align}
where $\mathbf{1}\{\cdot\}$ is the indicator function. $Q_{\mathrm{tot}}^\star$ is
monotone in each $Q_i$.

First, $Q_{\mathrm{tot}}^\star \notin \mathcal{F}_{\mathrm{agn}}$.
Fix $\bm Q=(1,0)$ and consider two delivery relations $G$ and $G'$ differing
only by edge $(1\!\to\!2)$.
Then $Q_{\mathrm{tot}}^\star(\bm Q,G)=1$ and $Q_{\mathrm{tot}}^\star(\bm Q,G')=0$,
whereas any $F(\bm Q,s)$ must output the same value for both.

Second, $Q_{\mathrm{tot}}^\star \in \mathcal{F}_{\mathrm{cond}}$.
Instantiate the GNN mixer with one aggregation round and sum pooling:
let node~2 aggregate incoming neighbor values by
$g(\{Q_j:j\in\mathcal{N}_2\})=\sum_{j\in\mathcal{N}_2} Q_j$; the pooled
representation equals $Q_2+\mathbf{1}\{1\in\mathcal{N}_2\}Q_1$, realizing~\eqref{eq:witness}.

Therefore $\mathcal{F}_{\mathrm{agn}} \subsetneq \mathcal{F}_{\mathrm{cond}}$.
\end{proof}

\section{Augmented Dec-POMDP: Complete Formal Definition}
\label{appendix: aug mdp def}

\begin{definition}[\textbf{Augmented Dec-POMDP}]
The aligned game and wireless environments jointly define the tuple $\paren{\mathcal{S}, \bm{\mathcal{A}}, \mathcal{T}, \bm{\Omega}, \mathcal{O}, \mathcal{R}, \gamma}$, where:
\begin{itemize}
 \item $\mathcal{S} = \mathcal{S}^{\mathcal{T}} \times \mathcal{S}^{\mathcal{C}}$ is the \textit{augmented state space}. $\mathcal{S}^{\mathcal{T}}$ denotes the game state (\eg, agent positions and prey location). $\mathcal{S}^{\mathcal{C}}$ denotes the wireless state, including the realized message buffers $\{\bm{c}_i^{t-1}\}_{i=1}^{N}$ and channel descriptors (\eg, path-loss coefficients and obstacle layout).
 \item $\bm{\mathcal{A}} = \bm{\mathcal{A}}^{\mathcal{T}} \times \bm{\mathcal{A}}^{\mathcal{C}}$ is the \textit{augmented joint action space}. $\bm{\mathcal{A}}^{\mathcal{T}}$ contains the game actions; $\bm{\mathcal{A}}^{\mathcal{C}}$ contains communication actions, primarily binary transmit $a_i^C\!\in\!\{0,1\}$ (the formulation accommodates extensions such as variable transmission power or contention parameters).
 \item $\mathcal{T}: \mathcal{S} \times \bm{\mathcal{A}} \times \mathcal{S} \rightarrow [0,1]$ is the transition kernel over the augmented state.
 \item $\bm{\Omega} = \bm{\Omega}^{\mathcal{T}} \times \bm{\Omega}^{\mathcal{C}}$ is the \textit{augmented joint observation space}. $\bm{\Omega}^{\mathcal{T}}$ contains game observations; $\bm{\Omega}^{\mathcal{C}}$ contains local wireless measurements (\eg, RSS $P_s$, ACK-related statistics).
 \item $\mathcal{O}: \mathcal{S} \times \bm{\mathcal{A}} \rightarrow \Delta(\bm{\Omega})$ is the observation function on the augmented spaces.
 \item $\mathcal{R}: \mathcal{S} \times \bm{\mathcal{A}} \rightarrow \mathbb{R}$ is the shared game reward. The wireless environment does not provide an explicit reward; it shapes the return by dictating how shared information influence future game actions.
 \item $\gamma \in (0,1)$ is the discount factor.
\end{itemize}
\end{definition}

\section{Additional Wireless Observations}
\label{appendix: other measurements}

The augmented MDP formulation (Section~\ref{sec: method mdp}) supports wireless observations beyond the RSS values used in our experiments (Section~\ref{sec: exp}). We describe two ACK-based extensions here; they are natural candidates for the richer communication actions noted in Section~\ref{sec: conclusion}.

\parag{Bandwidth estimation via ACK}
When an agent broadcasts, it requests acknowledgements from other agents and records how many are received.
The ratio of received ACKs to total agents yields an estimate of the packet reception rate (PRR), which serves as a proxy for available bandwidth under the current contention level.

\parag{Distributed RSS via ACK payload}
Additional network measurements can be piggybacked on ACK packets.
For example, after agent~$i$ receives agent~$j$'s message, $i$ measures the local RSS and includes it in the ACK sent back to~$j$.
As a result, both $i$ and $j$ obtain the RSS at agent~$i$'s position, enabling each agent to build a richer picture of the spatial link-quality distribution.

\section{Details in Experiments}

\subsection{Communication Protocols}
\label{appendix: communication protocols}
We summarize the $p$-CSMA protocol and reception model here for completeness.
Each transmitting agent draws a random back-off counter uniformly from $\{0,\ldots,W{-}1\}$ and, upon expiry, senses the channel: if the received energy is below the carrier-sense threshold $\Theta_f$, it transmits with probability $p$; otherwise it redraws.
Decoding is governed by SINR: a packet with signal power $P_s$ is decoded correctly when $P_s \ge \Theta_r(P_s'+N_0)$, where $P_s'$ is aggregate interference and $N_0$ is the noise floor; otherwise it is either detected but not decoded ($P_s \ge \Theta_f$) or entirely missed ($P_s < \Theta_f$).
Here $\Theta_f$ and $\Theta_r$ are physical-layer threshold constants (carrier-sense and SINR thresholds, respectively), not to be confused with the network parameters~$\theta$.

\subsection{Parameters of Wireless Environment}
\label{appendix: wireless parameters}
\parag{Default settings used in Section~\ref{sec: exp}}
Table~\ref{tab: wireless params} lists the wireless simulation parameters.

\begin{table}[h]
\centering
\caption{Default wireless simulation parameters.}
\label{tab: wireless params}
\begin{tabular}{lcc}
\toprule
Parameter & Symbol & Value \\
\midrule
Obstacle attenuation & $\att_0$ & 4.5 \\
Noise floor & $N_0$ & $-95$ dBm \\
Carrier-sense threshold & $\Theta_f$ & $-78$ dBm \\
SINR threshold & $\Theta_r$ & 15 dB (20 dB for \ppw{n}) \\
$p$-CSMA contention probability & $p$ & 0.3 \\
$p$-CSMA counter window & $W$ & 15 time slots \\
\bottomrule
\end{tabular}
\end{table}

\subsection{Training Hyperparameters and Neural Network Architecture}
\label{appendix: training hyperparameters}
As stated in Section~\ref{sec: exp}, all experiments are repeated with 5 random seeds, consistent with the MARL literature~\cite{smac,qmix_journal}. We report the mean and standard deviation of each metric over these seeds. Shaded regions in learning curves and $\pm$ values in tables denote $\pm 1$ standard deviation.

\parag{Agent-level architecture}
For all algorithms, we use a GRU cell with a 2-layer MLP of hidden dimension 128 as the agent-level $Q$-value network. The message encoder is a 2-layer MLP with hidden dimension 128 (Equation~\ref{eq: how gnn}).

\parag{Mixer architecture}
For VDN, QMIX, TarMAC+VDN, and TarMAC+QMIX, we use a 3-layer MLP with hypernetwork~\cite{hypernet} as the mixing network. For \coolname{}, the mixing network is the GNN-based mixer of Section~\ref{sec: method enhanced mixer} with the following specifications:
\begin{itemize}
\item \textbf{GNN hidden dimension:} $d = 64$.
\item \textbf{Message-passing rounds:} $L=2$, which covers the graph diameter for the agent counts tested (3--5 agents). We use 2 rounds; sensitivity to GNN depth is an interesting direction for future work.
\item \textbf{Neighbor aggregation:} Linear aggregation $g(\{\bm{z}_j^{l-1}\mid j\in\mathcal{N}_i\};\xi_i^l) = \xi_i^l \cdot \frac{1}{|\mathcal{N}_i|}\sum_{j\in\mathcal{N}_i}\bm{z}_j^{l-1}$, with $\xi_i^l = \textit{\textbf{abs}}(\bm{\psi}_B^l(\hat{s}_i)) \ge 0$, satisfying the Jacobian condition of Theorem~\ref{theorem: monotonicity}. We use linear aggregation for simplicity; attention-based aggregation is a natural extension but would violate the non-negative Jacobian requirement (see Section~\ref{sec: method enhanced mixer}).
\item \textbf{PEHypernet:} Each of $\psi_A^l$ and $\psi_B^l$ is a 2-layer MLP with hidden dimension 64 and ReLU activation. Input dimension equals the concatenated global state and local observation dimension; output dimensions are $d \times d$ (for $\bm{W}_i^l$) and $d$ (for $\xi_i^l$), respectively.
\item \textbf{Second-stage MLP:} A 2-layer MLP with hidden dimension 32 and non-negative weights maps the pooled vector $\bm{v}$ to scalar $Q_\text{tot}$.
\item \textbf{Pooling:} Sum pooling (Equation~\ref{eq: pooling}).
\end{itemize}

\parag{Training}
All algorithms are trained with Adam~\cite{adam} at learning rate $5\times10^{-4}$.
The target network is updated via hard replacement every 200 episodes.